\documentclass[twoside]{article}

\usepackage[accepted]{aistats2023}

\usepackage{amsmath, amssymb, xcolor, amsthm}
\usepackage{bbold}
\usepackage{algorithm}
\usepackage{algorithmic}
\usepackage{hyperref}
\usepackage{graphicx}
\usepackage{subcaption}
\usepackage{comment}

\def \be*{\begin{eqnarray*}}
\def \e*{\end{eqnarray*}}
\def \beg{\begin{eqnarray}}
\def \en{\end{eqnarray}}
\def \di{\displaystyle}
\def \bit{\begin{itemize}}
\def \eit{\end{itemize}}

\def \P{\mathbb P}
\def \ind{\mathbb 1}
\def \N{\mathbb N}

\def \w {\widehat}

\newtheorem{prop}{{\sc Proposition}}
\newtheorem{lem}{{\sc Lemme}}

\usepackage[round]{natbib}

\begin{document}

%

%

\twocolumn[

\aistatstitle{Data Augmentation for Imbalanced Regression}

\aistatsauthor{ Samuel Stocksieker \And Denys Pommeret \And Arthur Charpentier }

\aistatsaddress{Univ. Lyon, UCBL, ISFA LSAF  \And Aix-Marseille Univ., CNRS, I2M  \And UQAM - Montréal } ]

\begin{abstract}

In this work, we consider the problem of imbalanced data in a regression framework when the imbalanced phenomenon concerns continuous or discrete covariates. Such a situation can lead to biases in the estimates.  
In this case, we propose a  data augmentation algorithm that combines a weighted resampling (WR) and a data augmentation (DA) procedure. 
In a first step, the DA procedure permits exploring a wider support than the initial one. In a second step, the WR method drives the exogenous distribution to a target one. 
We discuss the choice of the DA procedure 
through a numerical study that illustrates the advantages of this  approach. Finally, an actuarial application is studied.  

\end{abstract}


\section{INTRODUCTION} 

Many real world forecasting problems 
are based on predictive models in a supervised learning framework. Most of these algorithms fail when the variable of interest is imbalanced. This situation also occurs when the population observed is significantly different from the true population. This is the case when certain parts of the support are observed less than in theory or are even not observed. We  may face unrepresentative training data due to sampling noise or due to the sampling method, even for large samples. Such selection bias or sampling bias, could affect the models and impact the results, by biasing the estimates.  
We naturally find this problem in clinical trial data (\cite{Rubin2012rerandom}, \cite{Friede2006}, \cite{Kahan2015}, \cite{Ciolino2011}, \cite{Ciolino2013}) and in survey sampling theory (\cite{Yang2021},\cite{GUPTA2022101513}, \cite{tille2022some}). We also have such an issue in Regression Discontinuity (\cite{caughey2011elections}, \cite{Peng2019RD}, \cite{Frolich2019RD}, \cite{Cattaneo2021RD}). 
This problem also appears  in actuarial science as illustrated in our application. 

This kind of situation can strongly impact standard learning algorithms, particularly if the rarely observed (or unobserved) values are the most relevant for the modeling. The learning from imbalanced data concerns many problems with numerous applications in different fields, with still some open issues and challenges (\cite{Krawczyk2016LearningFI}). This  research topic, very active, has mostly focused on solving classification tasks where many solutions have been discussed (\cite{Branco2016Survey}, \cite{fernandez2018learning}, \cite{fernandez2018smote}). 

However, very few works exist in a  regression framework. One of the best solutions currently proposed is to use the concept of utility-based regression defining the notion of relevance for a continuous target variable (\cite{torgo2015resampling}, \cite{branco2016ubl}, \cite{Brancophdthesis}, \cite{Ribeiro2020}). \cite{yang2021delving} propose to use distribution smoothing to handle this issue.
All these different techniques offer specific treatments to deal with the imbalanced distribution of variables of interest. 
Several sampling techniques for regression models (\textit{Pair Bootstrap}, \textit{Stratified Bootstrap}, \textit{Residuals Resampling - Wild Bootstrap}, etc.) have already been proposed in order to improve learning or get/reduce confidence intervals, without directly dealing with the imbalanced phenomenon \cite{Horowitz}, \cite{flachaire}. There also exists some sample adjustment techniques that can be used to create a synthetic dataset with statistical similarities to the true population of interest \cite{templ2017simulation} but the covariates considered are categorical or discrete.

Finally, there does not seem to be any work handling specifically the case of imbalanced continuous covariates and no method proposes to adjust the sample according to a continuous target covariate distribution.

In this work, we consider the problem   of imbalanced continuous covariates  in a regression context and we propose two solutions: first, a weighted resampling (WR), driving the distribution of the covariates  to the target one; second, a two-step data augmentation algorithm that combines a data augmentation (DA) procedure, exploring a wider support than the initial one, with the WR procedure. Interestingly, that method is valid for continuous, or discrete, as well as categorical covariates.
A numerical study illustrates the advantages of combining this method with six different DA approaches through measures of predictive performance indicators in regression. As a particular case, we also study the WR method which corresponds to a one-step version of the algorithm. 

The main contributions of this paper are: i) analyzing the potential impact of an imbalance of a covariate in a regression context; ii) introducing a WR method to deal with this phenomenon; and iii) presenting a combination of this method with several data generators to avoid an overfitting phenomenon and try to improve the regression.
The paper is organized as follows: In Section 2 we describe the WR procedure and we introduce the DA-WR algorithm.  
Section 3 contains simulation results and Section 4  presents an analysis of a dataset from the literature. The perspectives of this work are presented in Section 5.  Code,
data and results are available at: \url{https://github.com/sstocksieker/DAIR}.

\section{WR AND DA-WR ALGORITHMS}

\subsection{Notation}
Consider a sequence of 
iid random variables $ \{(\boldsymbol{X}_1,Y_1), \cdots, (\boldsymbol{X}_n,Y_n)\}$, which are realizations of $(\boldsymbol{X},Y)$, where the variable of interest $Y$ is univariate and the covariates $\boldsymbol{X} \in \mathcal{X}$ is a $p$-dimensional vector. $\boldsymbol{X}$ and $Y$ are supposed to be continuous or discrete and we consider the regression framework where the objective is to explain and predict $Y$ 
according to the following structure:
\begin{equation*}
    \label{regression} \mathbb{E}(Y|\boldsymbol{X}) = m(\boldsymbol{X}), 
\end{equation*}
for some  function $m$, parametric or not.  
Here we consider a random design where $\boldsymbol{X} = ({X_1,\cdots,X_p})$ 
has an unknown joint cumulative distribution function (cdf)  $F$ with  probability $\mathbb{P}$. We assume a target distribution $F_0$, which can be the distribution function of $\boldsymbol{X}$ in the true population, which can be known, or a distribution of interest.

When the observations of $\boldsymbol{X}$ are far from the target distribution we call this situation an imbalanced covariates regression, or simply an imbalanced regression here. This issue is slightly different from the imbalanced situation in a regression framework, in literature, because it handles one or more covariates ($X$) and not the target variable ($Y$). This case allows us to easily get a target distribution to drive the data generation (for example, from the population in survey analyses). 
More precisely, we should say that the imbalanced phenomenon occurs when $F \neq F_0$. To measure the degree of imbalance we can denote by  $\widehat{F}$ and $\widehat{\mathbb{P}}$ the empirical estimators and we propose the following definition: we  face a $(\alpha, \beta)$-imbalanced regression problem if there is a set $\chi\subset\mathcal{X}$ with  $\mathbb{P}_0(\boldsymbol{X}\in\chi) \geq \beta$ such that $\lvert \frac{\widehat{\mathbb{P}}(\boldsymbol{X}\in\chi)}{\mathbb{P}_0(\boldsymbol{X}\in\chi)} - 1 \rvert > \alpha$.  
In other words, an imbalanced regression means having a  sample significantly different from the target population for at least a significant part of the support of $\boldsymbol{X}$. 
Clearly, the larger the values of  $(\alpha, \beta)$, the greater the degree of imbalance.
This problem also depends of the sample size. If $F$ and $F_0$ have the  same support, then  the phenomenon will diminish with $n$. On the other hand, if a part of the support of $\boldsymbol{X}$ is never observed then the problem may persist even for large $n$.   

To handle this situation, we want to draw a $n^*$-sample, $\{(\boldsymbol{X}^*_i,Y^*_i)_{i=1,\cdots,n^*}\}$ from the initial $n$-sample $\{(\boldsymbol{X}_i,Y_i)_{i=1,\cdots,n}\}$, such that  the cdf $F^*$ of  $\boldsymbol{X^*}$ converges to the target  $F_0$ with associated probability $\mathbb{P}_0$.

We consider here the case where  the random vector $\boldsymbol{X}$ is composed of discrete or   continuous variables and in both cases, we denote by   $f$ its joint probability  density function (pdf) or probability mass function. 
For simplicity of notation, we consider the case $p=1$ and  from now on we then write $X$ instead of $\boldsymbol{X}$. The method can be extended to the multivariate case $p>1$ by using joint density estimators. Moreover, the case where X is qualitative could be treated in exactly the same way.

\subsection{WR algorithm}

We consider a classical kernel estimator of $f$ 
\begin{align*}
\w{f}(x) & = \di\frac{1}{n h_{n}}\di\sum_{j=1}^{n}{\cal K}\left(\frac{X_j-x}{h_{n}}\right),
\end{align*}
with an appropriate bandwidth $h_{n}$. 
To avoid unstable  numerical results we add a trimming  sequence $e_{n}$, such that $e_n \to 0$, as $n \to \infty$. Write \begin{align*}
    \w{f}_{e_{n}}=\max(\w{f},e_{n}) & \; {\rm  and \ }{f}_{e_{n}}=\max({f},e_{n}),
\end{align*}
and we now consider $\w f_{e_n}$ instead of $\widehat f$. We write $f_0$ the target probability density function associated to $F_0$. 

Define \\ 
- $\omega_i = \frac{f_0(X_{i})}{\w{f}_{e_n}(X_{i})}$ the drawing weight of observation  $i$, and \\
- $q_i = \frac{\omega_i}{\sum_j \omega_j}$ the normalized drawing weight of observation $i$ such that $\sum_i q_i = 1$

The  \textit{weighted resampling} is inspired by \cite{smith1992bayesian} and consists in sampling the observations according to their drawing probabilities $q_i$. This variant of the classical Bootstrap \cite{efron1982jackknife} is close to the Monte Carlo method \textit{Sampling Importance Resampling} \cite{rubin1988using}.  
We draw a random variable $X^*$ from $X_1,\cdots, X_n$ using the probabilities $q_i$; that is $\P(X^*=X_i) = q_i$.   
By construction the cdf of $X^*$  given $X_1,\cdots, X_n$ is   
\begin{eqnarray*}
F^*(x) & = & \mathbb{P}(X^* \leq x) \\ 
& = & \sum_{i=1}^n \mathbb{P}(X^*=X_i)\ind_{[-\infty,x]}(X_i) \\
& = & \sum_{i=1}^n q_i \ind_{[-\infty,x]}(X_i).
\end{eqnarray*}
Clearly, $X^*$ and $F^*$ depend on the sample size $n$ but we omit the index $n$ in the notation. 

We introduce a classical assumption 
\begin{description}
\item {\bf (C)} : $h_n^k + \frac{\log(n)}{nh_n} =O( e_n^2)$ holds and  $f \in {\cal C}^k$ ($k$ times derivable) for some $k\in \N^*$.
\end{description}
\begin{prop}
\label{propWR}
Assume that {\bf (C)} holds and that the support of  $F$ contains the support of  $F_0$. 
Then for all $x \in {\cal X}$, the cdf $F^*(x)$ of $X^*$ 
converges in probabilities to $F_0(x)$ as $n \to +\infty$.
\end{prop}

\label{proofWR}
\begin{proof}

We recall a result of Collomb (1984): 
\begin{lem}[Collomb, G. 1984]
\label{lem1}
Assume that $f \in {\cal C}^k$, then we have  
\begin{align*}
     \sup_x | \w f(x) -f(x) | = O_{\mathbb P}(h_n^k+log(n)/(nh_n))
\end{align*}
\end{lem}

   We have     $\forall x \in \mathcal{R}$ : \\ 
       \begin{eqnarray*}
       F^*(x) &  = & \sum_{i=1}^{n} q_i \ind_{[-\infty,z]}(X_i)\\ 
       &  = & \frac{\frac{1}{n}\sum_{i=1}^{n} \omega_i \ind_{[-\infty,x]}(X_i)}{\frac{1}{n}\sum_{i=1}^{n} \omega_i} \\
       &  = & \frac{\frac{1}{n}\sum_{i=1}^{n} \frac{f_0(x_{i})}{\w{f}_{e_n}(X_{i})} \ind_{[-\infty,x]}(X_i)}{\frac{1}{n}\sum_{i=1}^{n} \frac{f_0(X_{i})}{\w{f}_{e_n}(X_{i})}}
       \\
      & :=  \displaystyle \frac{\w u_n}{\w v_n}, 
        \end{eqnarray*}
where 
\begin{align*}
\widehat u_n & =     \frac{1}{n}\sum_{i=1}^{n} \frac{f_0(X_i)}{\widehat{f}_{e_n}(X_i)} \ind_{[-\infty,x]}(X_i)\\
\widehat v_n & = \frac{1}{n}\sum_{i=1}^{n} \frac{f_0(X_i)}{\widehat{f}_{e_n}(X_i)} \; > \; 0.
\end{align*}
Write
\begin{align*}
u_n & =     \frac{1}{n}\sum_{i=1}^{n} \frac{f_0(X_i)}{{f}(X_i)} \ind_{[-\infty,x]}(X_i)
\\
v_n & = \frac{1}{n}\sum_{i=1}^{n} \frac{f_0(X_i)}{{f}(X_i)} >0.
\end{align*}
We want to prove that 
\begin{align*}
    \widehat u_n - u_n  \to_{\mathbb{P}} 0 & {\rm \ \ and \ \ } 
    \w v_n - v_n  \to_\mathbb{P} 0.
\end{align*}
We can write
\begin{align*}
    \widehat u_n - u_n & = 
    \frac{1}{n}\sum_{i=1}^{n} f_0(x_i) \ind_{[-\infty,x]}(X_i)\big( \frac{1}{\widehat{f}_{e_n}(X_i)}- \frac{1}{f(X_i)} \big).
\end{align*}
Noting that
$
\frac{1}{\hat{f}_{e_{n}}}=\frac{1}{f}+\frac{f-f_{e_{n}}}{f_{e_{n}}f}+\frac{f_{e_{n}}-\hat{f}_{e_{n}}}{f_{e_{n}}\hat{f}_{e_{n}}}
=\frac{1}{f}+\frac{f-e_{n}}{f_{e_{n}}f}\mathbf{1}_{\{f<e_{n}\}}+\frac{f_{e_{n}}-\hat{f}_{e_{n}}}{f_{e_{n}}\hat{f}_{e_{n}}}$,
and that $|f_{e_{n}}(x)-\hat{f}_{e_{n}}(x)|\leq|f(x)-\hat{f}(x)|\,$ we obtain
\begin{align*}
    |\widehat u_n - u_n| & \leq 
    \frac{1}{n}\sum_{i=1}^{n} |f_0(X_i) \ind_{[-\infty,x]}(X_i)| | {\frac{1}{\widehat{f}_{e_n}(X_i)}}-\frac{1}{ f(X_i) }|
    \\
    \leq & 
    \frac{1}{n}\sum_{i=1}^{n} |f_0(X_i) \ind_{[-\infty,x]}(X_i)| \times 
    \bigg\lbrace \\
    &   | {\frac{f(X_i)-e_n}{f(X_i)\widehat{f}(X_i)}}\ind_{ f(X_i)<e_n} |
    + \frac{| f_{e_n}(X_i) -\widehat{f}_{e_n}(x_i)| }{\widehat{f}_{e_n}(X_i)f_{e_n}(X_i)}  \bigg\rbrace
    \\ 
    &  : = A + B.
\end{align*}
We have 
\begin{align*}
    A & \leq 
    \frac{1}{n}\sum_{i=1}^{n} |g_0(X_i) \ind_{[-\infty,x]}(X_i)| \frac{e_n}{f^2(X_i)}\ind_{f(X_i)<e_n}
    \\
    & \leq 
    \frac{1}{n}\sum_{i=1}^{n} |f_0(X_i) \ind_{[-\infty,x]}(X_i)| \frac{1}{f^2(X_i)}e_n
    \\
    & = 
    A_n \; e_n 
      \to_{\mathbb{P}} 0, { \rm \ \ as \ } n \to \infty,
\end{align*}
since $A_n$ converges by the Law of Large Numbers   and $e_n \to 0$. 
We now consider $B$
\begin{align*}
      B & \leq 
    \frac{1}{n}\sum_{i=1}^{n} |f_0(X_i) \ind_{[-\infty,x]}(X_i)| |\frac{\widehat{f}_{e_n}(X_i)-{f}_{e_n}(X_i)}{\widehat{f}_{e_n}(X_i){f}_{e_n}(X_i)}|
    \\
    & \leq
    \frac{1}{n}\sum_{i=1}^{n} |f_0(X_i) \ind_{[-\infty,x]}(X_i)| \frac{\sup|{\widehat{f}(x)-{f}(x)}|}{e_n^2}
    \\
    & = 
    B_n \; \frac{\sup|{\widehat{f}(x)-{f}(x)}|}{e_n^2}
\end{align*}
which converges  to zero in probabilities since $B_n$ converges by law of large numbers, 
and by Lemma \ref{lem1} combined with Assumption {\bf (C)}.  
Finally, we deduce  that 
$|\widehat u_n - u_n| \to_{\mathbb{P}} 0$. Similar considerations apply to $|\widehat v_n -v_n|$ and we conclude that 
\begin{align*}
F^*(x) -\frac{u_n}{v_n}     & = \frac{\widehat u_n}{\widehat v_n} -\frac{u_n}{v_n} 
 \to_{\mathbb{P}} 0, {\rm \ \ as \ } n \to \infty. 
\end{align*}
The study of $\frac{u_n}{v_n}$ is done in \cite{smith1992bayesian} where the authors proved that it converges to $F_0(x)$, which achieves the proof. 
\end{proof}

The use of the WR procedure in regression  yields a new sample with $X^*$ which tends to the target distribution.  
This oversampling can be applied with the following algorithm.

\begin{algorithm}[H]
   \caption{Weighted Resampling (WR) algorithm}
   \label{algoWE}
\begin{algorithmic}
    \STATE {\bfseries Input :} data $\{(\boldsymbol{X}_i,Y_i)_{i=1,\cdots,n}\}$; synthetic sample size $n^*$ ($=n$ if not specified); target density $f_0$; kernel $\cal K$ (including trimming $e_n$ and $h_n$)
    \STATE 
    {\bf Weighted Resampling} : 
    \STATE $w_i = f_0(X_i) / \widehat{f}_{e_n}(X_i)$; $i=1\cdots, n$
    \STATE $q_i = w_i / sum(w_i)$; $i=1\cdots, n$
    \STATE Draw $(X^*,Y^*) $  from $(X_1,Y_1),\cdots, (X_n,Y_n)$ with probabilities $q_1,\cdots, q_n$
    \STATE {\bfseries Output :} $(X^*,Y^*)$
\end{algorithmic}
\end{algorithm}

With this method, the cdf of $X^*$  converges to the target one from Proposition \ref{propWR}. The conditional distribution of $Y$ given ${X}$ remains unchanged since  the drawing is done on the whole observation. It is important to note that if there are auxiliary variables, they are drawn as $Y$.  

The particularity of the WR algorithm 
 is that it generates on the same support as the observations, which is imbalanced. 
 Indeed, with  a finite sample size,  bootstrap procedures naturally return the same values as the sample. There is thus an important risk of over-fitting for the segments poor in observations with a high draw weight. 

\subsection{DA-WR algorithm}

We propose to combine the previous WR algorithm with a method of DA generating synthetic data. This artificial data simulation allows to extend the support of the distribution and  possibly reduces the risk of over-fitting associated with oversampling. 

In order to enlarge the support of the covariates, we apply the DA first. 
We obtain the following algorithm, adding a DA step to the previous WR algorithm.

\begin{algorithm}[ht]
   \caption{Data Augmentation - Weighted Resampling (DA-WR) algorithm}
   \label{CWR}
\begin{algorithmic}
   \STATE {\bfseries Input :} data $\{(\boldsymbol{X}_i,Y_i)_{i=1,\cdots,n}\}$; DA sample size $N$; synthetic sample size $n^*$; target density $f_0$; kernel $\cal K$ (including trimming $e_n$ and $h_n$); DA generator
    \STATE  
    {\bf DA step} : 
    \STATE  
    $\{(X_j^s,Y_j^s)\}_{j=1,\cdots,N}$ = DA($\{(X_i,Y_i)\}_{i=1,\cdots,n}$)
    \STATE  
    {\bf WR step} : 
    \STATE $w_i = f_0(X_i^s) / \widehat{f}_{e_n}(X_i^s)$; $i=1\cdots, N$
    \STATE $q_i = w_i / sum(w_i)$; $i=1\cdots, N$
    \STATE Draw $(X^*,Y^*) $  from $(X_1^s,Y_1^s),\cdots, (X_N^s,Y_N^s)$ with probabilities $q_1,\cdots, q_N$
    \STATE {\bfseries Output :} $(X^*,Y^*)$
\end{algorithmic}
\end{algorithm}

As we can see, this approach does not require complex parameterization: only a target distribution for the covariate of interest and a density estimator for the empirical density (e.g kernel density estimate).
In practice, a preliminary step of WR can be used  to increase the performance of the DA-WR algorithm that becomes a WR-DA-WR algorithm. The  last step of the algorithm must be a WR step to bring closer the synthetic distribution to  the target one. It is important to note that if there are auxiliary variables then they are generated and drawn as $Y$. 
Note that the result of Proposition \ref{propWR} now depends on the DA procedure and additional conditions are needed to get the convergence to the target distribution. We give here an illustration (the proof is relegated in the Supplement) by considering 
 $X_i'$ the  new data obtained from the DA procedure, for $i=1\cdots, N=n$,  respectively, and by  assuming that for all $x \in {\cal X}$, 
 \begin{description}
\item {\bf (C')} : 
$\max_{i=1,\cdots,n} |\mathbb{I}_{X_i\leq x} - 
\mathbb I_{X_i'\leq x}| = o(1/n)$ 
\item {\bf (C'')} : 
$\max_{i=1,\cdots,n} |q_i -q_i' | =o(1/n)$,
\end{description}
where $q_i$ and $q_i'$ denote the weights associated with the initial and the augmented dataset, respectively.
\begin{prop}
\label{propDAWR}
Assume that {\bf (C)-(C'')} hold and that the support of  $F$ contains the support of  $F_0$. 
Then for all $x \in {\cal X}$, the cdf resulting from the DA-WR algorithm   
converges in probabilities to $F_0(x)$ as $n \to +\infty$.
\end{prop}

For the DA step, we consider the following six 
approaches:  
\bit
\item Perturbation approaches: \\
- \emph{Gaussian Noise}, inspired by \cite{Lee2000GN}: \textit{GN} \\
- \emph{ROSE}, a Smoothed Boostrap inspired by \cite{Menardi2014ROSE}: \textit{ROSE} \\
- \emph{Smoothed Boostrap}, another Smoothed Boostrap: \textit{KDE}
\item Interpolation approaches: \emph{k Nearest Neighbors} inspired by \cite{Chawla2002Smote}: \textit{SMOTE}
\item Latent structure model approaches: \\
- \emph{Gaussian Mixture Models}: \textit{GMM} \\
- \emph{Factor Analysis}: \textit{FA} 
\item Copula approach: \emph{Gaussian Copula Model} \cite{PatkiWV16}: \textit{Copula}
\item Deep Learning approach: \emph{Conditional Generative Adversarial Networks} \cite{DBLP:journals/corr/abs-1907-00503}: \textit{GAN}
\item Machine Learning approach: \emph{Ramdom forest} \cite{Nowok2016synthpopBC}: \textit{RF}
\eit
More details on these generators are given in \ref{Synthetic Data Generators} in the Supplement.
All these methods have been used as synthetic data generators, that is,  to construct samples with only artificial data. This choice has the advantage of defining the original observations as a test sample and  it also avoids the loss of information by separating learning and testing. In our study, this allows us to compare the different techniques. The existing algorithms in imbalanced regression cannot be applied in this context because they do not focus on the covariate distribution. More precisely, they do not propose to manage the selection bias although our WR algorithm tries to correct the lack of representativeness of the sample compared to the population (convergence to the target distribution).

All these generators were tested before and after a clustering (via a Gaussian Mixture Model) in order to eventually improve the accuracy of the data generation. Typically, this means performing a cluster-conditioned learning. Indeed, the structure (variance, dependence, etc.) of the data can be different according to certain subparts of the data space. It can be relevant to apply a resampling by cluster rather than on the whole sample.

\section{NUMERICAL ILLUSTRATION}
We illustrate our approach numerically by first showing the different generated data obtained with the WR and the DA-WR algorithms. Then, we analyze the impact of  bias selection in regression when an imbalanced sample is used for training the learning model. At last, we measure the benefit obtained with both WR and  DA-WR algorithms and compare the results obtained by the different generators. In this illustration, the algorithms are  intentionally evaluated on a complicated case with two inflection points on the border of the observed support to get a strong impact on the regression in order to assess  the method.

\subsection{Dataset design}


We consider a bi-dimensional initial population $\mathcal{D}^p = (X^p,Y^p)$, of size $n^p = 10,000$ such that  $ X \sim F_0 := \mathcal{B}(5,5)$ and $Y \sim \mathcal{N}(\sin(7X -0.5)+10,0.1)$, 
where ${\cal B}$ denotes the Beta distribution and ${\cal N}$ denotes the Gaussian distribution. 


From this population, we uniformly draw a test sample $\mathcal{D}^t$. From the remaining population, $\mathcal{D}^p$ \textbackslash $\mathcal{D}^t$, we uniformly draw a balanced sample $\mathcal{D}^b$, supposed to be representative of the population. Finally, we draw an imbalanced sample $\mathcal{D}^i$ from this remaining population. The draw weights to construct this imbalanced sample are defined by the distribution $F = \mathcal{B}(9,9)$. 
The test, balanced, and imbalanced samples are all of size $n=1,000$.

\begin{figure}[ht]
\centering
    \begin{subfigure}{0.49\textwidth}
    \includegraphics[width=\textwidth]{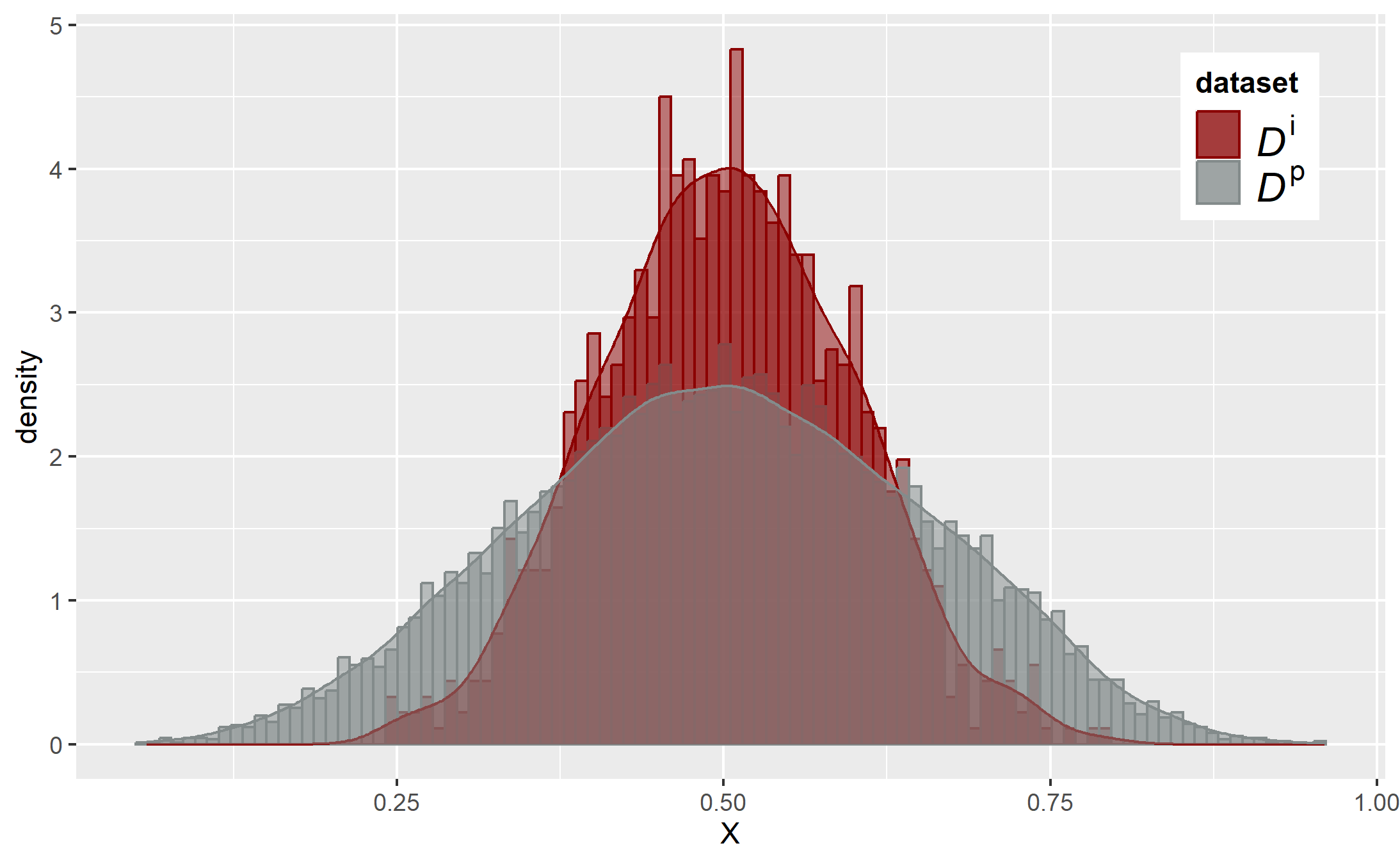} 
    \caption{Histograms of $X$ and density estimation from population and imbalanced sample}
    \label{comp_X_Ech0-vs-Pop}
    \end{subfigure}
\hfill
\begin{subfigure}{0.49\textwidth}
    \includegraphics[width=\textwidth]{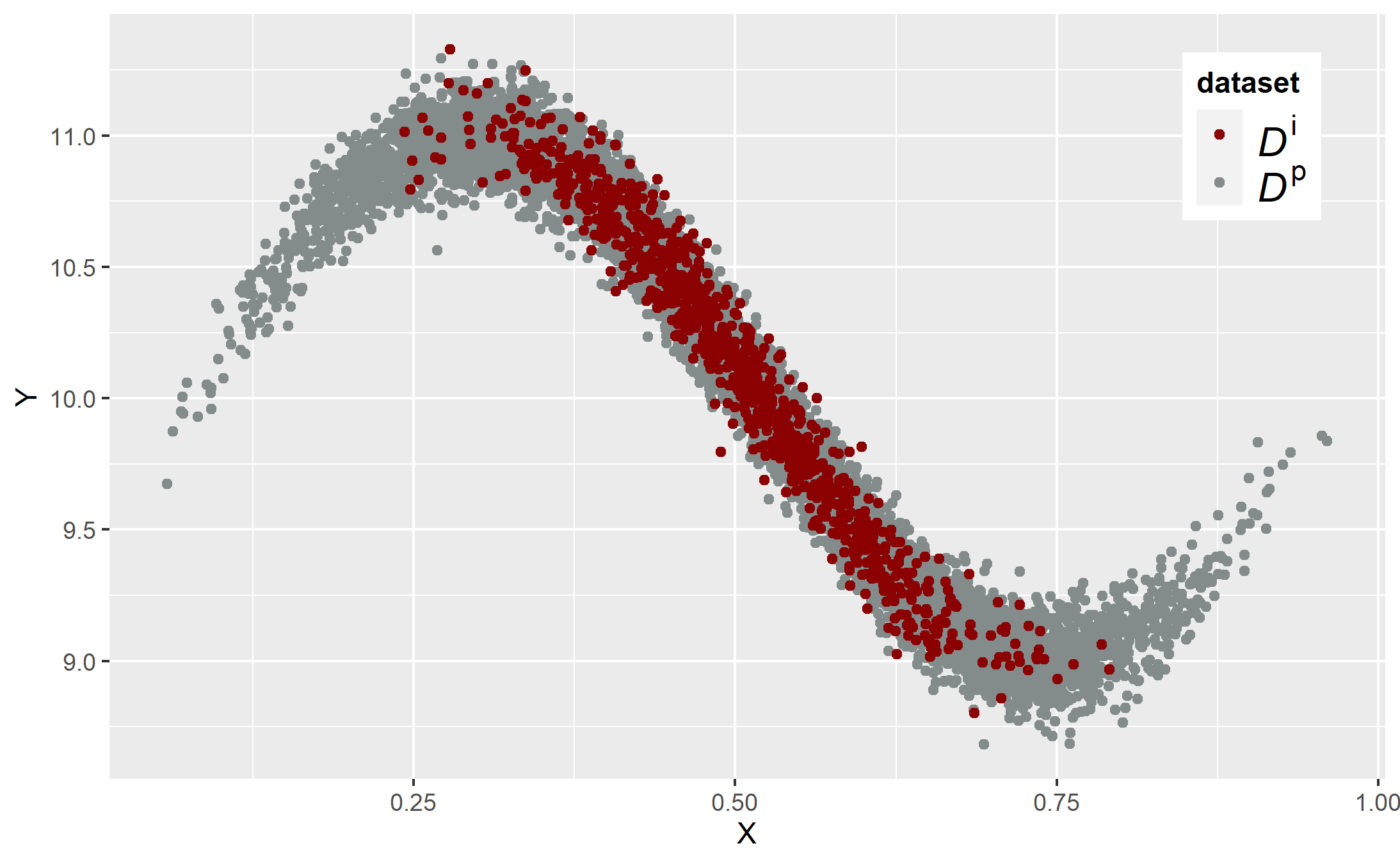}
    \caption{scatterplot $(X,Y)$ from population and imbalanced sample}
    \label{comp_Y_Ech0-vs-Pop}
\end{subfigure}
\caption{Histograms and density estimations of  balanced (in grey) and imbalanced (in red) populations.}
\end{figure}

Figure \ref{comp_X_Ech0-vs-Pop} shows  the empirical densities of $X$ in the population and in the imbalanced sample $\mathcal{D}^i$. Figure  \ref{comp_Y_Ech0-vs-Pop} shows the scatter plot $(X,Y)$ from   $\mathcal{D}^i$. As we can see, the imbalanced sample is more centered than the population and poorly covers the whole support of $X$: there is less  data on the sides and $Y$ is different on these parts of the space.
We face an imbalanced regression: a $(\alpha, \beta)$-imbalanced problem with, for example, $\lvert \frac{\widehat{\mathbb{P}}(\boldsymbol{X}\in\chi)}{\mathbb{P}_0(\boldsymbol{X}\in\chi)} - 1 \rvert > \alpha$ when  $\chi := [0,0.3]$ or $\chi := [0.7,1]$ and ($\alpha \leq 0.59, \beta \leq 0.09$). Here we are in the situation where $F$ and $F_0$  have the same support and the imbalanced problem will be less with $n$ large.

\subsection{Data generation and resampling analysis}

The aim is to build a new sample $\mathcal{D}^* := \{(X_i^*,Y_i^*)_{i=1,\cdots,n^*}\}$ having a cdf $F^*$ close to the target one $F_0$, as observed in Figure \ref{Hist_X_Ech0-vs-Tgt} in the Supplement. We define $n^*=n=1,000$. We therefore want to obtain a wider distribution with more observations on the sides.

\subsubsection{WR algorithm}

We first use the one-step WR algorithm. 
The draw weights calculated by the WR are represented in Figure \ref{weights-ech0} in the Supplement. 
Figure \ref{hist_X_ech_add-vs-Tgt} shows the distribution of $X$ in the WR sample $\mathcal{D}^*$  which is closer to the target distribution than the initial one. The values on the sides, quite rare, are naturally often drawn.  We see there are still some parts of the support without observation, especially on the right side. 
This situation could lead to an overfitting phenomenon because the same observations are replicated several times and could lead to an over-fitting effect.

\begin{figure}[ht]
\centering
\includegraphics[width=0.49\textwidth]{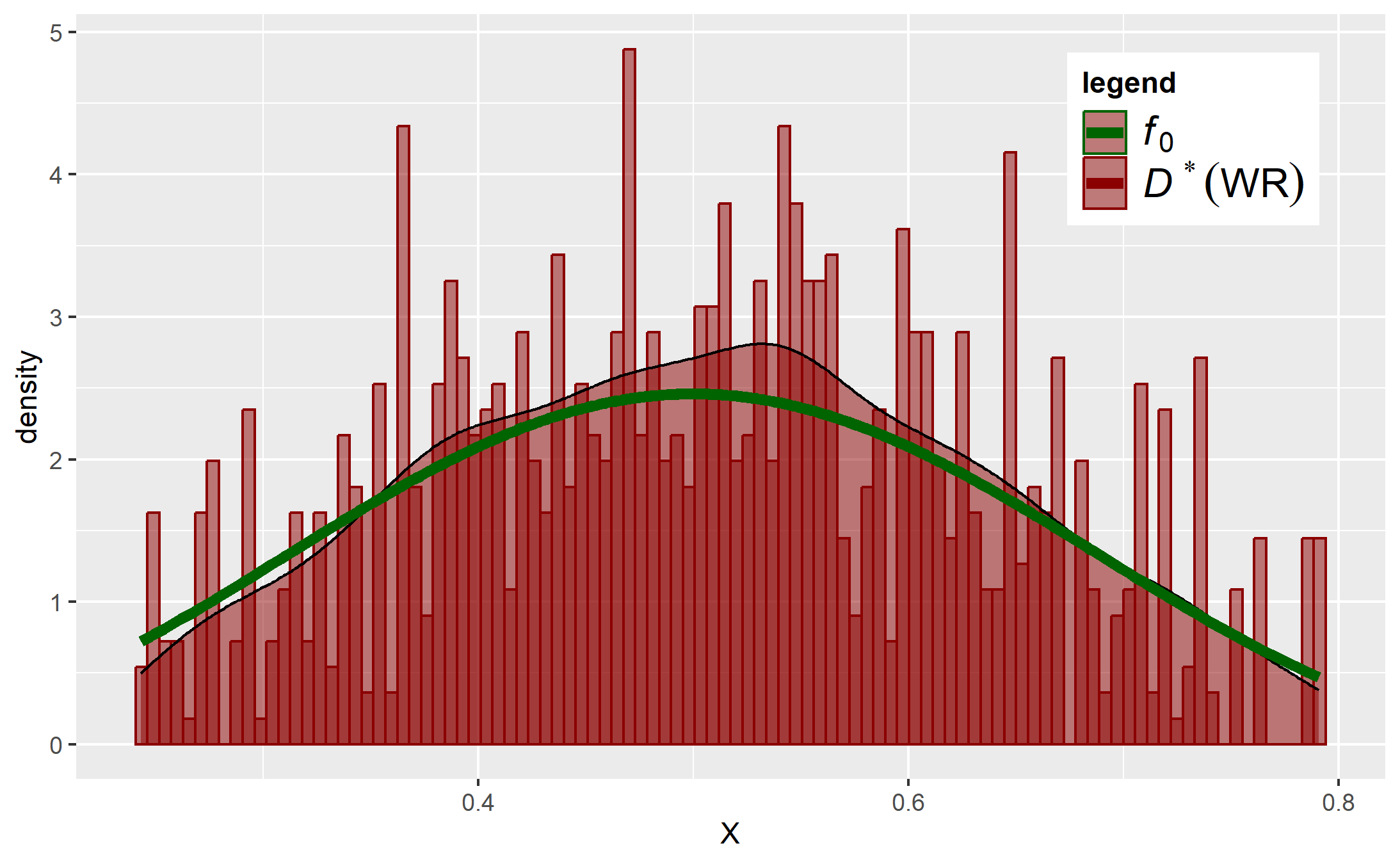}
\caption{Histogram of $X$ obtained with weighted resampling method vs target distribution}
\label{hist_X_ech_add-vs-Tgt}
\end{figure}

\subsubsection{DA-WR algorithm}

The different data generators are applied directly on the imbalanced sample or by a clustering obtained with a Gaussian Mixture Model (GMM) in order to generate synthetic observations within each cluster. The results of this clustering on the initial sample $\mathcal{D}^i$ are presented in Figure \ref{clustering} in the Supplement.

Figure \ref{Hist_X_ech_GN_SC-vs-ech_add} shows the histograms of $X$ obtained both with WR and DA-WR algorithms (with a Gaussian Noise method). As expected, the observations are more extended on the sides and cover more the support with Da-WR: some initial parts without observations are filled in. We also observe  that the DA-WR distribution is closer to the target one than the WR sample. Figure \ref{scatterplot_ech_GN_SC-vs-ech_add} presents the scatterplot$(X^*,Y^*)$ obtained with a generator versus the initial (or with the weighted resampling that draws the same samples).  


\begin{figure}[ht]
\centering
    \begin{subfigure}{0.49\textwidth}
    \includegraphics[width=\textwidth]{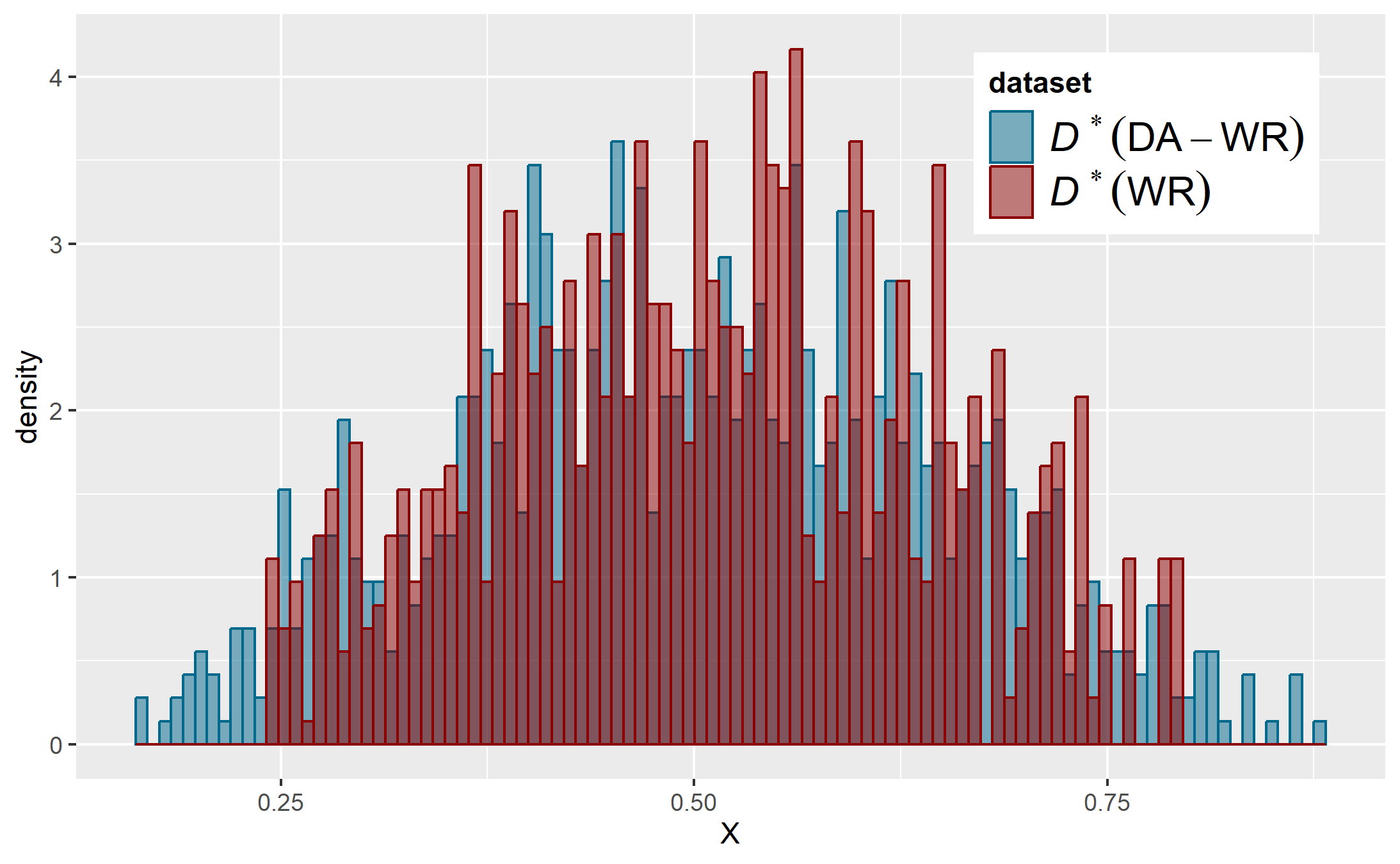} 
    \caption{Histogram of $X$ obtained with Gaussian Noise method vs WR}
    \label{Hist_X_ech_GN_SC-vs-ech_add}
    \end{subfigure}
\hfill
\begin{subfigure}{0.49\textwidth}
    \includegraphics[width=\textwidth]{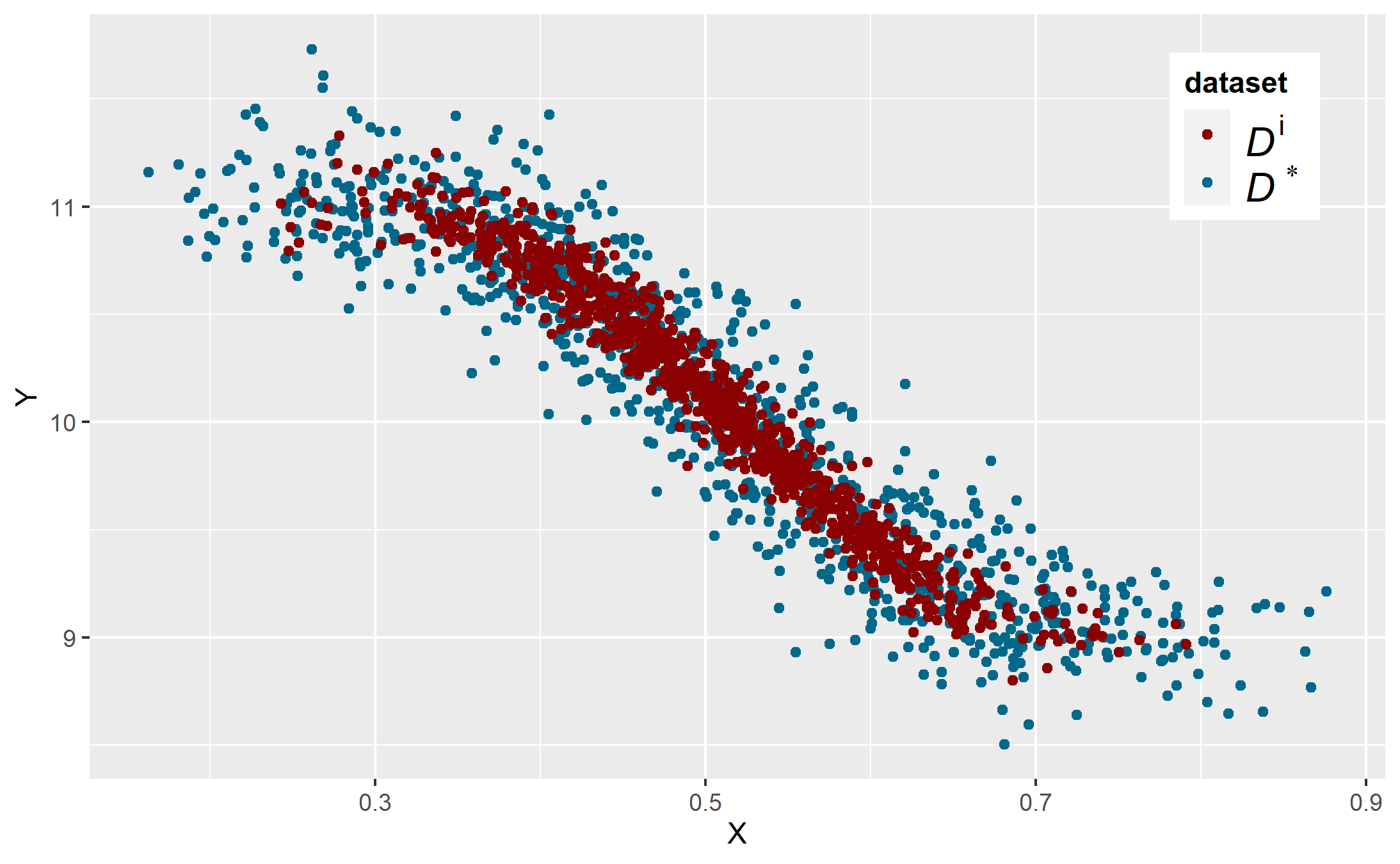}
     \caption{Scatterplot $(X^*,Y^*)$ obtained with Gaussian Noise method vs WR}
     \label{scatterplot_ech_GN_SC-vs-ech_add}
\end{subfigure}
\caption{Comparison between the WR and DA-WR algorithms}
\label{Comparison_ech_GN_SC-vs-ech_add}
\end{figure}

To compare the effect of the different generators, we analyze the histogram of $X$ from  $\mathcal{D}^*$, called "new" in the figures. The comparison is made according to the target distribution $F_0$ (\ref{Hist_X_ech_XXX-vs-Tgt} in the Supplement) and also according to the histogram of $X$  obtained with the weighted resampling (\ref{Hist_X_ech_XXX-vs-ech_add} in the Supplement). 
Figure \ref{illu_KS_dist_X} in the Supplement represents the Kolmogorov-Smirnov distance between   the distributions of $X$ for the balanced sample and the different samples. It confirms that the augmented samples are closer to the balanced sample.
At last, we analyze the scatter plot $(X^*,Y^*)$ from the sample $\mathcal{D}^*$ that we compare  to the imbalanced one from  $\mathcal{D}^i$ (\ref{comp_Y_ech_XXX-vs-ech0} in the Supplement).  
We can see in Figure  \ref{Hist_X_ech_XXX-vs-Tgt} in the Supplement that  all obtained distributions are quite close to the target one. In Figure \ref{Hist_X_ech_XXX-vs-ech_add} in the Supplement, we see that the DA-WR algorithm does not necessarily improve  the WR one: with the random forest it provides  the same values of $X$;  with copulas, Smote, or Smote-GMM, it generates values within the observed range. 
This is  a drawback in our illustration since we need to get more values on the sides. 

About $Y$ generation, we observe in Figure  \ref{comp_Y_ech_XXX-vs-ech0} in the Supplement that the generations obtained with the clustering are closer to the initial values. As previously, the copula, Smote, Smote-GMM approaches generate within the observed range.  
Random forests generate the same values for $X$ while the GAN and ROSE generation  are disappointing: the $Y$ values are too extended and scattered.  

\subsection{Predictive performance analysis}

We evaluate the impacts of the WR and DA-WR algorithms on three learning models: Generalized Additive Models (GAM), Random Forest (RF) and Multivariate Adaptative Regression Splines (MARS). More details on these algorithms are given in the supplement. We evaluate the prediction results of these three models on the test sample through  Root Mean Square Error (RMSE).
The different smoothed predictions on the test dataset ($\mathcal{D}^t$) are shown in Figures \ref{pred_Y_GAM_ech_XXX-vs-test}-
\ref{pred_Y_MARS_ech_XXX-vs-test} in the Supplement, for each learning model. 
The stability of the results for  the different generators is analyzed by applying the method on several training samples (Figures \ref{RMSE_GAM}-
\ref{RMSE_MARS} in the Supplement).

\subsubsection{Impact of an imbalanced dataset}

Compared to the balanced training dataset ($\mathcal{D}^b$), we observe, on the different figures, that the predictions with the imbalanced sample ($\mathcal{D}^i$) are quite far from the real values on the sides, whatever the model. In addition, we can see the increase of RMSE in Table \ref{recap_RMSE}. 
\subsubsection{Effect of the DA-WR algorithm}

Compared to the imbalanced training dataset ($\mathcal{D}^i$), we observe that the predictions with a rebalanced training sample are closer to the real values on the sides except when the random forest is used. Indeed, the random forest  algorithm with only one covariate is actually a bagging algorithm. This algorithm generally offers good predictive performance but the predictions on the test sample are constant on the sides and they are not better than the initial. 

Based on the RMSE given in Table \ref{recap_RMSE}, it can be  observed that, with GAM and MARS models, some approaches provide better results, especially WR, GN-GMM, ROSE, ROSE-GMM, KDE-GMM and Smote-GMM. However, some other approaches give worse results than the imbalanced sample: Copula at first, KDE (without GMM), GAN, RF and ROSE whatever the model. Others are slightly worse than the imbalanced sample for one of the two models: GN, GMM, Smote and FA-GMM. 
The clustering seems to improve the results. Even if the interpolation approaches give good results, their data generation technique could be a drawback because of their limit. Finally, the perturbations approach, with a kernel, seems to be effective.

\begin{table}[ht]
\centering
\includegraphics[width=0.49 \textwidth]{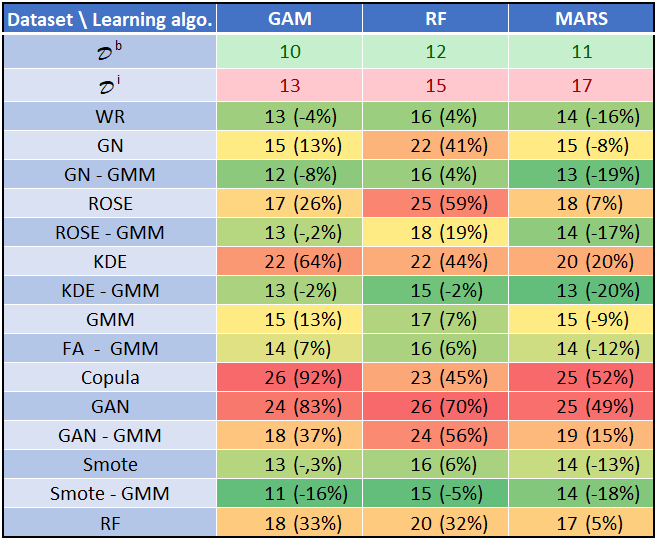}
\caption{RMSE (x100) on the test sample and gain or loss (\%) with respect to the RMSE for the imbalanced sample. From green to red we go from the best to the worst results. }
\label{recap_RMSE}
\end{table}

Figures \ref{pred_Y_GAM_ech_XXX-vs-test}-  
\ref{pred_Y_MARS_ech_XXX-vs-test} in the Supplement confirm the previous comments: with some generators, we manage to adjust the sample to provide better predictions, closer to the real values. However, with others, our methodology does not give the expected results.

To avoid a sampling effect and give more robustness to the results, we test our method on 100 imbalanced samples. Figures \ref{RMSE_GAM}-
\ref{RMSE_MARS} in the Supplement shows the RMSE boxplot for these multiple simulations. The purpose is to provide a better RMSE (median and range) than the imbalanced sample. 
We can see that with the GAM model, the WR, GN-GMM, ROSE-GMM, KDE-GMM and Smote-GMM are really well and confirm the previous analysis. The GAN generator is the worst. KDE and copula are also pretty bad. The others are not significantly better than the imbalanced sample. We obtain the same behavior with MARS model but Smote, RF and ROSE are also better than the imbalanced. 
The results with the RF model confirm these previously obtained with a single simulation: the combined weighted resampling methods do not provide better results on prediction.

\section{APPLICATION}

\subsection{Dataset design}

We test our approach on a portfolio  of automobile insurance described in a 
dataset of driver telematics 
from 
(\cite{telematics}).  This dataset contains $n^p = $ 100,000 observations considered as the whole population.  
The covariates are the driver's characteristics and some telematic information. Traditional non-life pricing is based on estimating  the claims frequencies which represent here our dependent variable  $Y$. 

The imbalanced dataset replicates a real case of sampling bias observed in insurance data, especially by the impact of the imbalanced covariate on the target variable $Y$. Typically, the distribution of the covariate in this application is quite asymmetric and we make the tail of the distribution poorer, just by removing some rare observations. We have thus intentionally applied a sampling bias on the datasets in order to obtain a significant impact of the imbalance on the regression.

As in the illustration, we construct different samples as follows. From the initial population, we construct a test sample, denoted by  $\mathcal{D}^t$, with a uniform draw. In order to have a training sample independent of the test one, the imbalanced  sample, $\mathcal{D}^i$ is drawn from the remaining population.  
The methodology of this drawing is given in the Supplement. We also draw, uniformly, a balanced sample, denoted by $\mathcal{D}^b$, from the remaining population to measure the impact of the imbalanced sample on the predictions. 

We aim to obtain a new sample  $\mathcal{D}^*$, from the imbalanced one, providing best predictions on  $\mathcal{D}^t$. 
The covariate of interest, that we wish to rebalance, is the total miles driven per year, denoted by $X$.  
The others covariates are the age of the car, the credit score and the years without a claim. We consider the duration as an offset in our model. 

At first, the remaining population 
is used to estimate the claim frequencies on the test sample. 
These estimates can be considered as reference values. 
Next,  the claim frequencies are estimated on $\mathcal{D}^t$  from the different training samples, the aim being to get as close as possible to the reference values. We evaluate the prediction results on the test sample with the Root Mean Square Error (RMSE) relative to the reference values. 

Figure \ref{appli_s_km_driven-vs-Y_Pop} in the Supplement shows the estimated effect of $X$ on $Y$  by the model (based on the whole population). We can observe that this effect increases until around 10,000 miles then becomes slightly constant. We used this information to construct our imbalanced sample. 

\begin{figure}[ht]
\centering
\includegraphics[width=0.49\textwidth]{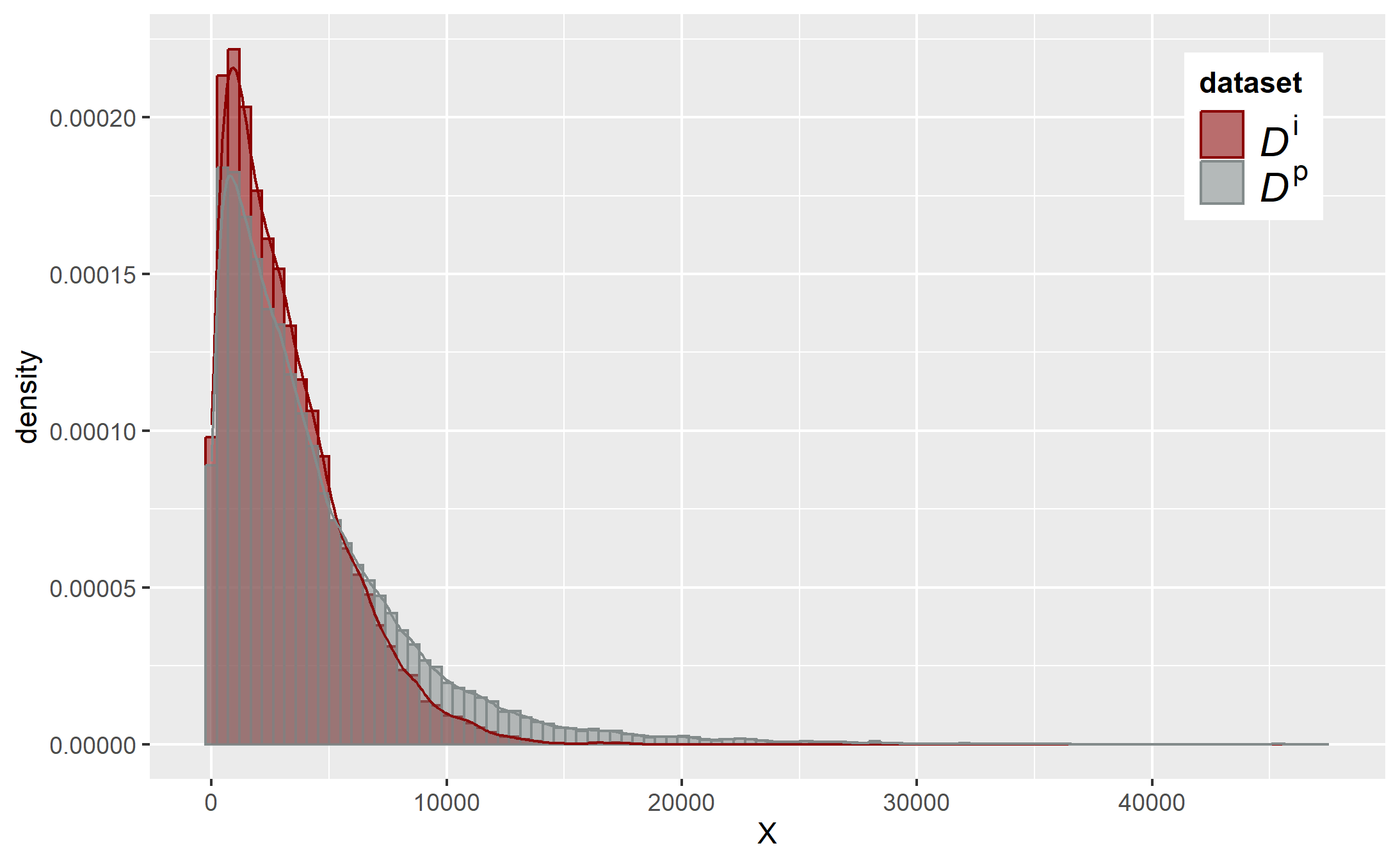} 
\caption{Histograms of $X$ from population and imbalanced sample}
\label{appli_comp_X_Ech0-vs-Pop-Dens}
\end{figure}

We can observe in Figure \ref{appli_comp_X_Ech0-vs-Pop-Dens} the distribution of $X$ 
 in the imbalanced sample is not as wide as the population. We can see there are fewer observations over 10,000 miles in the imbalanced sample than population.

\subsection{Data generation analysis}

We defined the target distribution from the population with a kernel density estimator. Figure  \ref{Appli_Hist_X_Ech0-vs-Tgt} in the Supplement,  compares the  distribution of $X$  in $\mathcal{D}^i$ with the target distribution. The drawing weights obtained by the WR algorithm are given in Figure \ref{Appli_weights-ech0} in the Supplement. We can see that the weights increase with $X$.

We can compare the WR method with the DA-WR method. Only the generators with the best results are considered. This comparison is done by the different histograms of $X$ in $\mathcal{D}^*$  versus the target distribution (\ref{Appli_Hist_X_ech_XXX-vs-Tgt} in the Supplement). We also compare these  histograms versus the histogram of $X$ obtained  from the WR algorithm  (\ref{Appli_Hist_X_ech_XXX-vs-ech_add} in the Supplement). We note that the data augmentation extends the  distribution of $X$ (except for the random forest generator) which is closer to the target one. 
For the distributions of $X$, it can be  observed on Figure \ref{appli_KS_dist_X} that   $\mathcal{D}^*$ is  much closer to $\mathcal{D}^b$ than $\mathcal{D}^i$. 

\subsection{Predictive performance analysis}

The performance of both WR and DA-WR  algorithms are compared on three learning models: a Generalized Additive Model with a Zero-Inflated Poisson distribution for $Y$ (noted GAM-ZIP), a Generalized Additive Model with a Poisson distribution for $Y$ (noted GAM-P) and a Random Forest (RF). For the first two models, we apply a regression spline on the insured driver's age and $X$.

The different predictions obtained with the GAM-ZIP are shown in Figure \ref{Appli_pred_Y_Pois_ech_XXX-vs-test} in the Supplement. 
Table \ref{Appli_recap_RMSE_ref} in the Supplement shows the RMSE for the balanced, imbalanced, WR and  best DA-WR samples. These results are summarized in Table \ref{Appli_recap_RMSE_ref_agg}.

\begin{table}[ht]
\centering
\includegraphics[width=0.49 \textwidth]{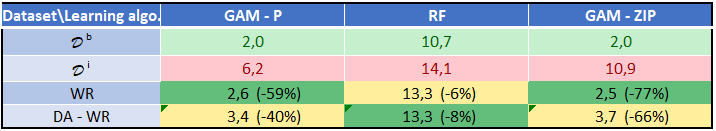}
\caption{RMSE relative to the reference values (x100) on the test sample and gain or loss (\%) with respect to the RMSE for the imbalanced sample}
\label{Appli_recap_RMSE_ref_agg}
\end{table}

\subsubsection{Impact of an imbalanced dataset}

The predictions on  $\mathcal{D}^t$ obtained from $\mathcal{D}^i$
are shown in Figure \ref{Appli_pred_Y_Pois_ech_XXX-vs-test} in the Supplement. We can see that the predictions obtained from the imbalanced sample are further from the observed values and that the confidence interval is larger in the distribution tail, whatever model. This result is quantified in Table \ref{Appli_recap_RMSE_ref} where it can be observed that the RMSE with $\mathcal{D}^i$ is strongly affected compared to $\mathcal{D}^b$.

\subsubsection{Effect of the DA-WR algorithm}

The WR and DA-WR samples yield much better predictions than $\mathcal{D}^i$. 
We can also see that the confidence intervals are reduced and that RMSE is  reduced for the three models.

\section{DISCUSSION AND PERSPECTIVES} \label{discussion}

The DA-WR algorithm is a new approach to balancing  a training sample in an imbalanced regression context. This approach could be used: i) with other types of continuous distributions (e.g multimodal); ii) with multivariate distributions;  iii) with other covariates, which would then be  treated as the variable of interest $Y$, as in the application. The WR can easily be used with categorical covariates but the DA-WR approach depends on the capacity of the DA generator to handle such variables.

Through the illustration and the application, we have seen the potential impacts of an imbalanced sample for prediction with various learning algorithms.
The WR algorithm can improve the learning and so the prediction by adjusting the sample to a target distribution. However, the results of the DA-WR approach are naturally dependent of the choice of the data generator. Some of the proposed generators do not provide the expected  results. This may be due to the fact that both variables $X$ and $Y$ are generated simultaneously under the  imbalanced  phenomenon. A local approach based on generators  restricted on local parts of the support could certainly improve the DA step, considering more local relation between $X$ and $Y$ and annihilating the imbalanced effect locally.   

Moreover, the analyses of the  results from multiple simulations show that some generators seem quite sensitive. Then the  
approach could also be improved by adding a treatment  
for large values, that is for the  extremities of the support of $X$, to avoid an extrapolation of atypical observations. 

Eventually, the DA-WR algorithm can be extended to balance simultaneously several covariates, by specifying a multivariate target distribution. It can also be extended to a conditional approach, that is applying this  technique within different subpopulations.
It could also be extended to mixed data (with possibly a restatement of categorical covariates to get a multivariate target distribution for the covariates of interest)   using a DA method dedicated to  mixed data.

\subsubsection*{Acknowledgements}
The authors thank the reviewers for their helpful comments which helped to improve the manuscript. This work benefited from the support of the Research
Chair DIALog under the aegis of the Risk Foundation, a joint initiative by
CNP Assurance

\bibliography{references}


\onecolumn
\aistatstitle{Data Augmentation for Imbalanced Regression: \\
Supplementary Materials}

\appendix

\section{\textsc{Synthetic Data Generators}} \label{Synthetic Data Generators}

We use the following notations for generators (in the same dorder as on the figures): 

\bit
\item \emph{WR}:  \emph{Weighted Resampling}
\item \emph{GN}:  \emph{Gaussian Noise}
\item \emph{GN - GMM}: \emph{GN} applied on GMM clusters
\item \emph{ROSE}:  \emph{Smoothed Bootstrap}, using the proposal of the algorithm \emph{ROSE} for the bandwidth matrix (\cite{Bowman1999AppliedST}) 
\item \emph{ROSE - GMM}: \emph{ROSE} applied on GMM clusters
\item \emph{KDE}: \emph{Smoothed Bootstrap}, using the R-package \emph{KernelBoot}, The bandwidth matrix being defined according to Silverman's proposal (\cite{Silverman86})
\item \emph{KDE - GMM}: \emph{KDE} applied on GMM clusters
\item \emph{GMM}: \emph{Gaussian Mixture Model}, using the R-package  \emph{MCLUST}
\item \emph{FA - GMM}: \emph{Factor Analysis}, using the python-package \emph{Scikit-learn}, applied on GMM clusters 
\item \emph{Copula}: \emph{Gaussian Copula Model}, using the python-package \emph{Synthetic Data Vault}  
\item \emph{GAN}: \emph{Conditional Generative Adversarial Networks}, using the python-package \emph{Synthetic Data Vault}
\item \emph{RF}: \emph{Random Forest},  using the R-package \emph{SynthPop}
\item \emph{RF - GMM}: \emph{RF} applied on GMM clusters
\item \emph{SMOTE}: interpolation by \emph{k nearest neighboors}, using the proposal of the algorithm SMOTE
\item \emph{SMOTE - GMM}: \emph{SMOTE} applied on GMM clusters
\eit

Other generators used but not selected because not relevant: \emph{SMOTER}, \emph{SMOGN}, some techniques adapted for regression tasks from the R-package \cite{branco2016ubl}, by defining the weights resampling as relevance function and some previous methods applied on GMM clusters or values of $Y = 0,1,2$ in the application. More details on these approaches are given on \ref{generatorMethod}.

\newpage

\section{\textsc{Illustration results}}

\subsection{On a single simulation}

For the illustration, we chose a trimming  sequence $e_{n} = \frac{1}{10 \times n}$

\subsubsection{Imbalanced sample}

\begin{figure}[H]
\centering
\begin{subfigure}{0.49\textwidth}
\includegraphics[width=\textwidth]{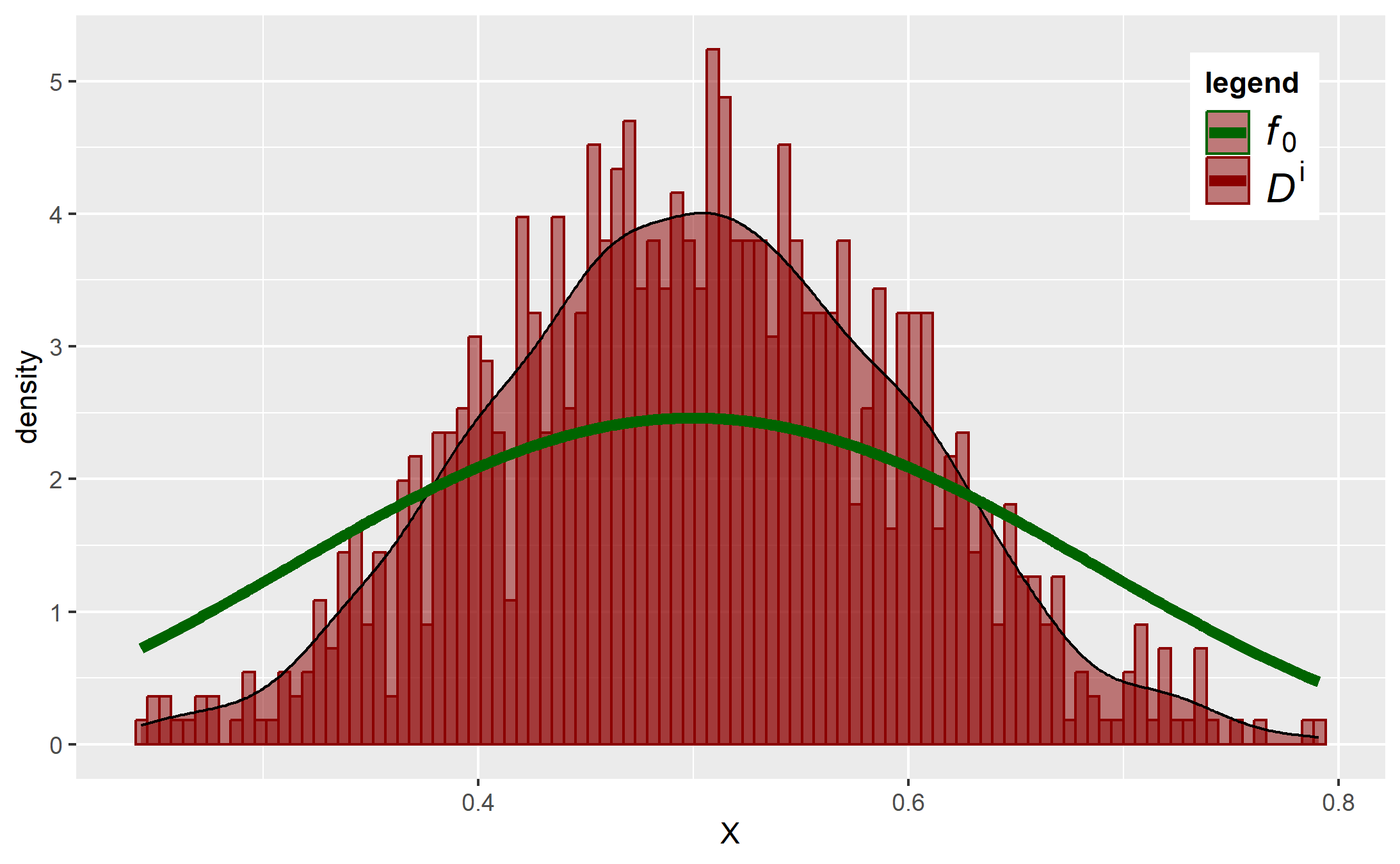} 
\caption{Histogram of $X$ in the imbalanced sample (red) vs the target distribution $f_0$ (green)}
\label{Hist_X_Ech0-vs-Tgt}
\end{subfigure}
\hfill
\begin{subfigure}{0.49\textwidth}
\includegraphics[width=\textwidth]{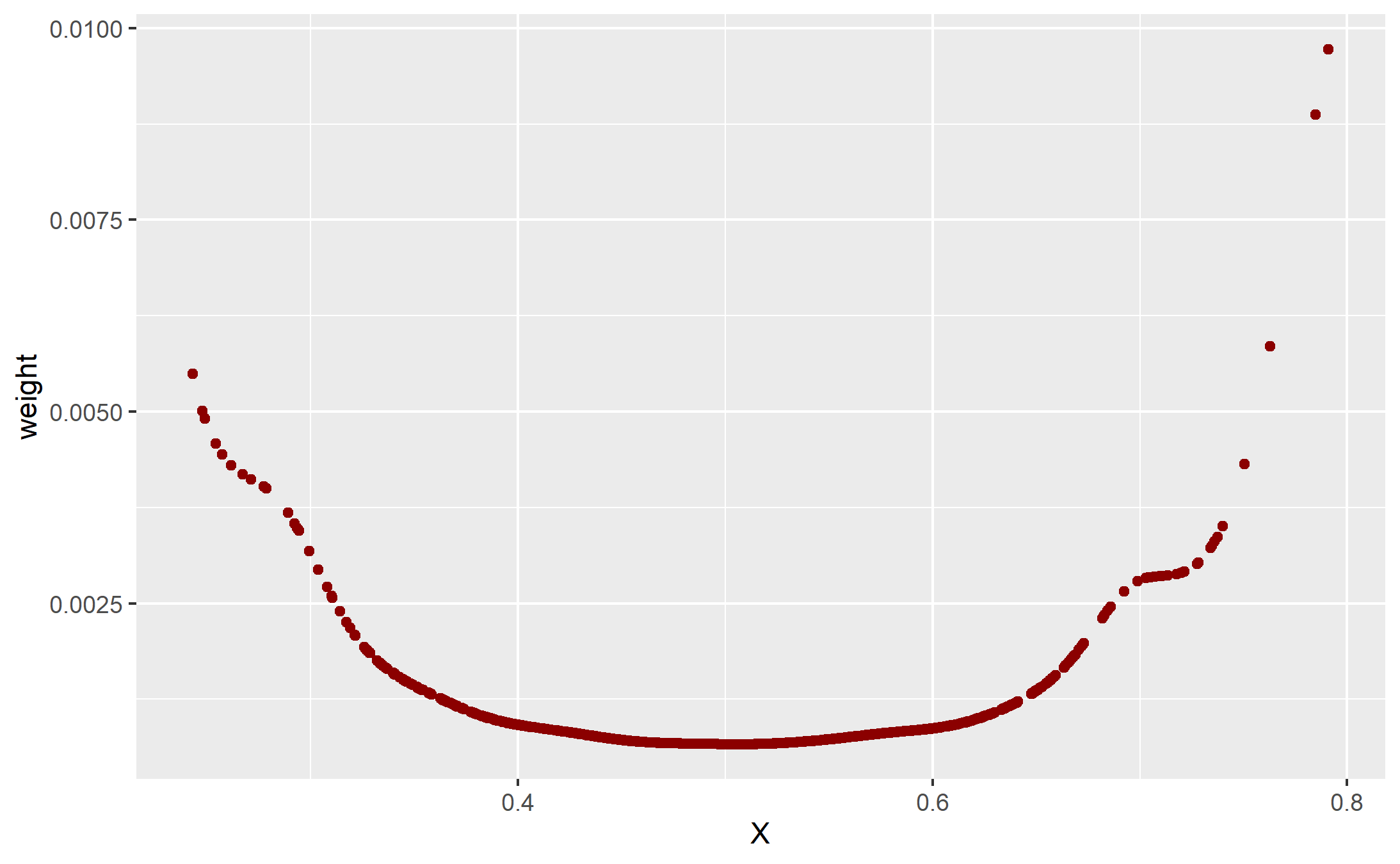}
\caption{Weights of $X$ on the imbalanced sample obtained with weighted resampling method}
\label{weights-ech0}
\end{subfigure}

\caption{Comparison between the imbalanced sample vs the target distribution and associated WR weights}
\label{ech0-vs-target}
\end{figure}

\begin{figure}[H]
\centering
\includegraphics[width=0.49\textwidth]{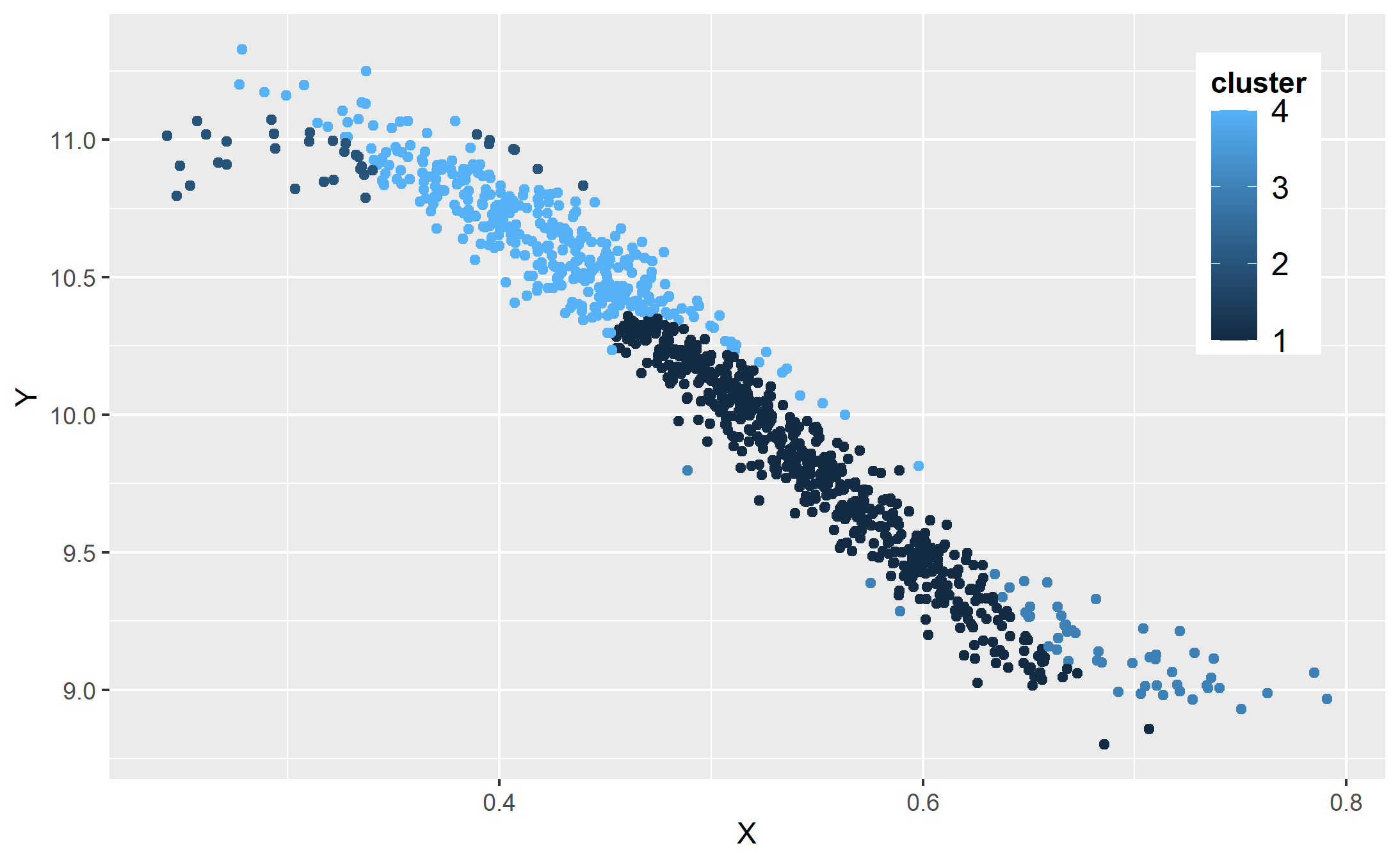}
\caption{GMM Clustering on imbalanced sample}
\label{clustering}
\end{figure}

\newpage
\subsubsection{Data Generation}

\textbf{Histogram of $X$ obtained in new samples vs target}

\begin{figure}[H]
  \centering
  \begin{subfigure}[b]{0.3\linewidth}
    \includegraphics[width=\linewidth]{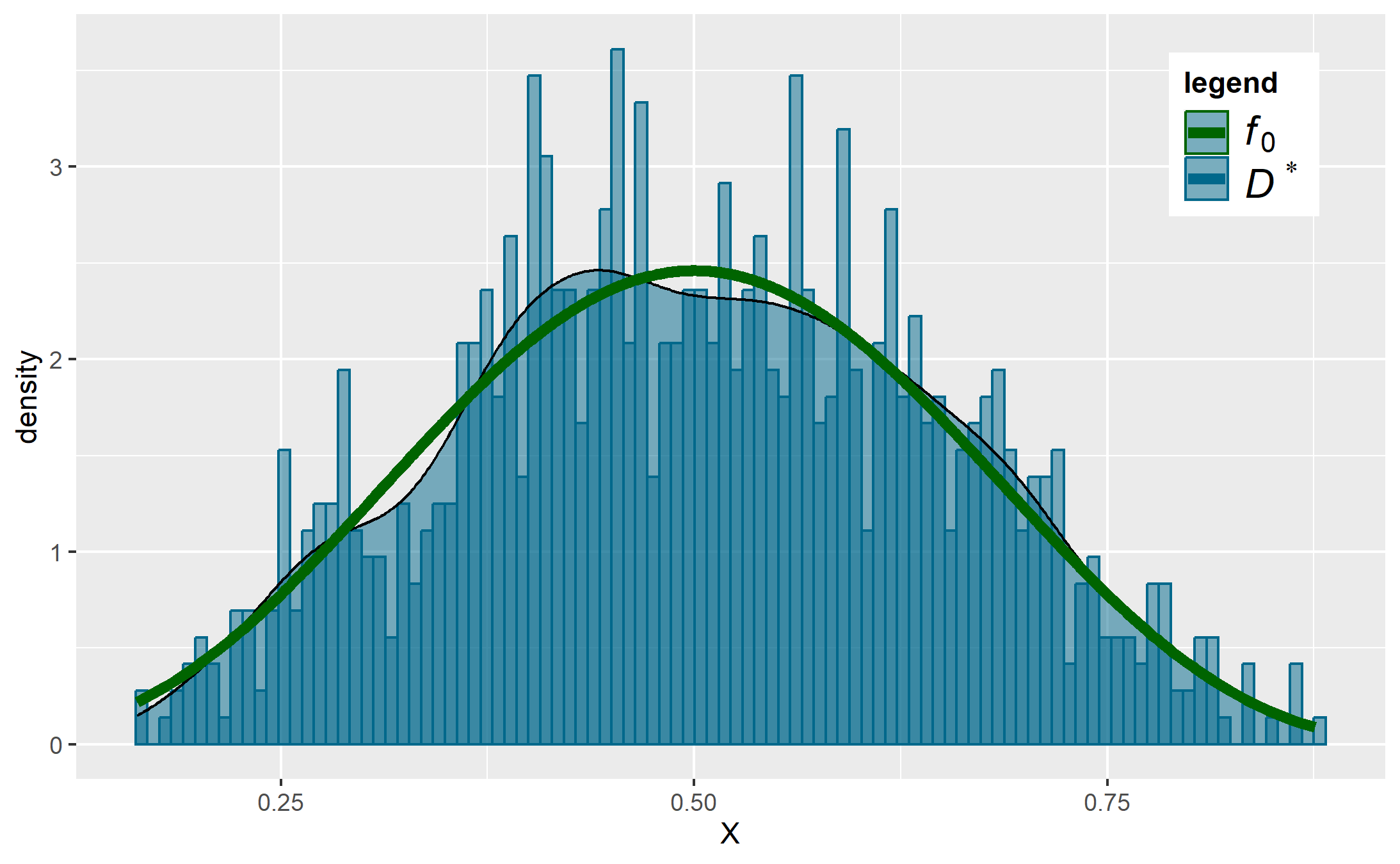}
     \caption{GN}
  \end{subfigure}
  \begin{subfigure}[b]{0.3\linewidth}
    \includegraphics[width=\linewidth]{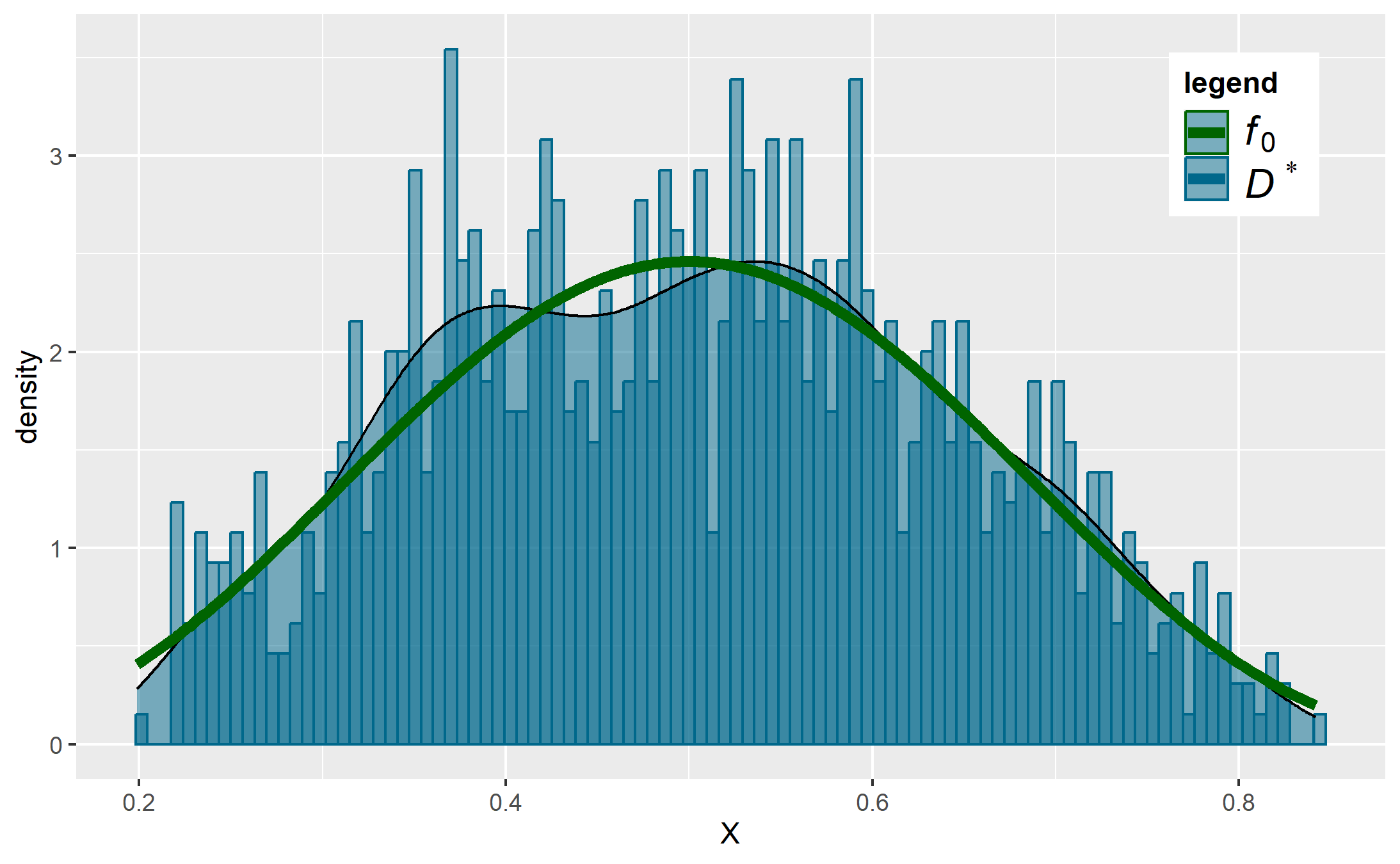}
    \caption{GN - GMM}
  \end{subfigure}
  \begin{subfigure}[b]{0.3\linewidth}
    \includegraphics[width=\linewidth]{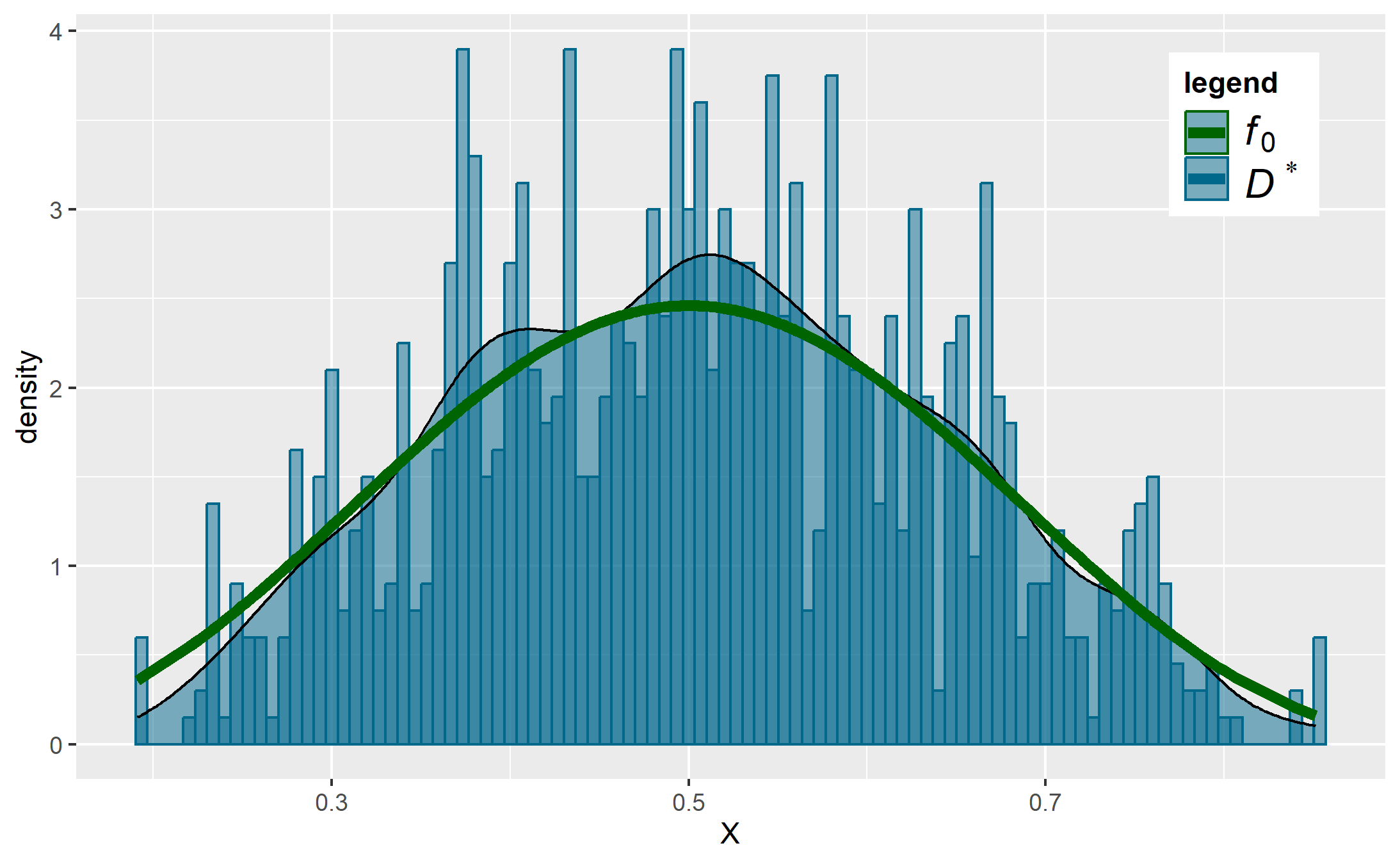}
    \caption{ROSE}
  \end{subfigure}

  \begin{subfigure}[b]{0.3\linewidth}
    \includegraphics[width=\linewidth]{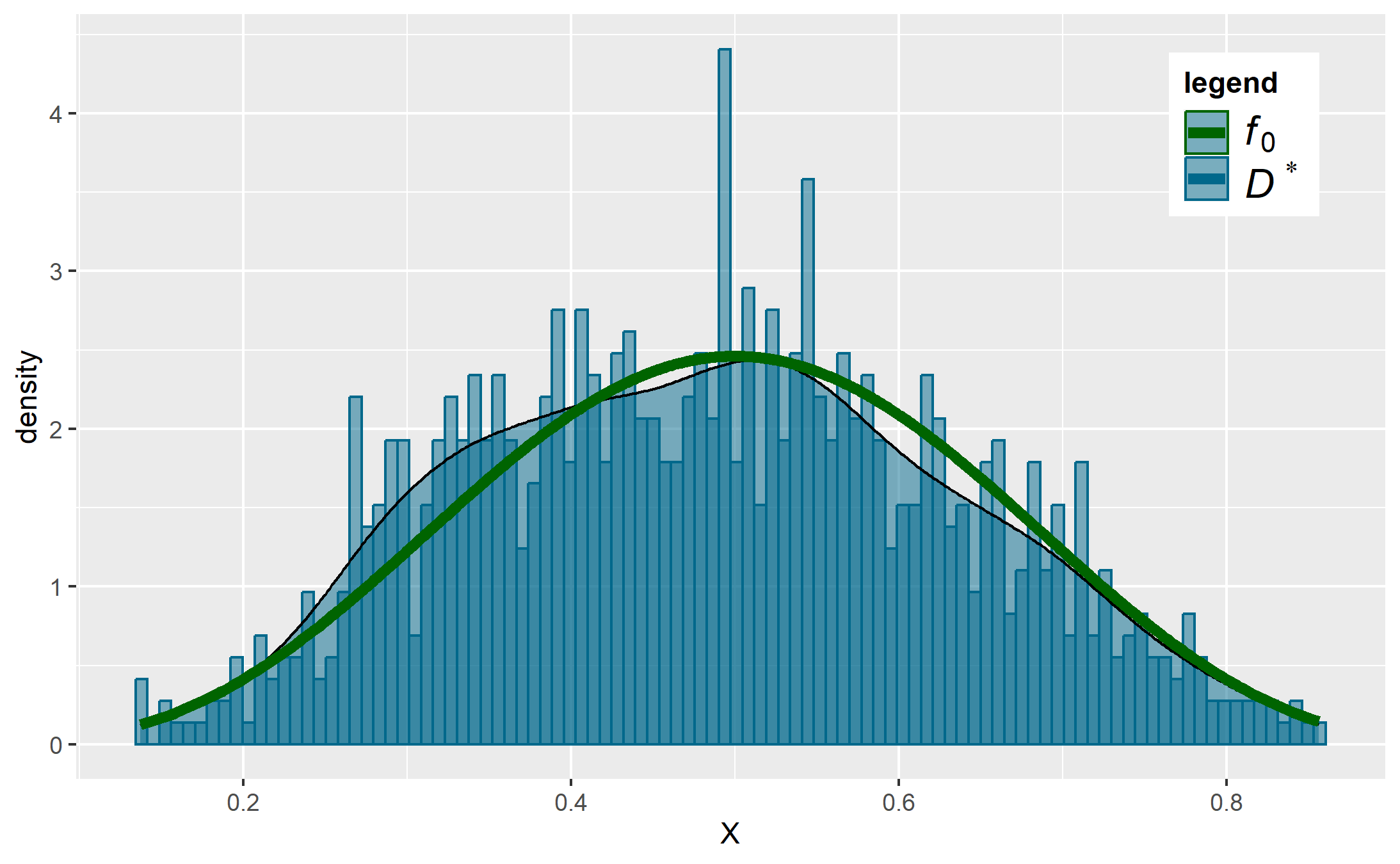}
     \caption{ROSE - GMM}
  \end{subfigure}
  \begin{subfigure}[b]{0.3\linewidth}
    \includegraphics[width=\linewidth]{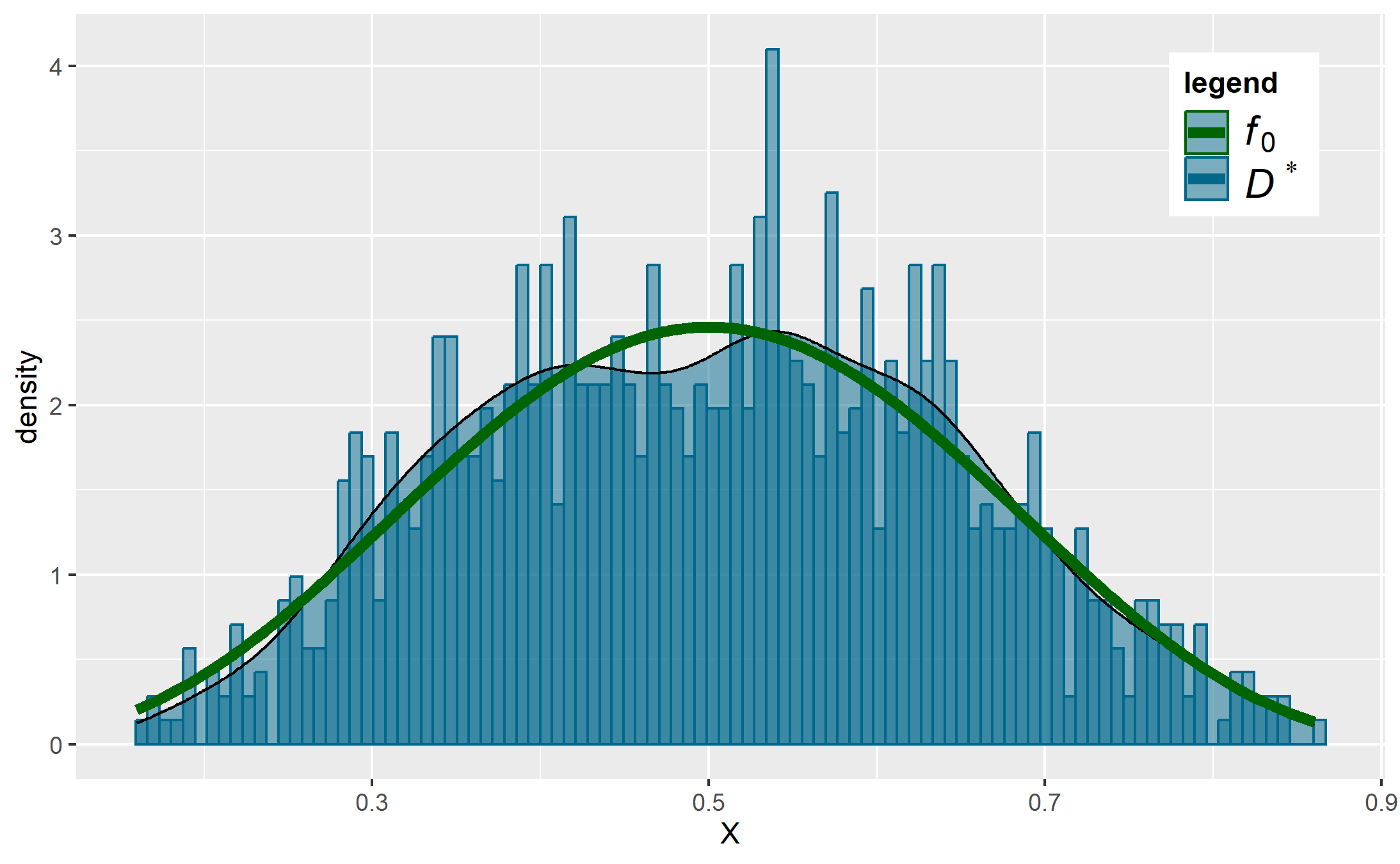}
    \caption{KDE - GMM}
  \end{subfigure}
  \begin{subfigure}[b]{0.3\linewidth}
    \includegraphics[width=\linewidth]{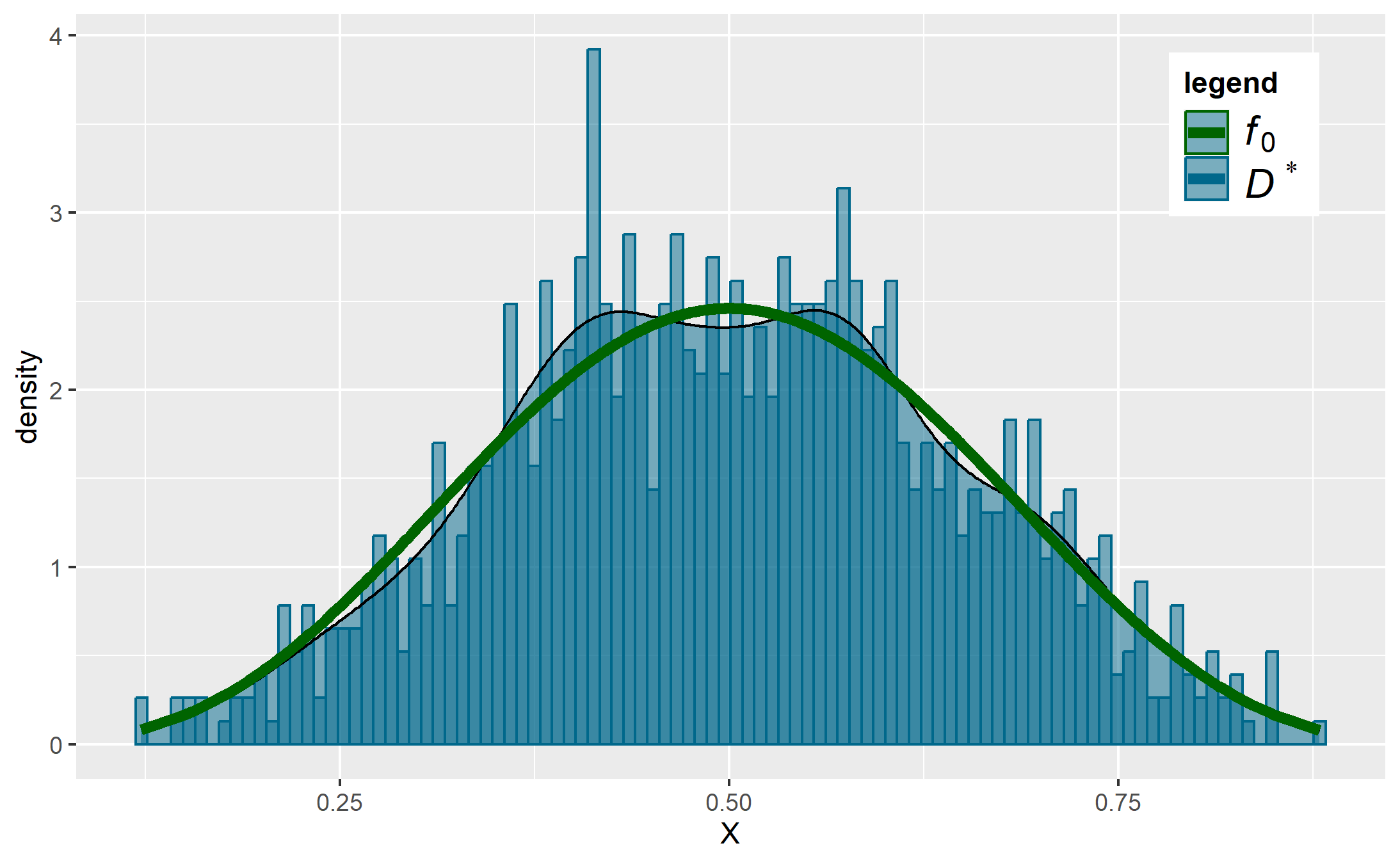}
    \caption{GMM}
  \end{subfigure}

  \begin{subfigure}[b]{0.3\linewidth}
    \includegraphics[width=\linewidth]{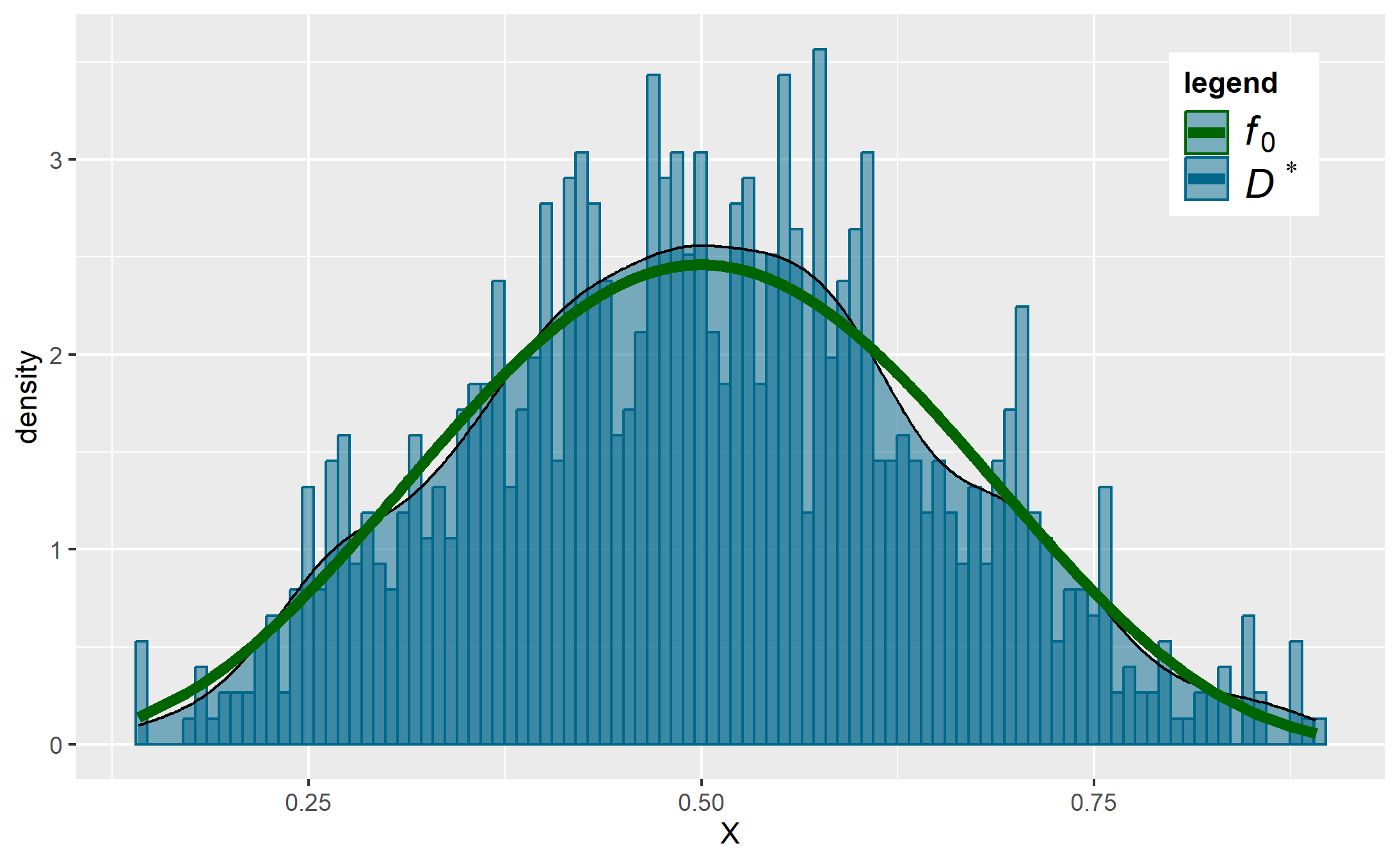}
     \caption{FA - GMM}
  \end{subfigure}
  \begin{subfigure}[b]{0.3\linewidth}
    \includegraphics[width=\linewidth]{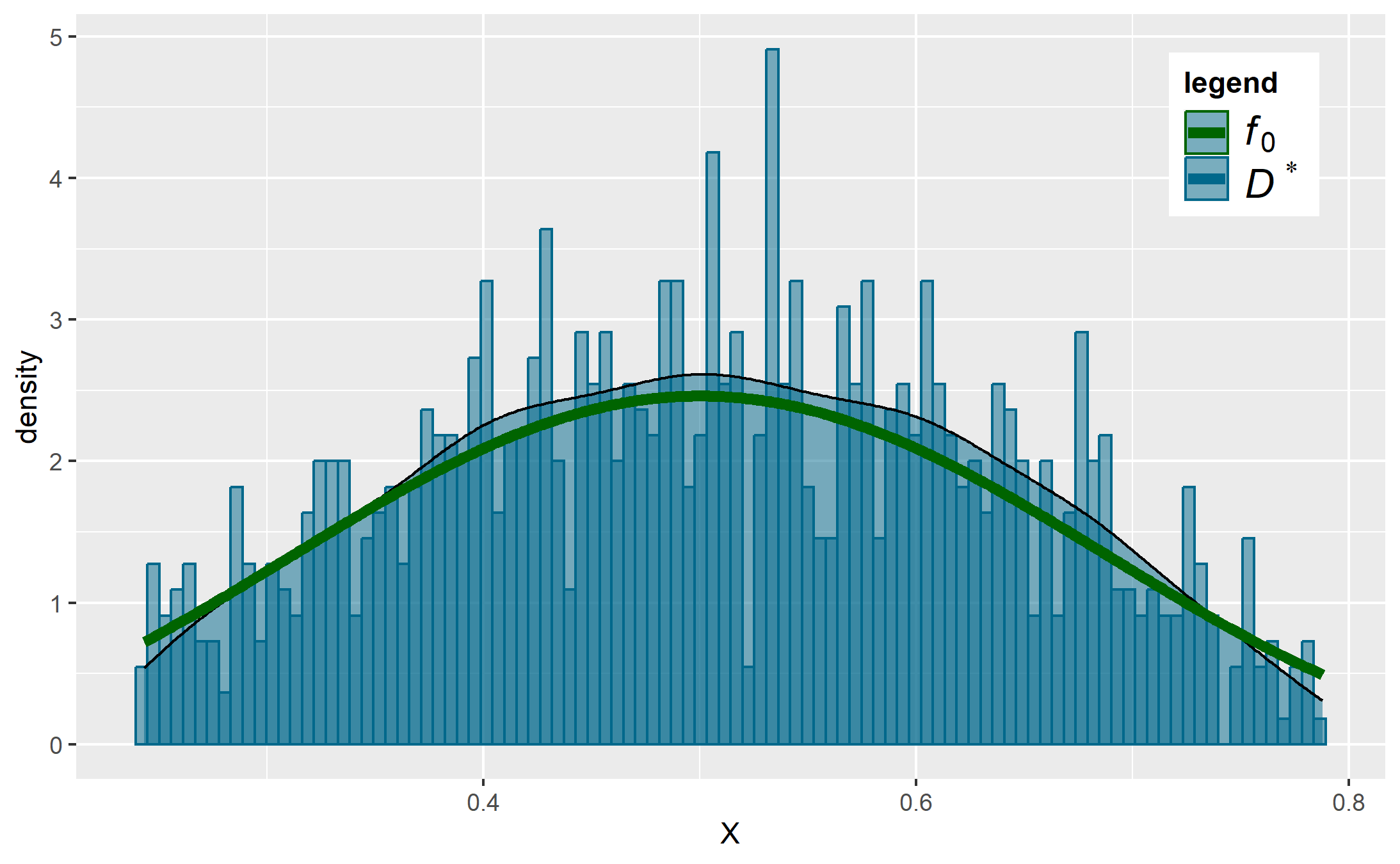}
    \caption{Copula}
  \end{subfigure}
  \begin{subfigure}[b]{0.3\linewidth}
    \includegraphics[width=\linewidth]{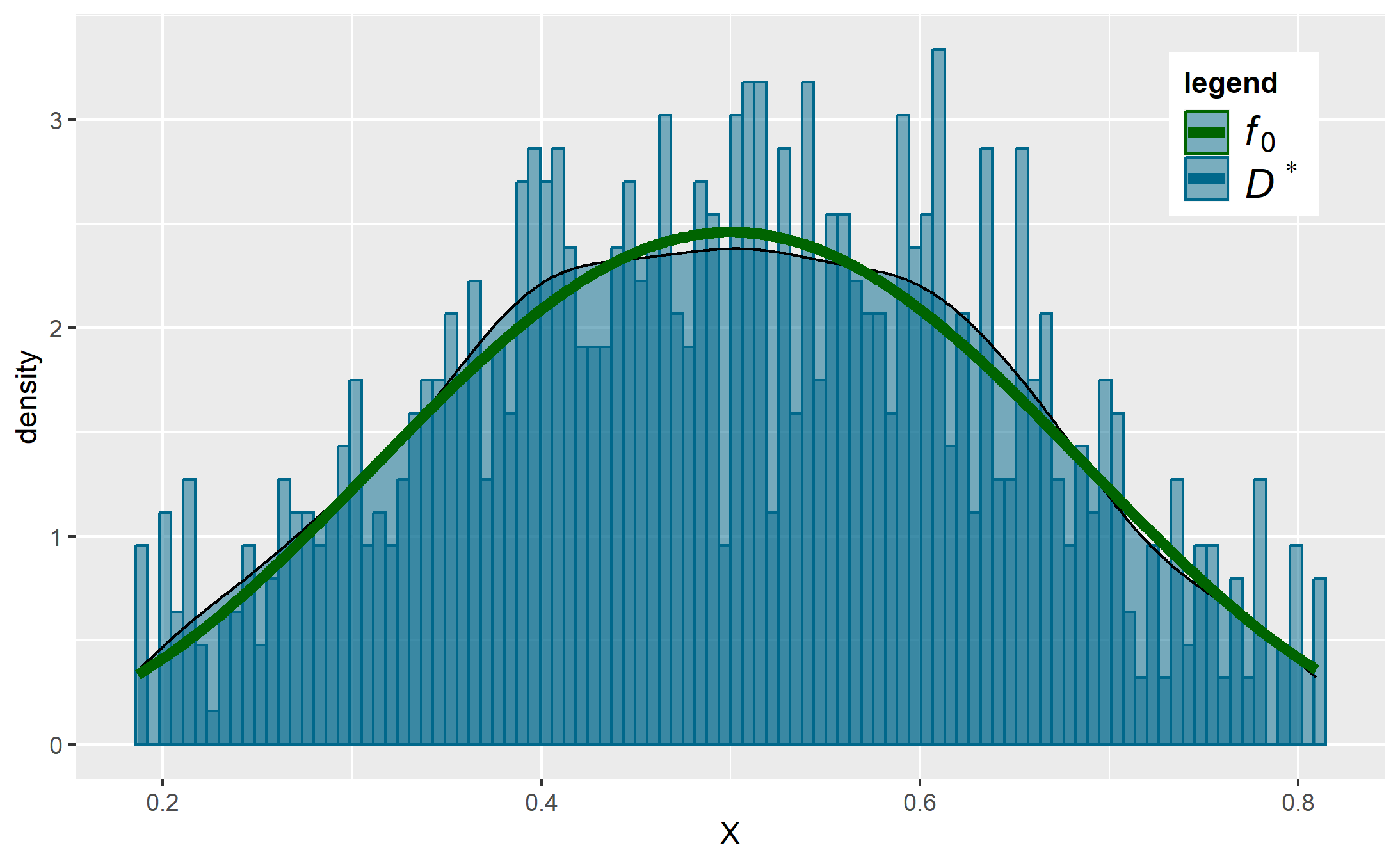}
    \caption{GAN}
  \end{subfigure}

  \begin{subfigure}[b]{0.3\linewidth}
    \includegraphics[width=\linewidth]{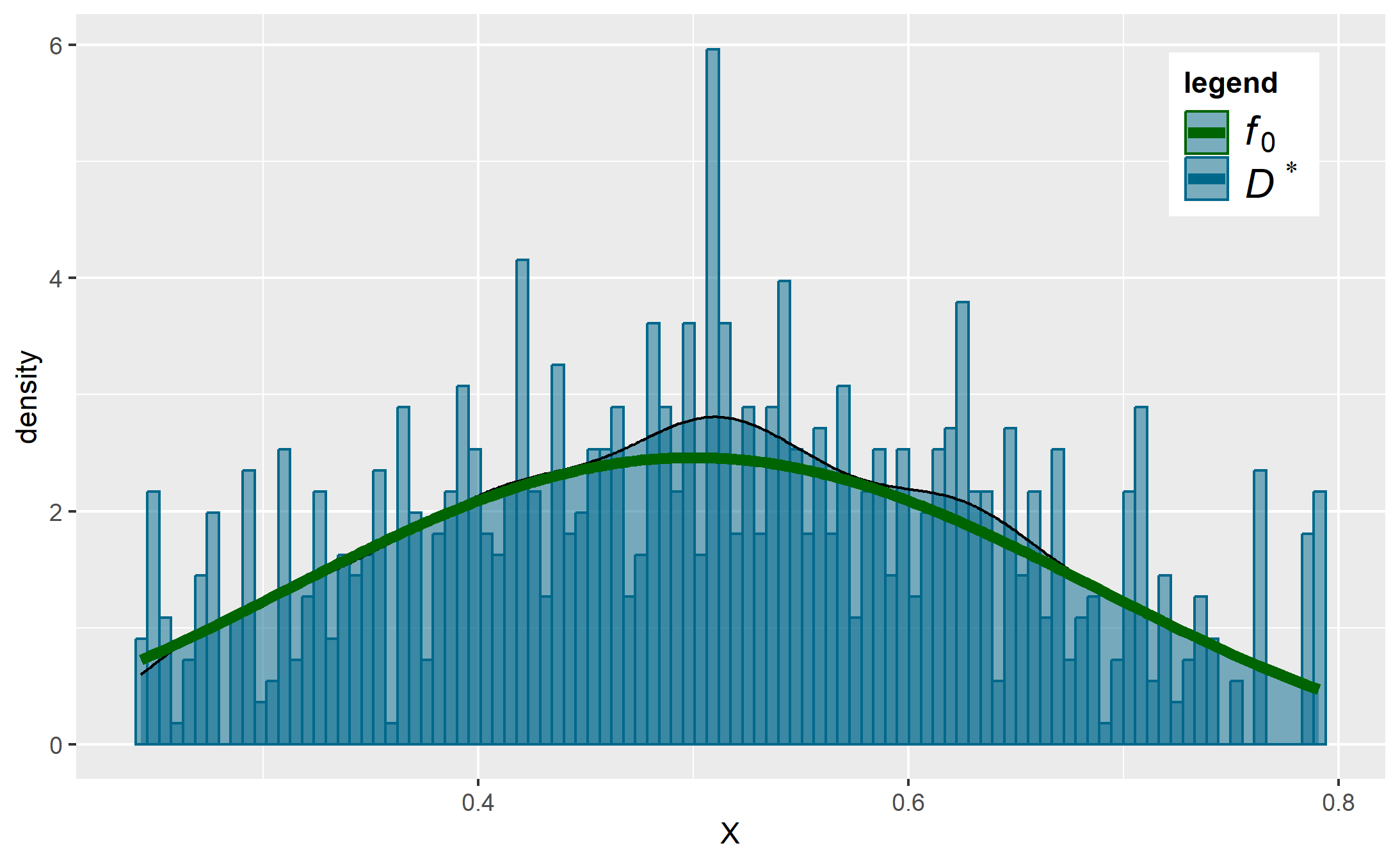}
     \caption{RF}
  \end{subfigure}
  \begin{subfigure}[b]{0.3\linewidth}
    \includegraphics[width=\linewidth]{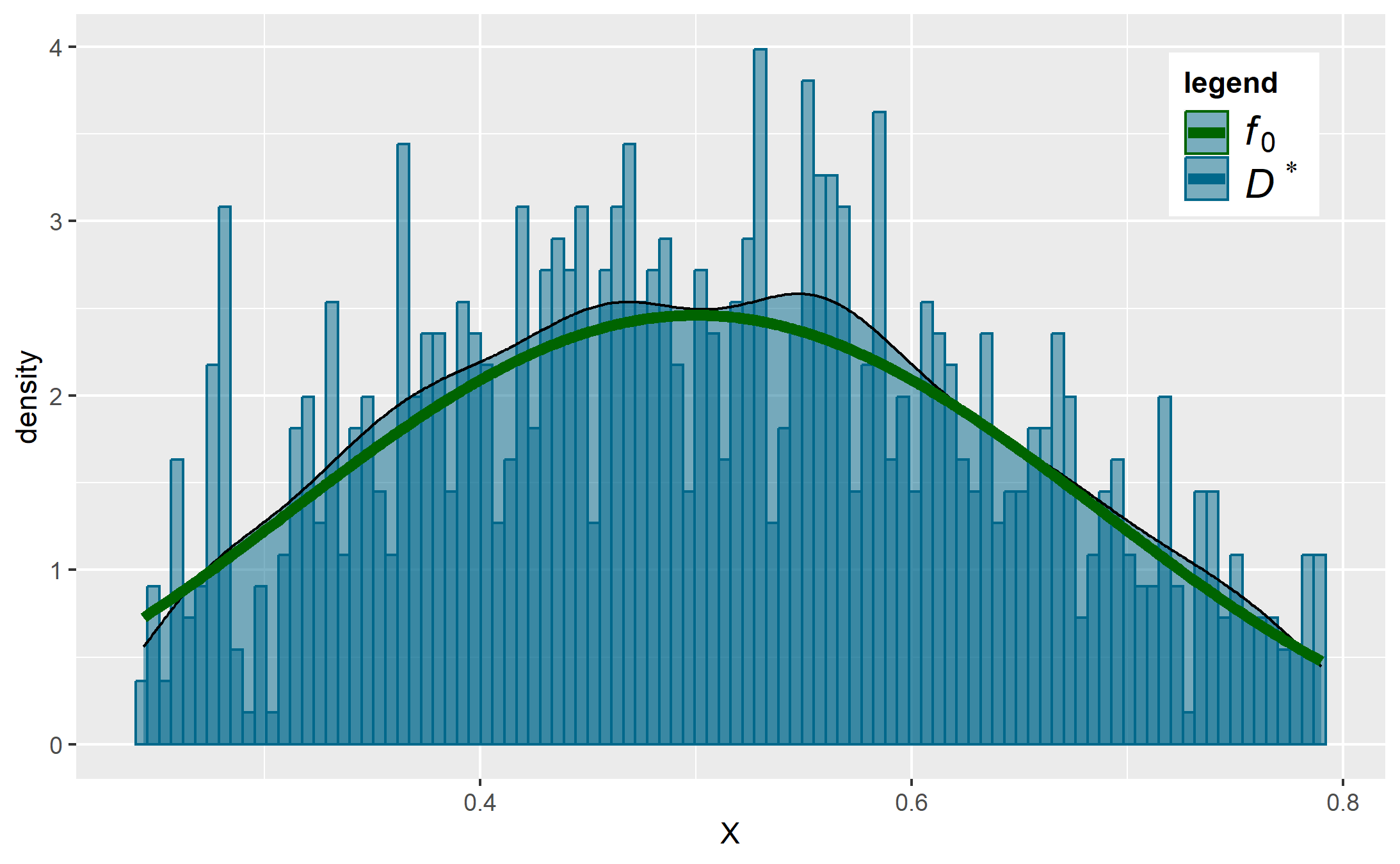}
    \caption{SMOTE}
  \end{subfigure}
  \begin{subfigure}[b]{0.3\linewidth}
    \includegraphics[width=\linewidth]{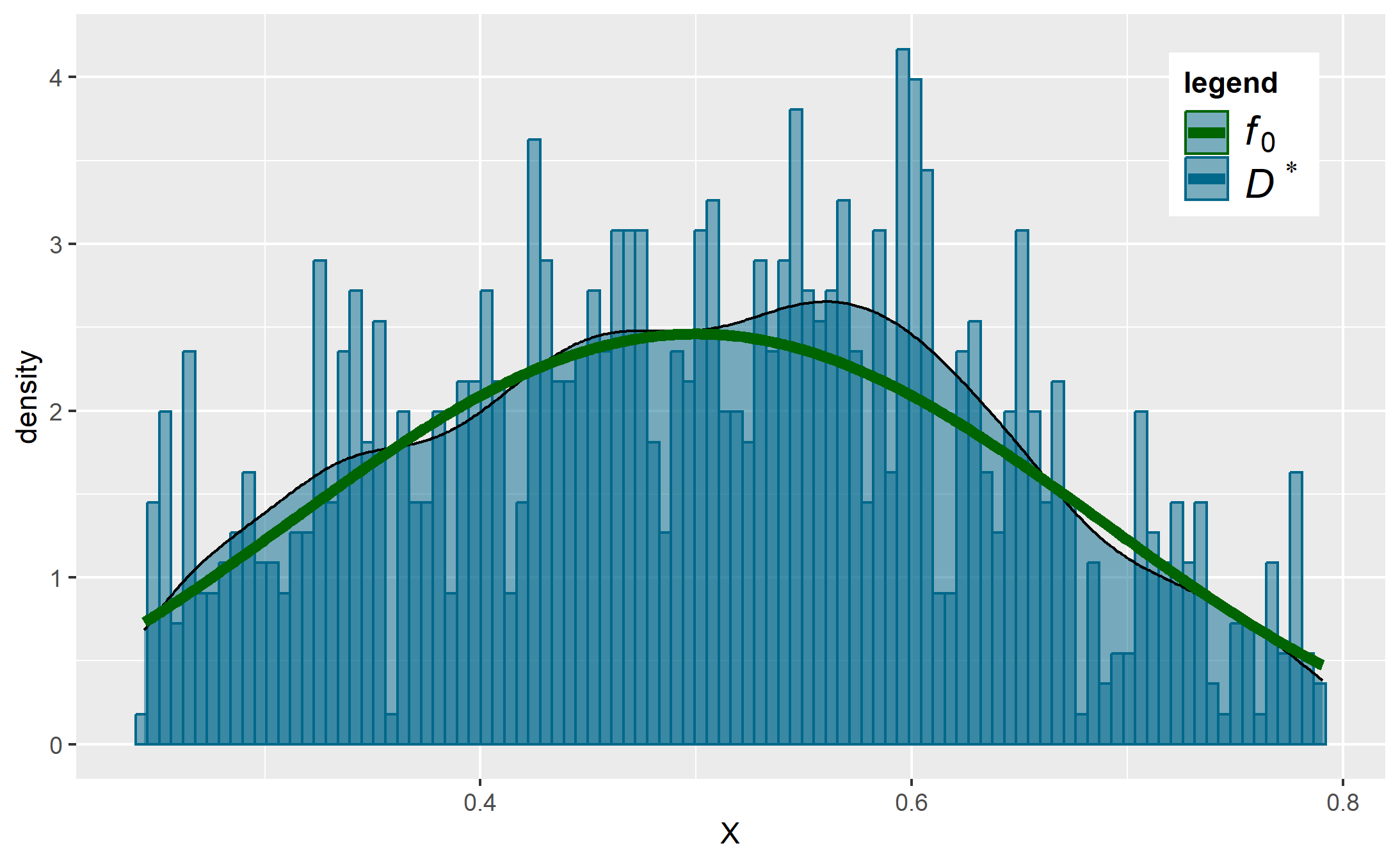}
    \caption{SMOTE - GMM}
  \end{subfigure}

  \caption{Histogram of $X$ obtained in new samples (new) vs target (f0)}
  \label{Hist_X_ech_XXX-vs-Tgt}
\end{figure}

\newpage
\textbf{Histogram of $X$ obtained in new samples vs WR}

\begin{figure}[H]
  \centering
  \begin{subfigure}[b]{0.3\linewidth}
    \includegraphics[width=\linewidth]{imgs/Illu/Hist_X_ech_GN_SC-vs-ech_add.png}
     \caption{GN}
  \end{subfigure}
  \begin{subfigure}[b]{0.3\linewidth}
    \includegraphics[width=\linewidth]{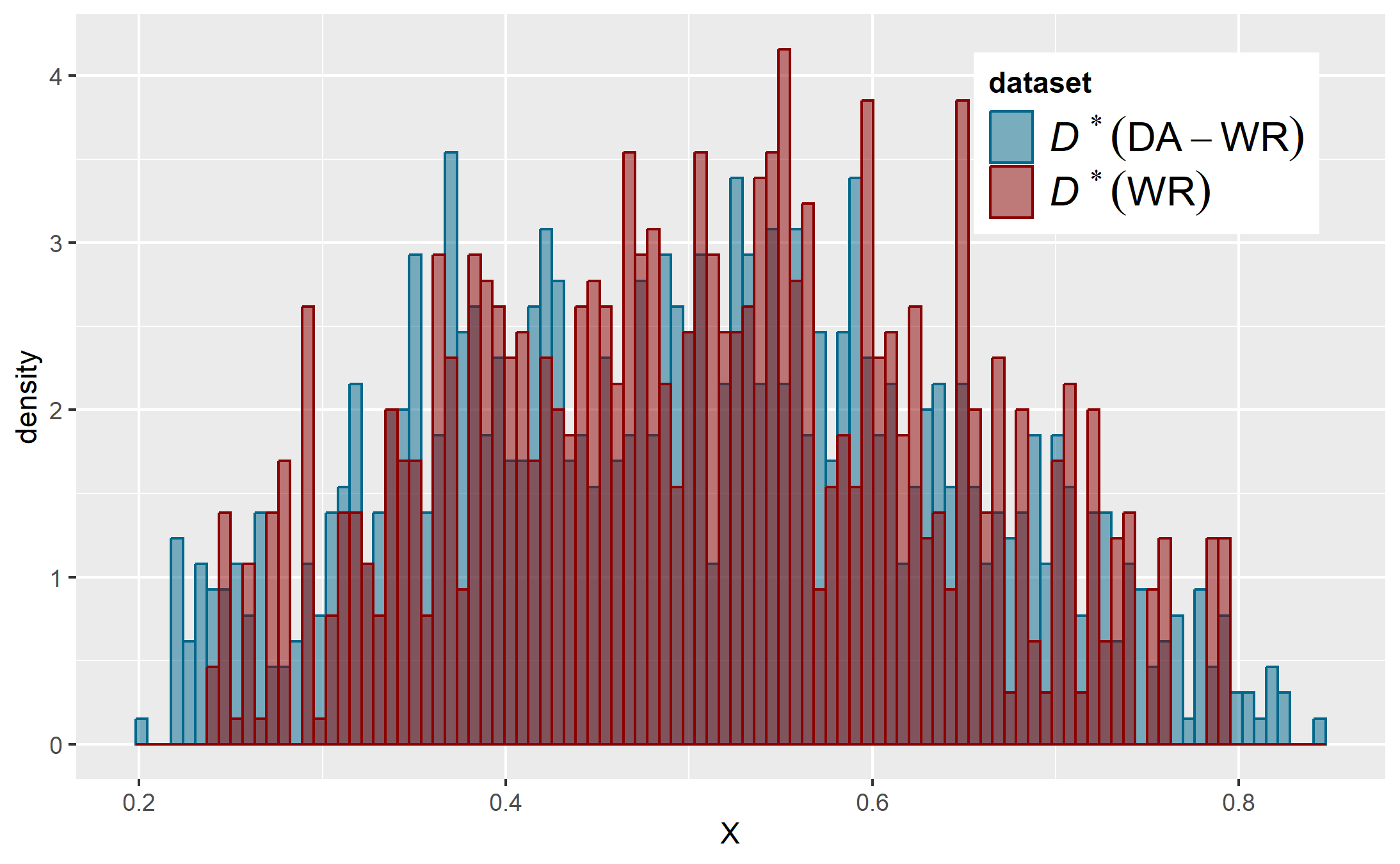}
    \caption{GN - GMM}
  \end{subfigure}
  \begin{subfigure}[b]{0.3\linewidth}
    \includegraphics[width=\linewidth]{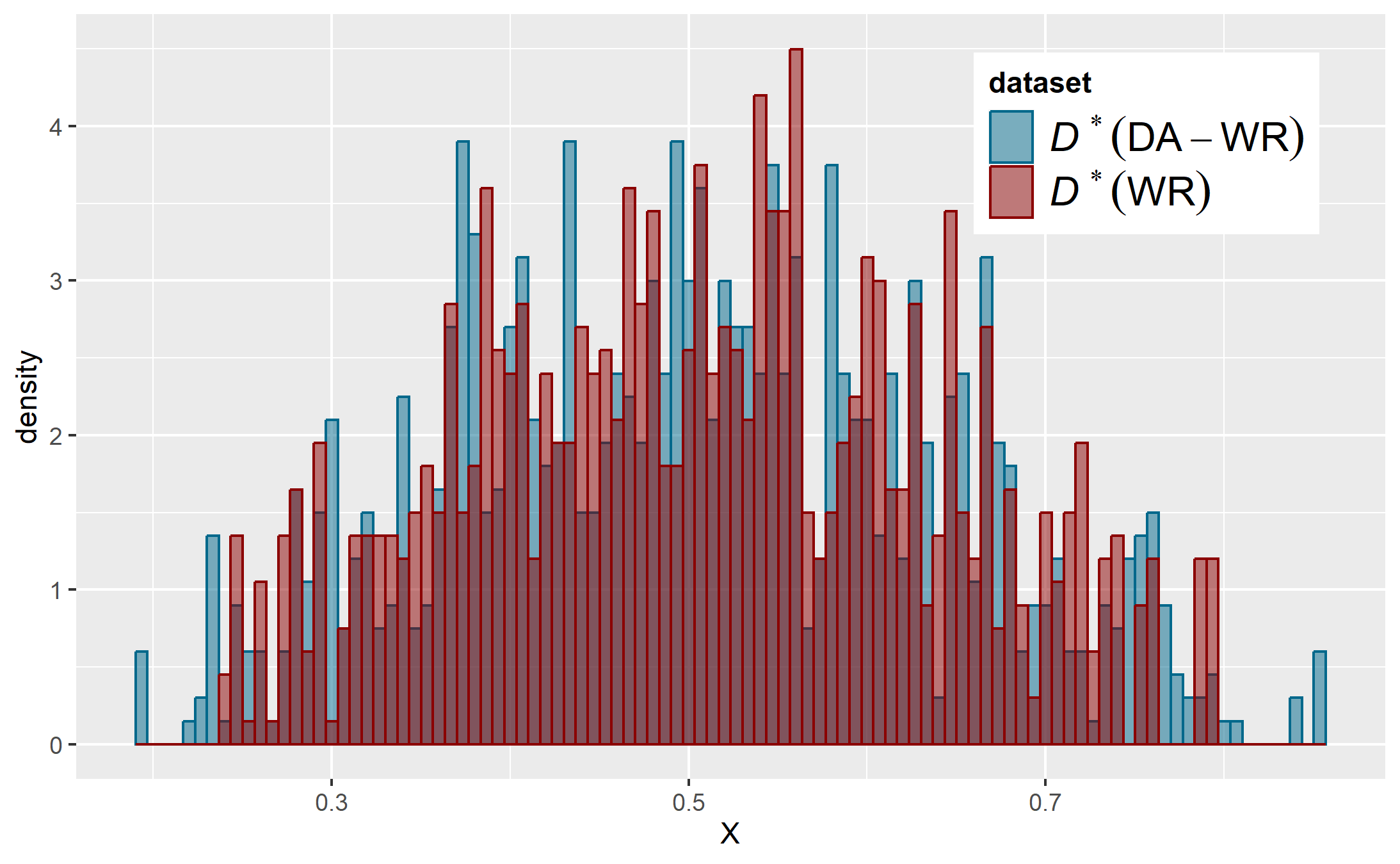}
    \caption{ROSE}
  \end{subfigure}

  \begin{subfigure}[b]{0.3\linewidth}
    \includegraphics[width=\linewidth]{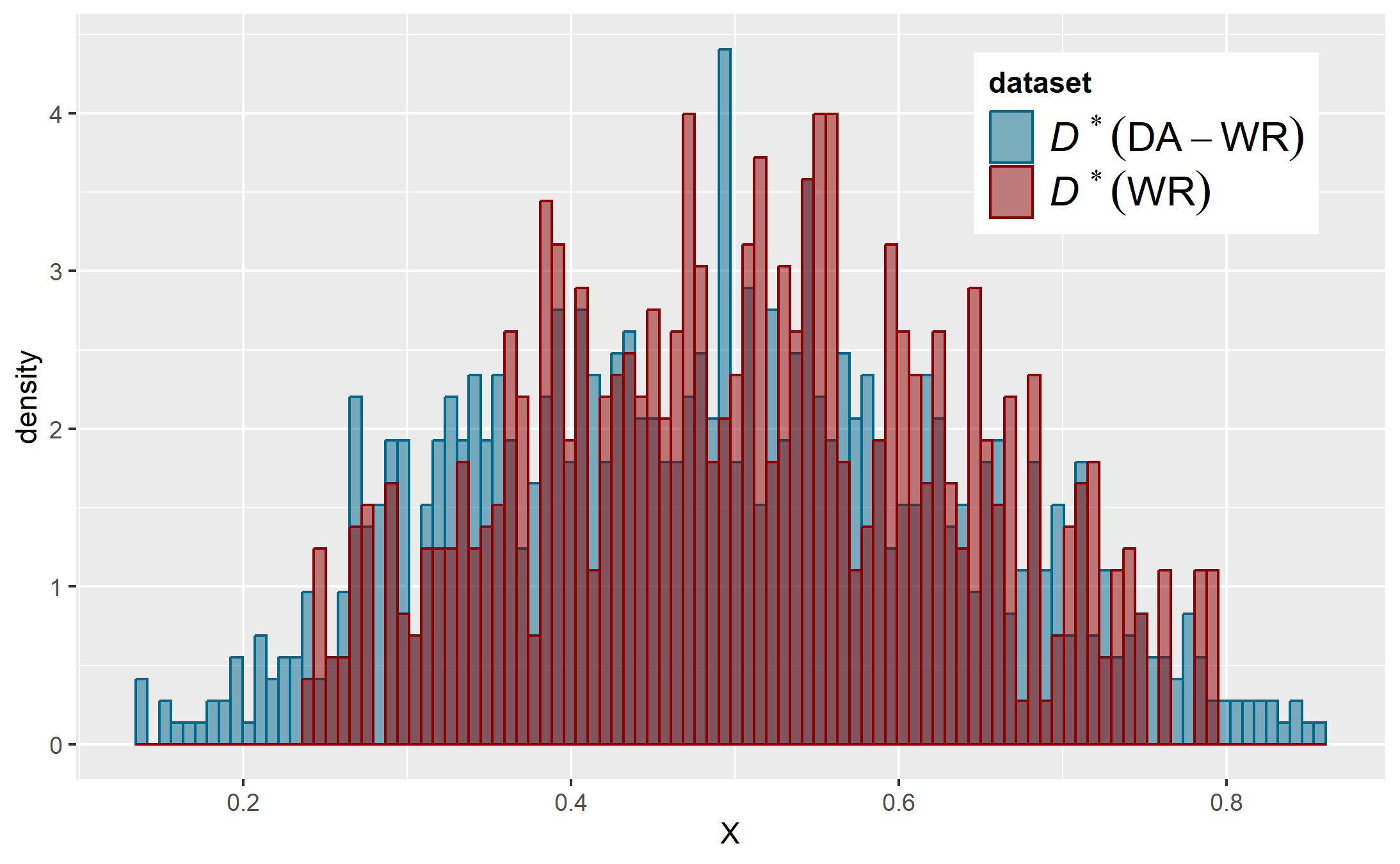}
     \caption{ROSE - GMM}
  \end{subfigure}
  \begin{subfigure}[b]{0.3\linewidth}
    \includegraphics[width=\linewidth]{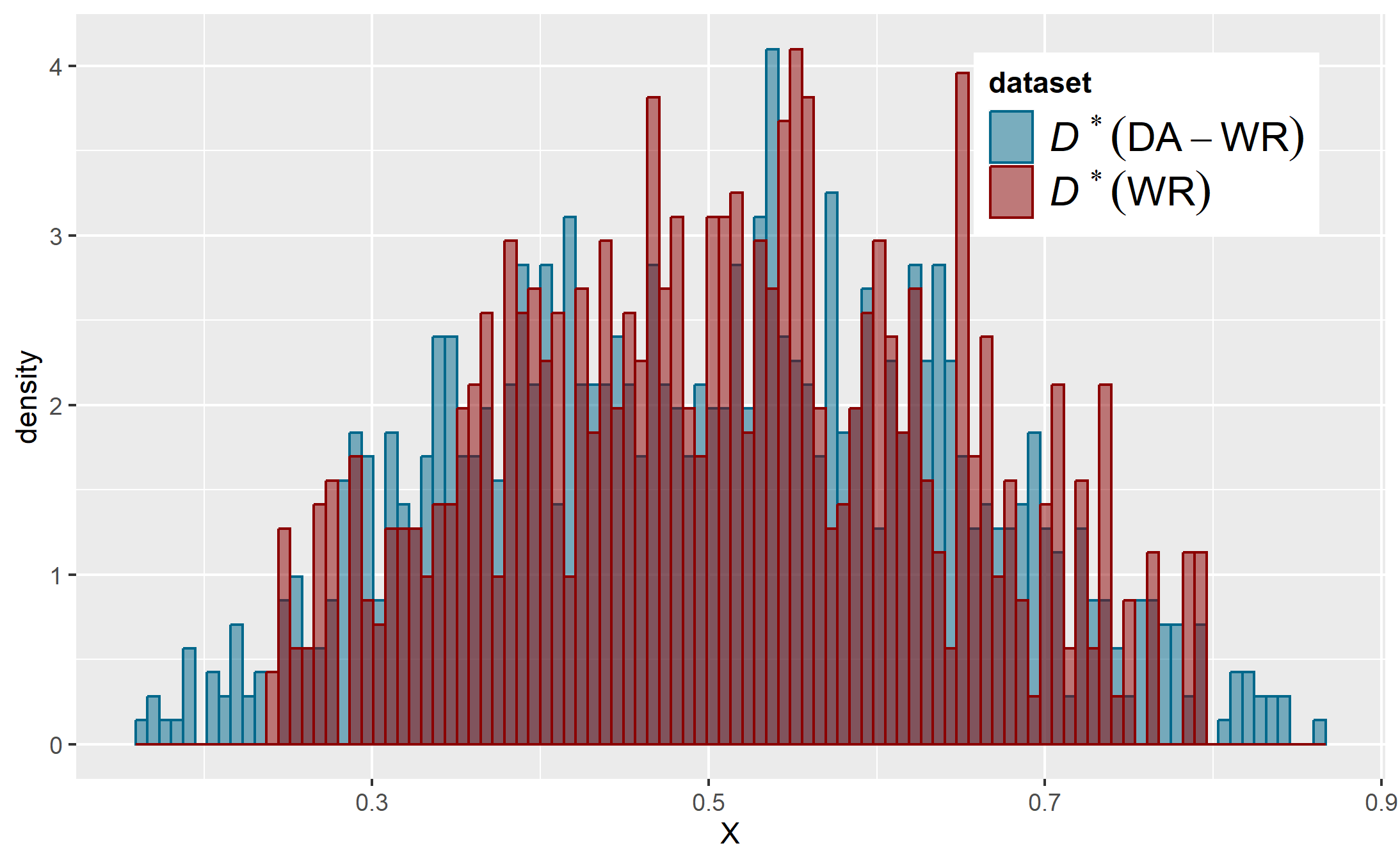}
    \caption{KDE - GMM}
  \end{subfigure}
  \begin{subfigure}[b]{0.3\linewidth}
    \includegraphics[width=\linewidth]{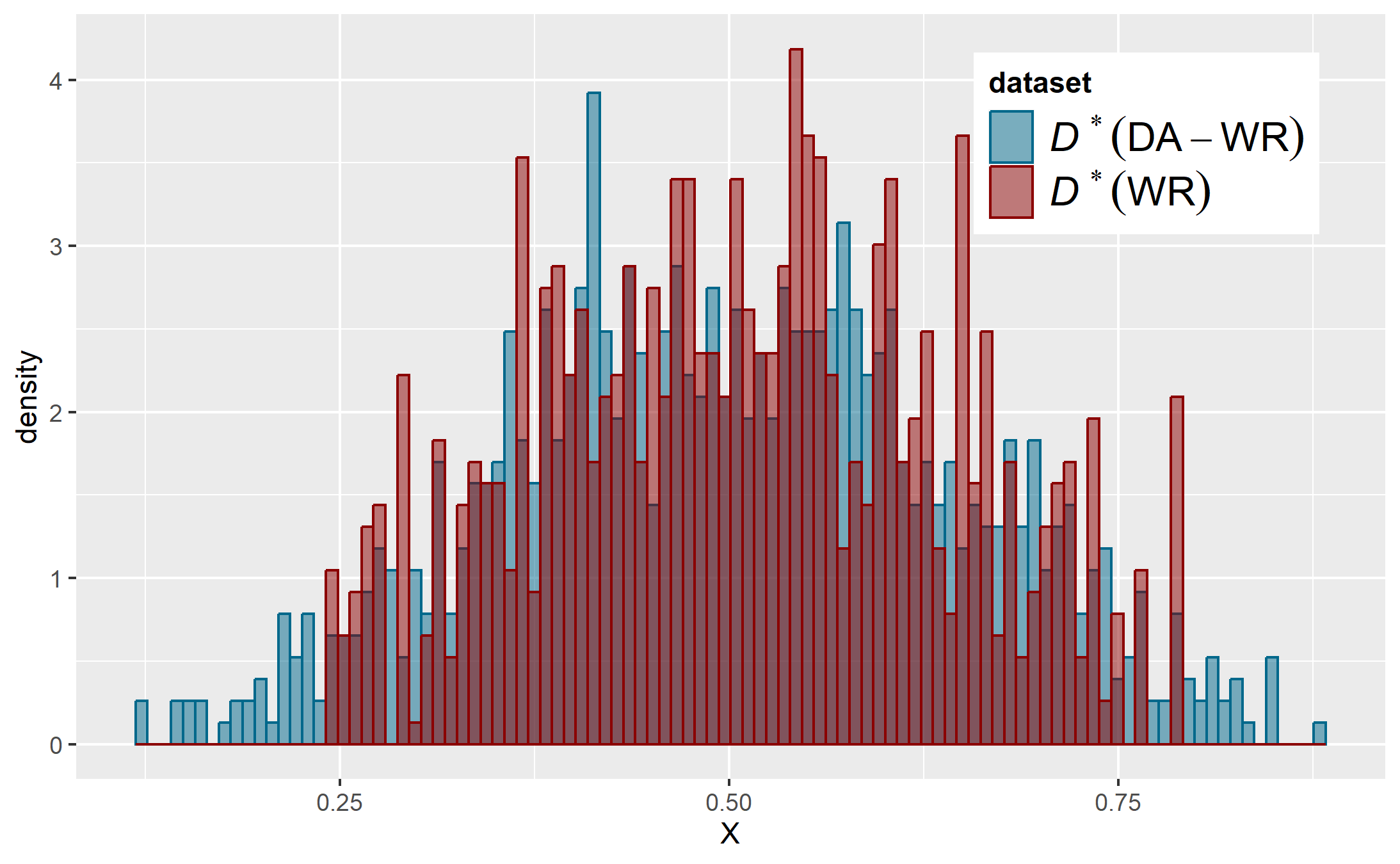}
    \caption{GMM}
  \end{subfigure}

  \begin{subfigure}[b]{0.3\linewidth}
    \includegraphics[width=\linewidth]{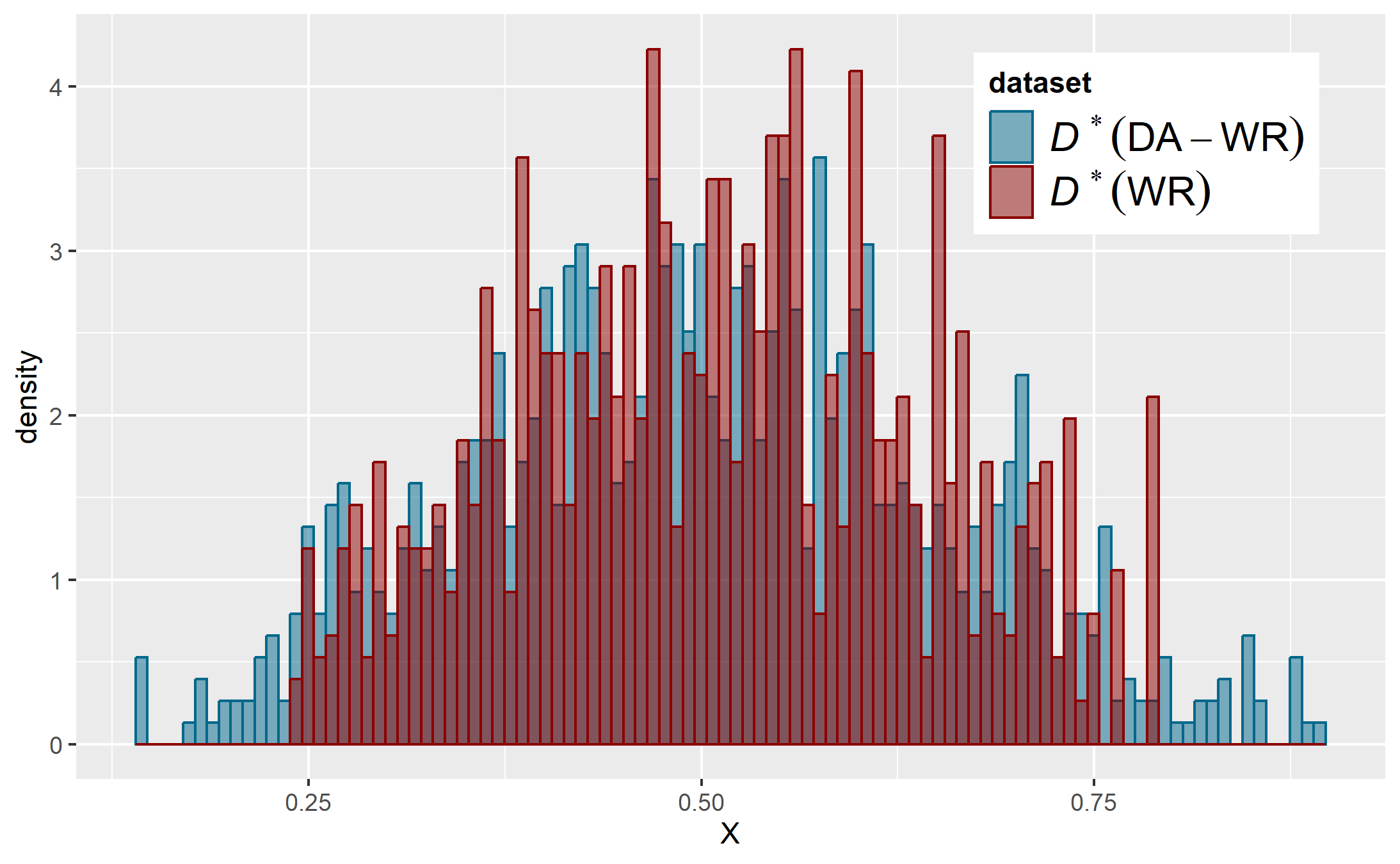}
     \caption{FA - GMM}
  \end{subfigure}
  \begin{subfigure}[b]{0.3\linewidth}
    \includegraphics[width=\linewidth]{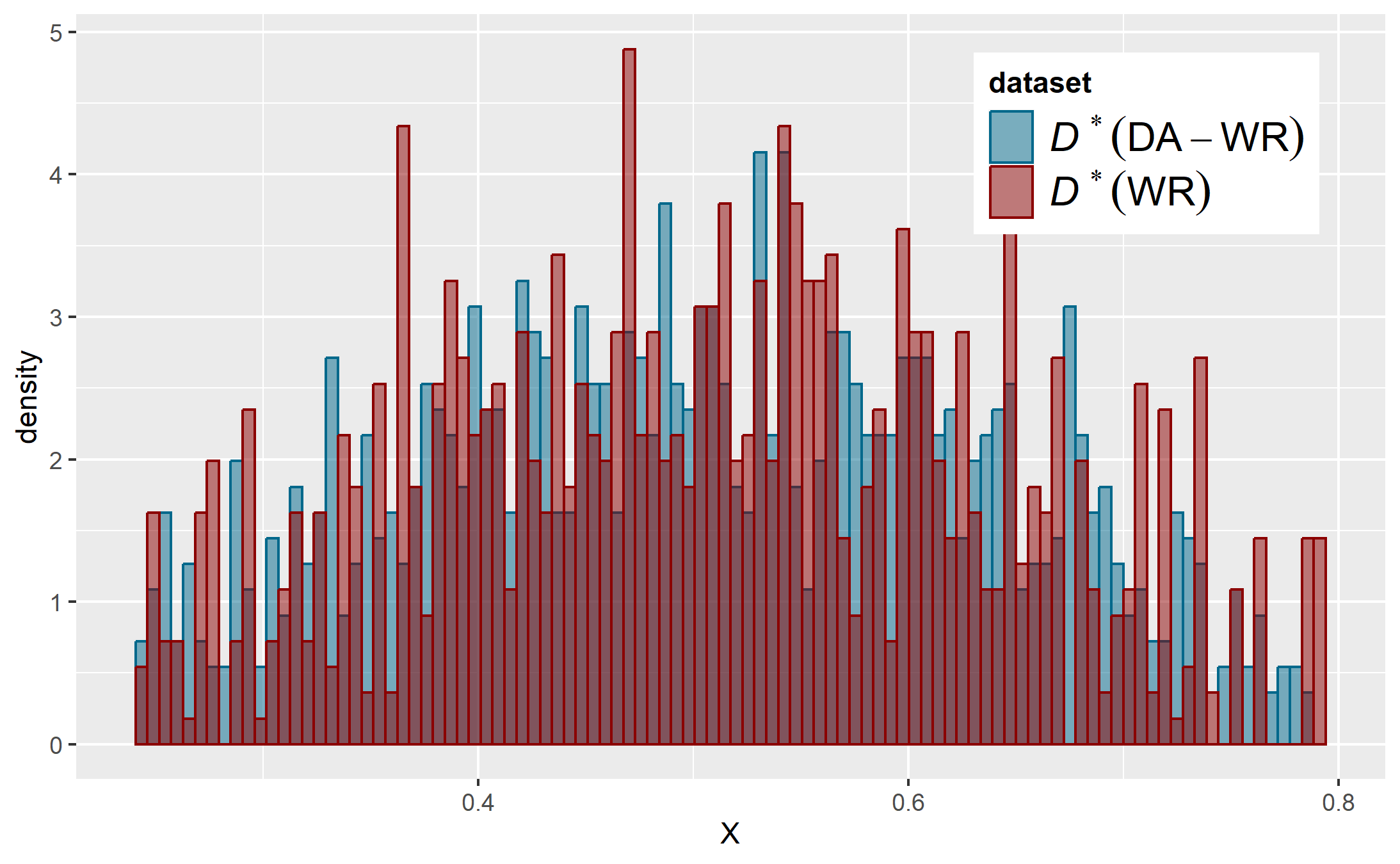}
    \caption{Copula}
  \end{subfigure}
  \begin{subfigure}[b]{0.3\linewidth}
    \includegraphics[width=\linewidth]{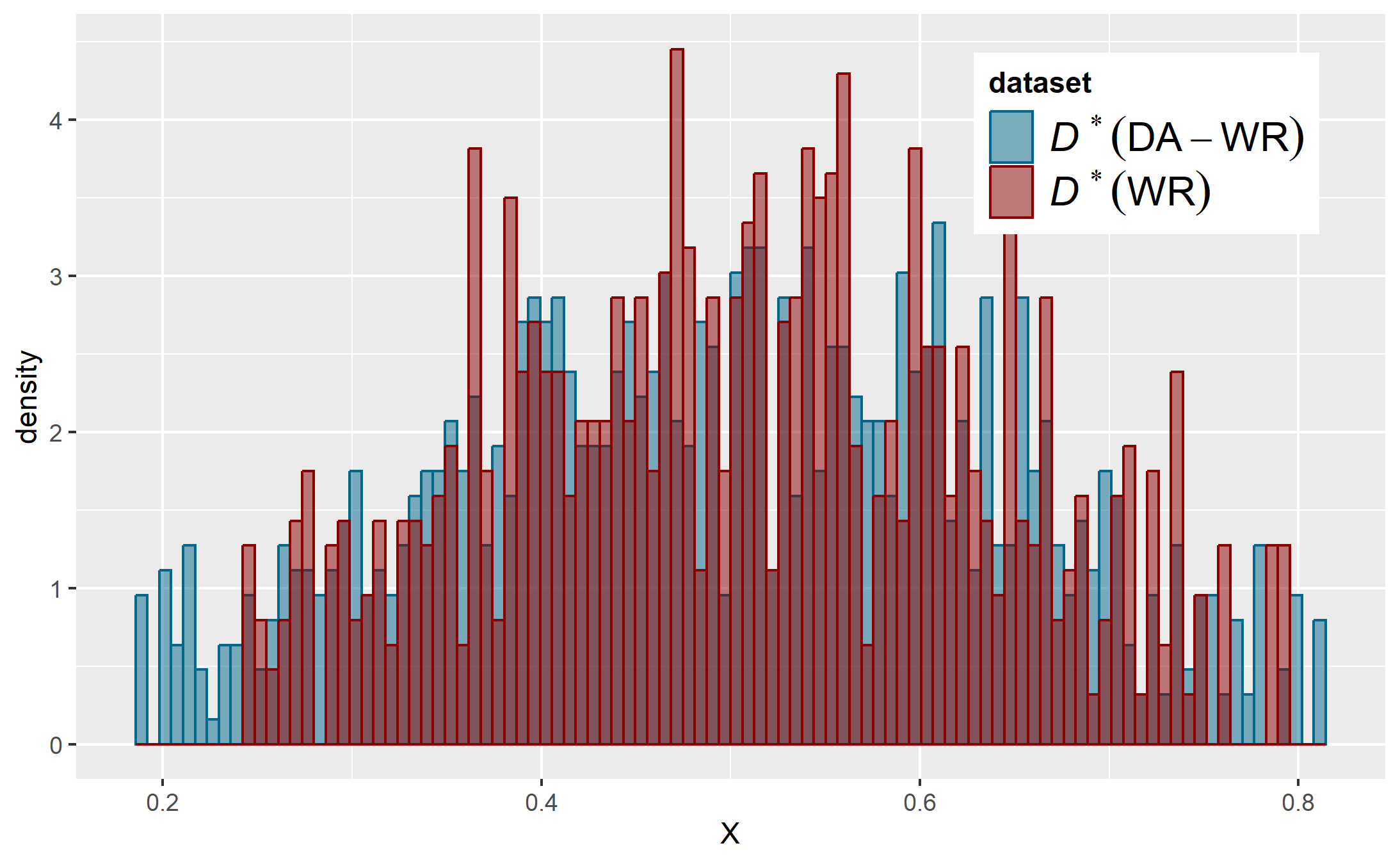}
    \caption{GAN}
  \end{subfigure}

  \begin{subfigure}[b]{0.3\linewidth}
    \includegraphics[width=\linewidth]{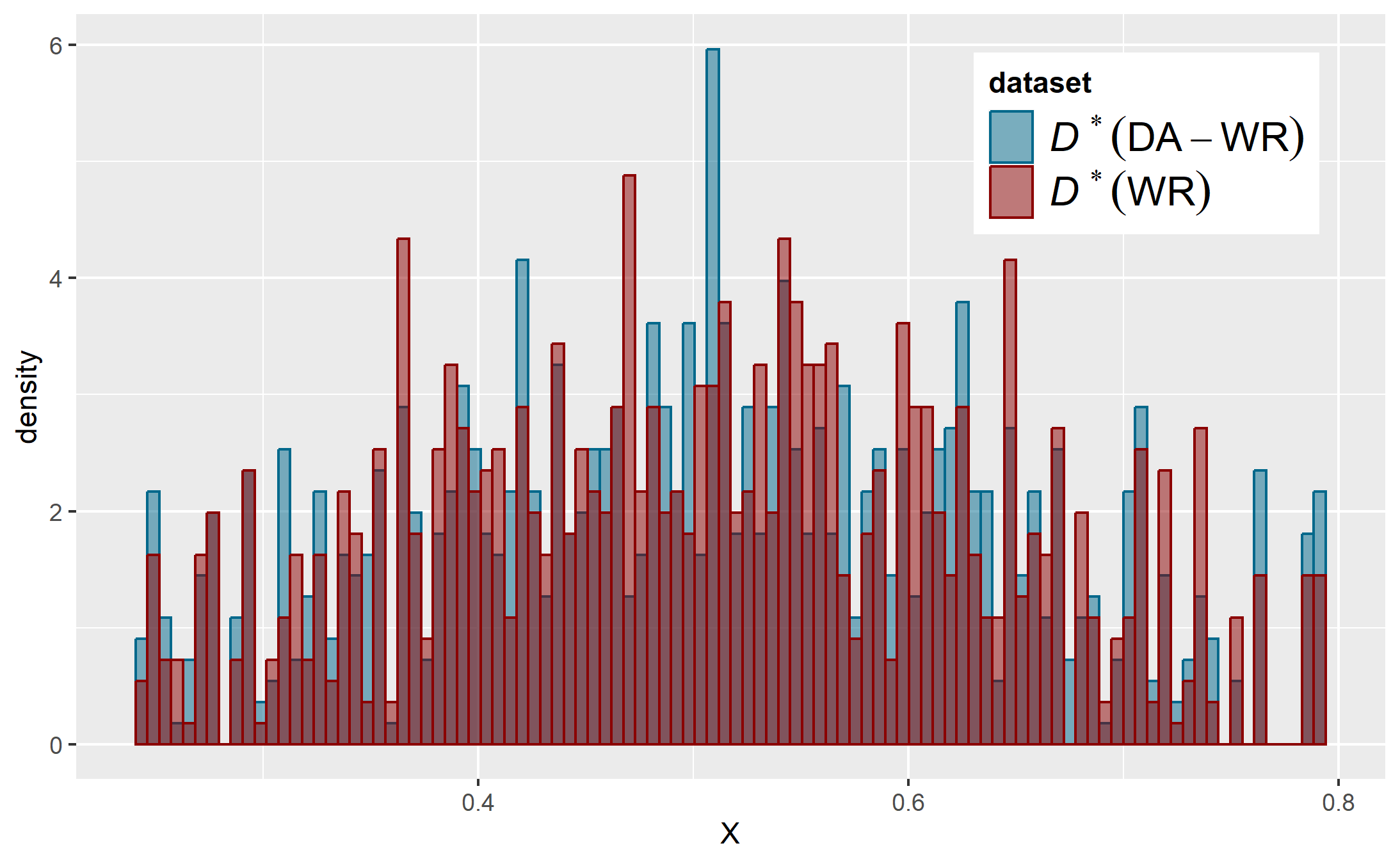}
     \caption{RF}
  \end{subfigure}
  \begin{subfigure}[b]{0.3\linewidth}
    \includegraphics[width=\linewidth]{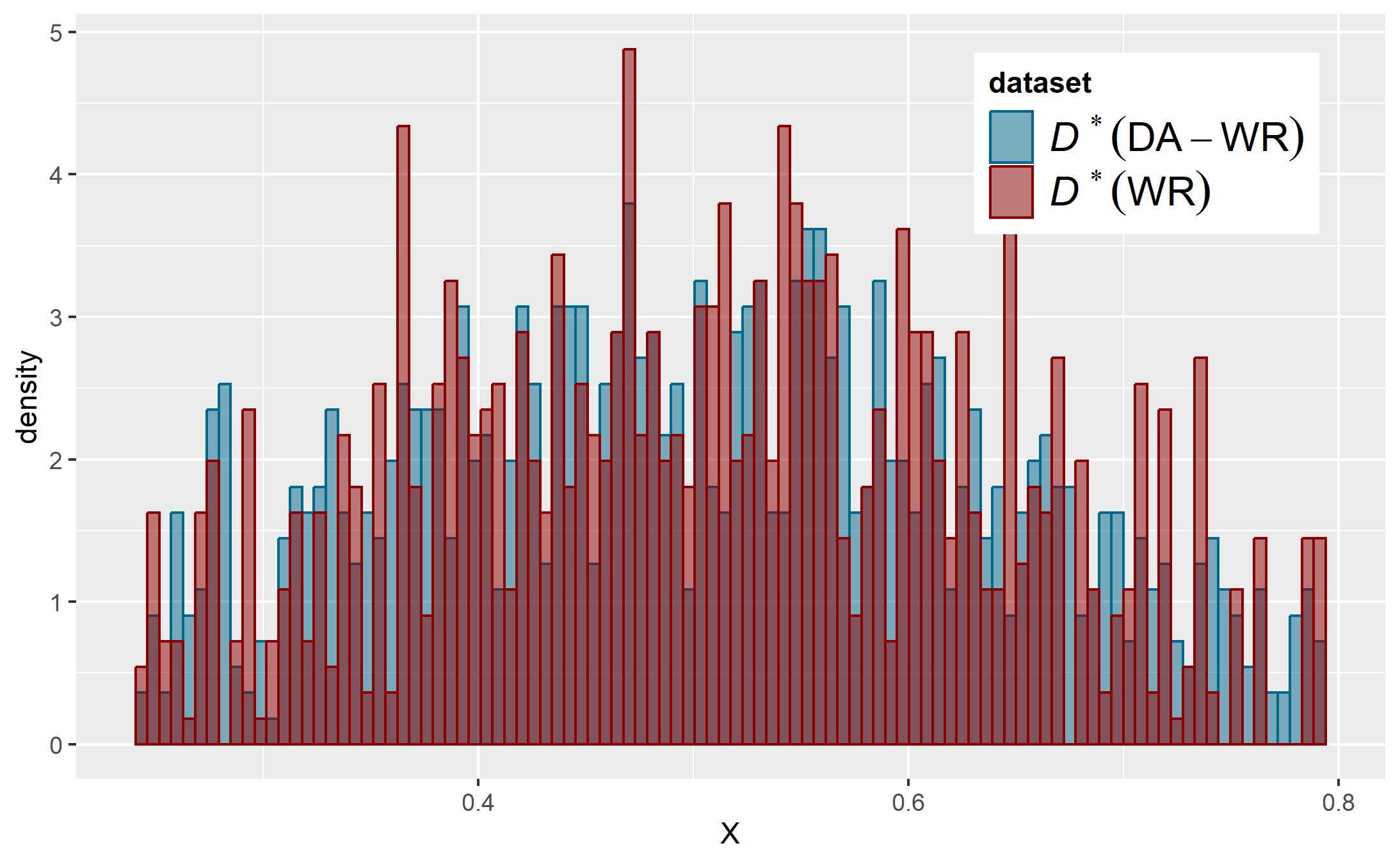}
    \caption{SMOTE}
  \end{subfigure}
  \begin{subfigure}[b]{0.3\linewidth}
    \includegraphics[width=\linewidth]{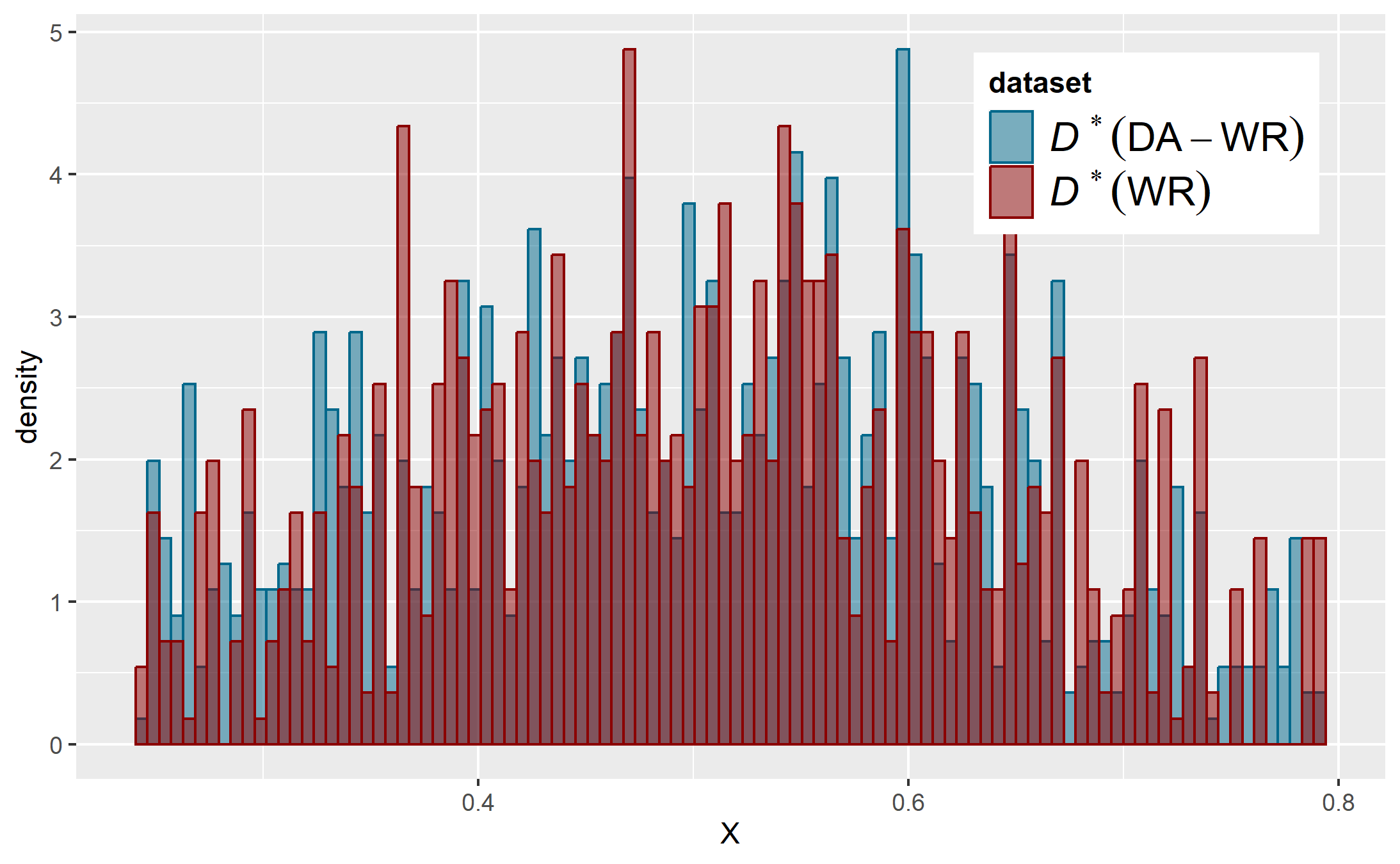}
    \caption{SMOTE - GMM}
  \end{subfigure}

  \caption{Histogram of $X$ obtained in new samples (new) vs WR}
  \label{Hist_X_ech_XXX-vs-ech_add}
\end{figure}

\newpage
\textbf{Scatterplot $(X,Y)$ obtained in new samples vs imbalanced sample}

\begin{figure}[H]
  \centering
  \begin{subfigure}[b]{0.3\linewidth}
    \includegraphics[width=\linewidth]{imgs/Illu/comp_Y_ech_GN_SC-vs-ech0.png}
     \caption{GN}
  \end{subfigure}
  \begin{subfigure}[b]{0.3\linewidth}
    \includegraphics[width=\linewidth]{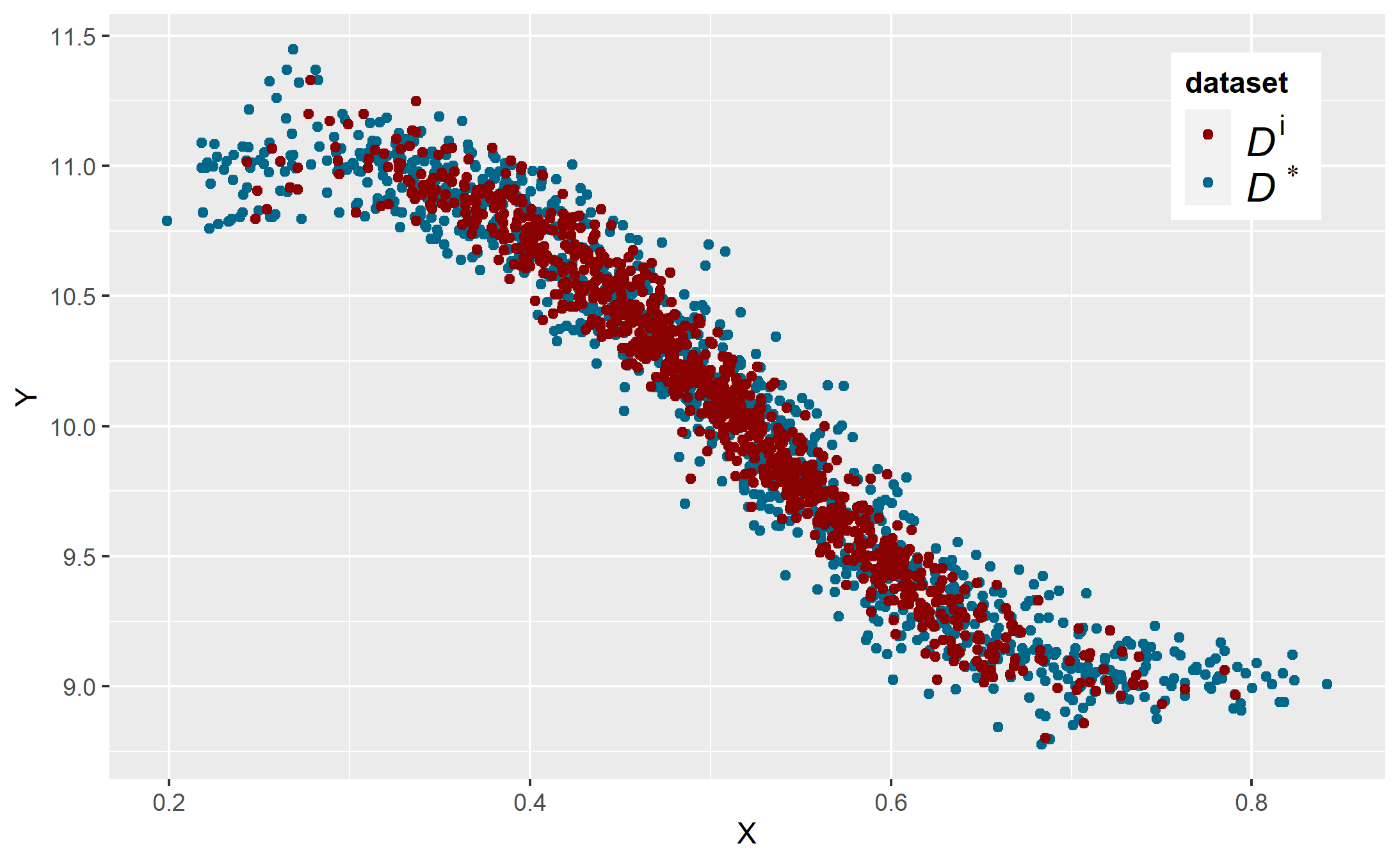}
    \caption{GN - GMM}
  \end{subfigure}
  \begin{subfigure}[b]{0.3\linewidth}
    \includegraphics[width=\linewidth]{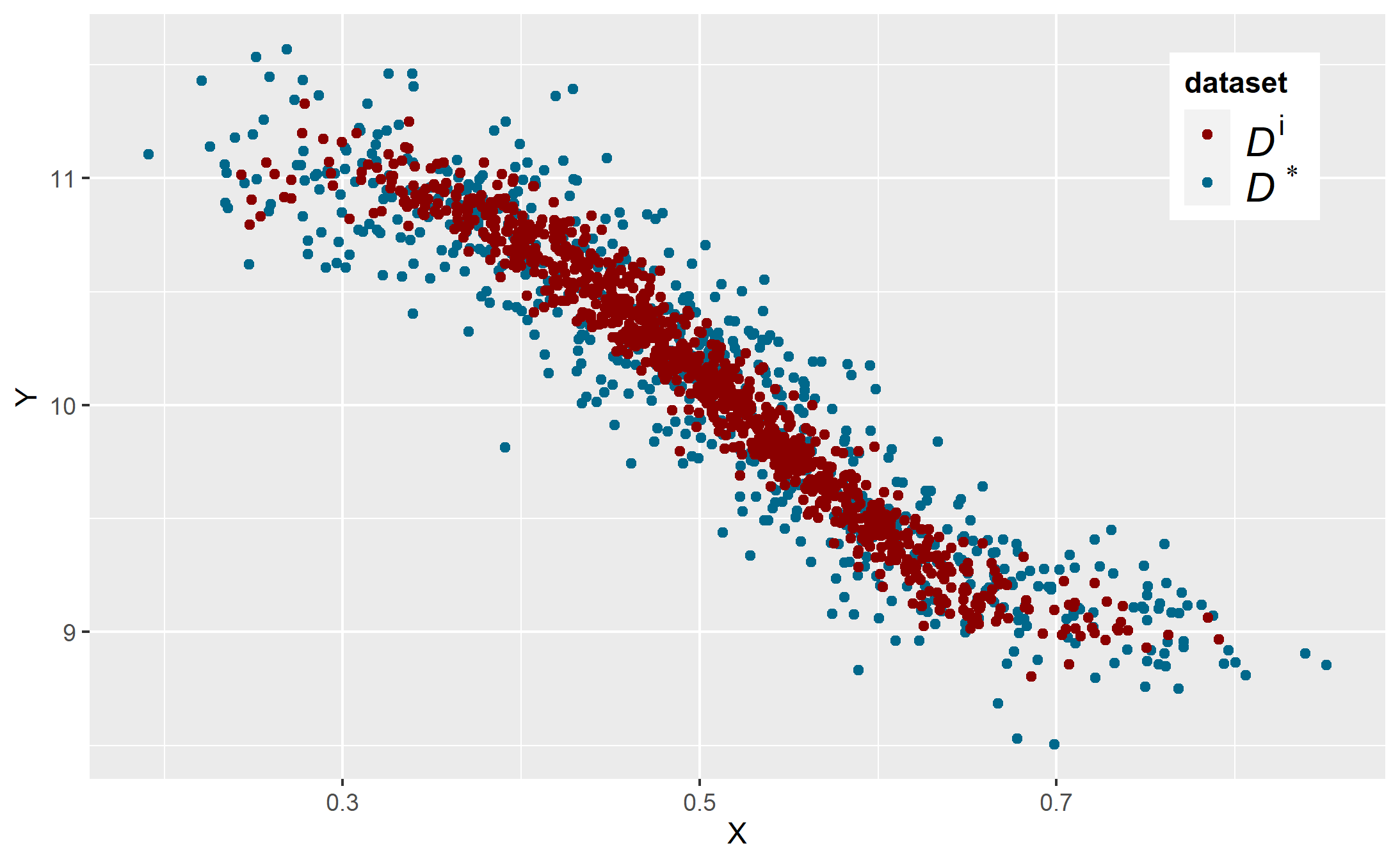}
    \caption{ROSE}
  \end{subfigure}

  \begin{subfigure}[b]{0.3\linewidth}
    \includegraphics[width=\linewidth]{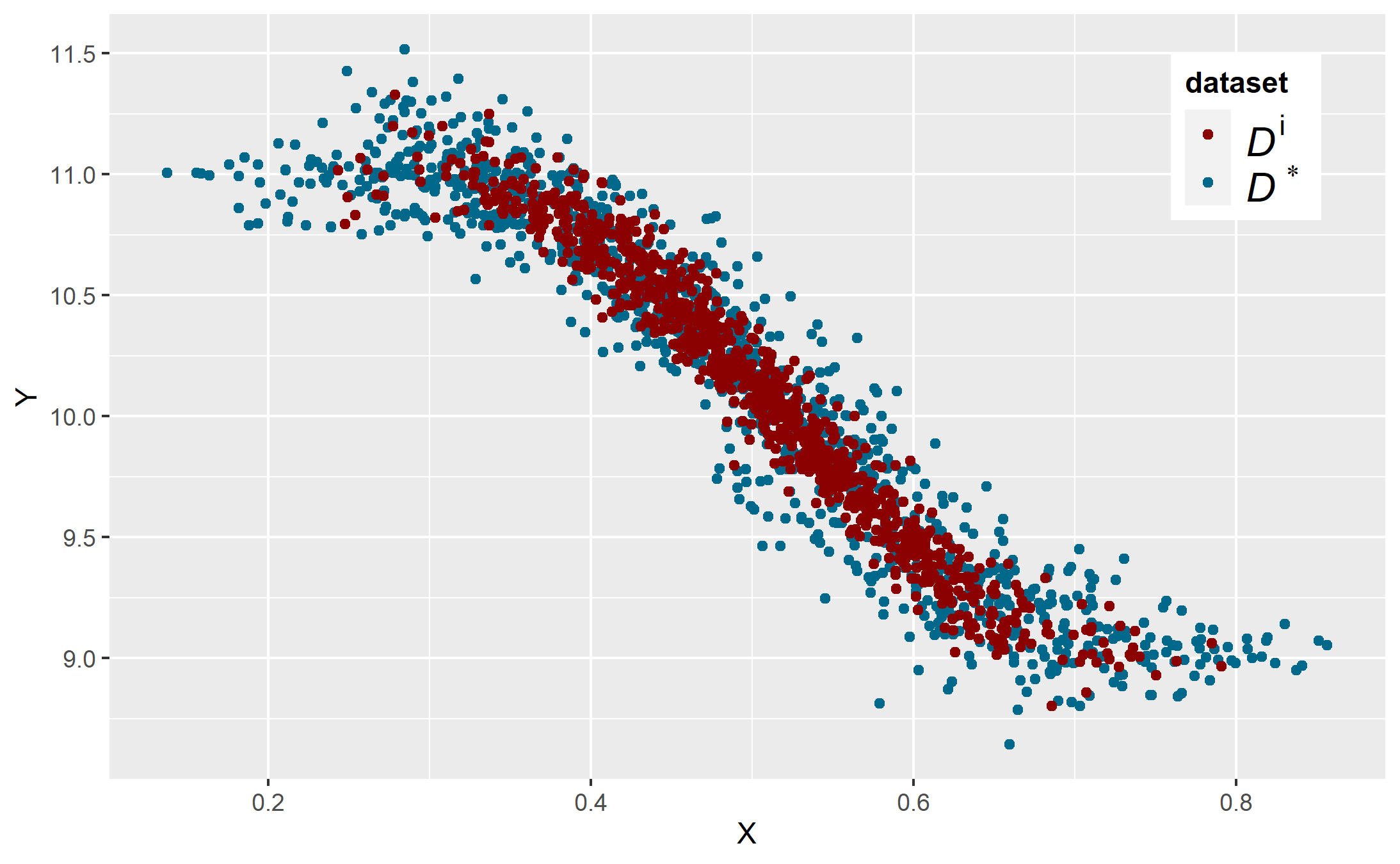}
     \caption{ROSE - GMM}
  \end{subfigure}
  \begin{subfigure}[b]{0.3\linewidth}
    \includegraphics[width=\linewidth]{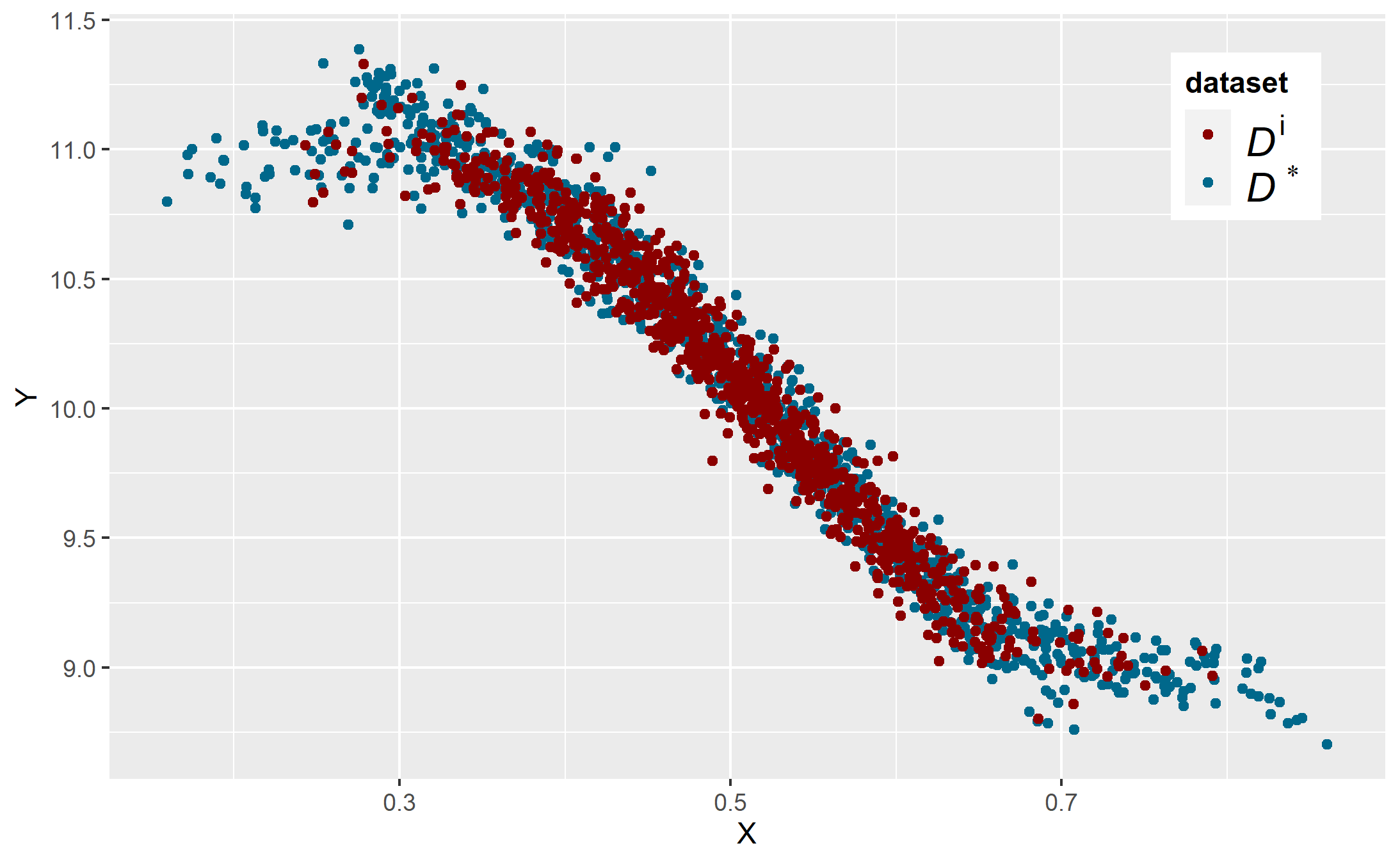}
    \caption{KDE - GMM}
  \end{subfigure}
  \begin{subfigure}[b]{0.3\linewidth}
    \includegraphics[width=\linewidth]{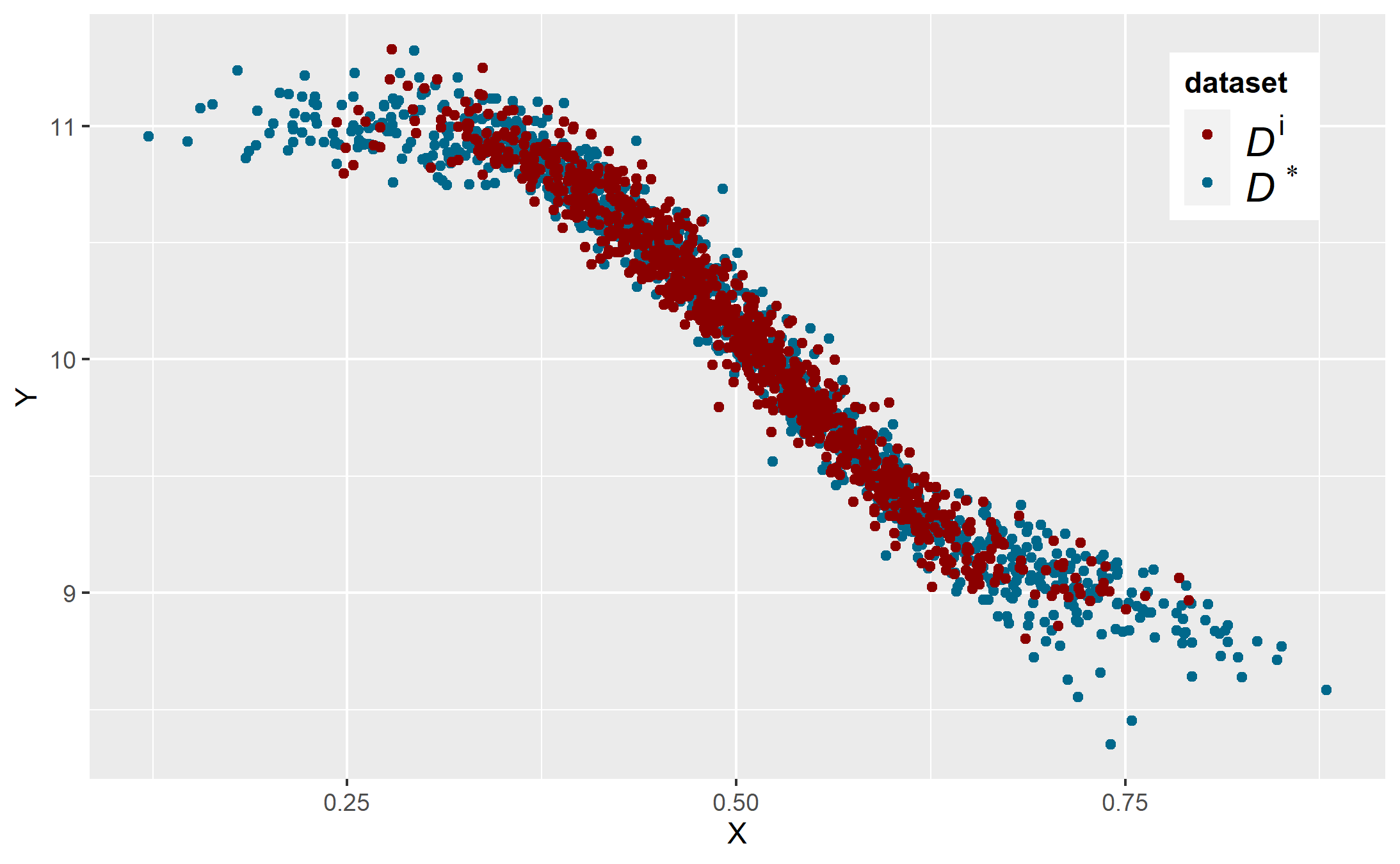}
    \caption{GMM}
  \end{subfigure}

  \begin{subfigure}[b]{0.3\linewidth}
    \includegraphics[width=\linewidth]{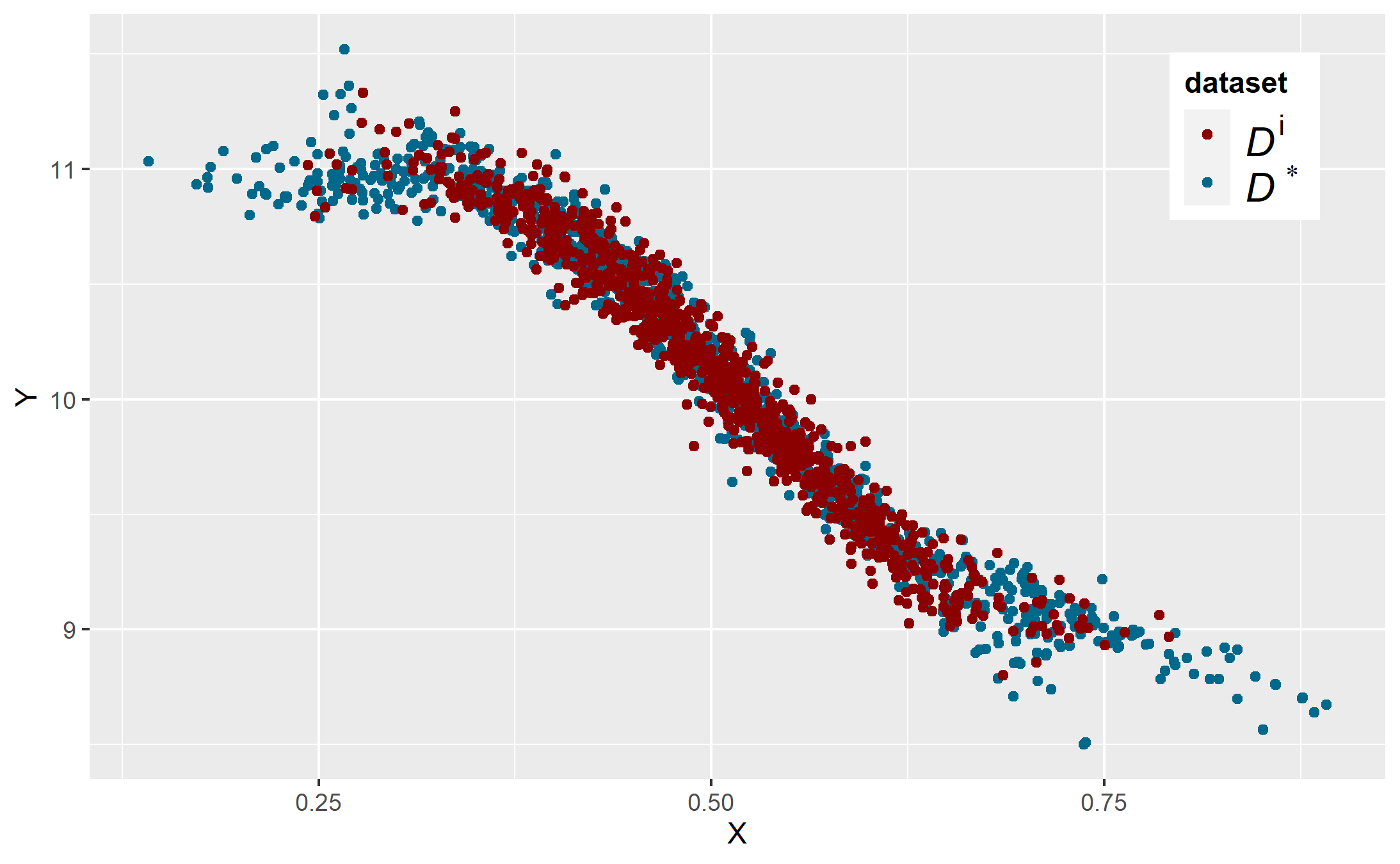}
     \caption{FA - GMM}
  \end{subfigure}
  \begin{subfigure}[b]{0.3\linewidth}
    \includegraphics[width=\linewidth]{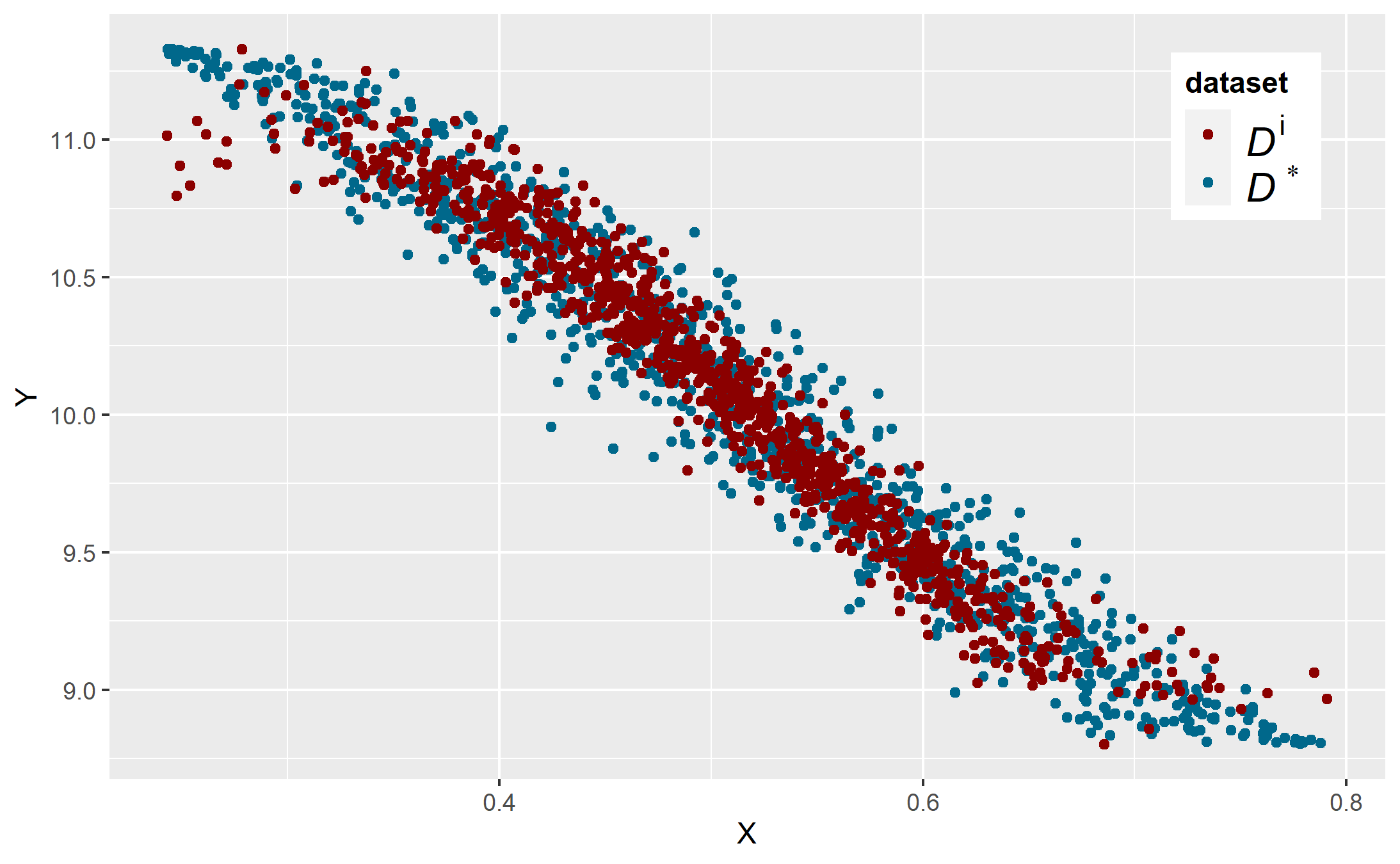}
    \caption{Copula}
  \end{subfigure}
  \begin{subfigure}[b]{0.3\linewidth}
    \includegraphics[width=\linewidth]{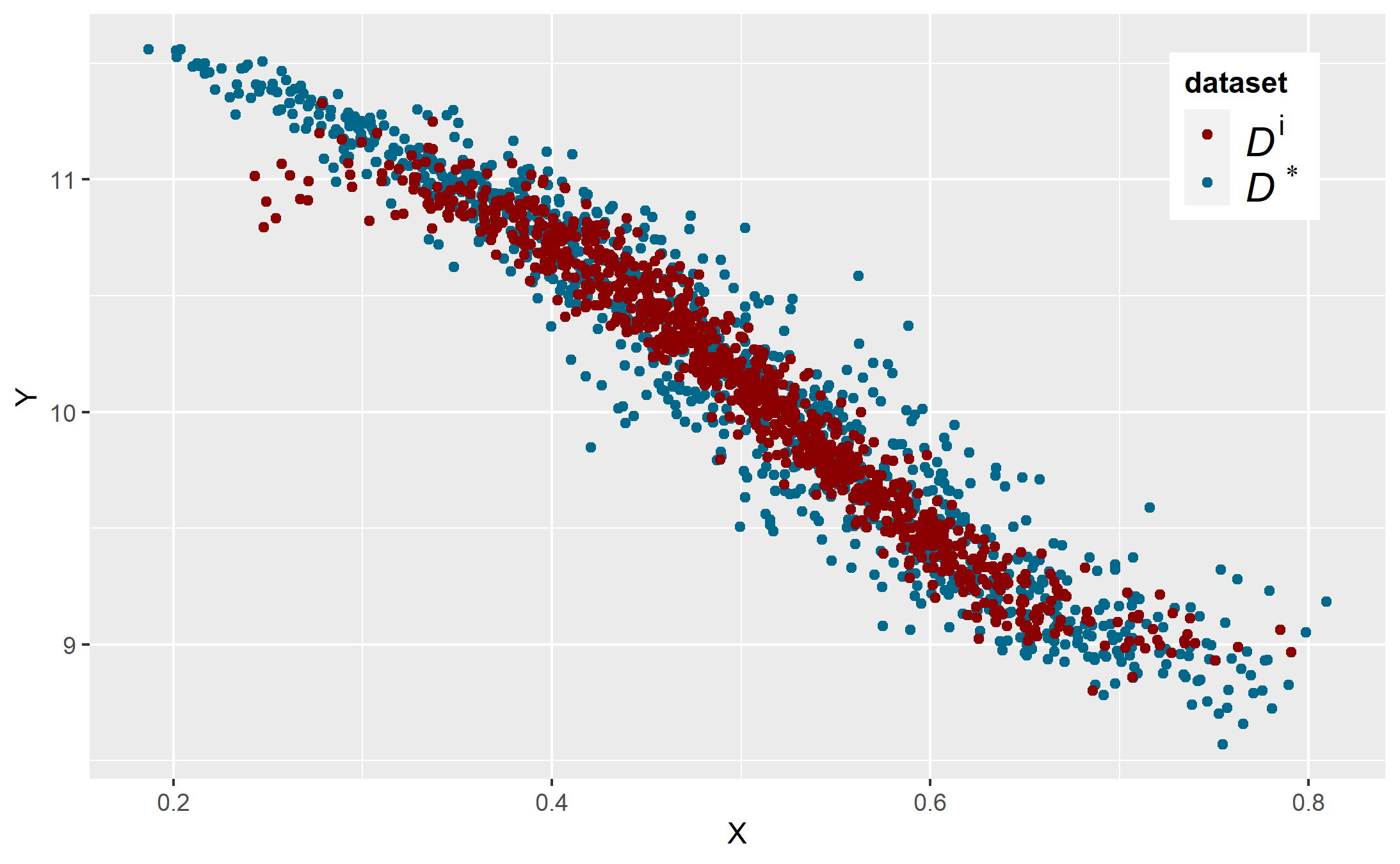}
    \caption{GAN}
  \end{subfigure}

  \begin{subfigure}[b]{0.3\linewidth}
    \includegraphics[width=\linewidth]{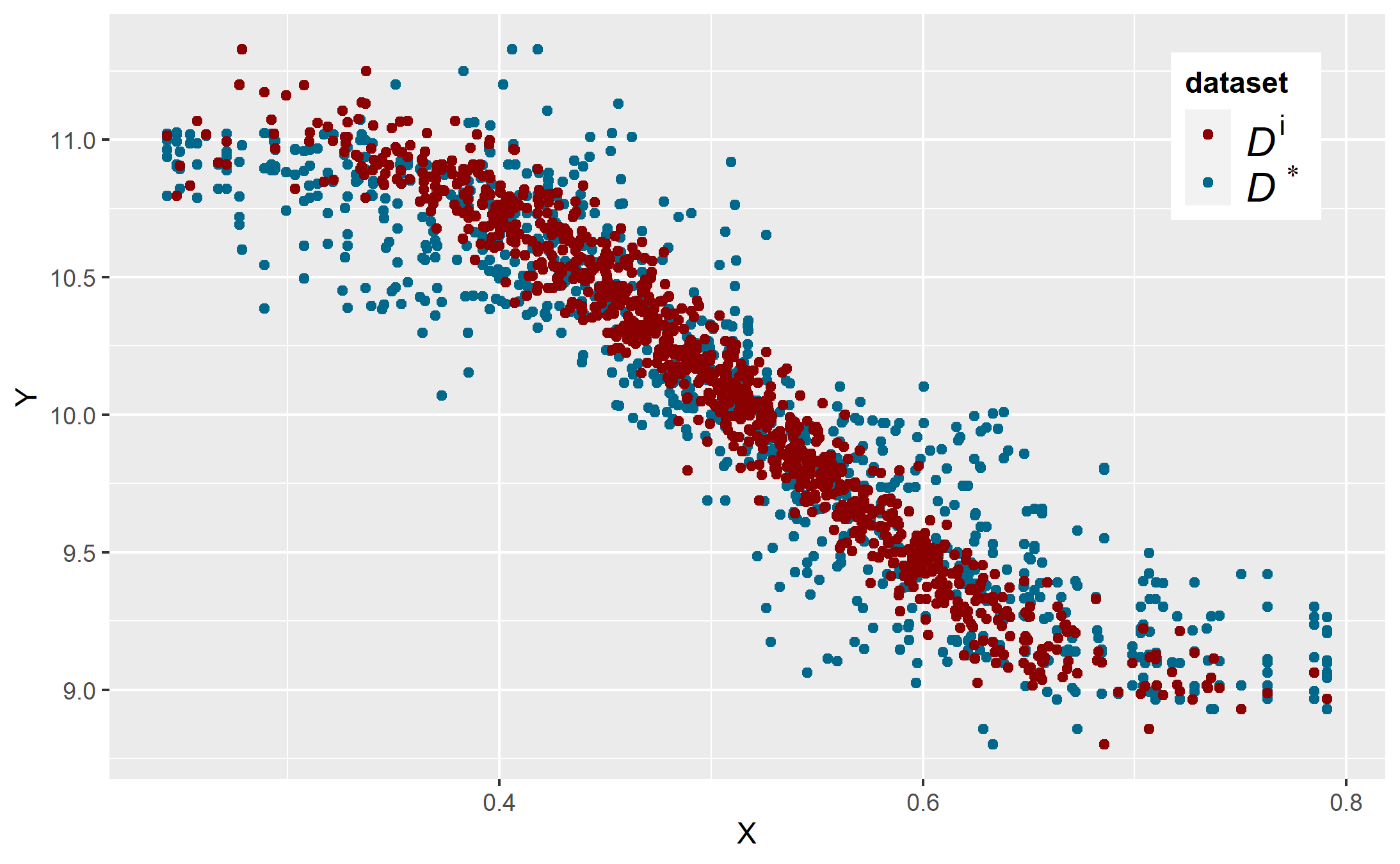}
     \caption{RF}
  \end{subfigure}
  \begin{subfigure}[b]{0.3\linewidth}
    \includegraphics[width=\linewidth]{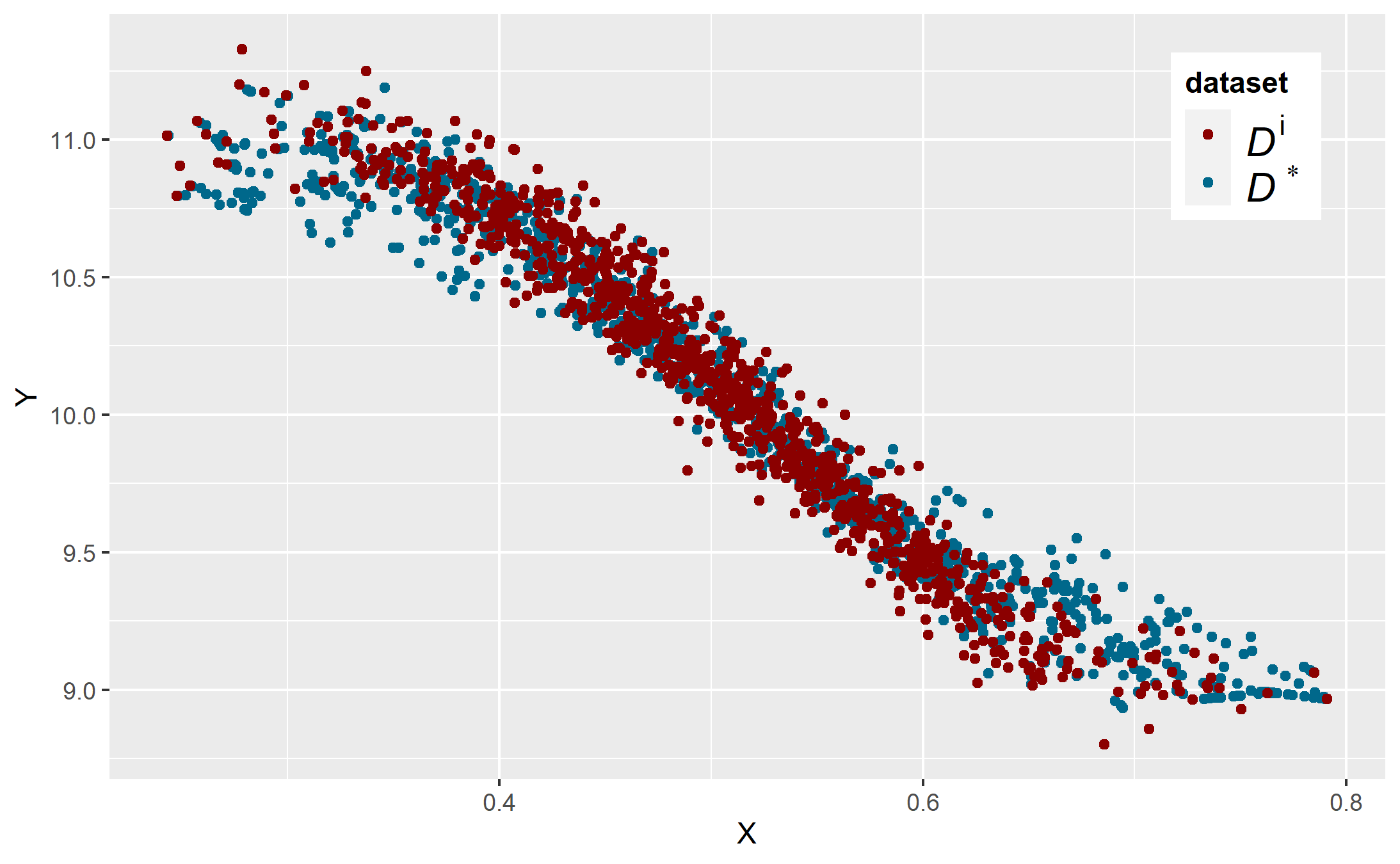}
    \caption{SMOTE}
  \end{subfigure}
  \begin{subfigure}[b]{0.3\linewidth}
    \includegraphics[width=\linewidth]{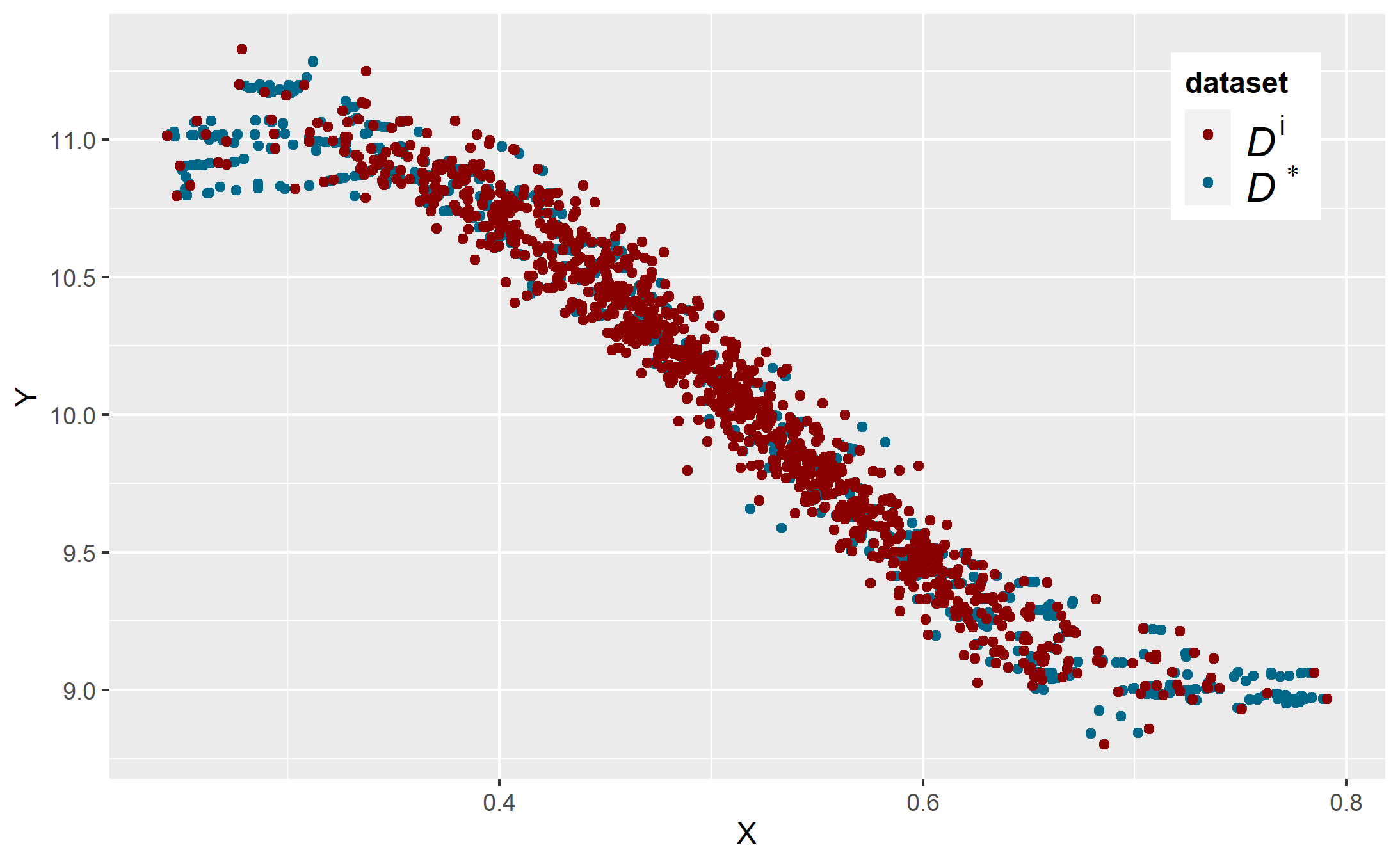}
    \caption{SMOTE - GMM}
  \end{subfigure}

  \caption{Scatterplot $(X,Y)$ obtained in new samples (new) vs imbalanced sample (imb)}
  \label{comp_Y_ech_XXX-vs-ech0}
\end{figure}

\newpage

\subsubsection{Predictions}

\textbf{Generalized Additive Model predictions}

\begin{figure}[H]
  \centering

  \begin{subfigure}[b]{0.3\linewidth}
    \includegraphics[width=\linewidth]{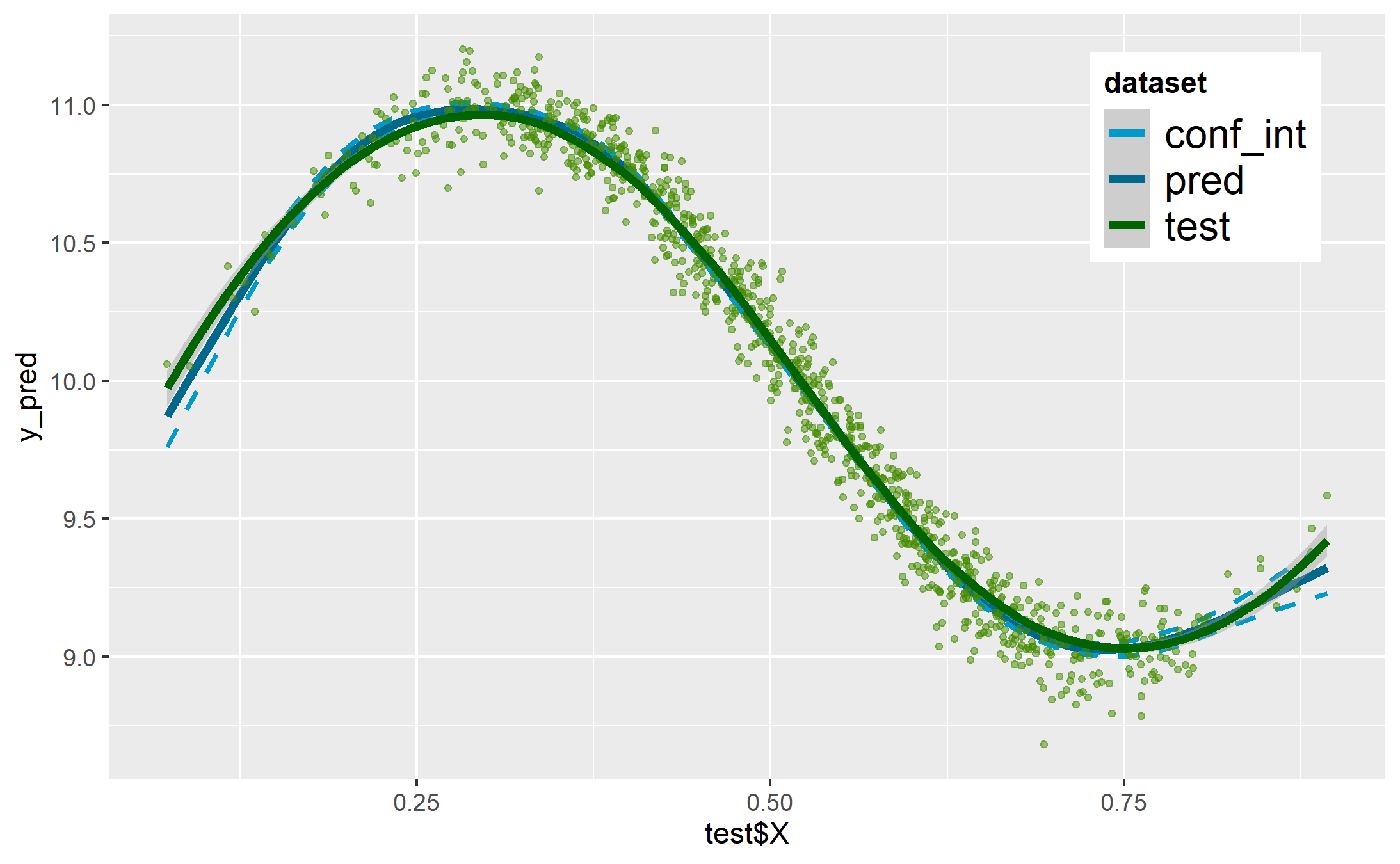}
     \caption{$\mathcal{D}^b$}
  \end{subfigure}
  \begin{subfigure}[b]{0.3\linewidth}
    \includegraphics[width=\linewidth]{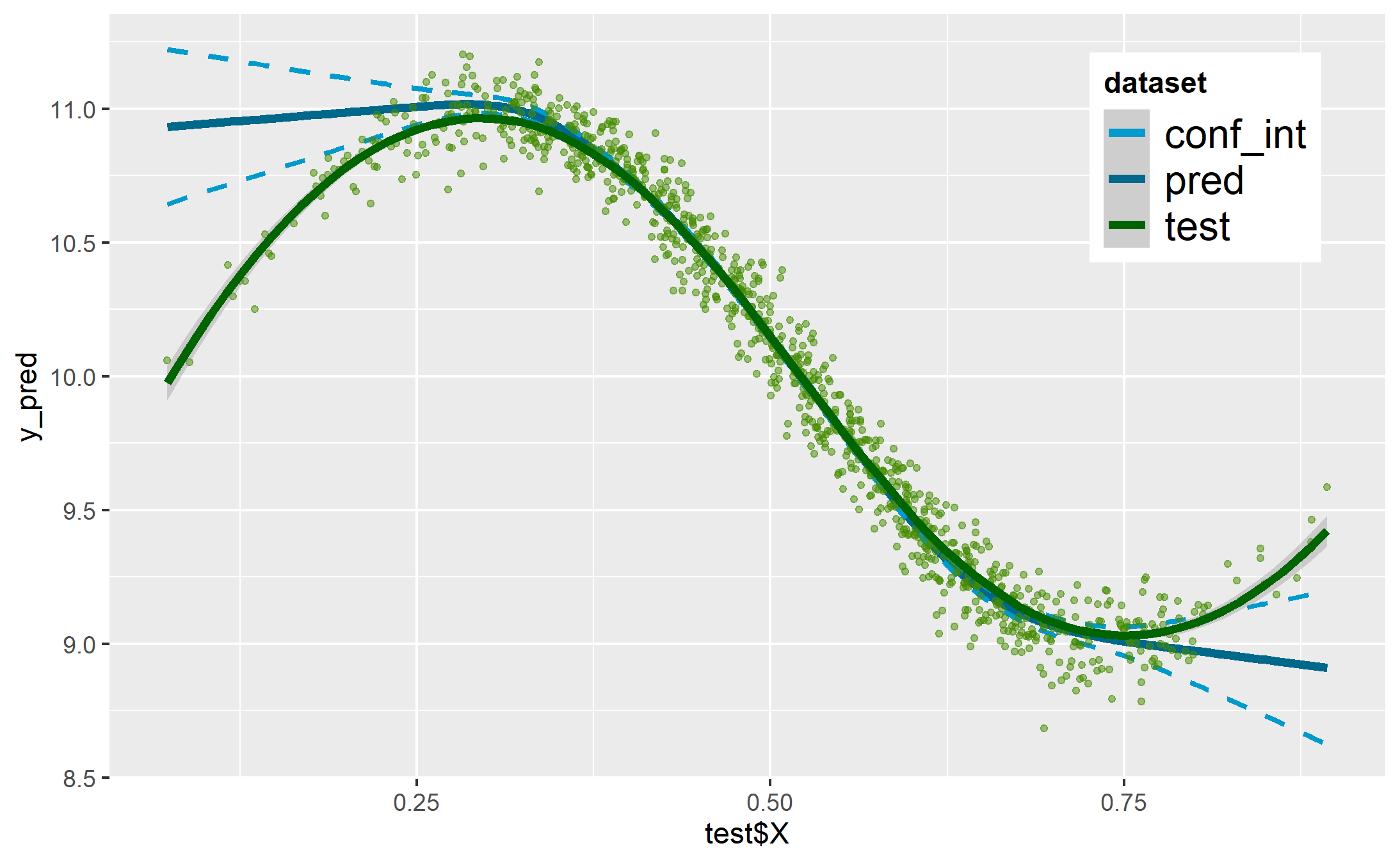}
    \caption{$\mathcal{D}^i$}
  \end{subfigure}
  \begin{subfigure}[b]{0.3\linewidth}
    \includegraphics[width=\linewidth]{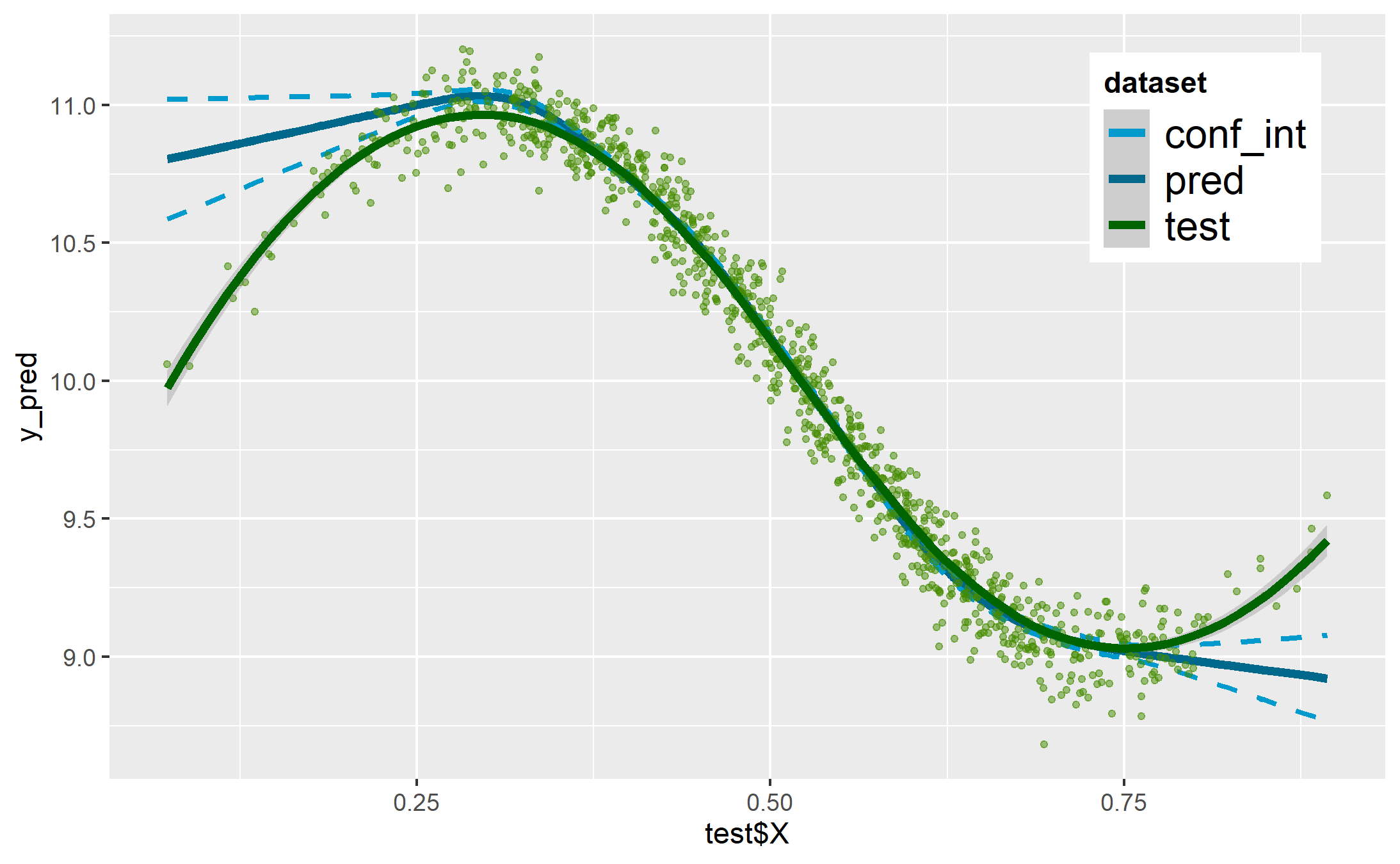}
    \caption{WR}
  \end{subfigure}

  \begin{subfigure}[b]{0.3\linewidth}
    \includegraphics[width=\linewidth]{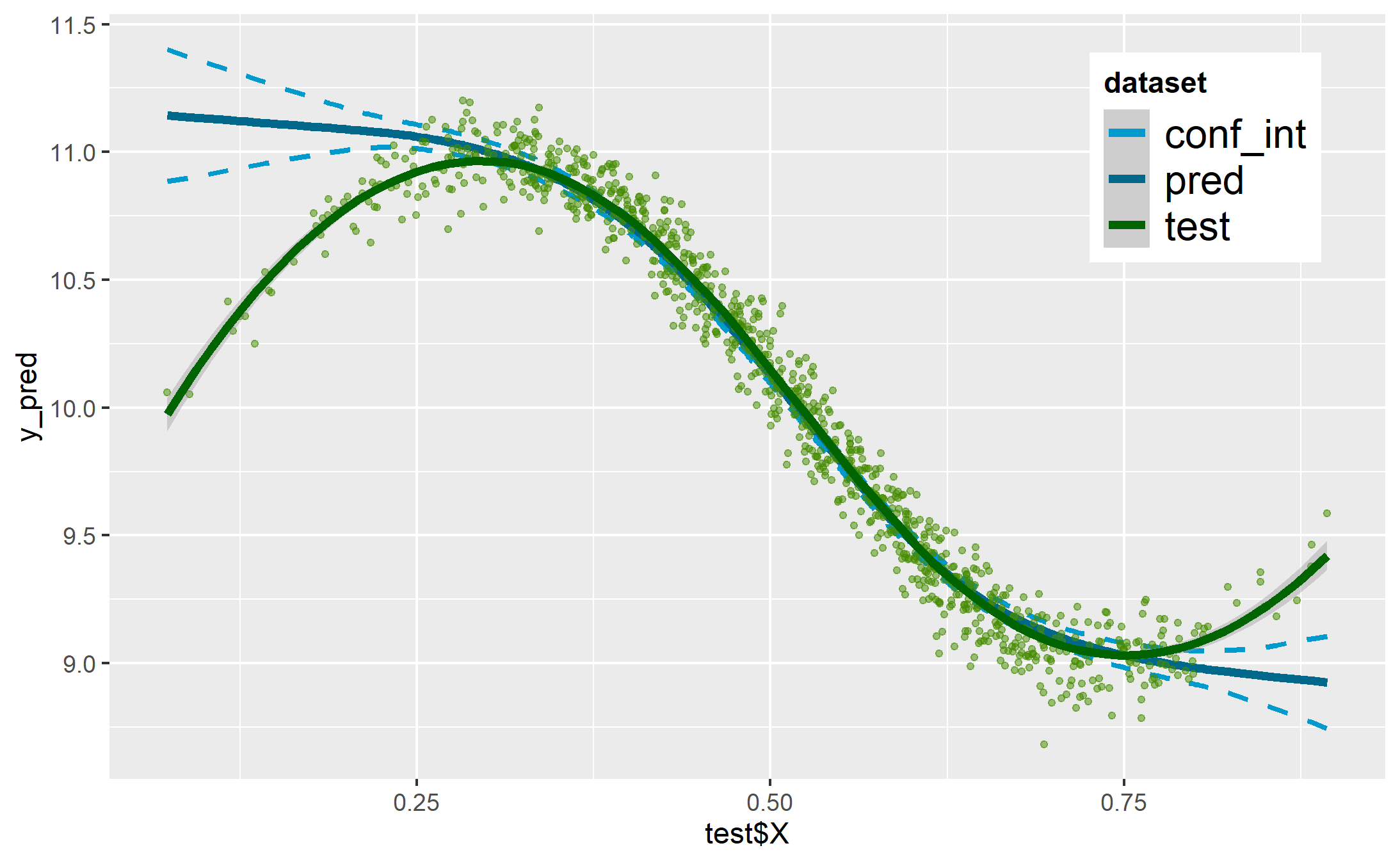}
     \caption{GN}
  \end{subfigure}
  \begin{subfigure}[b]{0.3\linewidth}
    \includegraphics[width=\linewidth]{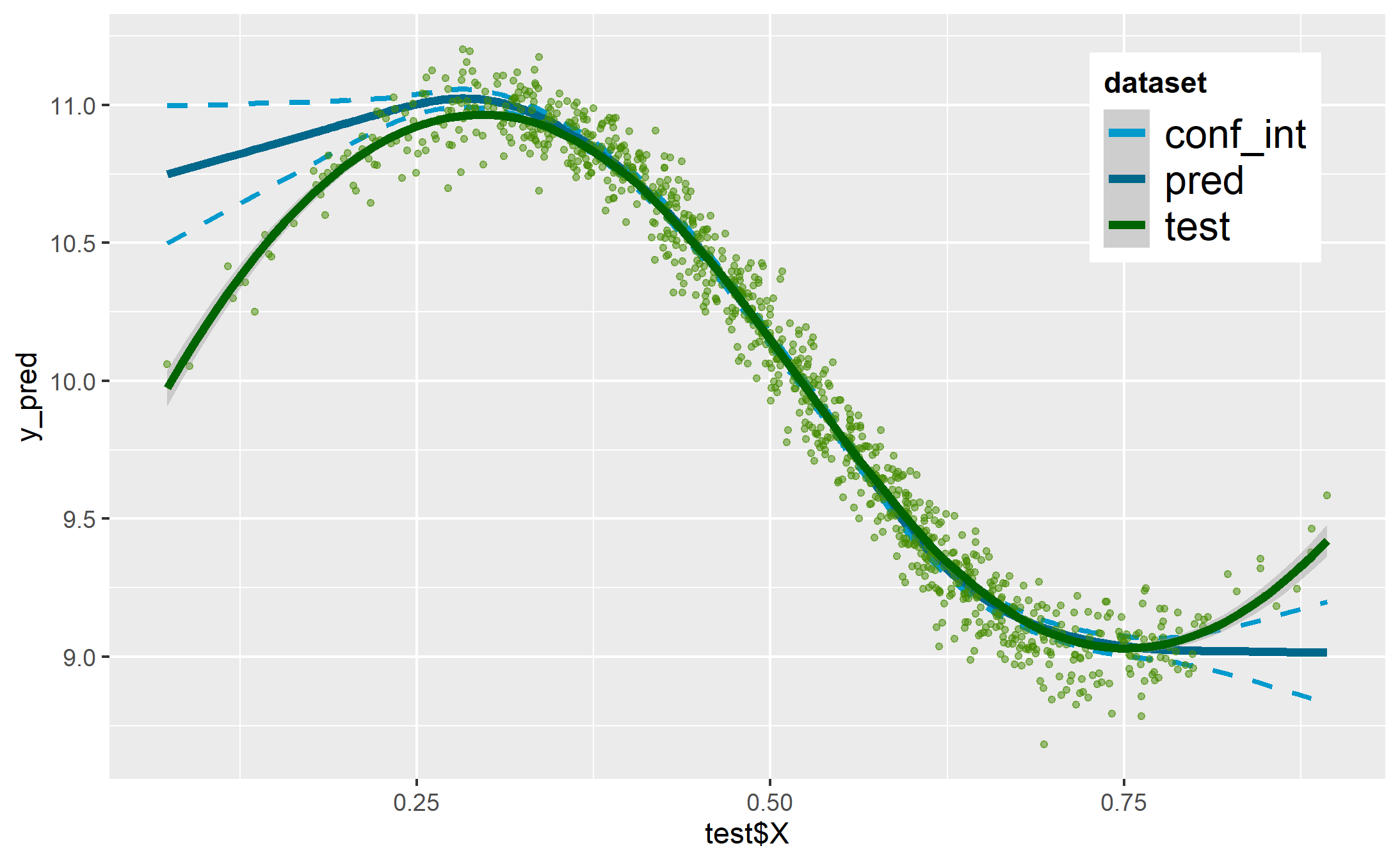}
    \caption{GN - GMM}
  \end{subfigure}
  \begin{subfigure}[b]{0.3\linewidth}
    \includegraphics[width=\linewidth]{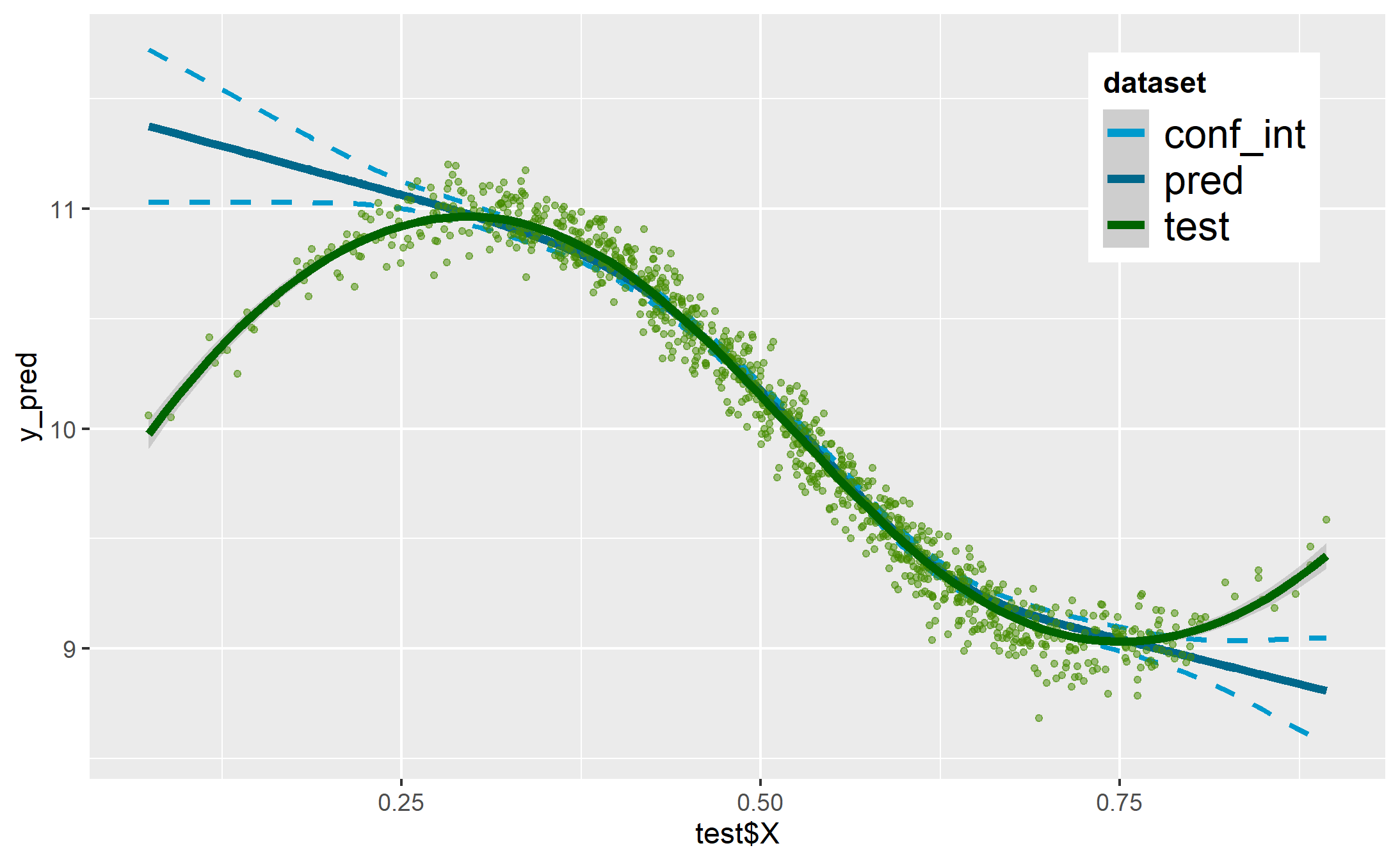}
    \caption{ROSE}
  \end{subfigure}

  \begin{subfigure}[b]{0.3\linewidth}
    \includegraphics[width=\linewidth]{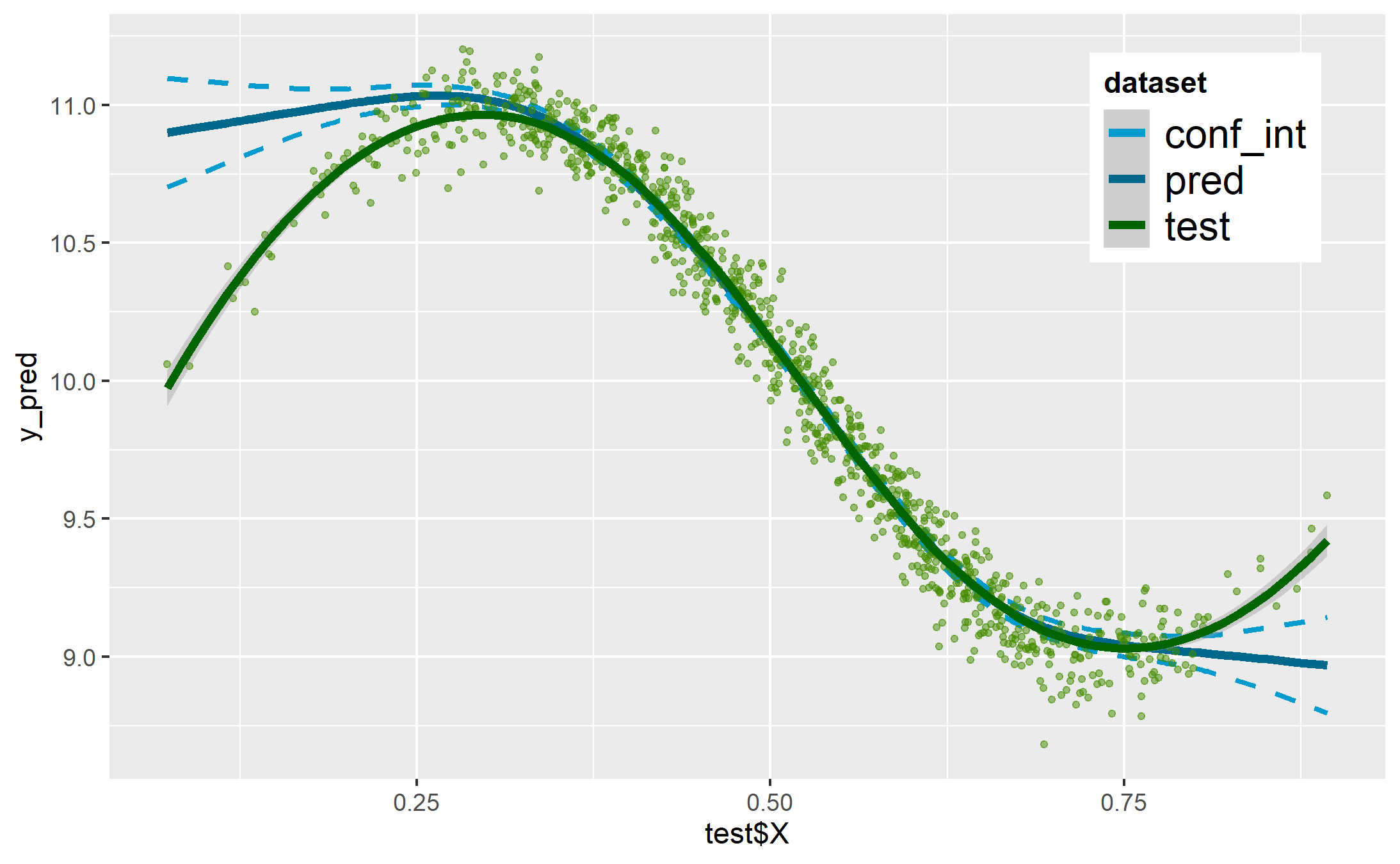}
     \caption{ROSE - GMM}
  \end{subfigure}
  \begin{subfigure}[b]{0.3\linewidth}
    \includegraphics[width=\linewidth]{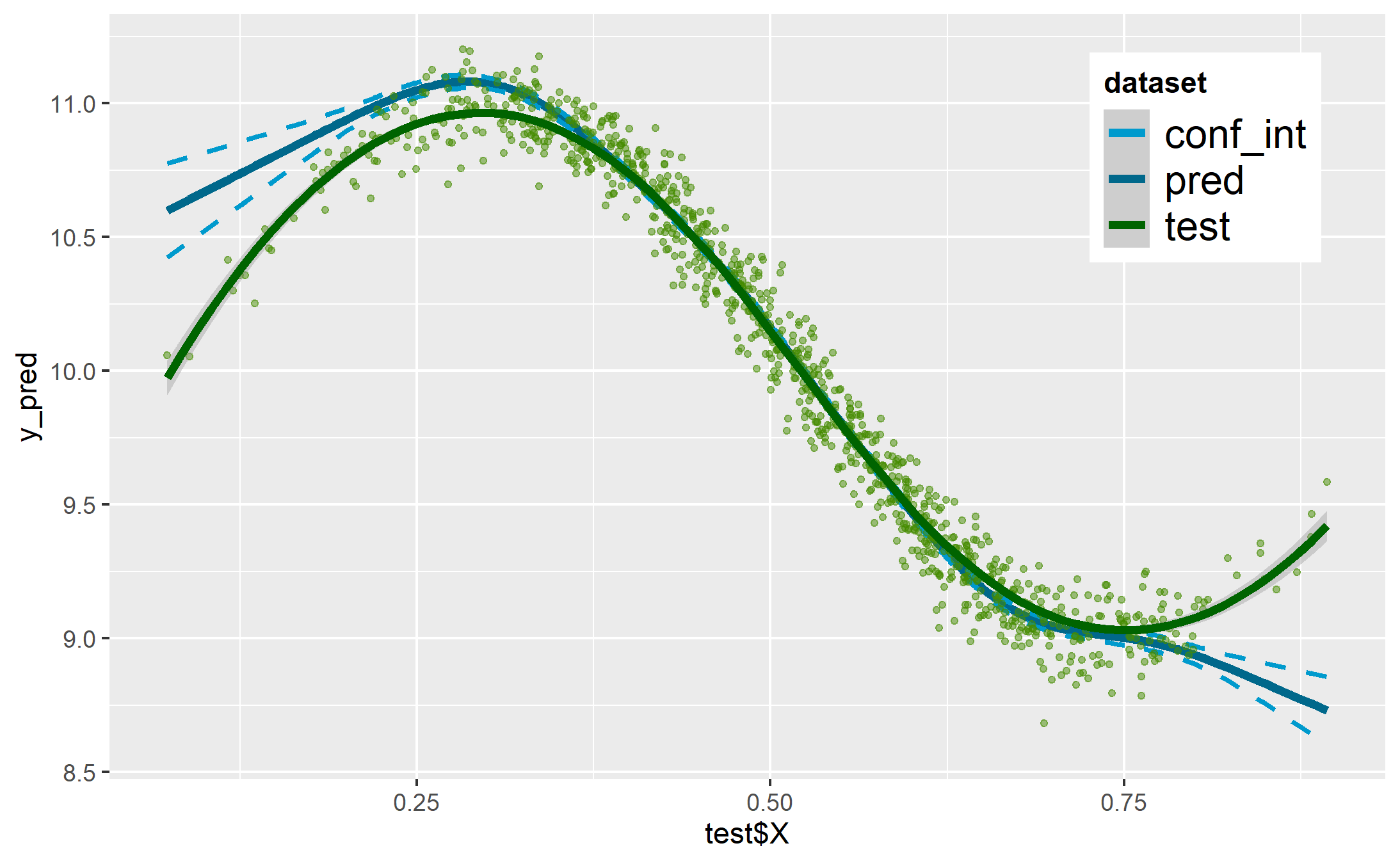}
    \caption{KDE - GMM}
  \end{subfigure}
  \begin{subfigure}[b]{0.3\linewidth}
    \includegraphics[width=\linewidth]{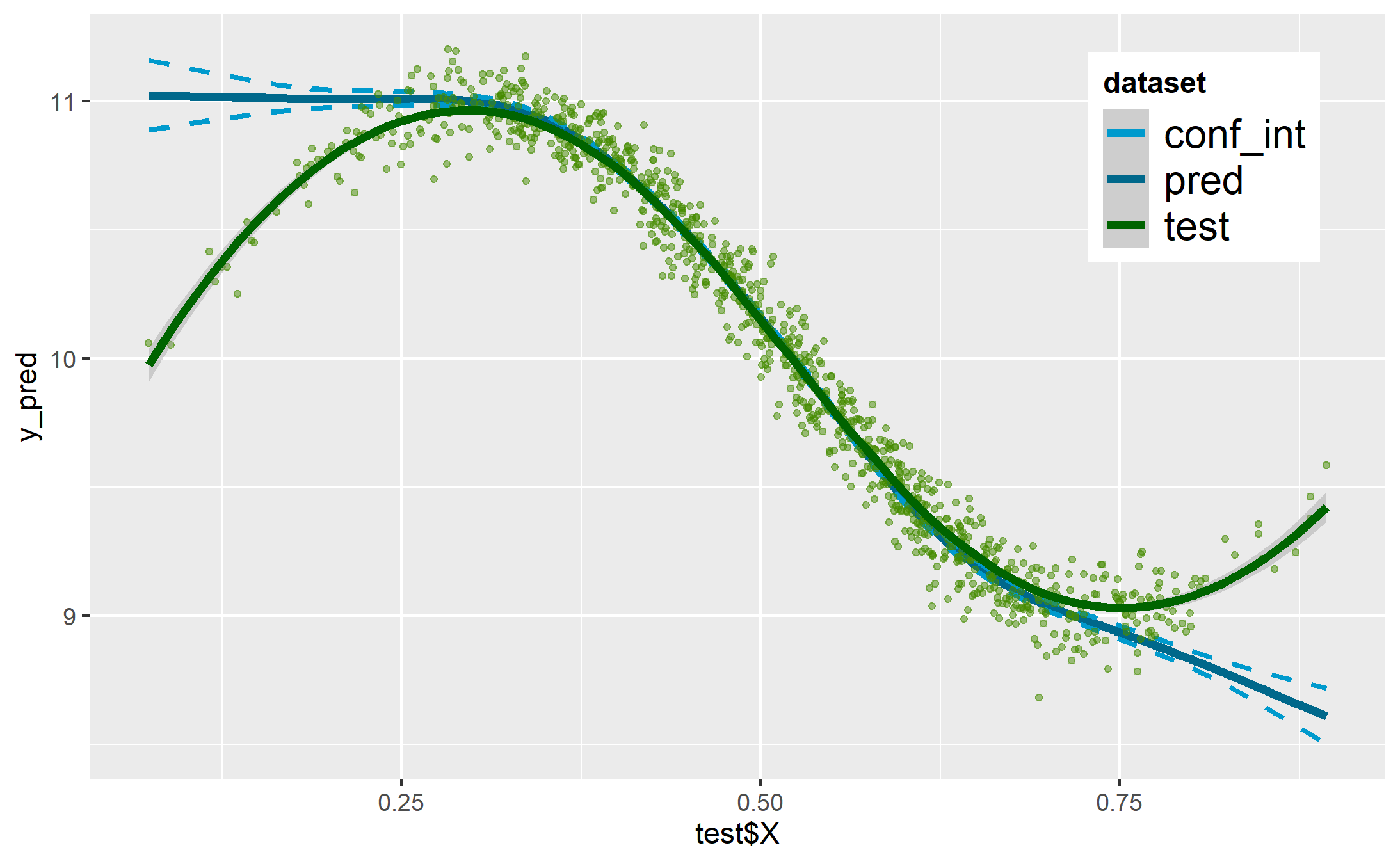}
    \caption{GMM}
  \end{subfigure}

  \begin{subfigure}[b]{0.3\linewidth}
    \includegraphics[width=\linewidth]{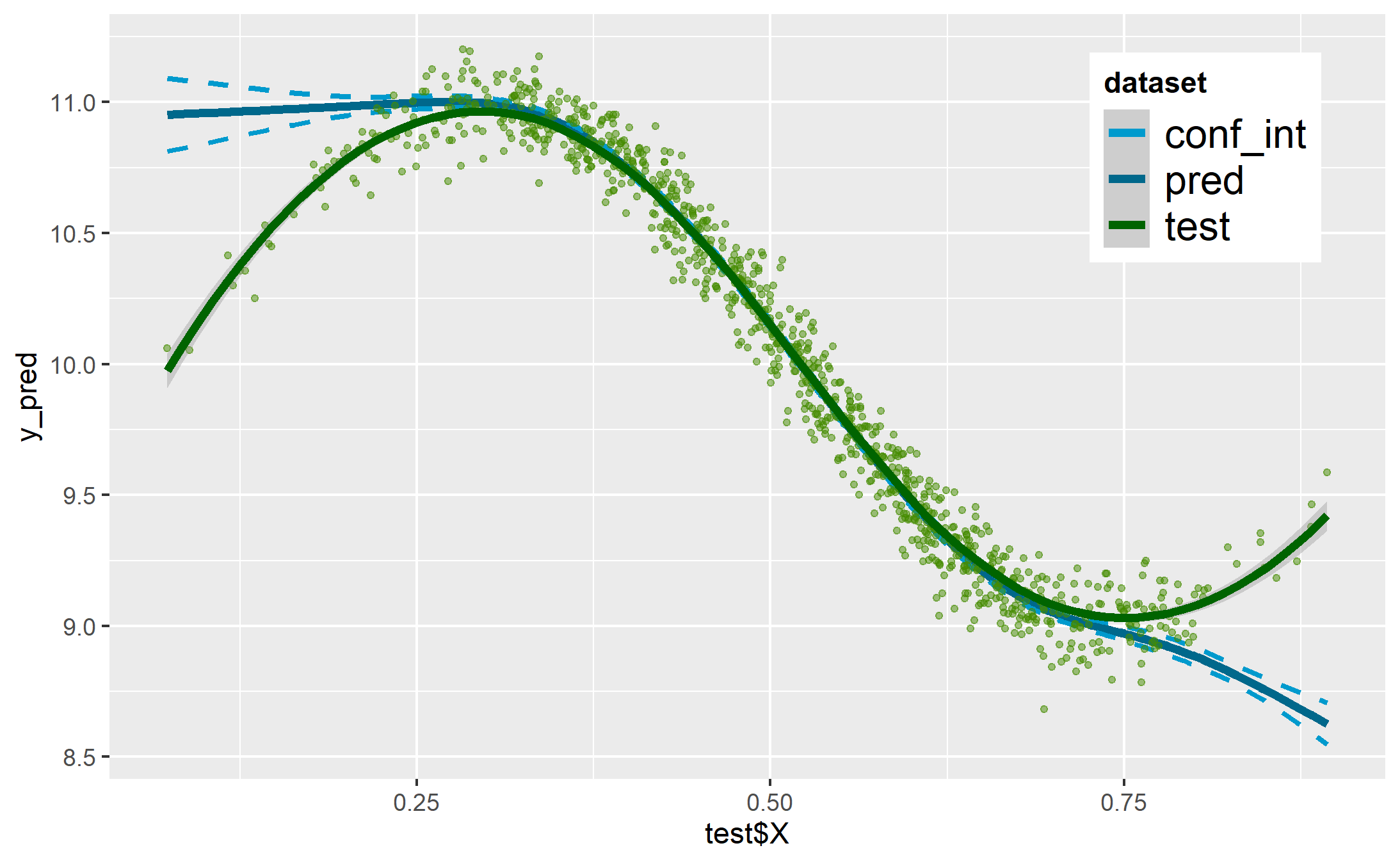}
     \caption{FA - GMM}
  \end{subfigure}
  \begin{subfigure}[b]{0.3\linewidth}
    \includegraphics[width=\linewidth]{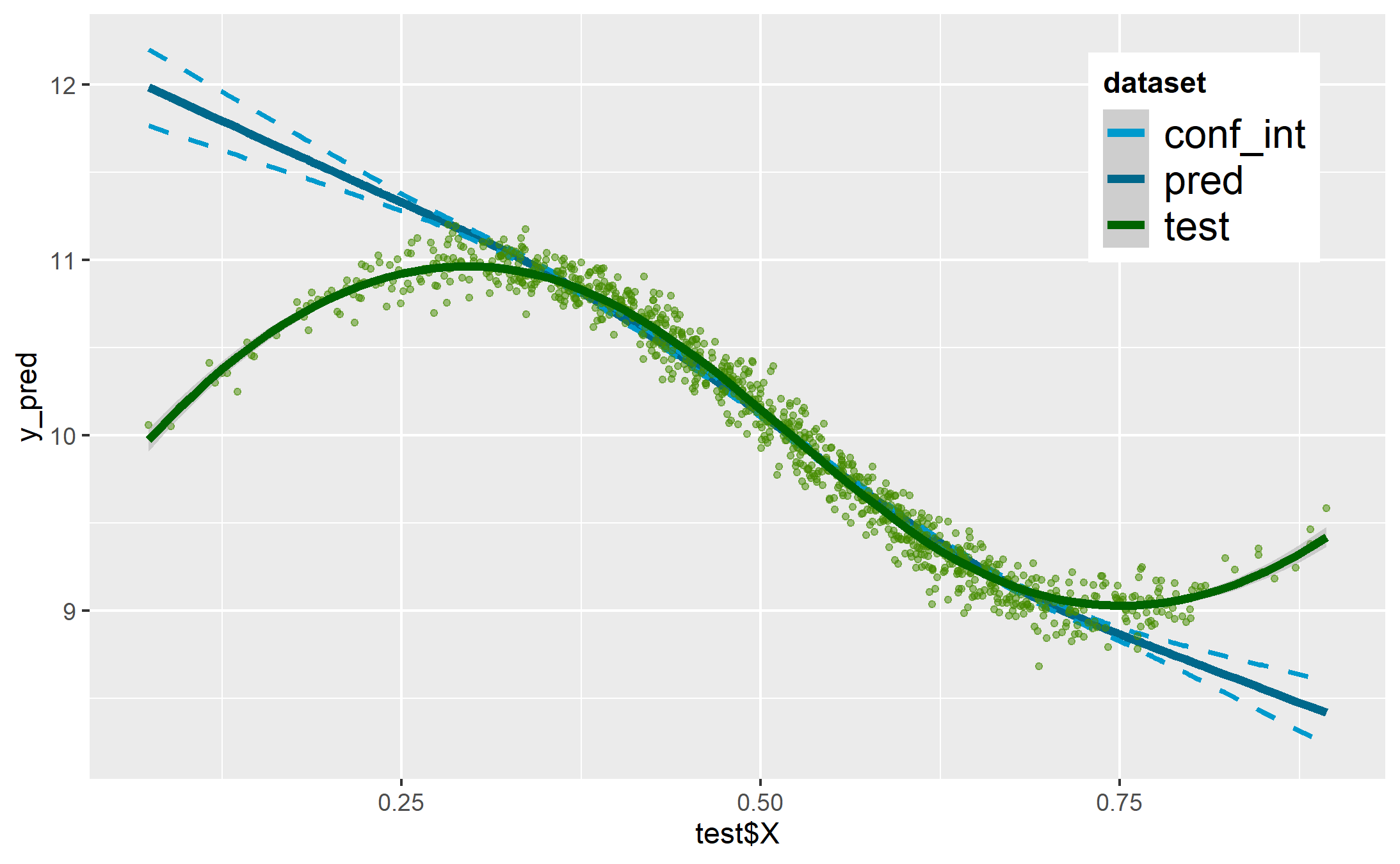}
    \caption{Copula}
  \end{subfigure}
  \begin{subfigure}[b]{0.3\linewidth}
    \includegraphics[width=\linewidth]{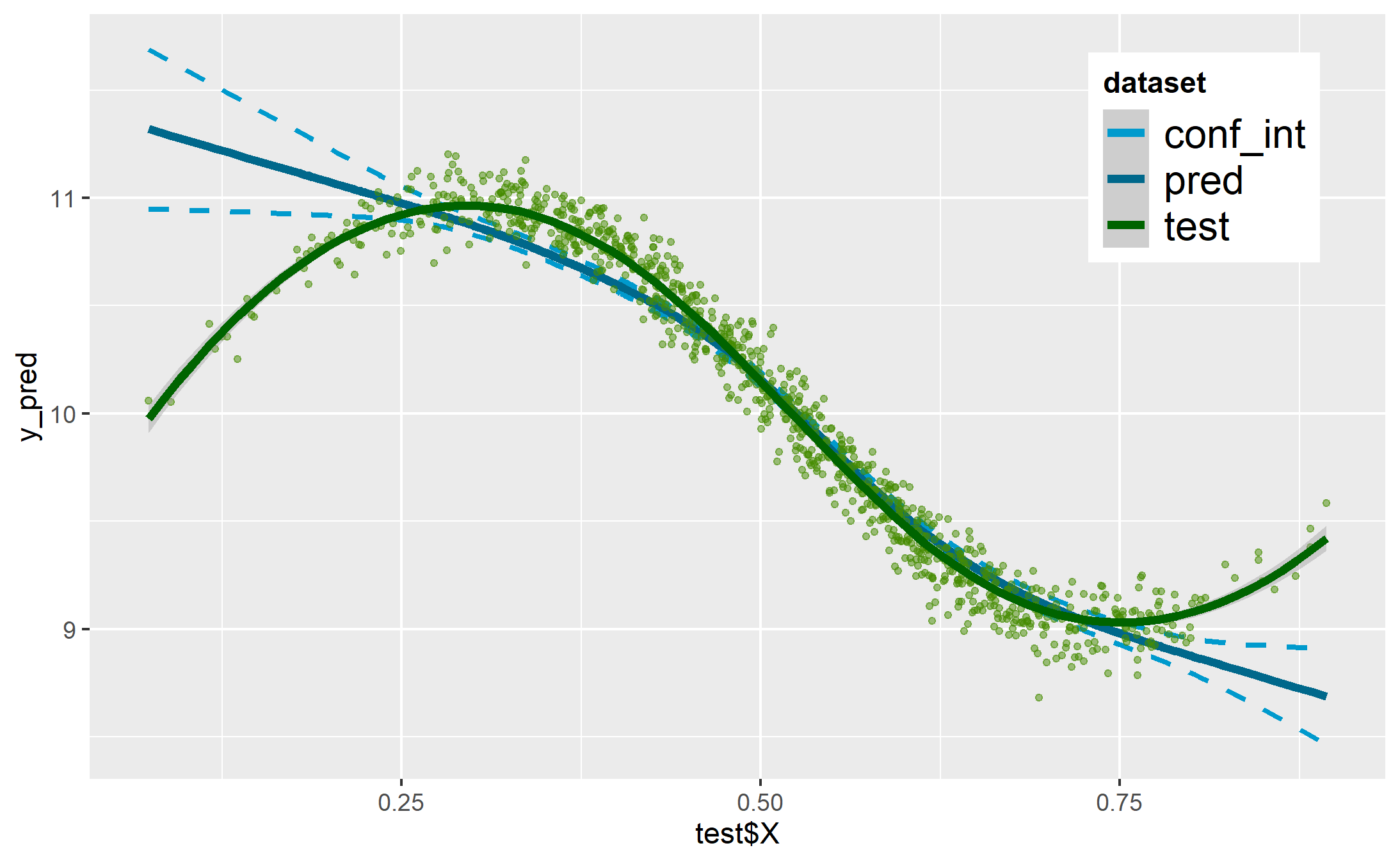}
    \caption{GAN}
  \end{subfigure}

  \begin{subfigure}[b]{0.3\linewidth}
    \includegraphics[width=\linewidth]{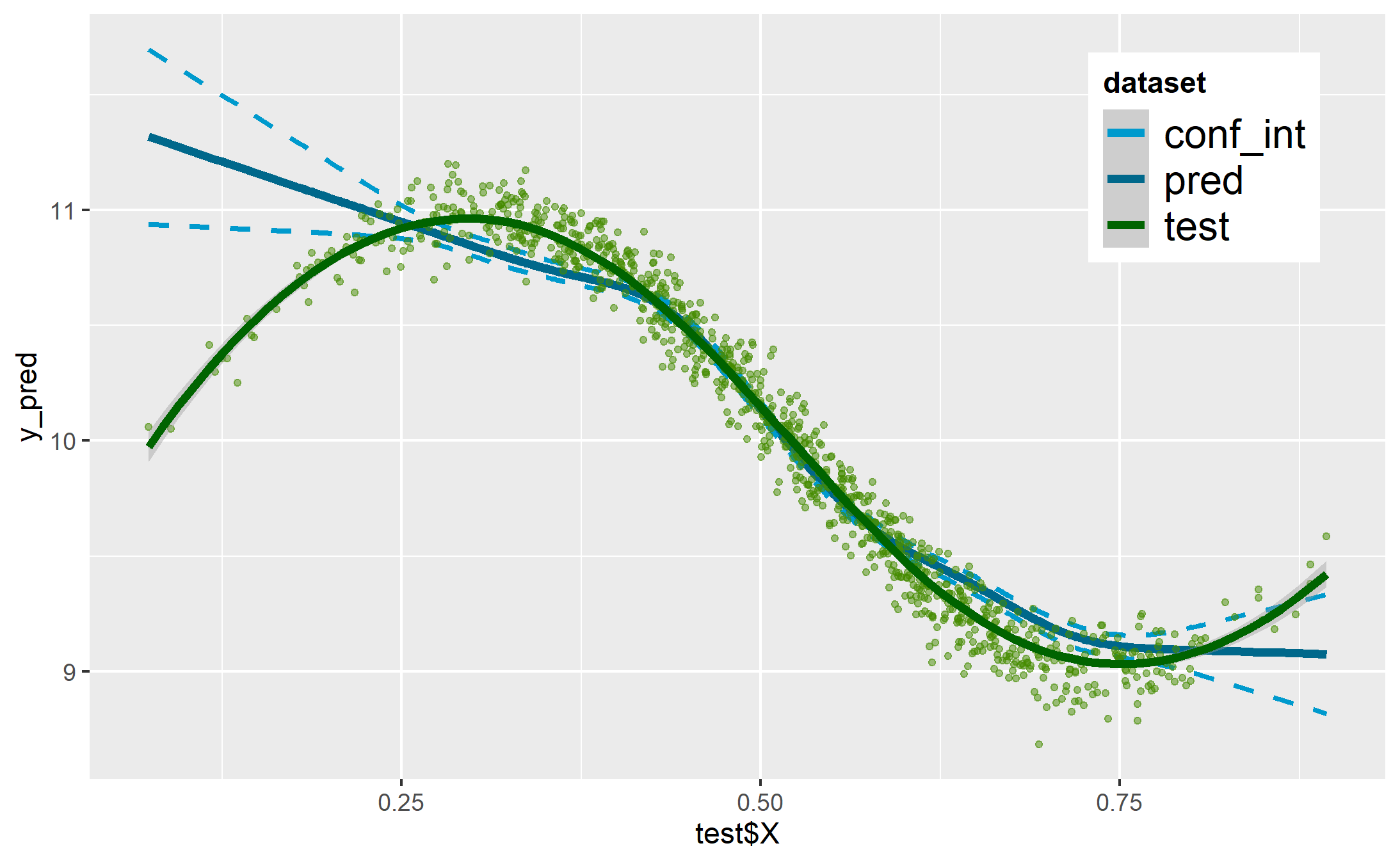}
     \caption{RF}
  \end{subfigure}
  \begin{subfigure}[b]{0.3\linewidth}
    \includegraphics[width=\linewidth]{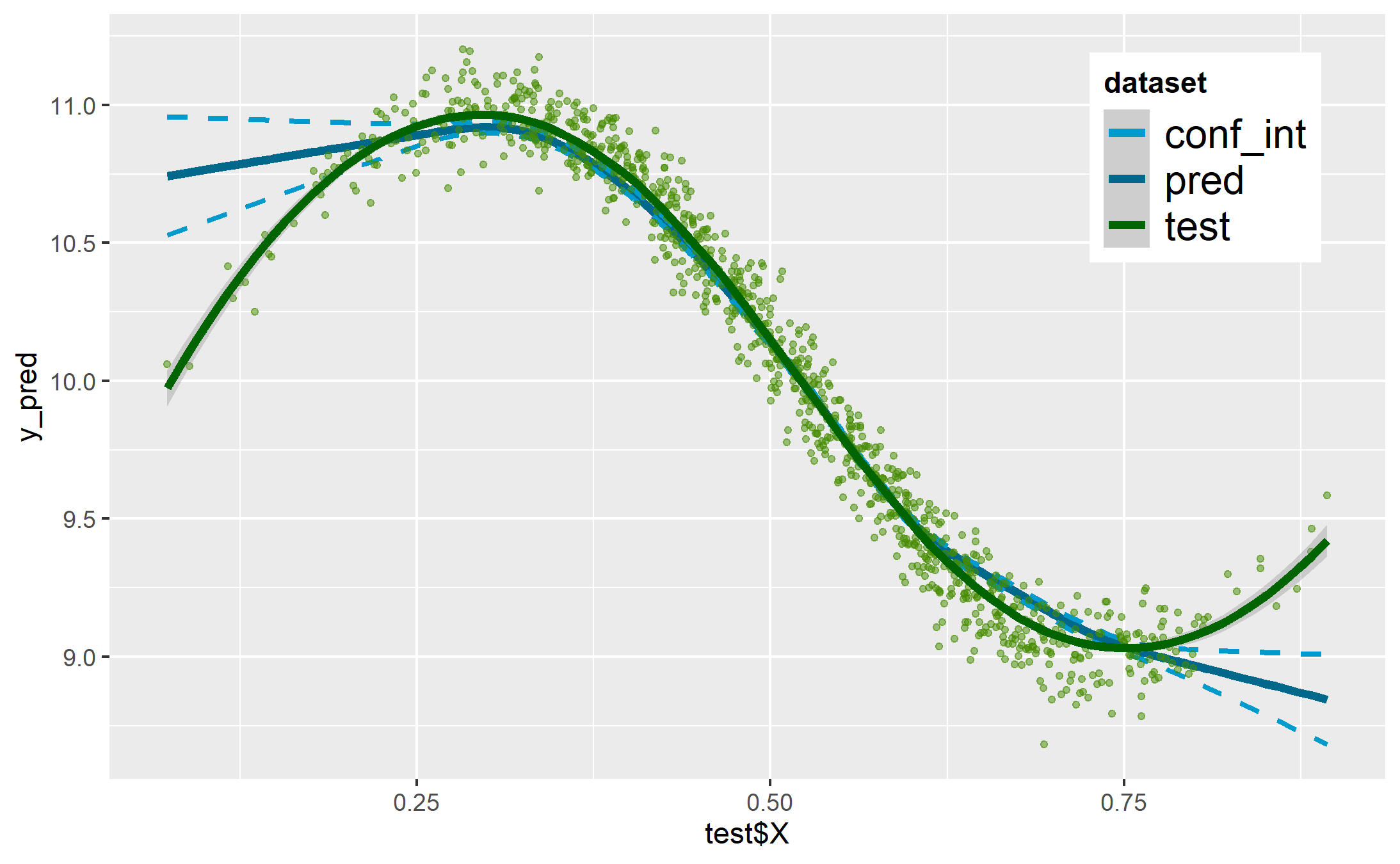}
    \caption{SMOTE}
  \end{subfigure}
  \begin{subfigure}[b]{0.3\linewidth}
    \includegraphics[width=\linewidth]{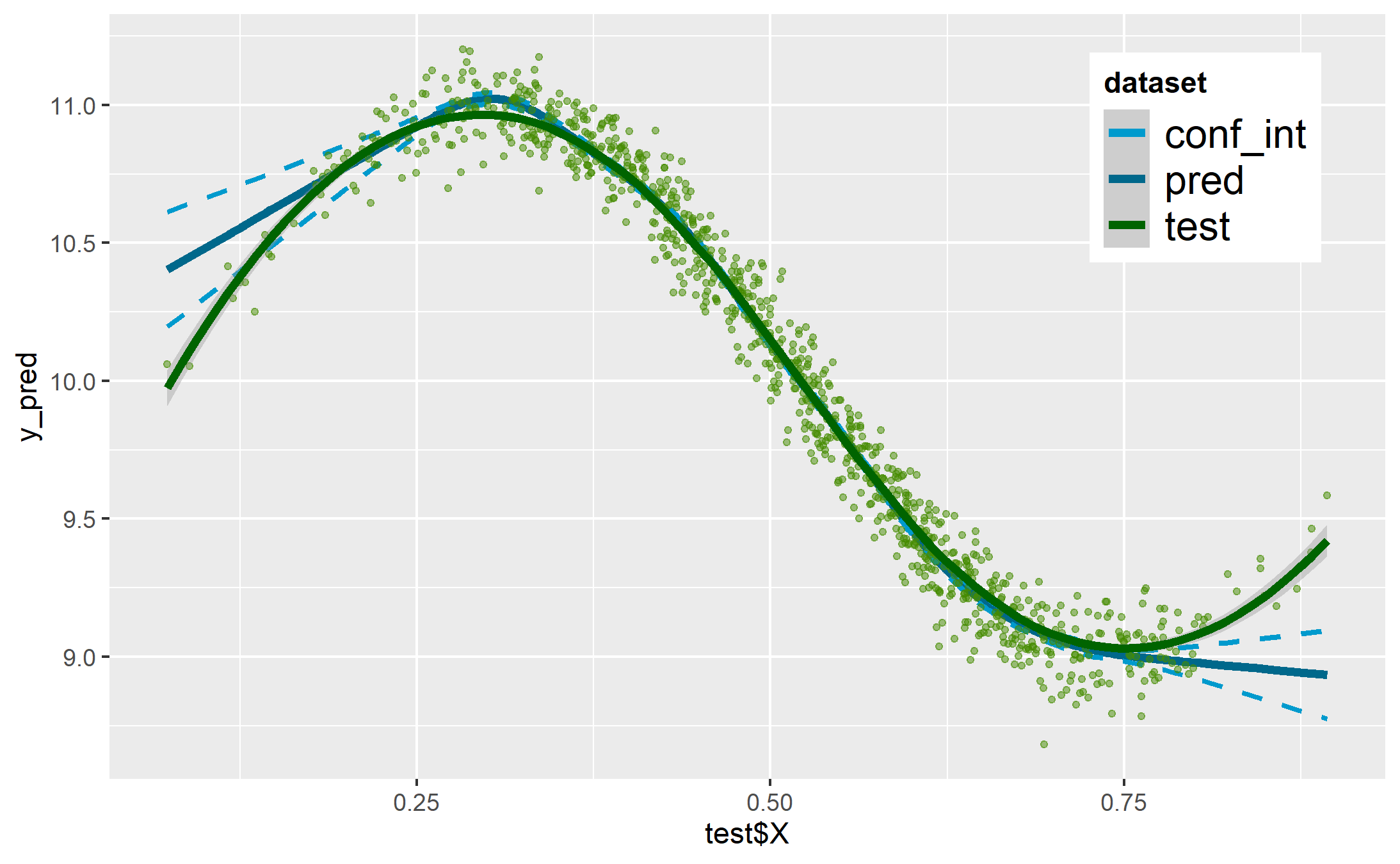}
    \caption{SMOTE - GMM}
  \end{subfigure}

  \caption{Smoothed predictions with GAM}
  \label{pred_Y_GAM_ech_XXX-vs-test}
\end{figure}

\newpage
\textbf{Random Forest predictions}

\begin{figure}[H]
  \centering

  \begin{subfigure}[b]{0.3\linewidth}
    \includegraphics[width=\linewidth]{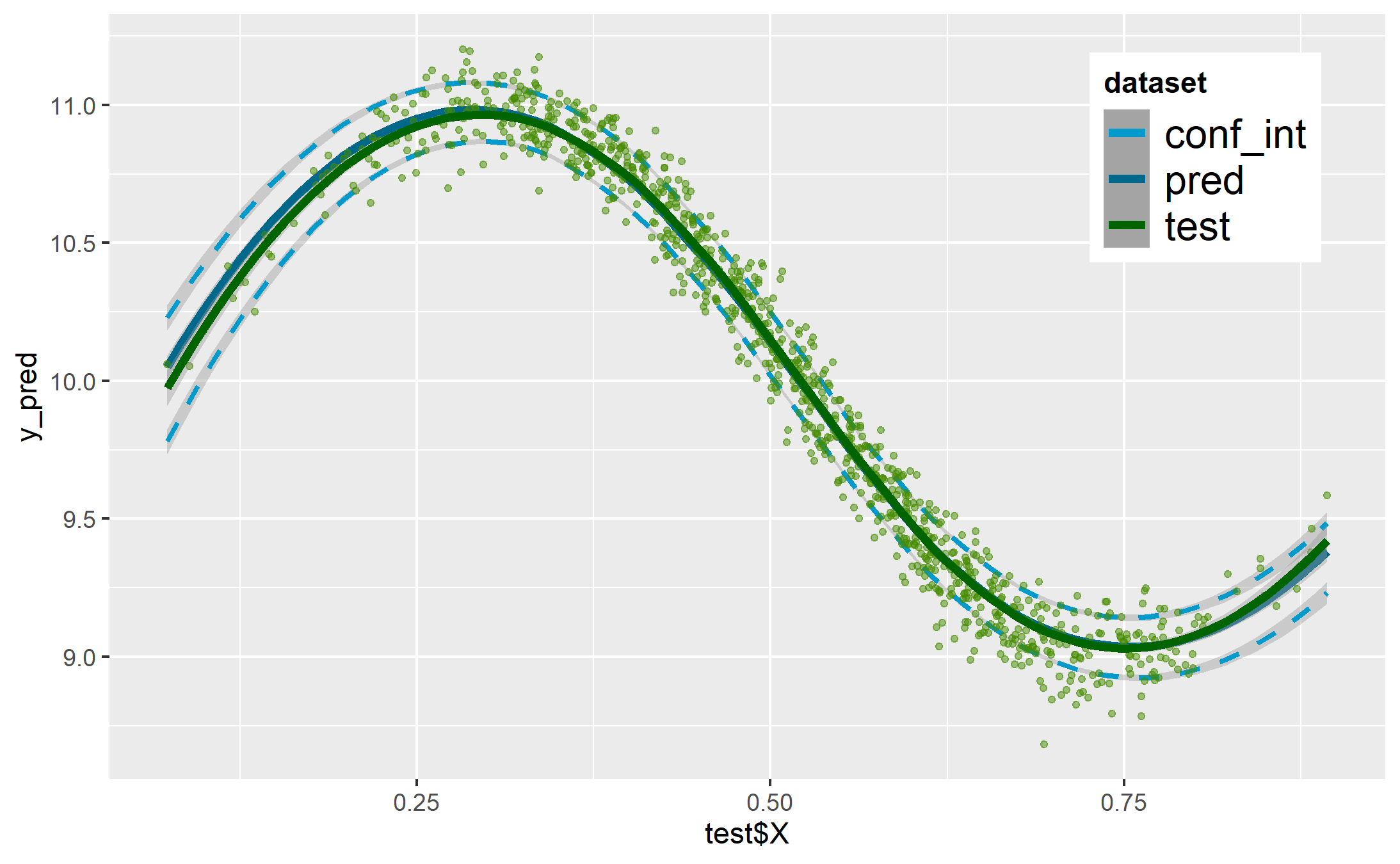}
     \caption{$\mathcal{D}^b$}
  \end{subfigure}
  \begin{subfigure}[b]{0.3\linewidth}
    \includegraphics[width=\linewidth]{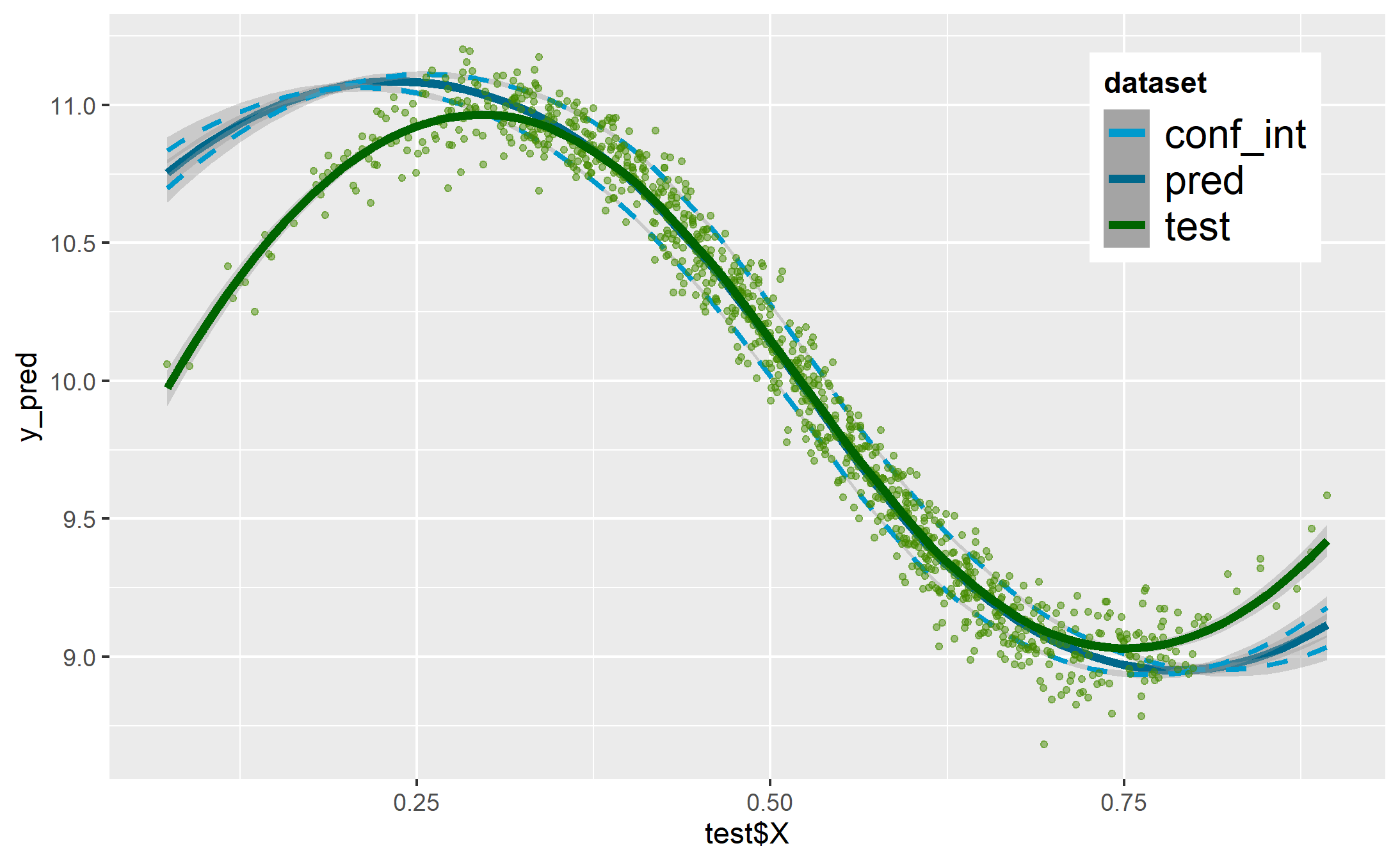}
    \caption{$\mathcal{D}^i$}
  \end{subfigure}
  \begin{subfigure}[b]{0.3\linewidth}
    \includegraphics[width=\linewidth]{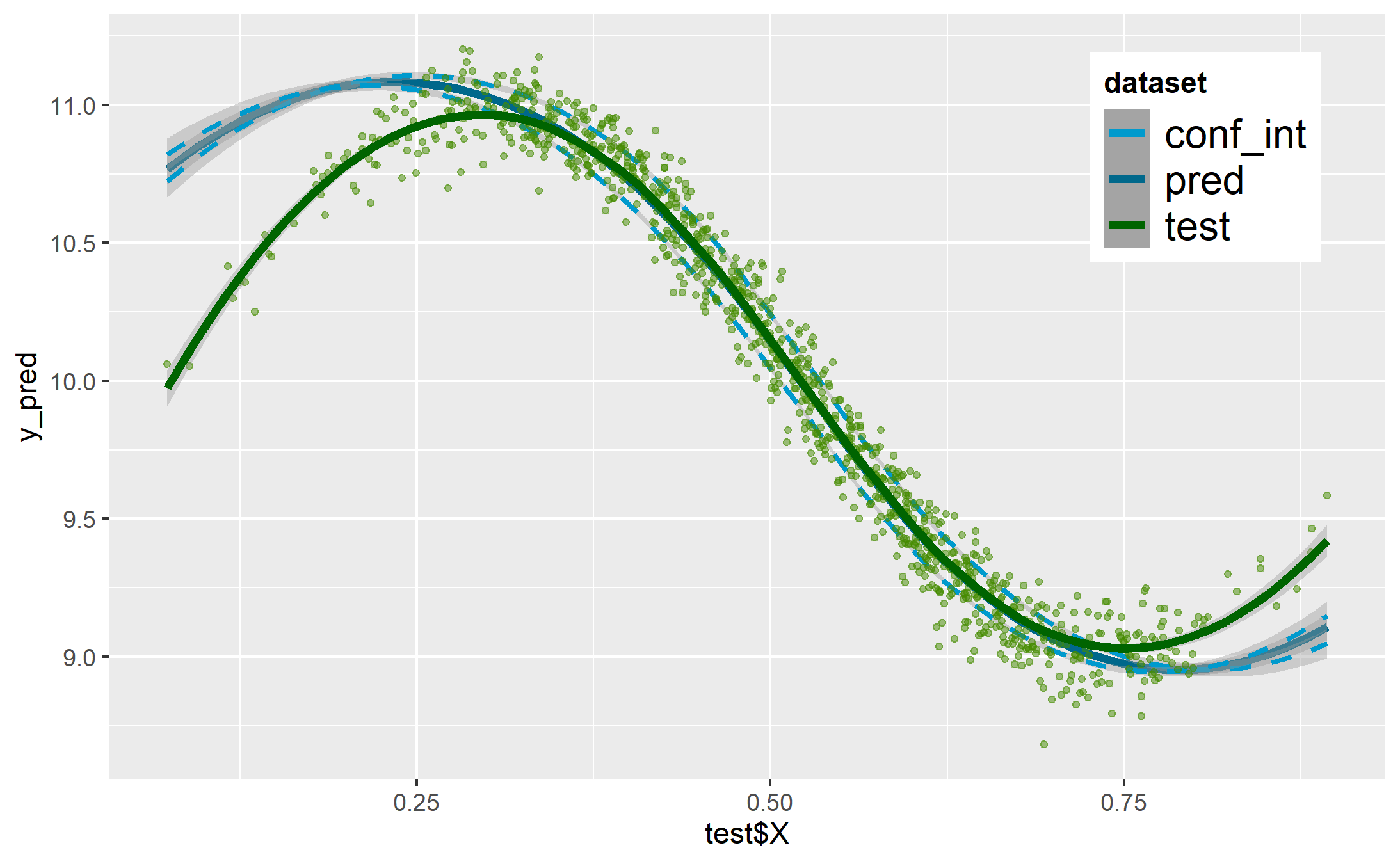}
    \caption{WR}
  \end{subfigure}

  \begin{subfigure}[b]{0.3\linewidth}
    \includegraphics[width=\linewidth]{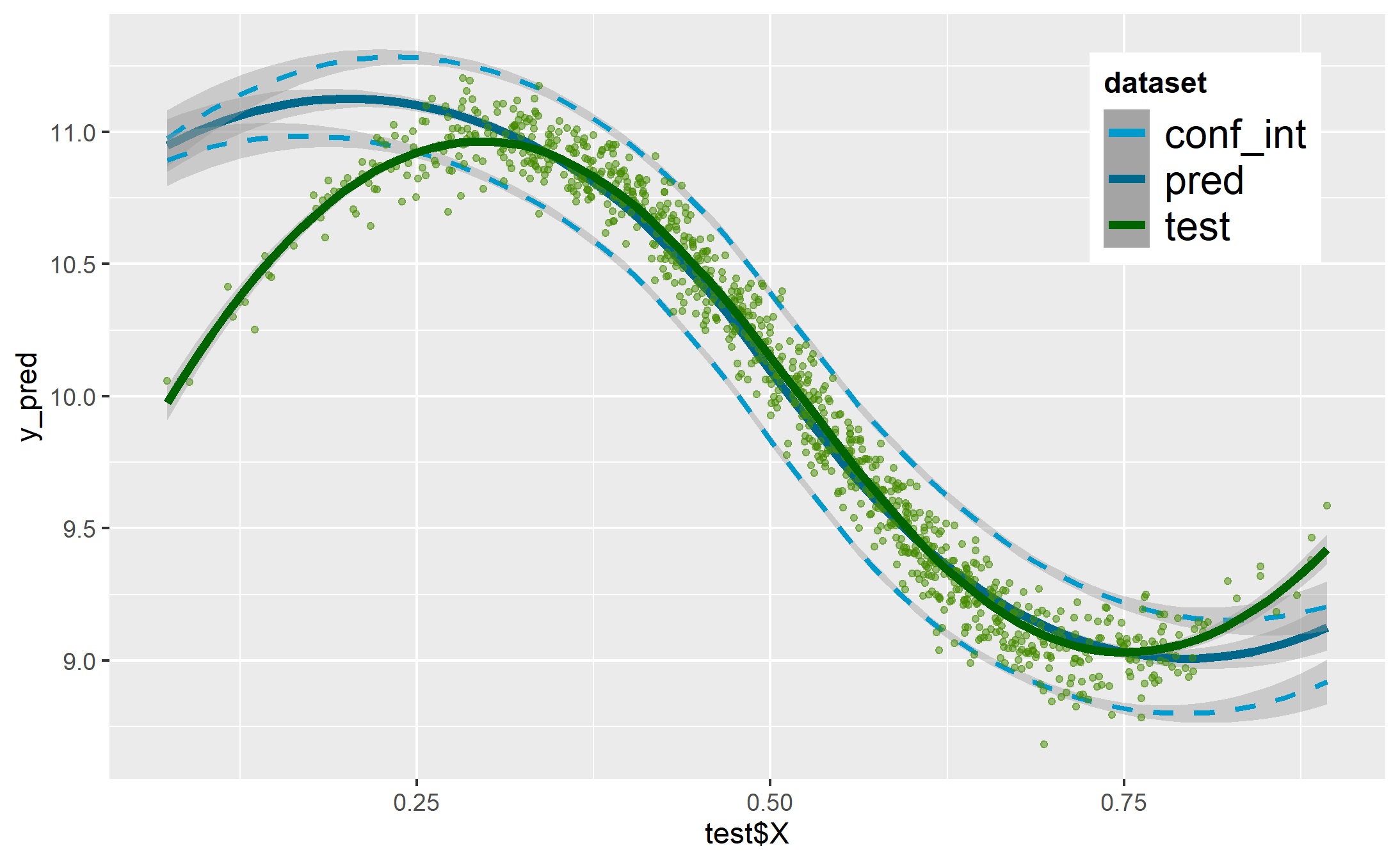}
     \caption{GN}
  \end{subfigure}
  \begin{subfigure}[b]{0.3\linewidth}
    \includegraphics[width=\linewidth]{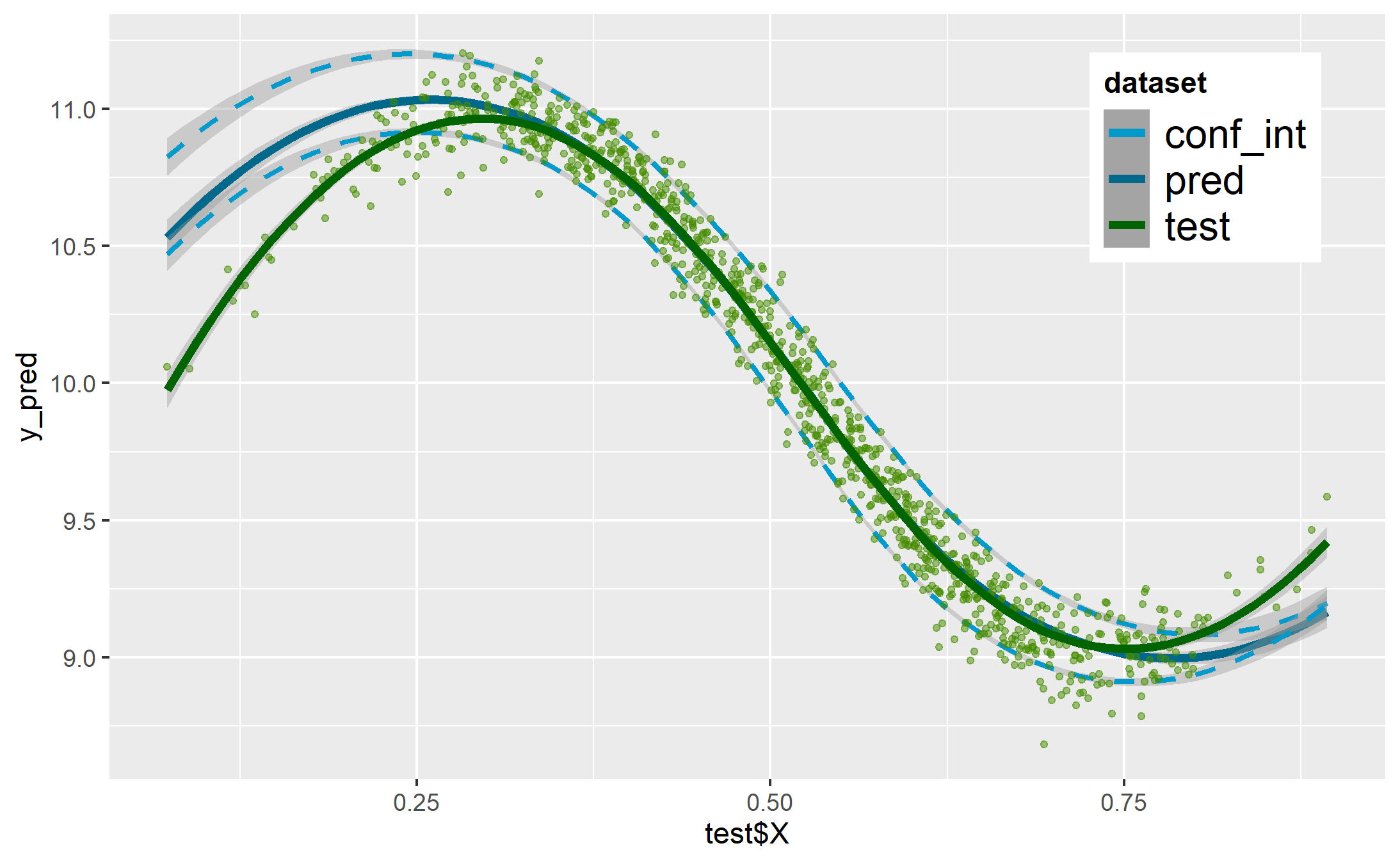}
    \caption{GN - GMM}
  \end{subfigure}
  \begin{subfigure}[b]{0.3\linewidth}
    \includegraphics[width=\linewidth]{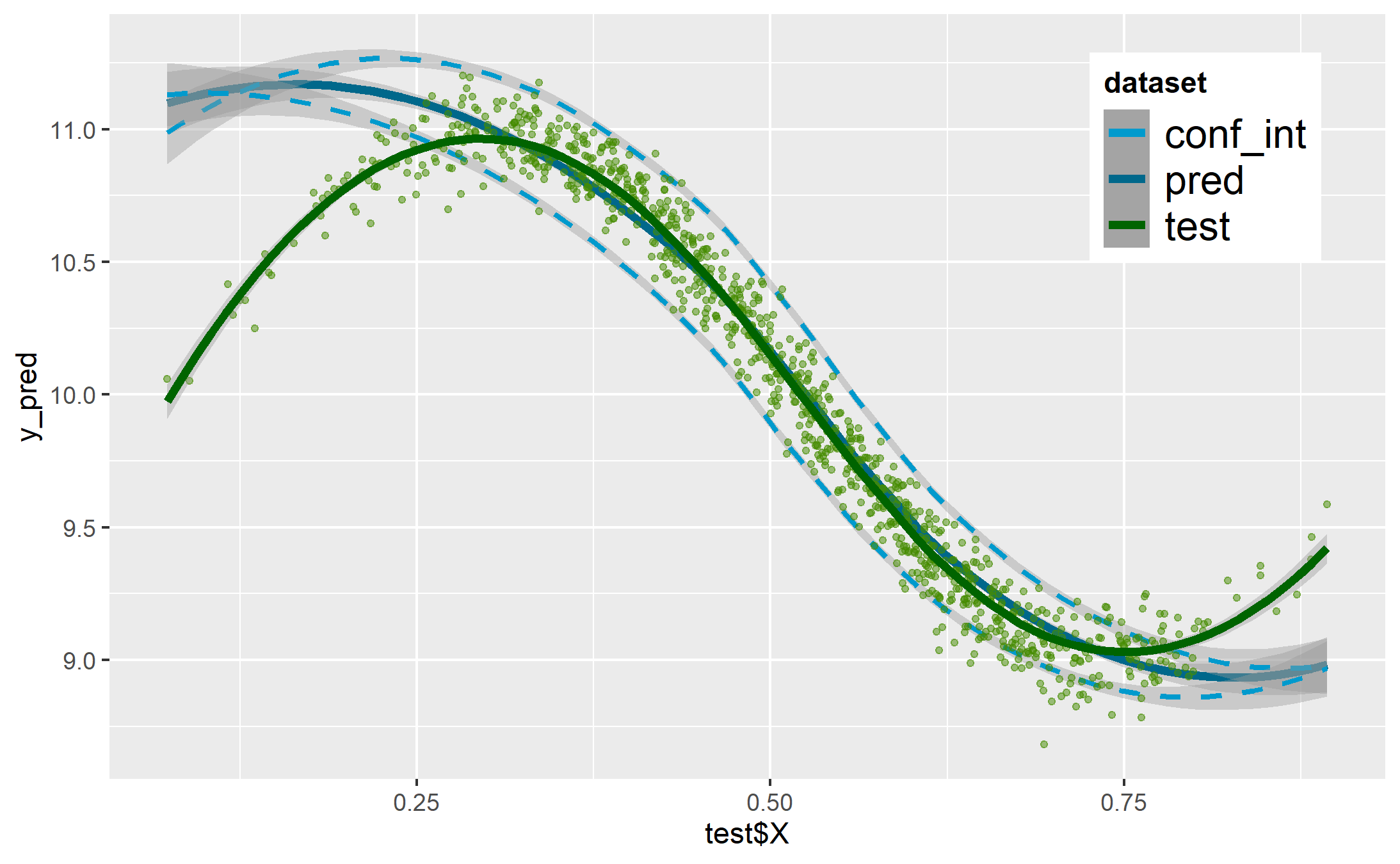}
    \caption{ROSE}
  \end{subfigure}

  \begin{subfigure}[b]{0.3\linewidth}
    \includegraphics[width=\linewidth]{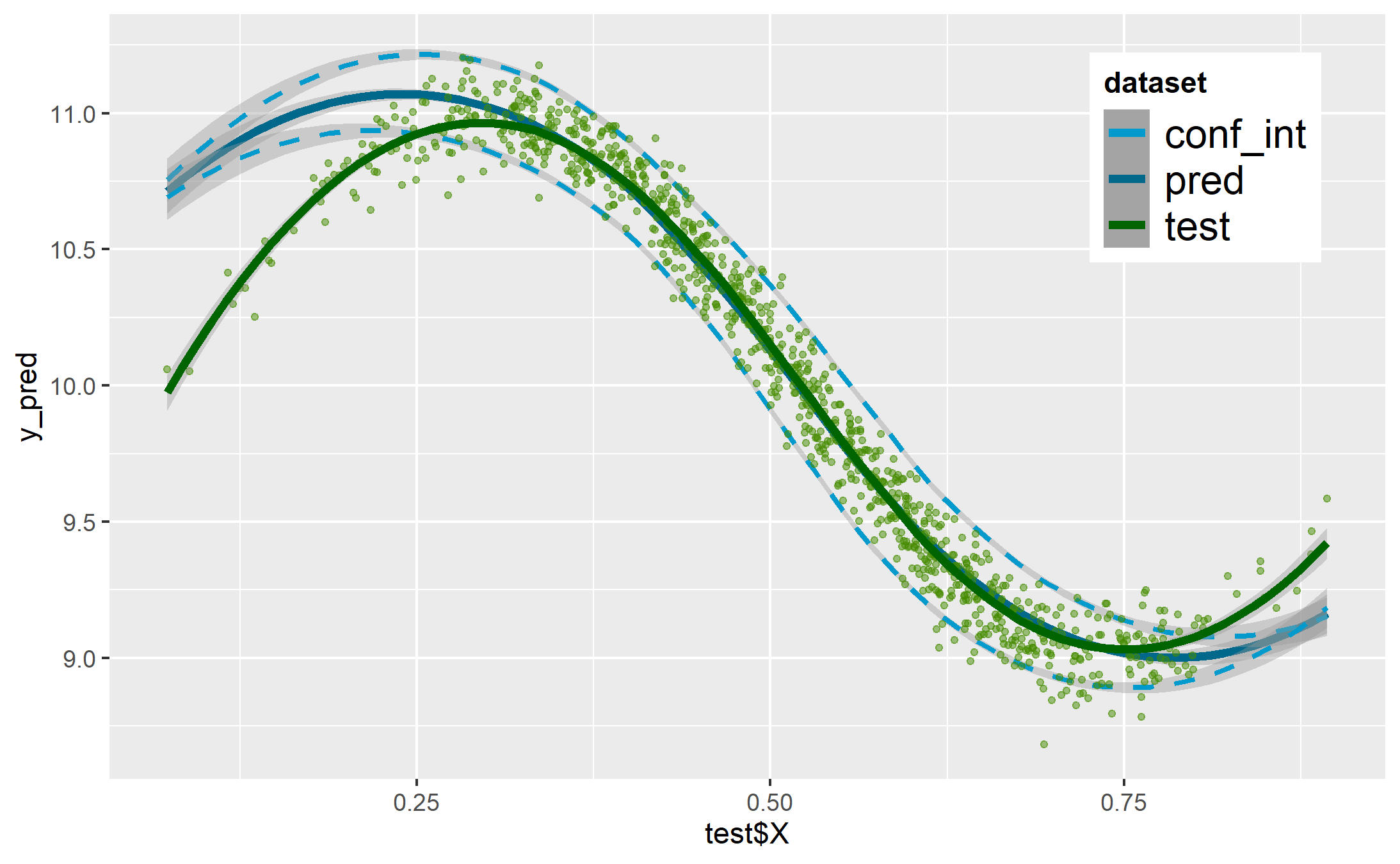}
     \caption{ROSE - GMM}
  \end{subfigure}
  \begin{subfigure}[b]{0.3\linewidth}
    \includegraphics[width=\linewidth]{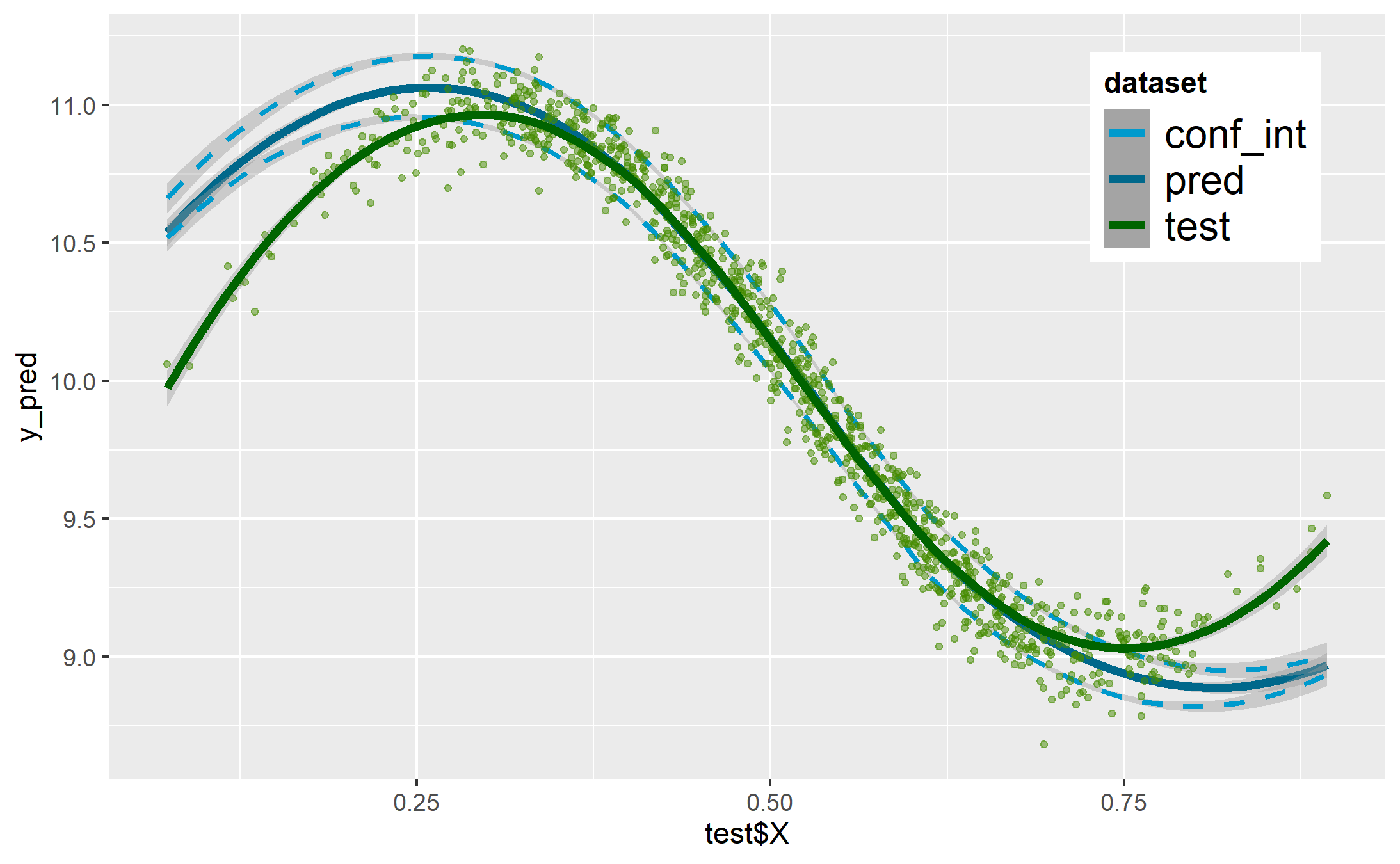}
    \caption{KDE - GMM}
  \end{subfigure}
  \begin{subfigure}[b]{0.3\linewidth}
    \includegraphics[width=\linewidth]{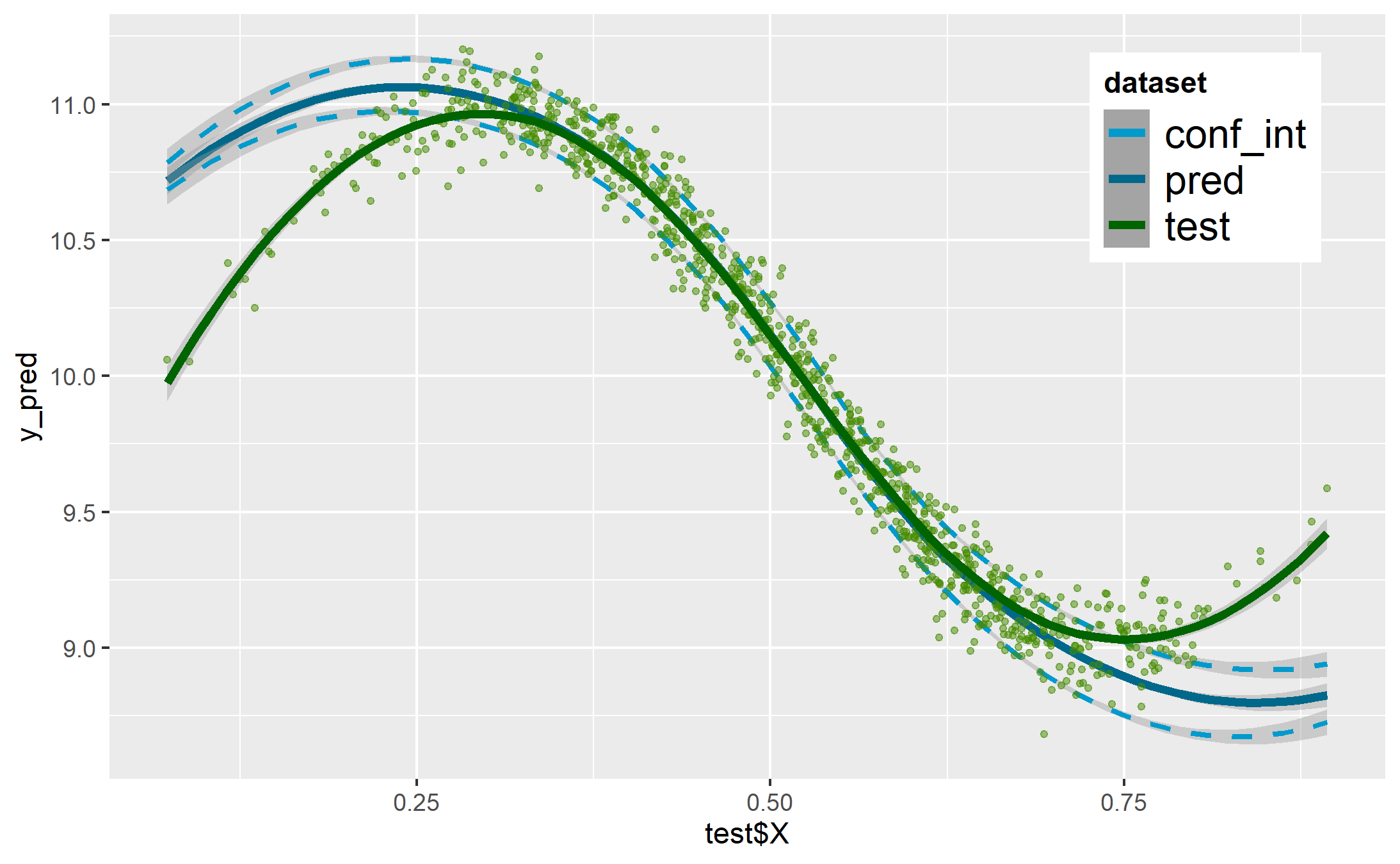}
    \caption{GMM}
  \end{subfigure}

  \begin{subfigure}[b]{0.3\linewidth}
    \includegraphics[width=\linewidth]{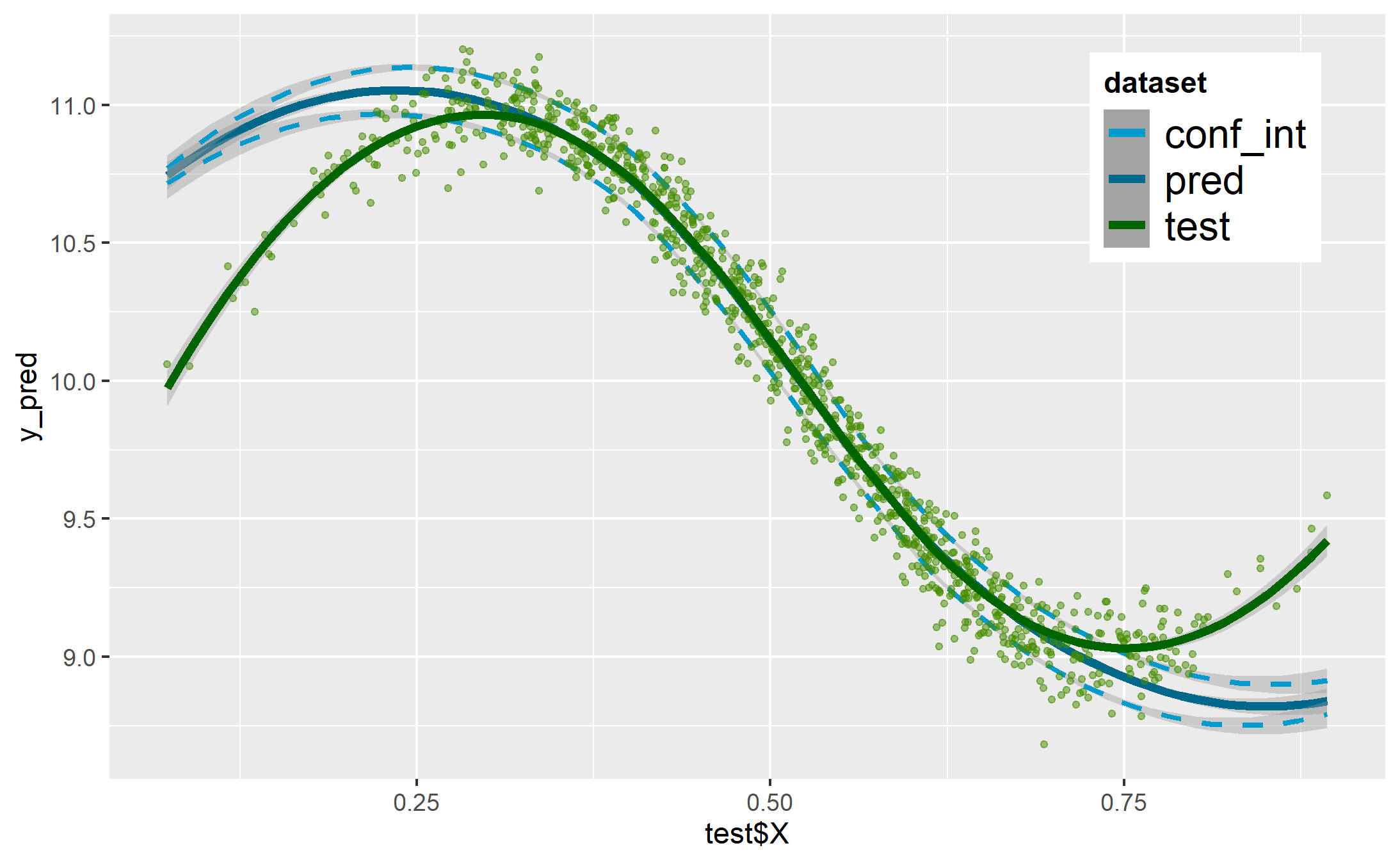}
     \caption{FA - GMM}
  \end{subfigure}
  \begin{subfigure}[b]{0.3\linewidth}
    \includegraphics[width=\linewidth]{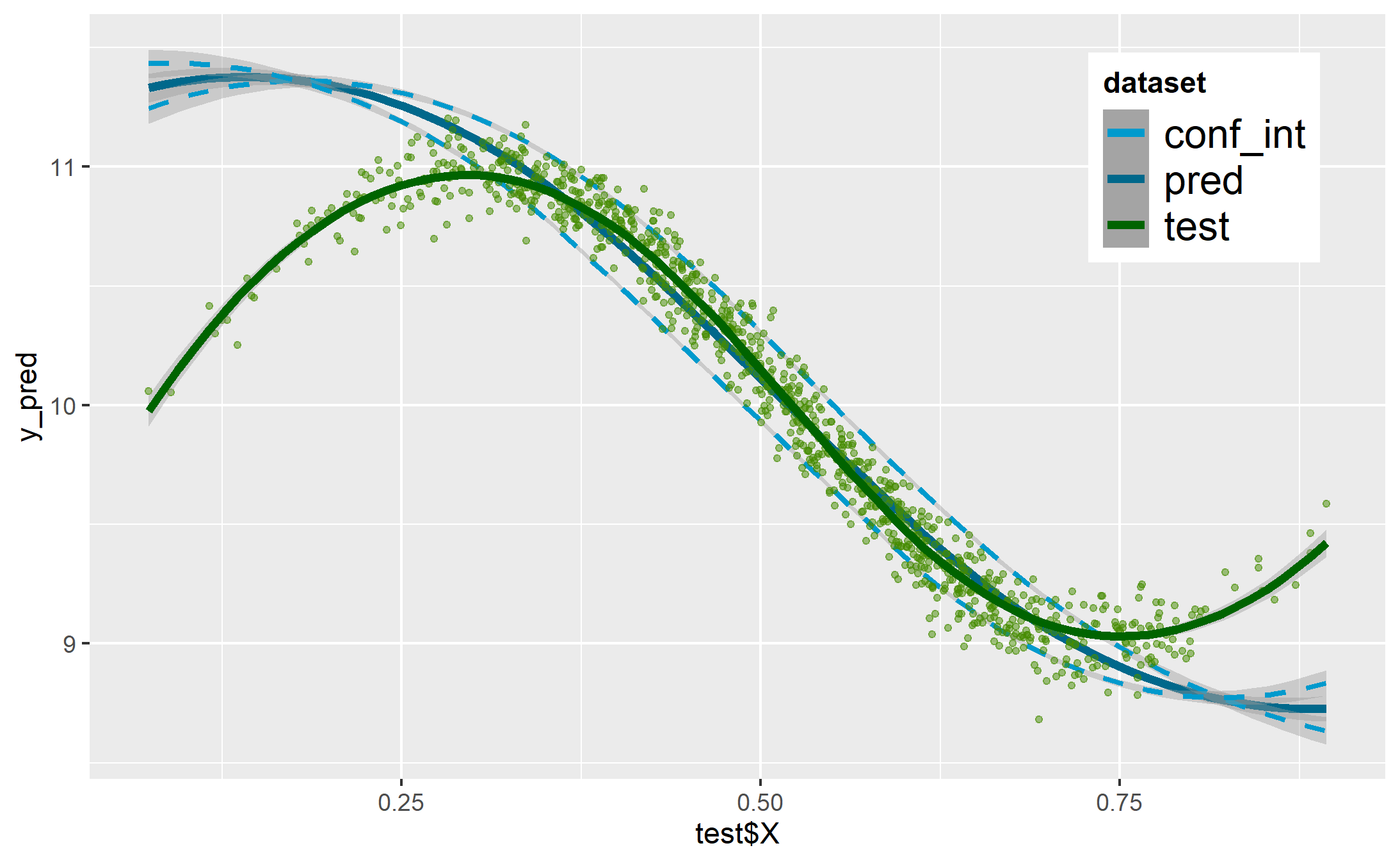}
    \caption{Copula}
  \end{subfigure}
  \begin{subfigure}[b]{0.3\linewidth}
    \includegraphics[width=\linewidth]{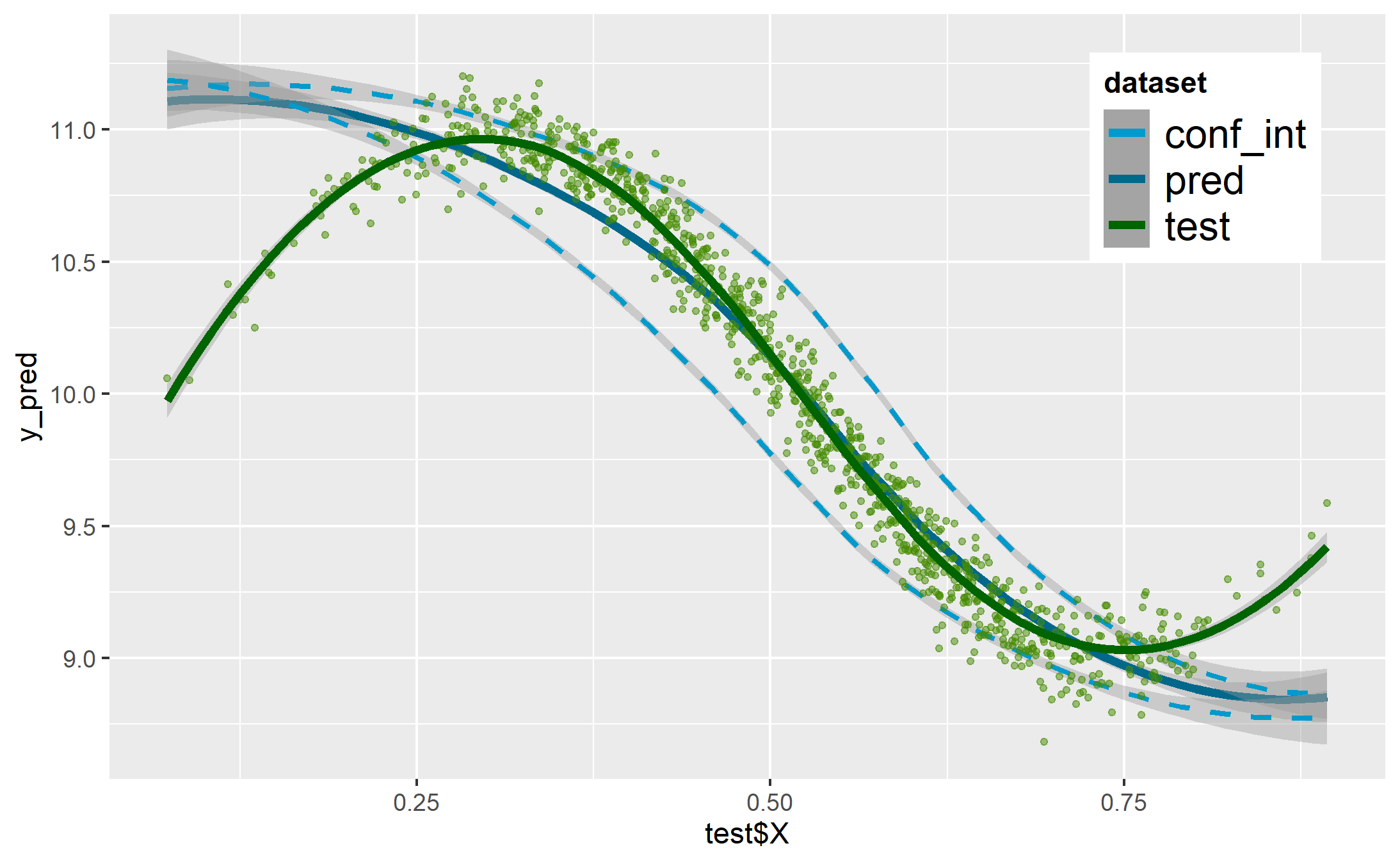}
    \caption{GAN}
  \end{subfigure}

  \begin{subfigure}[b]{0.3\linewidth}
    \includegraphics[width=\linewidth]{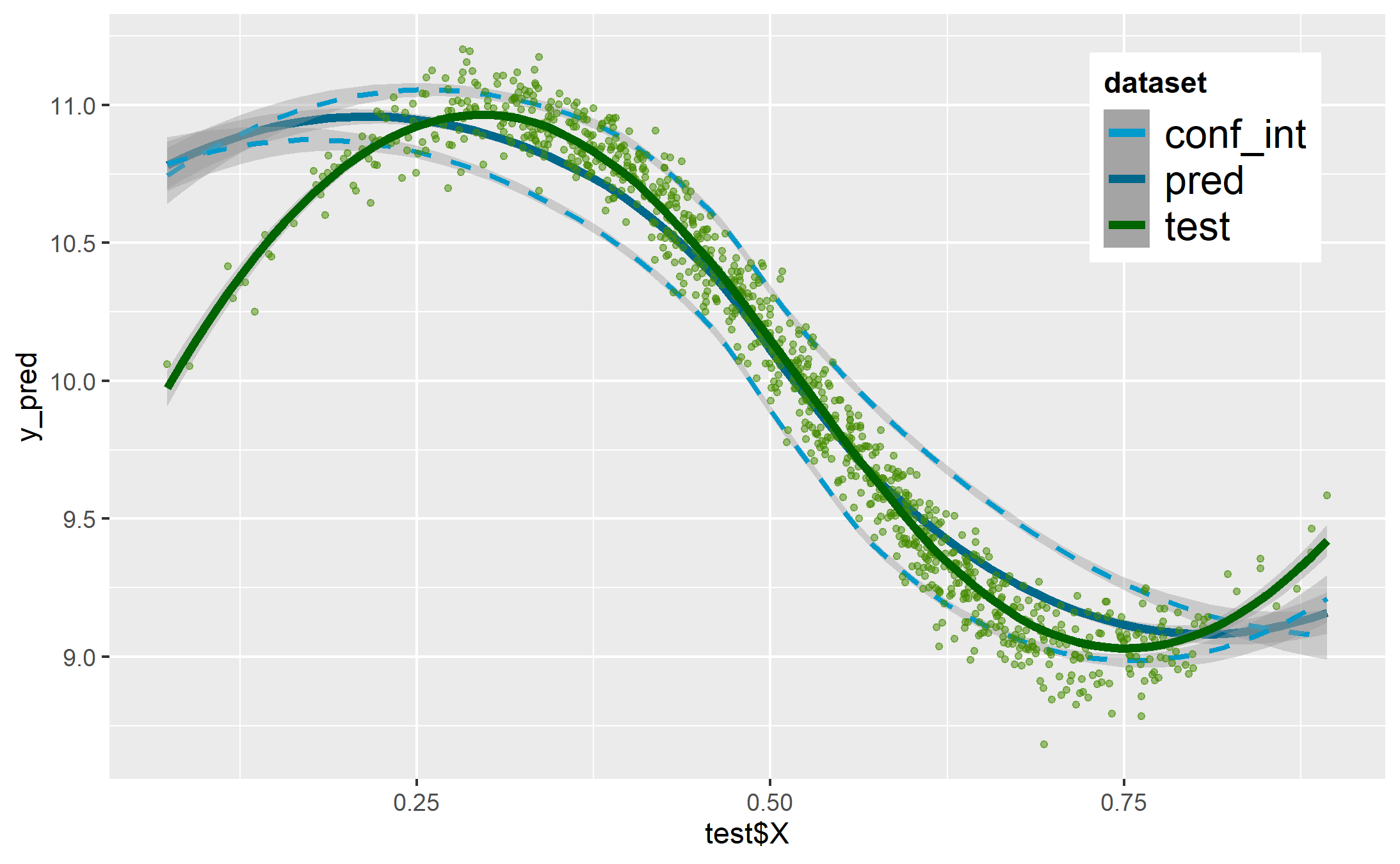}
     \caption{RF}
  \end{subfigure}
  \begin{subfigure}[b]{0.3\linewidth}
    \includegraphics[width=\linewidth]{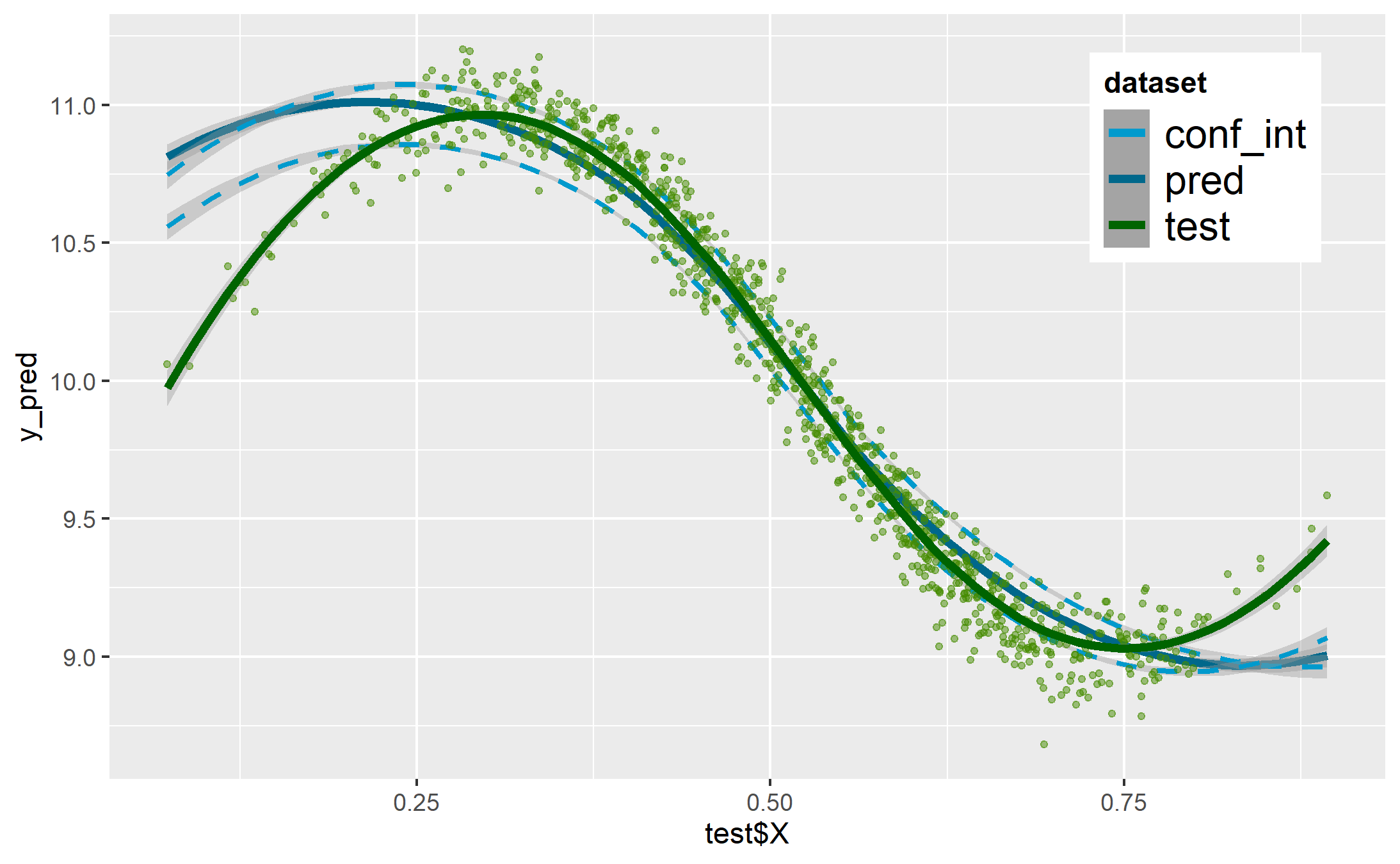}
    \caption{SMOTE}
  \end{subfigure}
  \begin{subfigure}[b]{0.3\linewidth}
    \includegraphics[width=\linewidth]{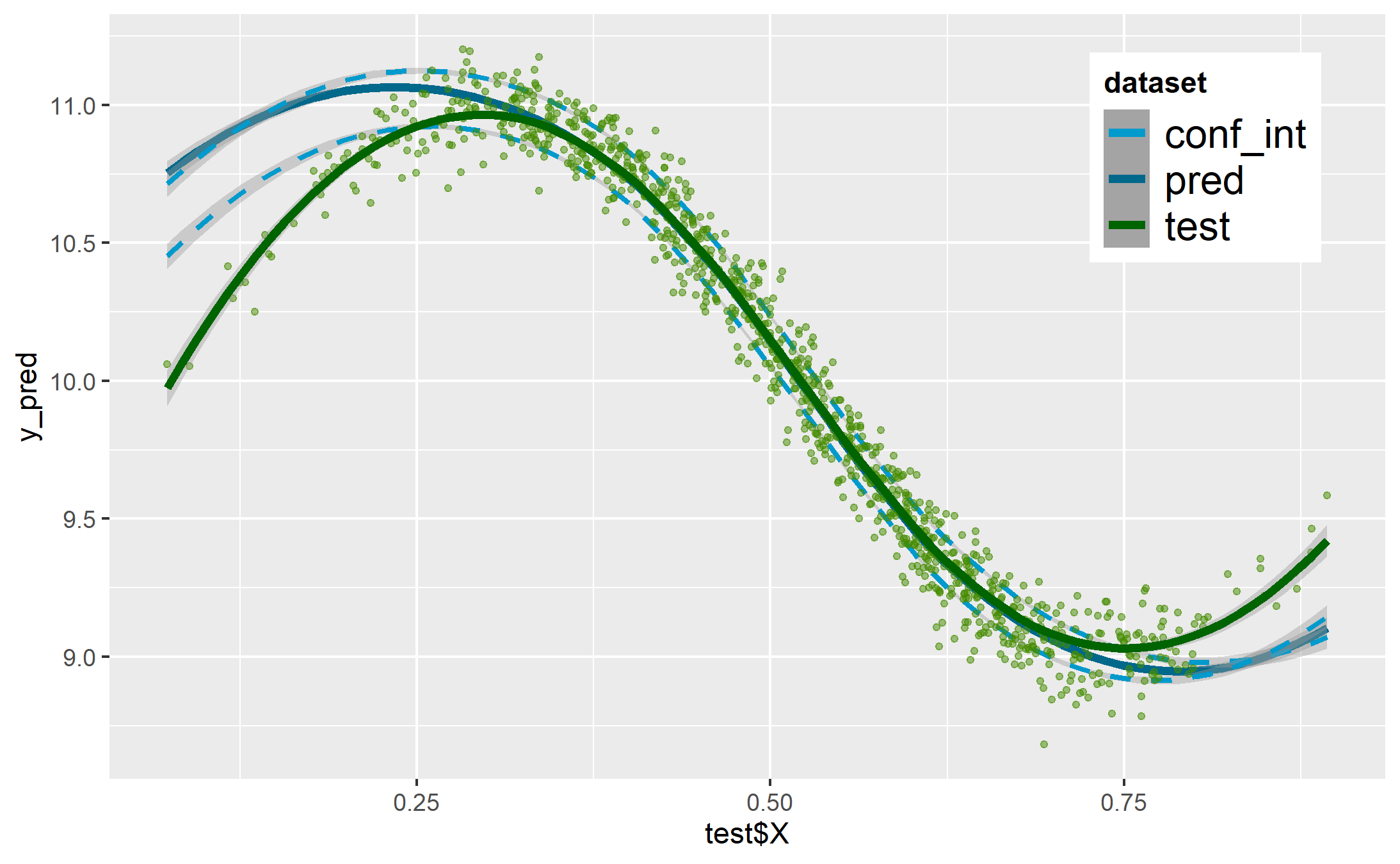}
    \caption{SMOTE - GMM}
  \end{subfigure}

  \caption{Smoothed predictions with RF}
  \label{pred_Y_RF_ech_XXX-vs-test}
\end{figure}

\newpage
\textbf{Multivariate Adaptative Regression Splines predictions}

\begin{figure}[H]
  \centering

  \begin{subfigure}[b]{0.3\linewidth}
    \includegraphics[width=\linewidth]{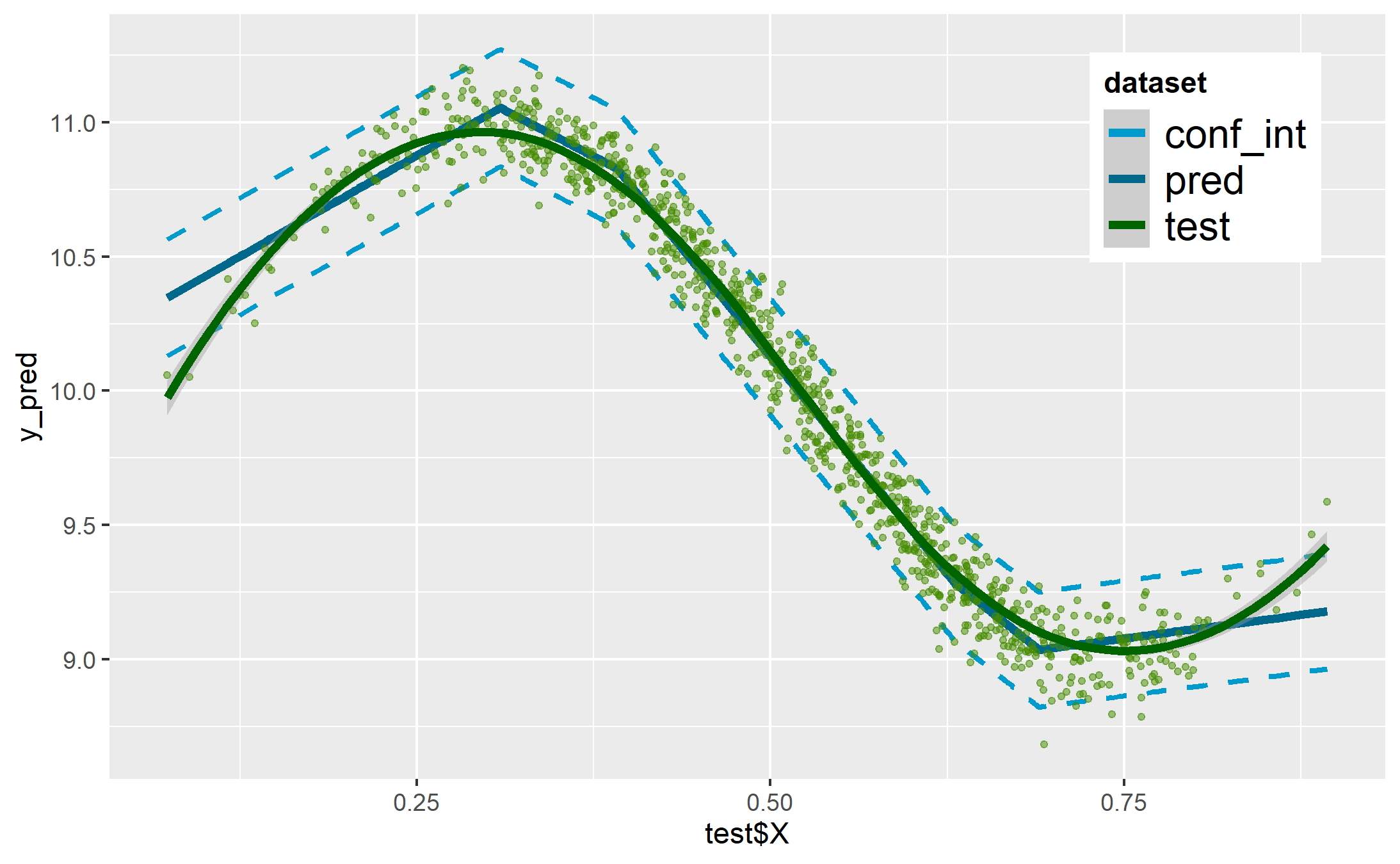}
     \caption{$\mathcal{D}^b$}
  \end{subfigure}
  \begin{subfigure}[b]{0.3\linewidth}
    \includegraphics[width=\linewidth]{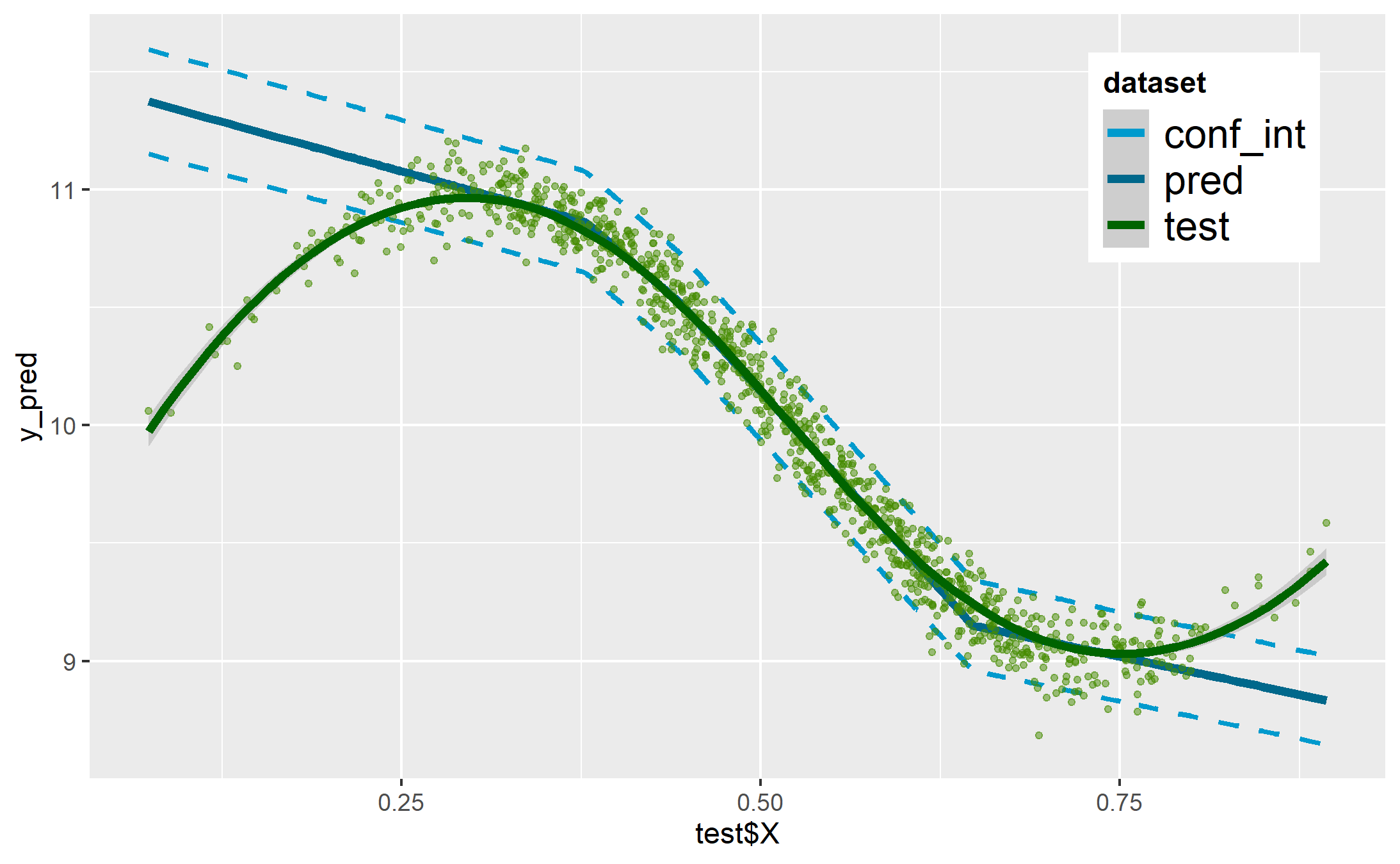}
    \caption{$\mathcal{D}^i$}
  \end{subfigure}
  \begin{subfigure}[b]{0.3\linewidth}
    \includegraphics[width=\linewidth]{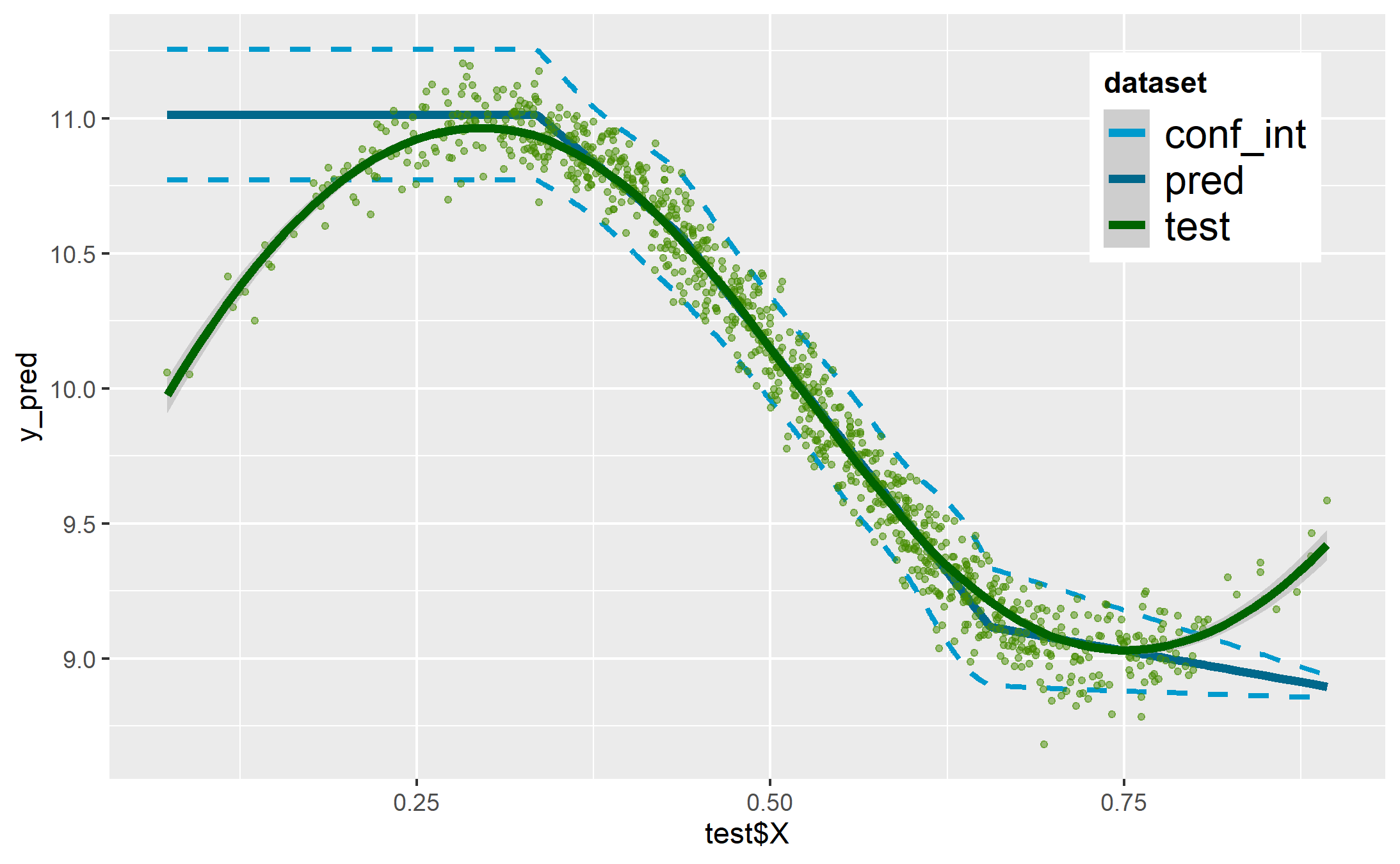}
    \caption{WR}
  \end{subfigure}

  \begin{subfigure}[b]{0.3\linewidth}
    \includegraphics[width=\linewidth]{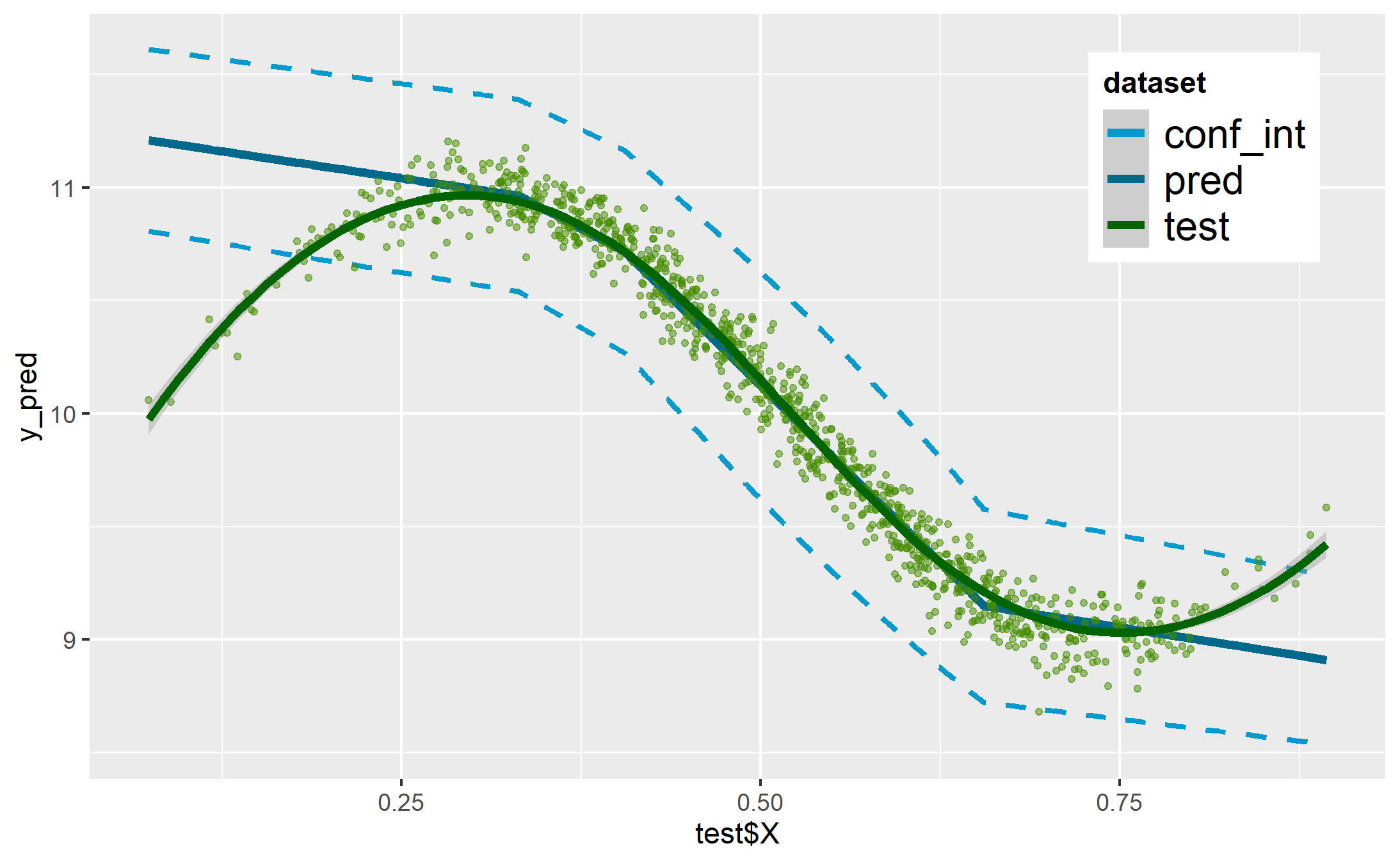}
     \caption{GN}
  \end{subfigure}
  \begin{subfigure}[b]{0.3\linewidth}
    \includegraphics[width=\linewidth]{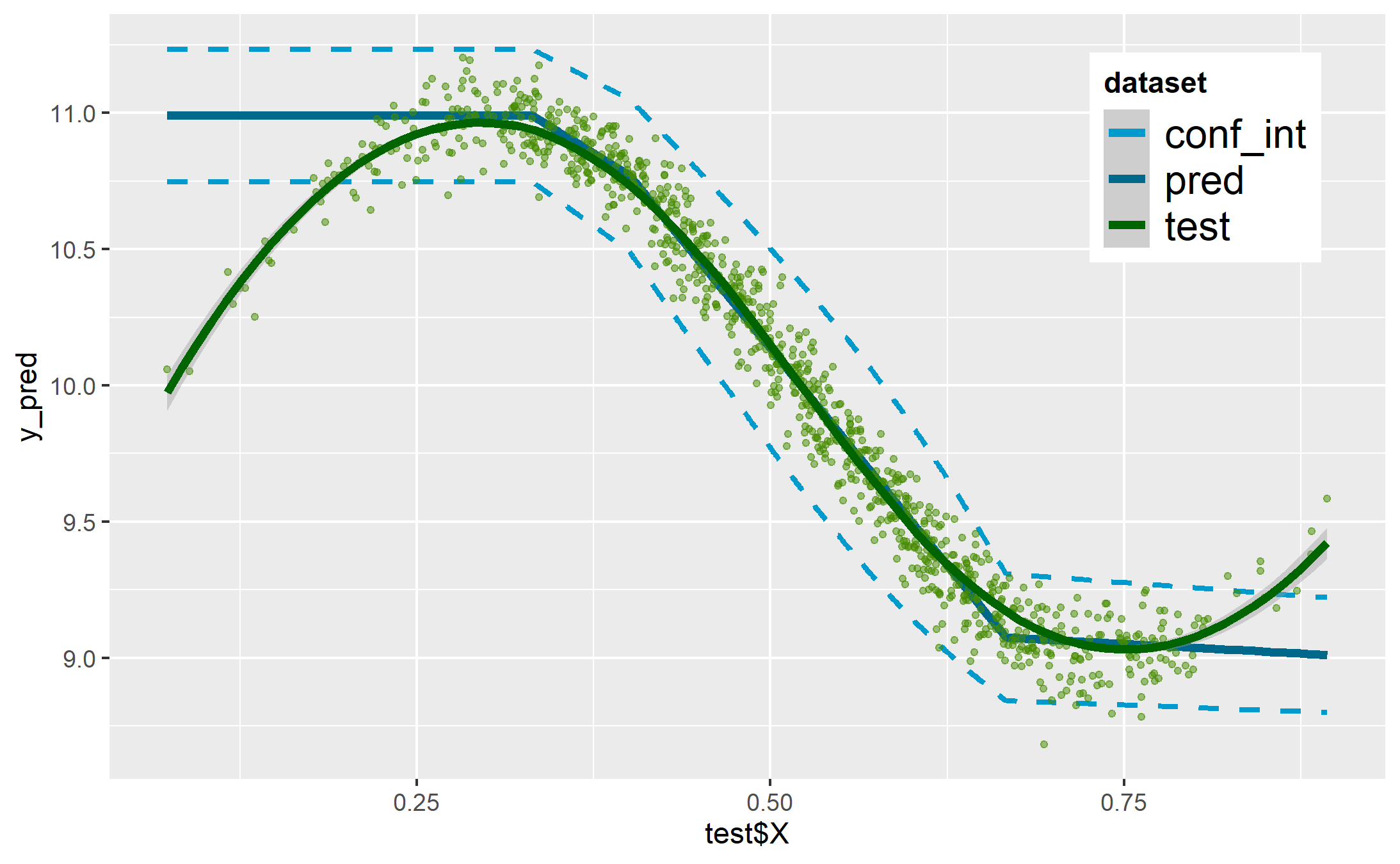}
    \caption{GN - GMM}
  \end{subfigure}
  \begin{subfigure}[b]{0.3\linewidth}
    \includegraphics[width=\linewidth]{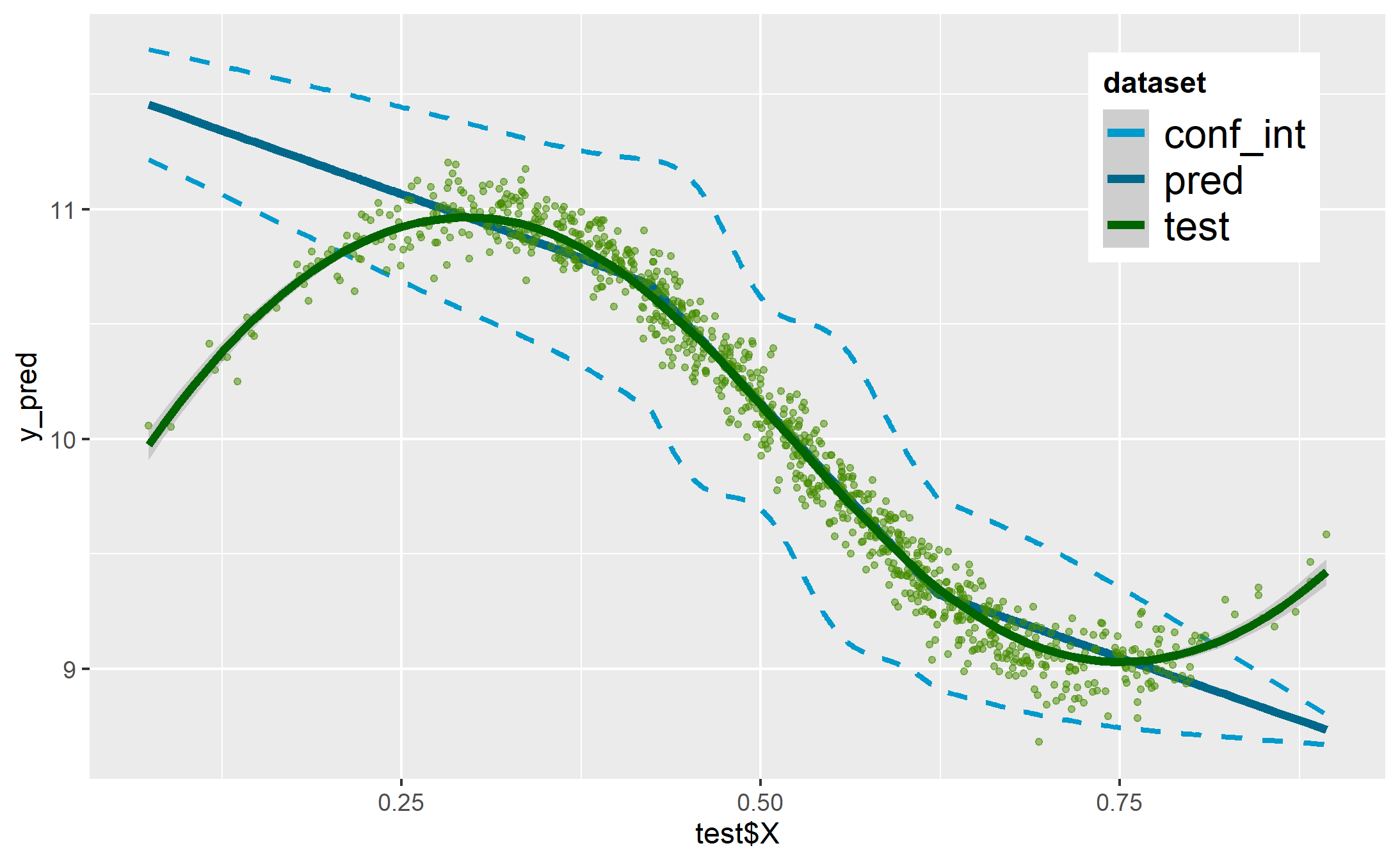}
    \caption{ROSE}
  \end{subfigure}

  \begin{subfigure}[b]{0.3\linewidth}
    \includegraphics[width=\linewidth]{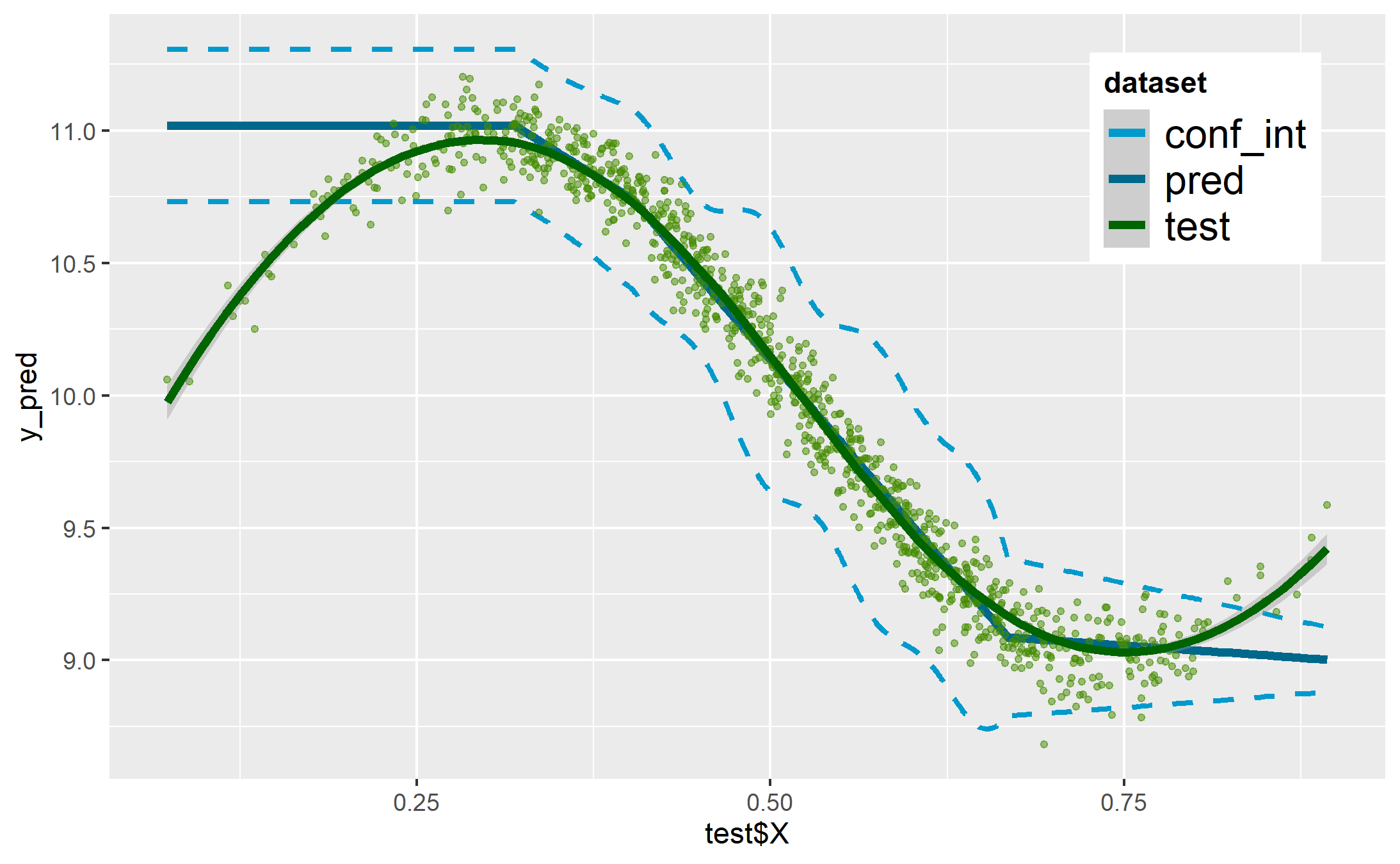}
     \caption{ROSE - GMM}
  \end{subfigure}
  \begin{subfigure}[b]{0.3\linewidth}
    \includegraphics[width=\linewidth]{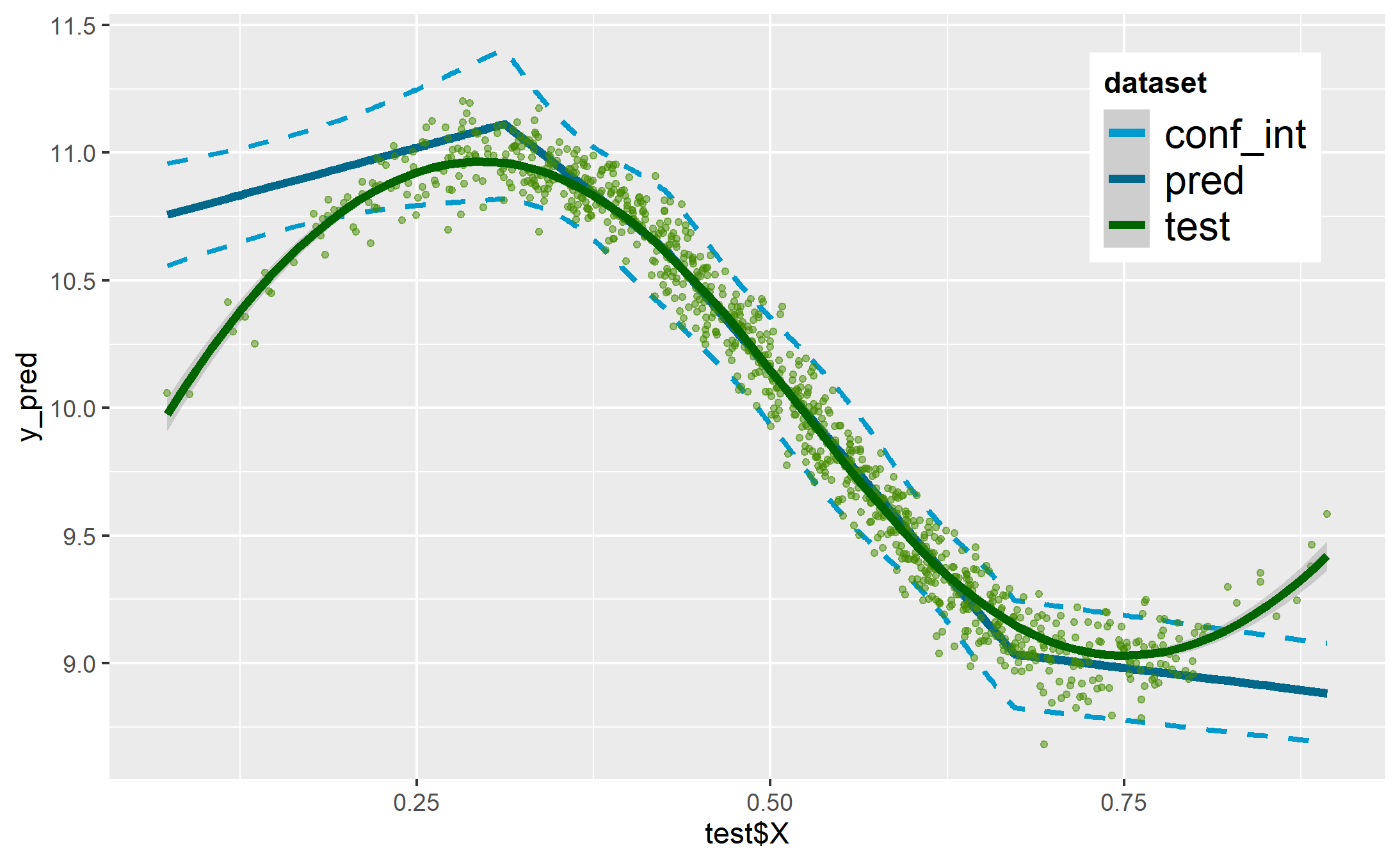}
    \caption{KDE - GMM}
  \end{subfigure}
  \begin{subfigure}[b]{0.3\linewidth}
    \includegraphics[width=\linewidth]{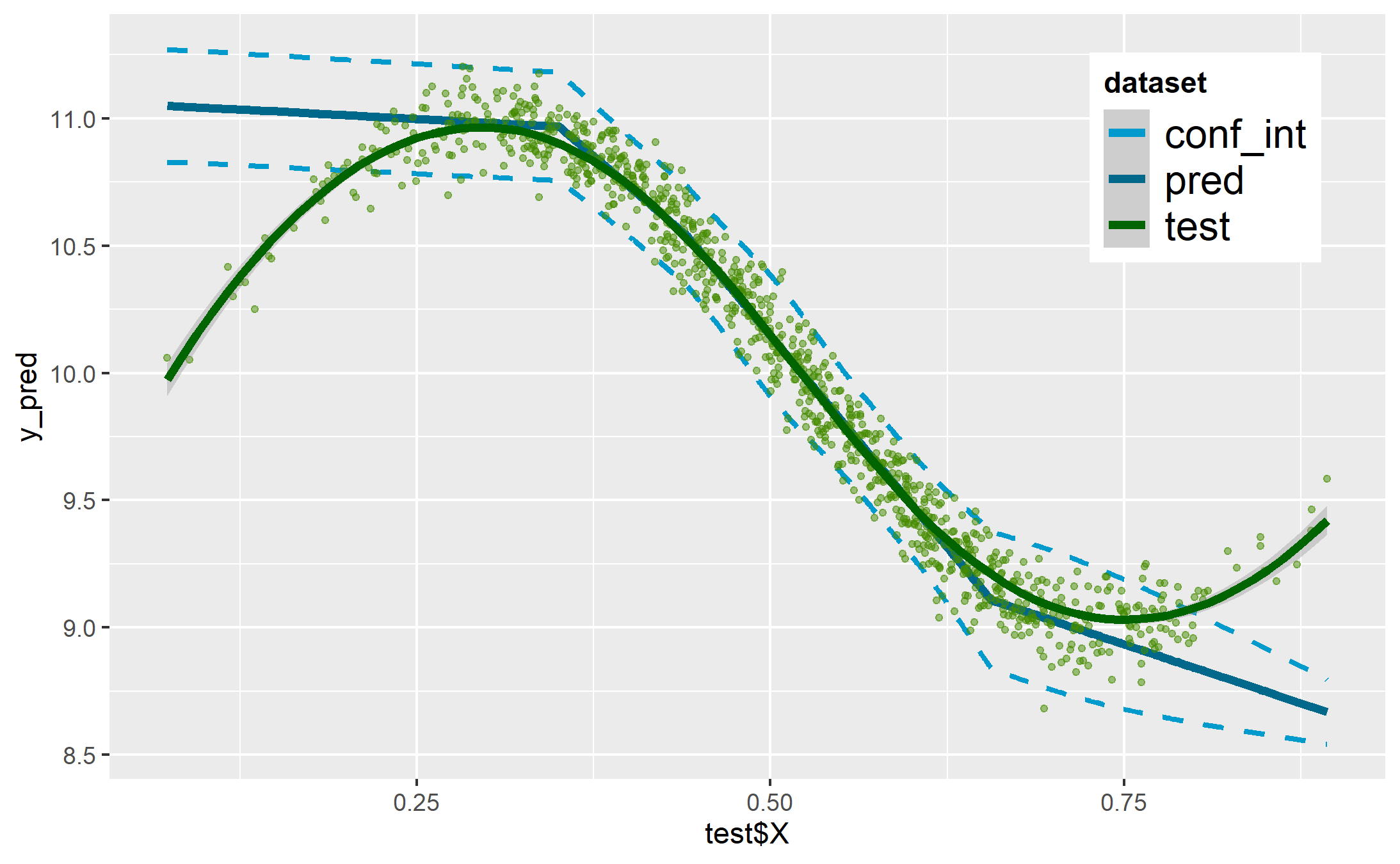}
    \caption{GMM}
  \end{subfigure}

  \begin{subfigure}[b]{0.3\linewidth}
    \includegraphics[width=\linewidth]{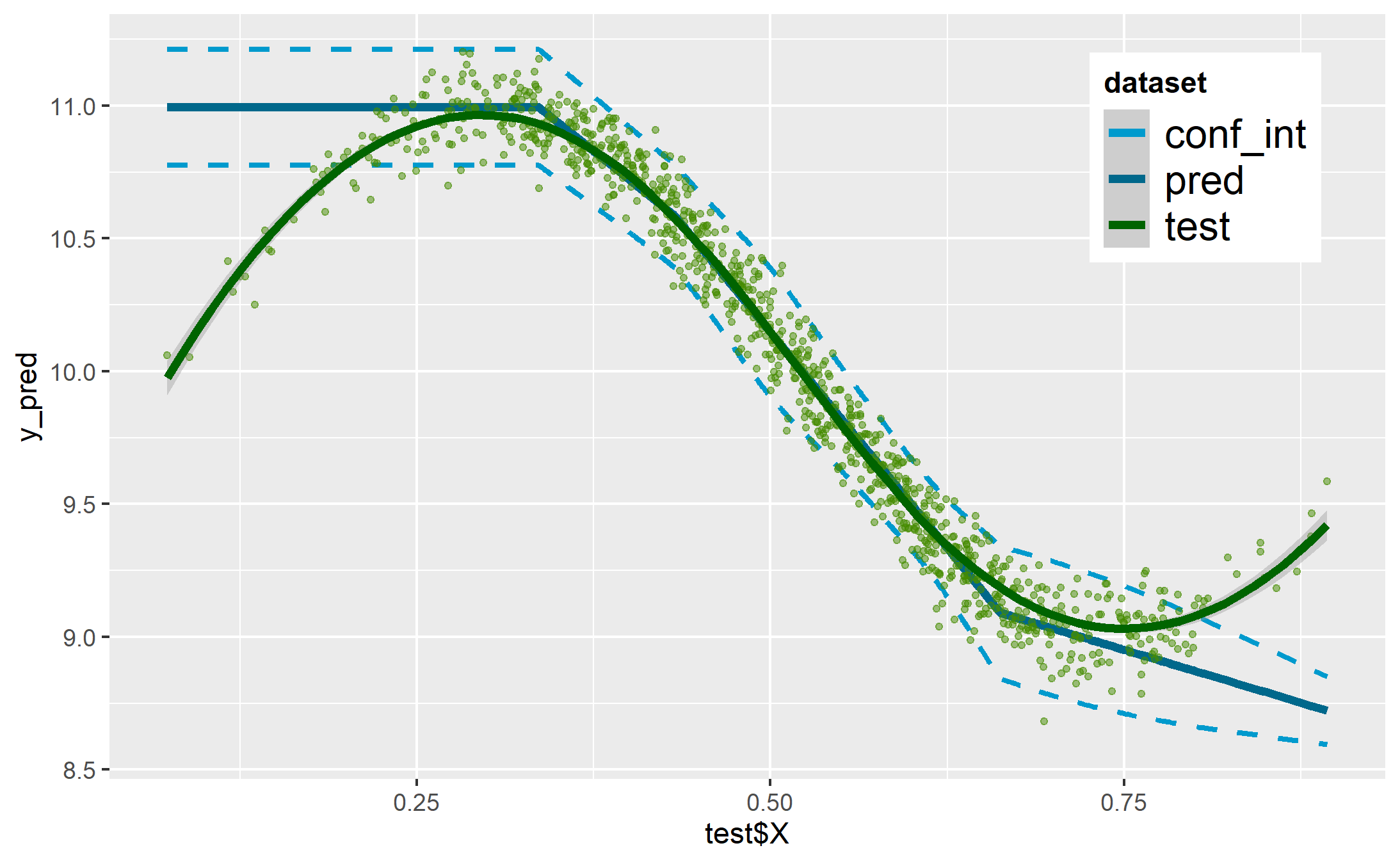}
     \caption{FA - GMM}
  \end{subfigure}
  \begin{subfigure}[b]{0.3\linewidth}
    \includegraphics[width=\linewidth]{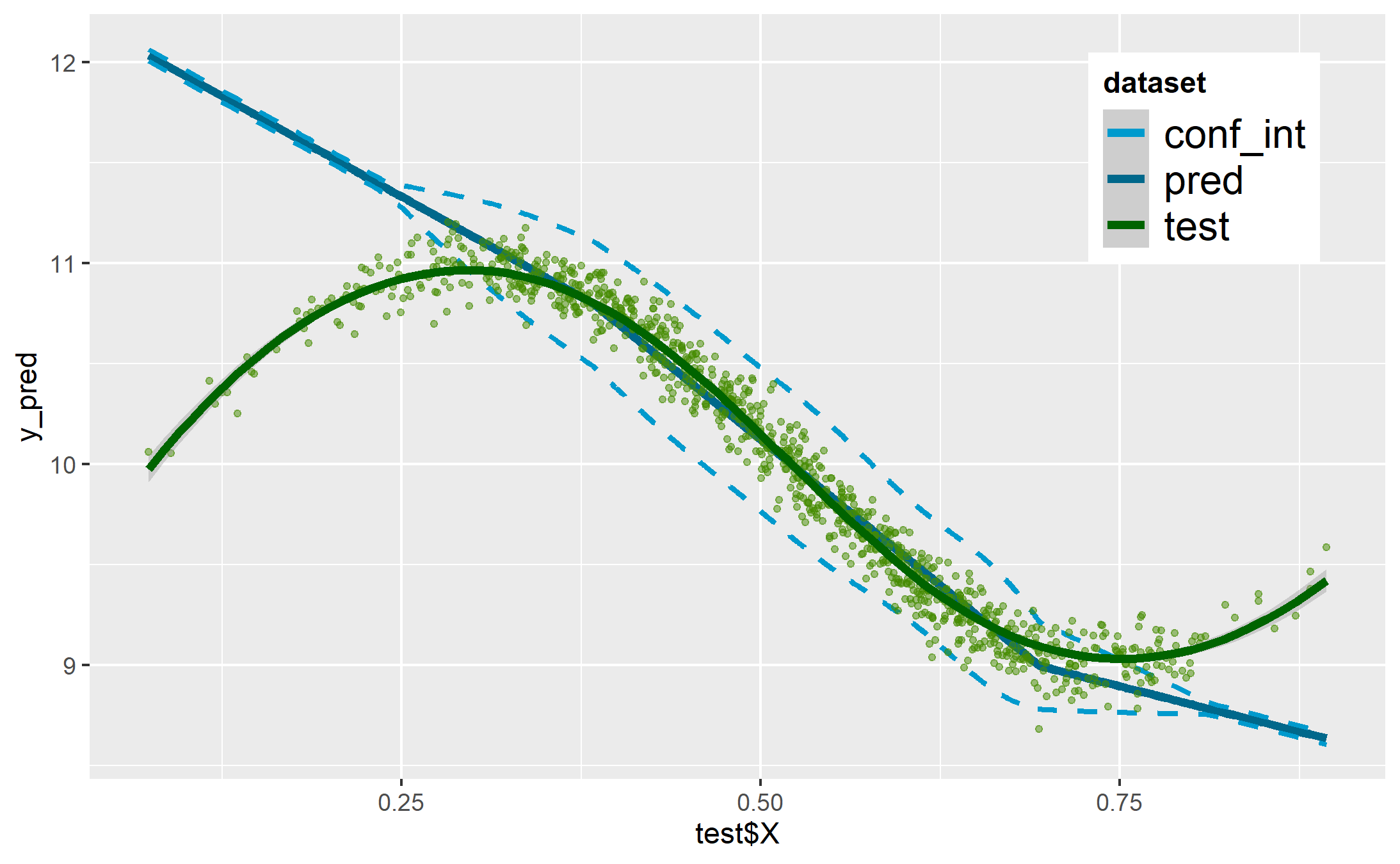}
    \caption{Copula}
  \end{subfigure}
  \begin{subfigure}[b]{0.3\linewidth}
    \includegraphics[width=\linewidth]{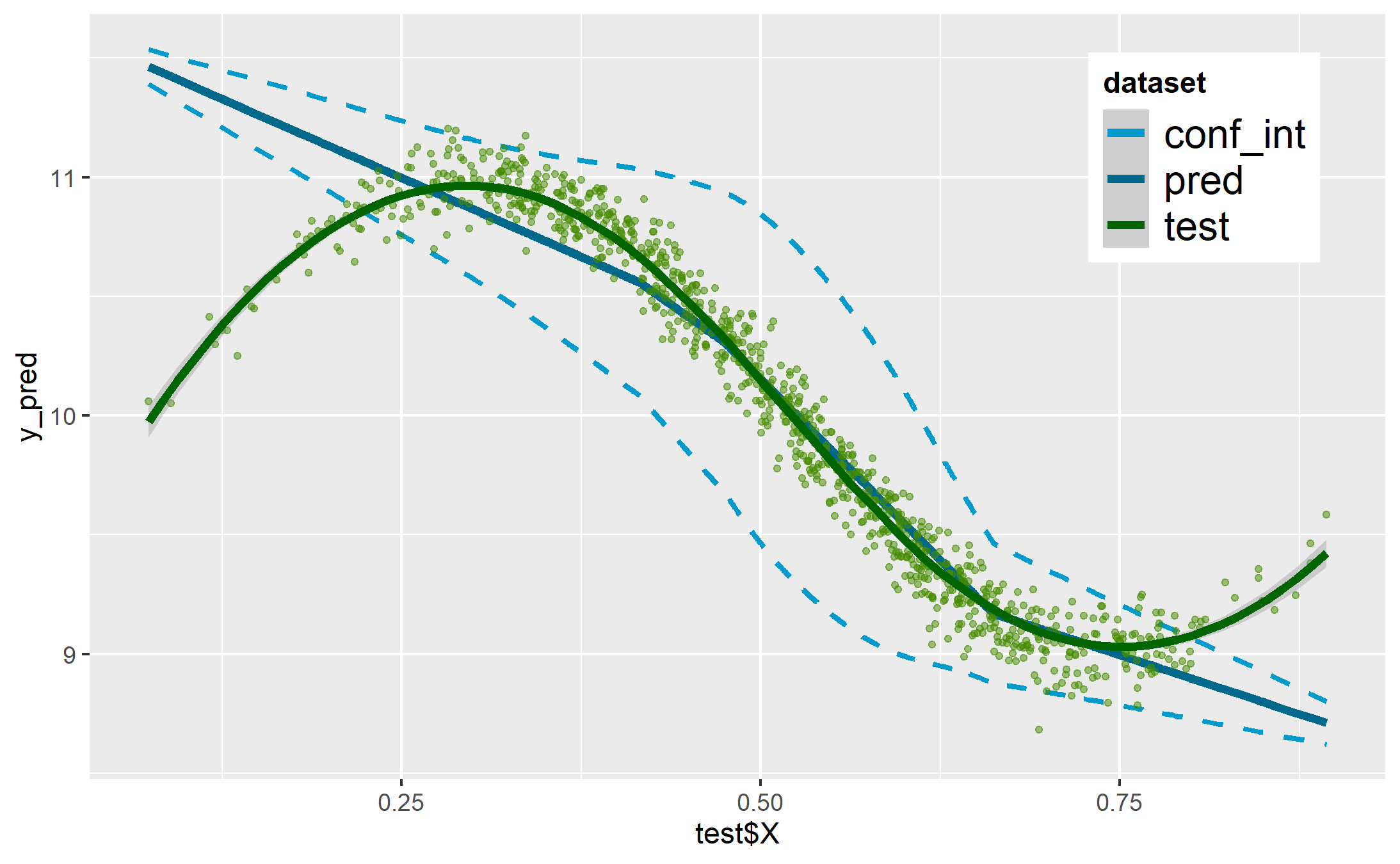}
    \caption{GAN}
  \end{subfigure}

  \begin{subfigure}[b]{0.3\linewidth}
    \includegraphics[width=\linewidth]{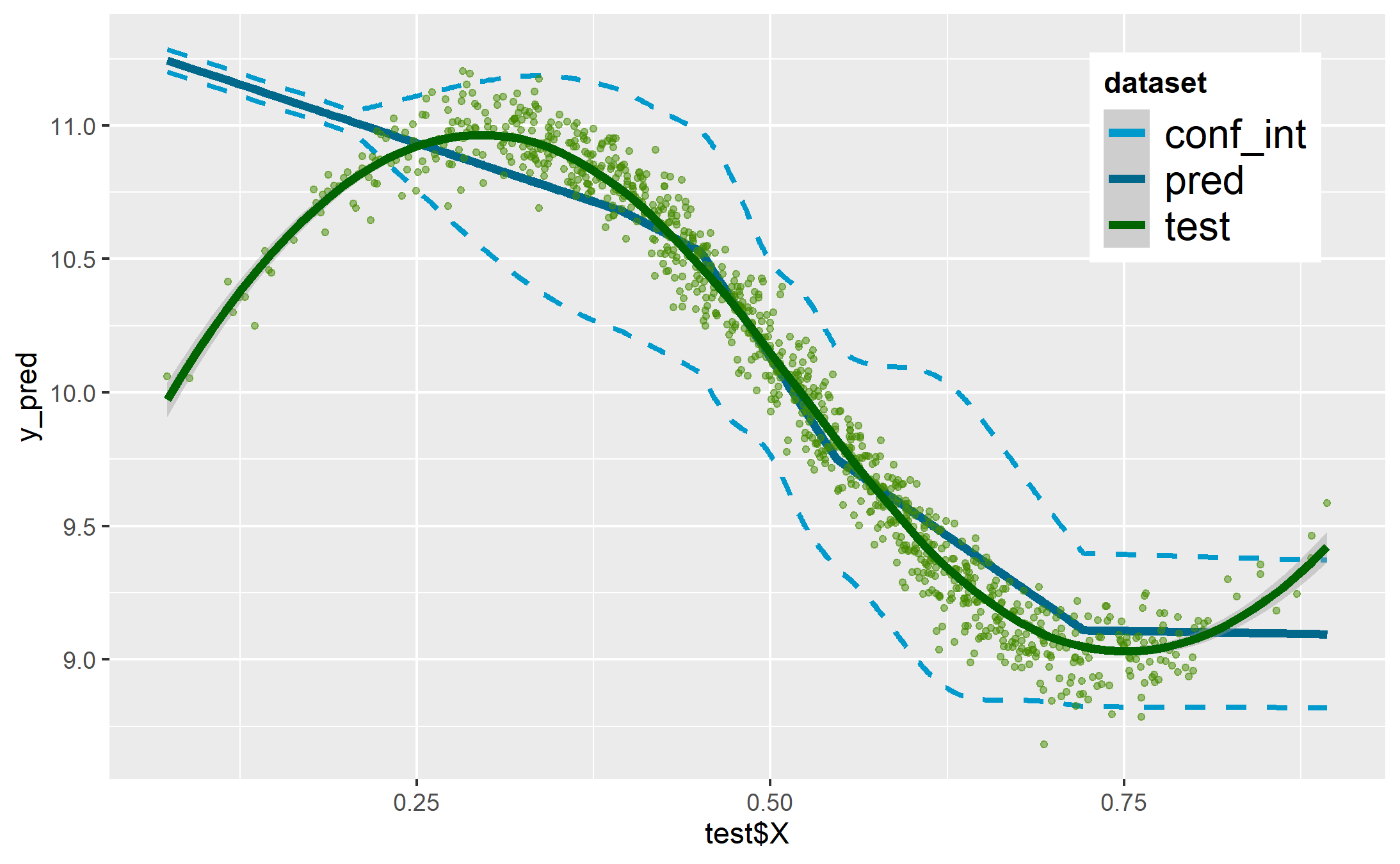}
     \caption{RF}
  \end{subfigure}
  \begin{subfigure}[b]{0.3\linewidth}
    \includegraphics[width=\linewidth]{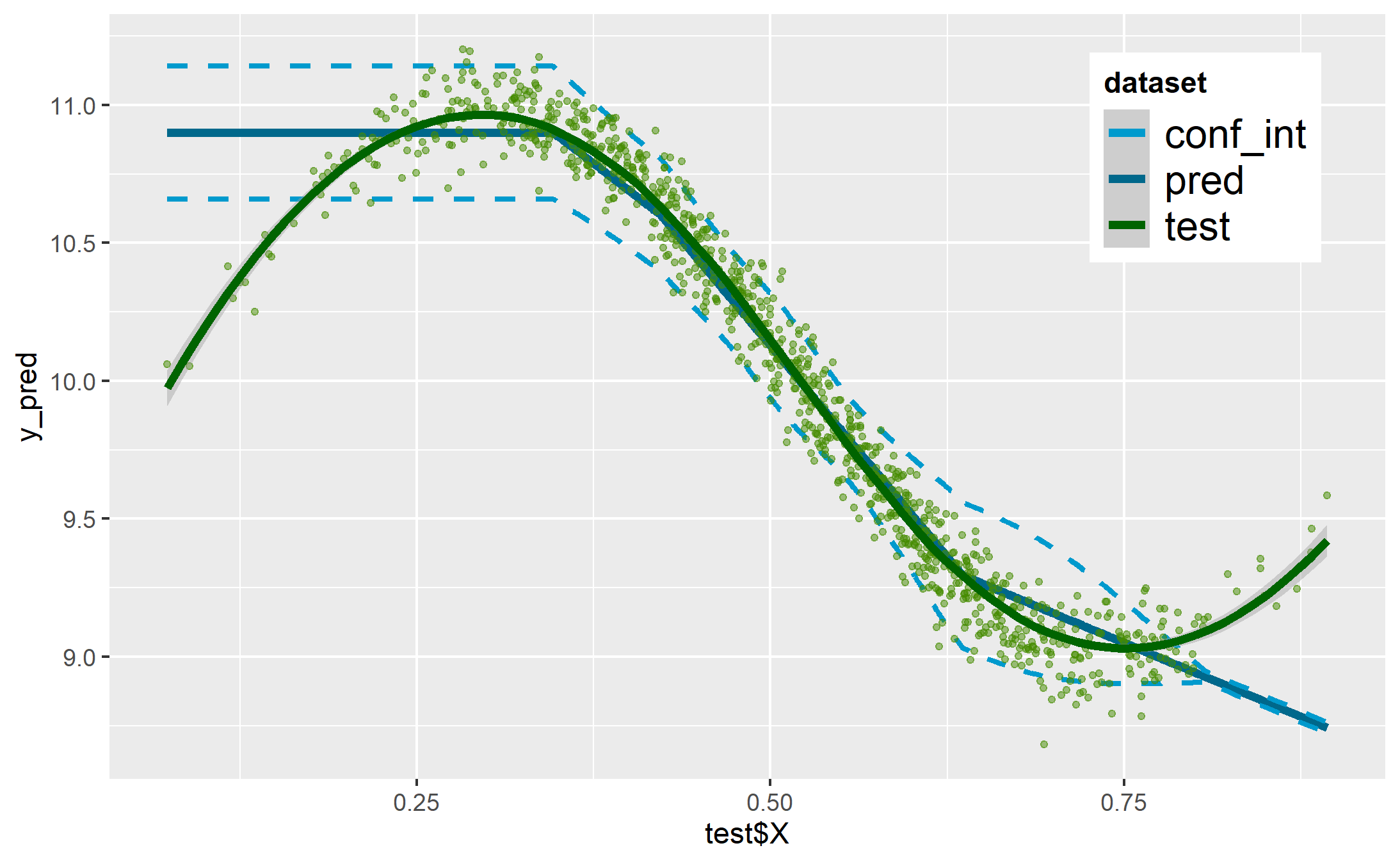}
    \caption{SMOTE}
  \end{subfigure}
  \begin{subfigure}[b]{0.3\linewidth}
    \includegraphics[width=\linewidth]{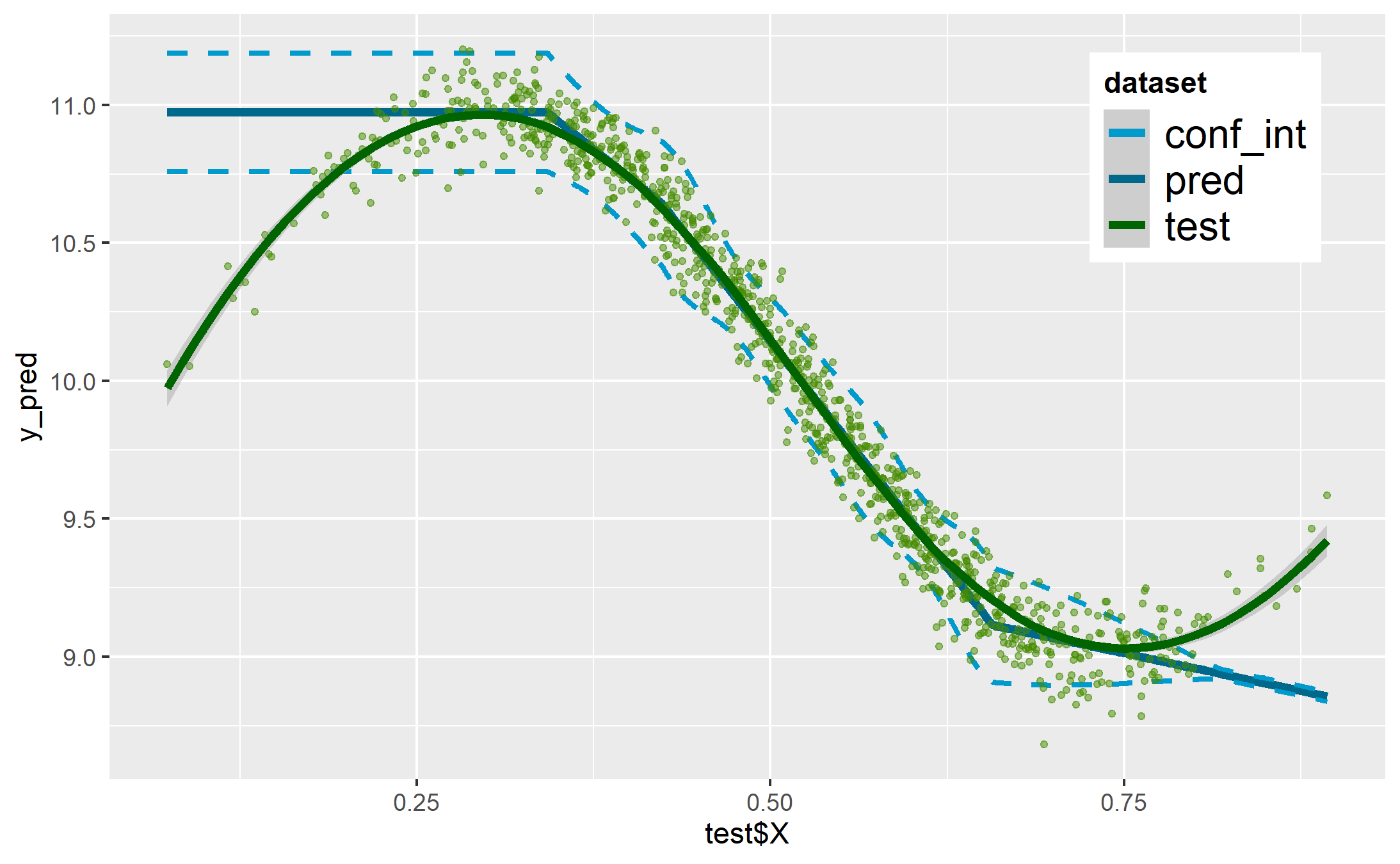}
    \caption{SMOTE - GMM}
  \end{subfigure}

  \caption{Smooth predictions with MARS}
  \label{pred_Y_MARS_ech_XXX-vs-test}
\end{figure}

\subsection{Analysis of distribution of X}

\begin{figure}[H]
\centering
\includegraphics[width= 0.9 \textwidth]{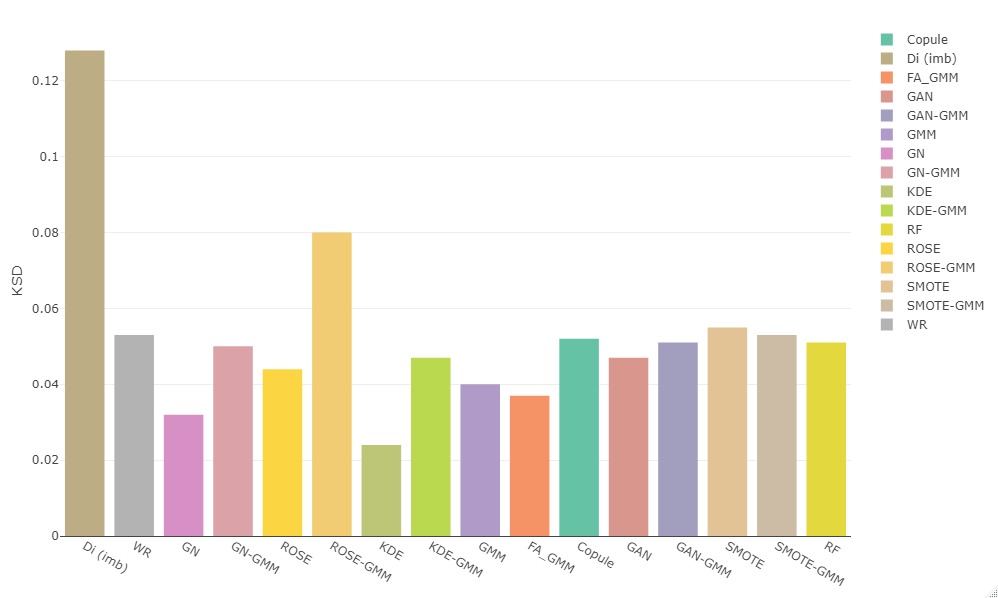}
\caption{Kolmogorv-Smirnov distance between the distribution of $X$ in the balanced sample and train samples}
\label{illu_KS_dist_X}
\end{figure}

\subsection{On multiple simulations}

\begin{figure}[H]
\centering
\includegraphics[width= 0.9 \textwidth]{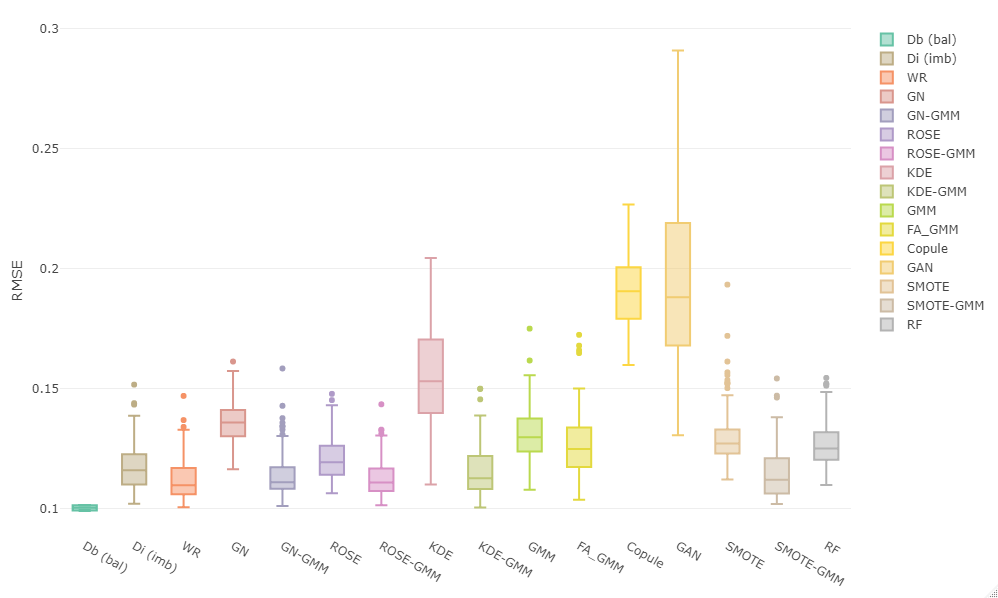}
\caption{RMSE boxplots on test datasets prediction with GAM}
\label{RMSE_GAM}
\end{figure}

\begin{figure}[H]
\centering
\includegraphics[width=0.9\textwidth]{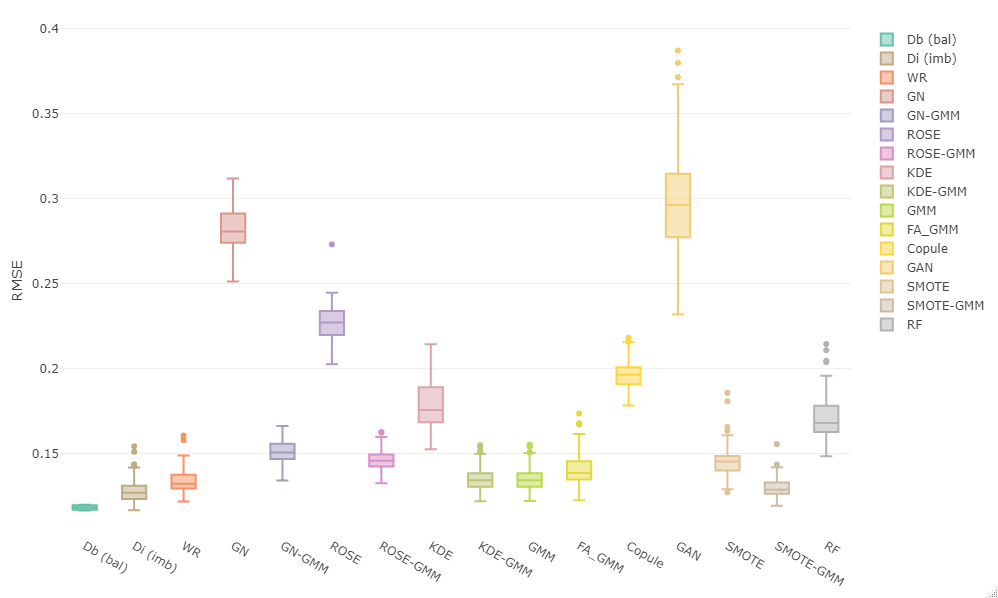}
\caption{RMSE boxplots on test datasets prediction with RF}
\label{RMSE_RF}
\end{figure}

\begin{figure}[H]
\centering
\includegraphics[width=0.9\textwidth]{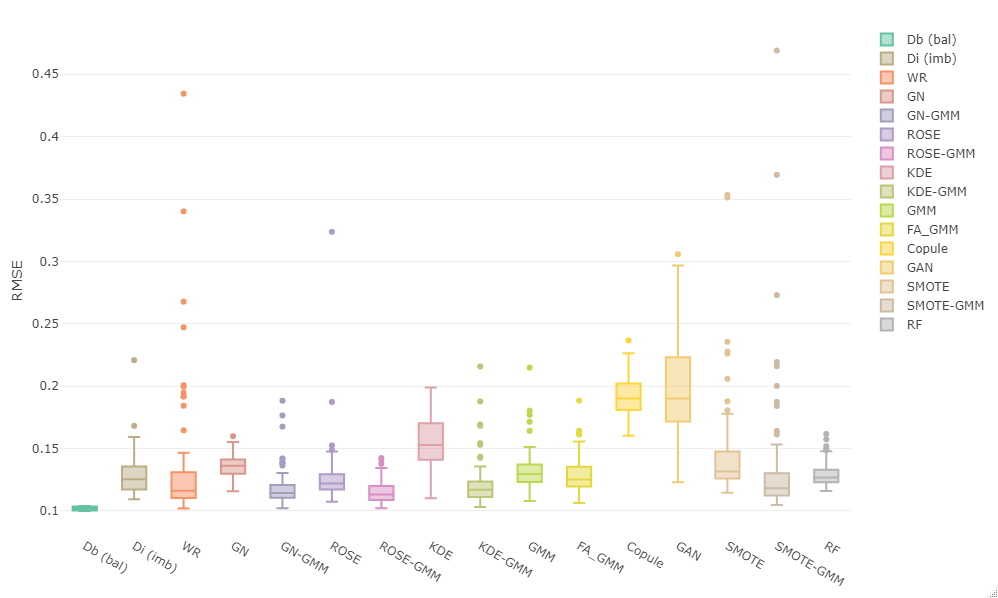}
\caption{RMSE boxplots on test datasets prediction with MARS}
\label{RMSE_MARS}
\end{figure}

Note: the results for the GAN is slightly worse than for the single simulation because of the lower epoch parameter.

\section{\textsc{Application}}

For the application, we chose a trimming  sequence $e_{n} = \frac{1}{10 \times n}$

\subsection{Dataset details}

Table \ref{dataset_variables} provides a brief description of the variables in the dataset, extracted from \cite{telematics}.

\begin{table}[H]
\begin{center}
\begin{tabular}{ c c c }
\textbf{Type}       &\textbf{Variable}  &\textbf{Description}\\
\hline
Traditional         &Duration           &Duration of the insurance coverage of a given policy, in days \\
                    &Insured.age        &Age of insured driver, in years \\
                    &Insured.sex        &Sex of insured driver (Male/Female) \\
                    &Car.age            &Age of vehicle, in years \\
                    &Marital            &Marital status (Single/Married) \\
                    &Car.use            &Use of vehicle: Private, Commute, Farmer, Commercial \\
                    &Credit.score       &Credit score of insured driver \\
                    &Region             &Type of region where driver lives: rural, urban \\
                    &Annual.miles.drive &Annual miles expected to be driven declared by driver \\
                    &Years.noclaims     &Number of years without any claims \\
                    &Territory          &Territorial location of vehicle \\
\hline
Telematics          &Annual.pct.driven  &Annualized percentage of time on the road \\
                    &Total.miles.driven &Total distance driven in miles \\
                    &Pct.drive.xxx      &Percent of driving day xxx of the week: mon/tue/. . . /sun \\
                    &Pct.drive.xhrs     &Percent vehicle driven within x hrs: 2hrs/3hrs/4hrs \\
                    &Pct.drive.xxx      &Percent vehicle driven during xxx: wkday/wkend \\
                    &Pct.drive.rushxx   &Percent of driving during xx rush hours: am/pm \\
                    &Avgdays.week       &Mean number of days used per week \\
                    &Accel.xxmiles      &Number of sudden acceleration 6/8/9/. . . /14 mph/s per 1000miles \\
                    &Brake.xxmiles      &Number of sudden brakes 6/8/9/. . . /14 mph/s per 1000miles \\
                    &Left.turn.intensityxx &Number of left turn per 1000miles with intensity 08/09/10/11/12 \\
                    &Right.turn.intensityxx &Number of right turn per 1000miles with intensity 08/09/10/11/12 \\
\hline
Response            &NB Claim           &Number of claims during observation \\
                    &AMT Claim          &Aggregated amount of claims during observation \\
\end{tabular}
\caption{Variable names and descriptions} \label{dataset_variables}
\end{center}
\end{table}

\begin{figure}[H]
\centering
\includegraphics[width=0.49\textwidth]{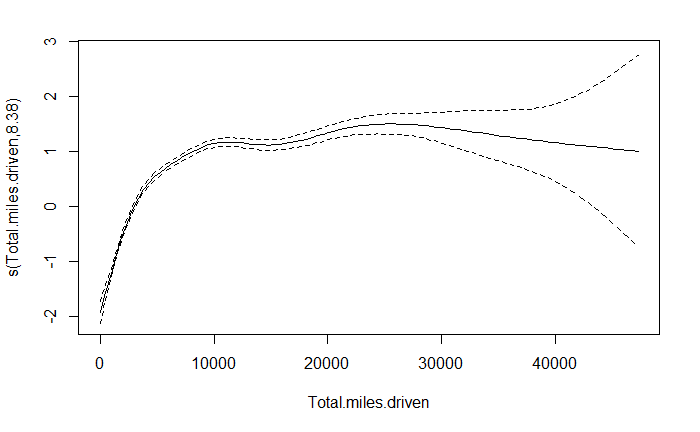}
\caption{Regression spline for $X$ with GAM-ZIP based on population}
\label{appli_s_km_driven-vs-Y_Pop}
\end{figure}

Figure \ref{appli_s_km_driven-vs-Y_Pop} shows the effect of $X$, total miles driven, on the claim frequency estimated by a GAM model. We can see that, with this model, the claim frequency increases with $X$ to about 10,000 miles then becomes quite constant.

As with the illustration, we defined three samples from population $\mathcal{D}^p$ of size $n^p=100,000$: balanced, imbalanced and test, all of a size $n=10,000$. The test sample $\mathcal{D}^t$ is uniformly drawn such as all observations in population have the same drawing weight ($\frac{1}{n^p}$). The balanced sample is then drawn uniformly (with a probability equal to $\frac{1}{n^p - n}$) from the remaining population ($\mathcal{D}^p$ \textbackslash $\mathcal{D}^t$).
The imbalanced sample is defined according to $X$, \textit{total miles driven}, such as: the larger $X$ is the smaller the drawing weight is. We want to get a more asymmetric distribution than the population with fewer observations after 10,000 miles. 
More precisely, the proposed drawing weight to construct the imbalanced sample is  defined by a Gaussian distribution $\mathcal{N}(2000,6000)$. As we can see on Figure \ref{appli_comp_X_Ech0-vs-Pop-Dens}, the distribution obtained has very few values above 10,000 miles.

\subsection{Imbalanced sample}

\begin{figure}[H]
\centering
\begin{subfigure}{0.49\textwidth}
\includegraphics[width=\textwidth]{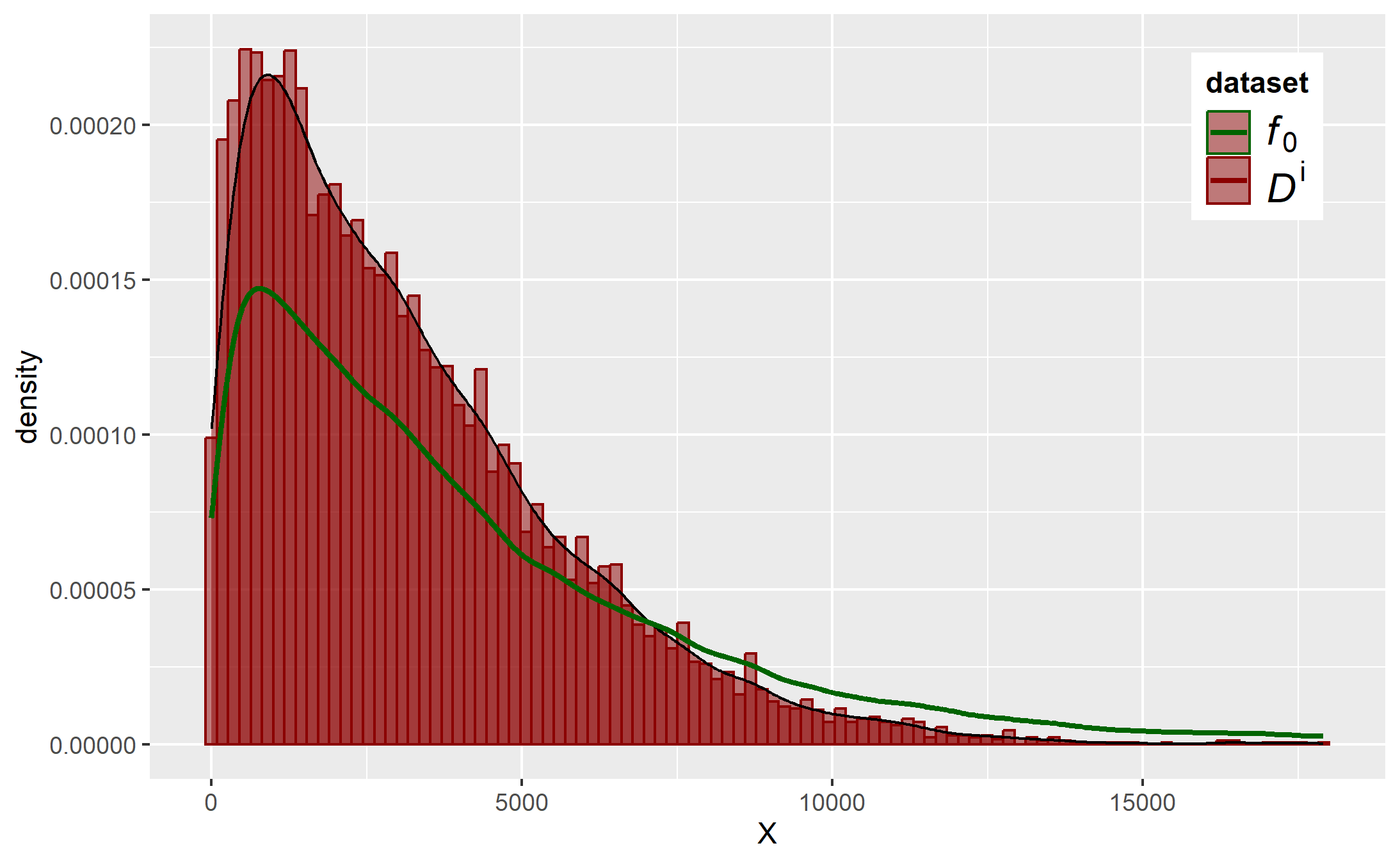} 
\caption{Histogram of $X$ in the imbalanced sample (red) vs the target distribution $f_0$ (green)}
\label{Appli_Hist_X_Ech0-vs-Tgt}
\end{subfigure}
\hfill
\begin{subfigure}{0.49\textwidth}
\includegraphics[width=\textwidth]{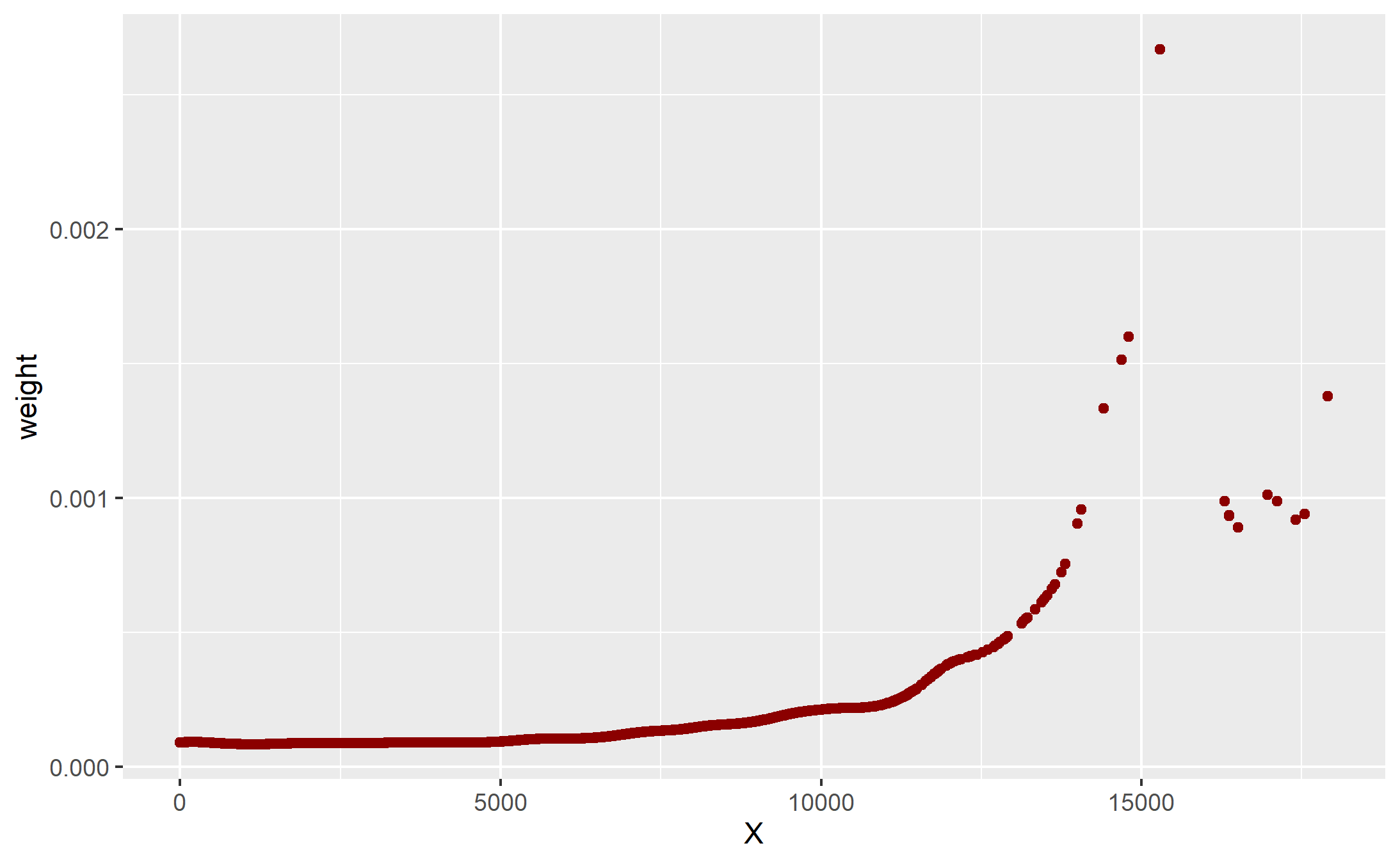}
\caption{Weights of $X$ obtained with weighted resampling method}
\label{Appli_weights-ech0}
\end{subfigure}
\caption{Comparison between the imbalanced sample vs the target distribution and associated WR weights}
\label{appli_ech0-vs-target}
\end{figure}

\subsection{Data Generation}

\textbf{Histogram of $X$ obtained in new samples vs target}

\begin{figure}[H]
  \centering

  \begin{subfigure}[b]{0.3\linewidth}
    \includegraphics[width=\linewidth]{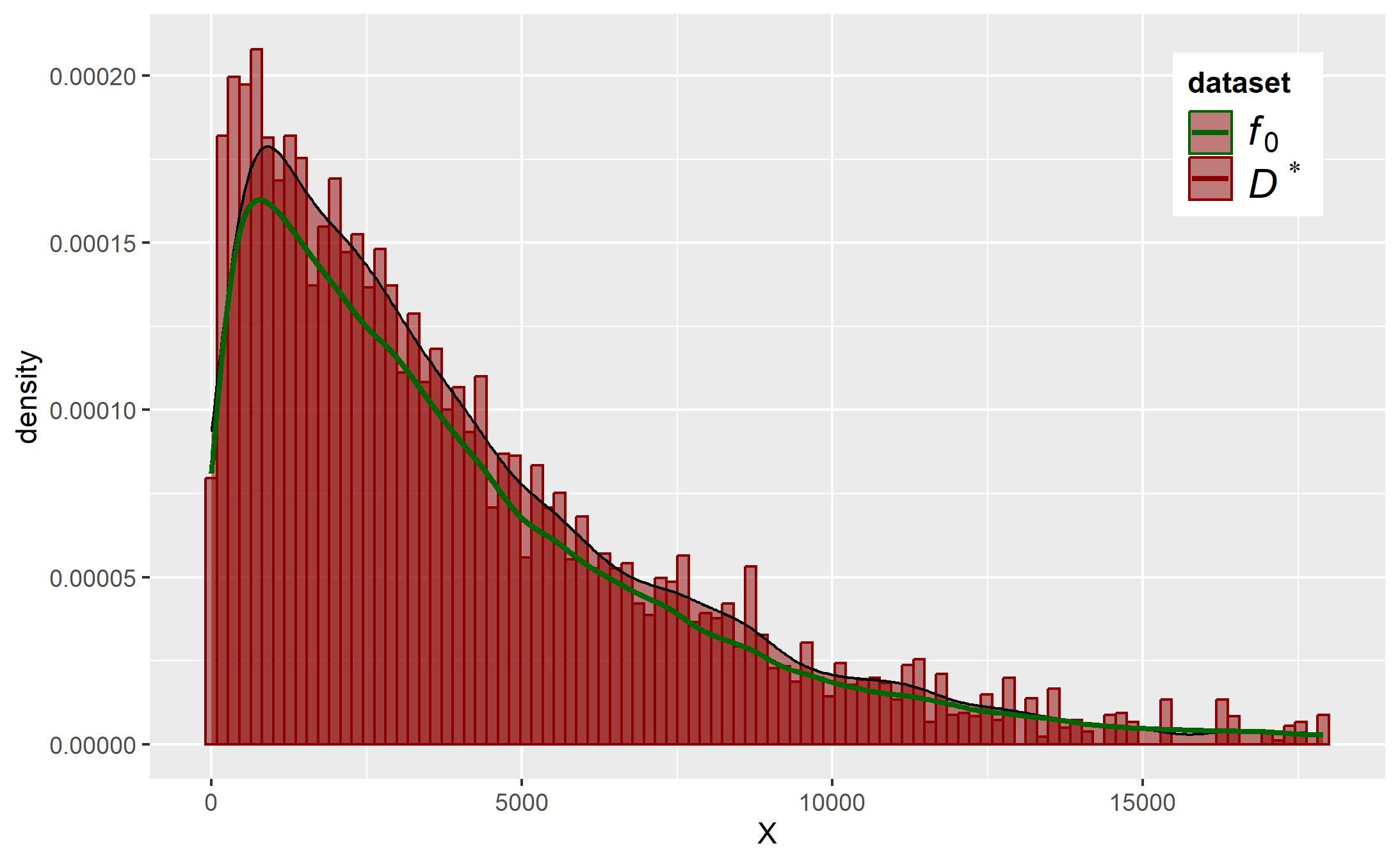}
     \caption{WR}
  \end{subfigure}
  \begin{subfigure}[b]{0.3\linewidth}
    \includegraphics[width=\linewidth]{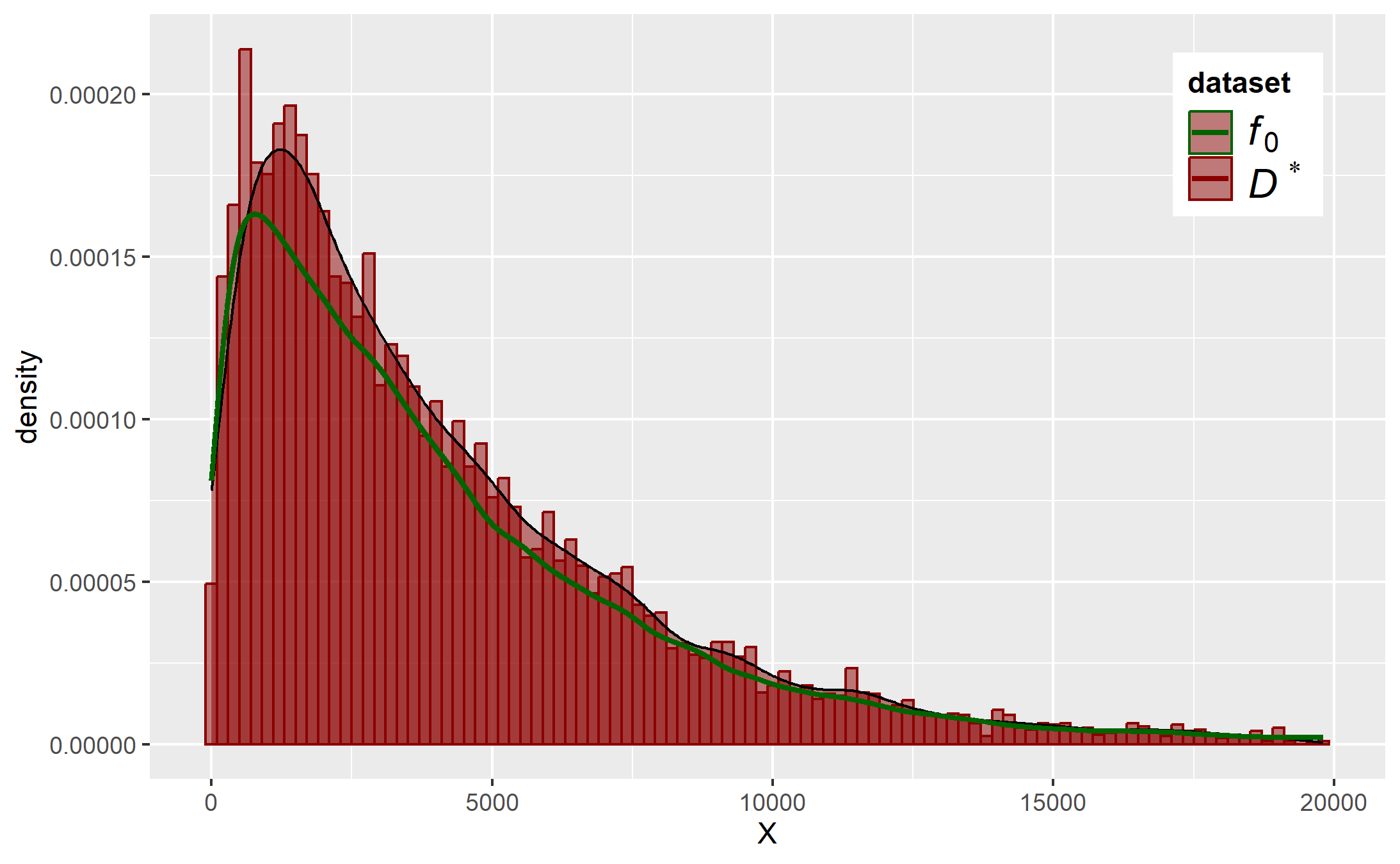}
    \caption{GN}
  \end{subfigure}
  \begin{subfigure}[b]{0.3\linewidth}
    \includegraphics[width=\linewidth]{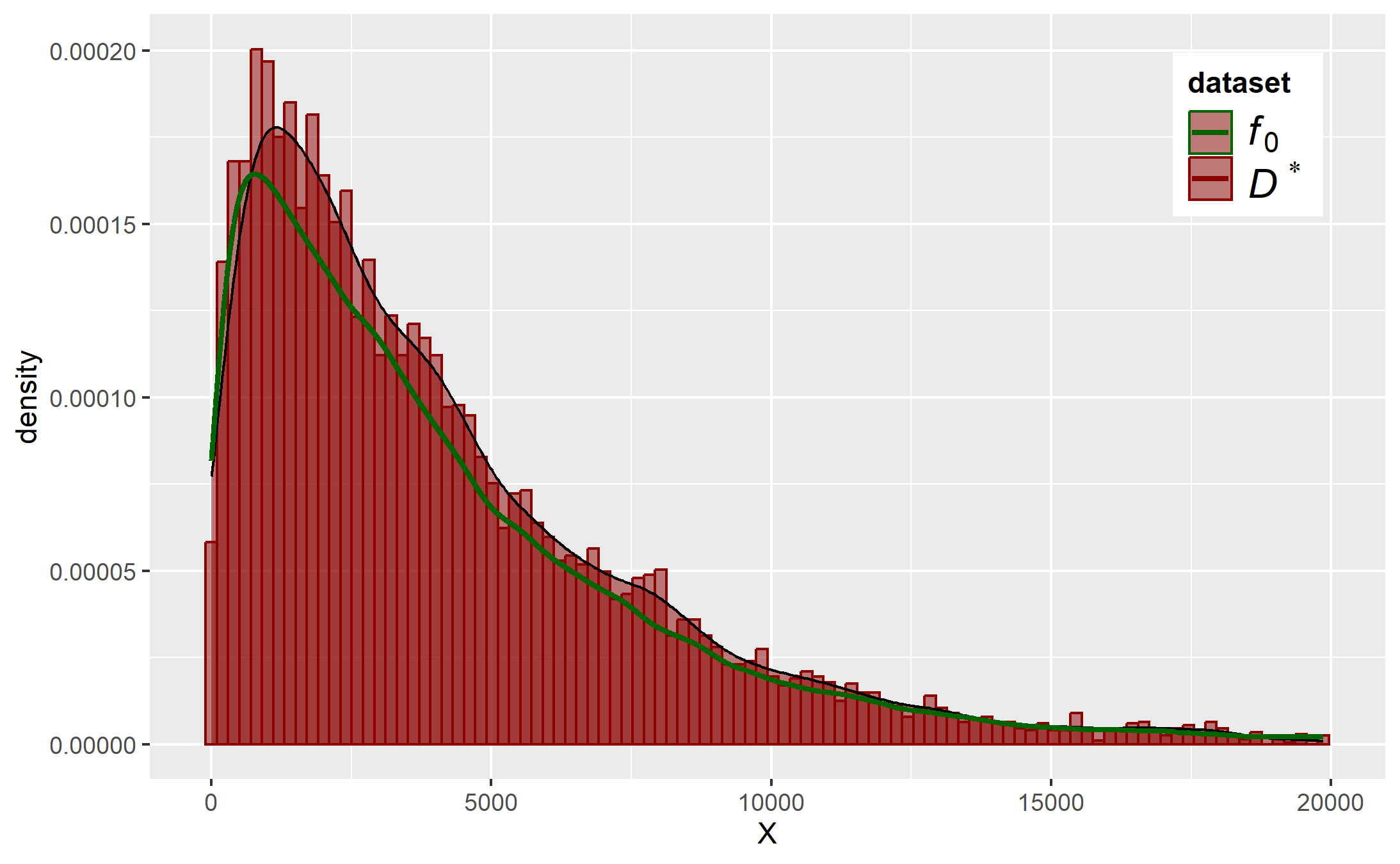}
    \caption{GN - GMM}
  \end{subfigure}
  \begin{subfigure}[b]{0.3\linewidth}
    \includegraphics[width=\linewidth]{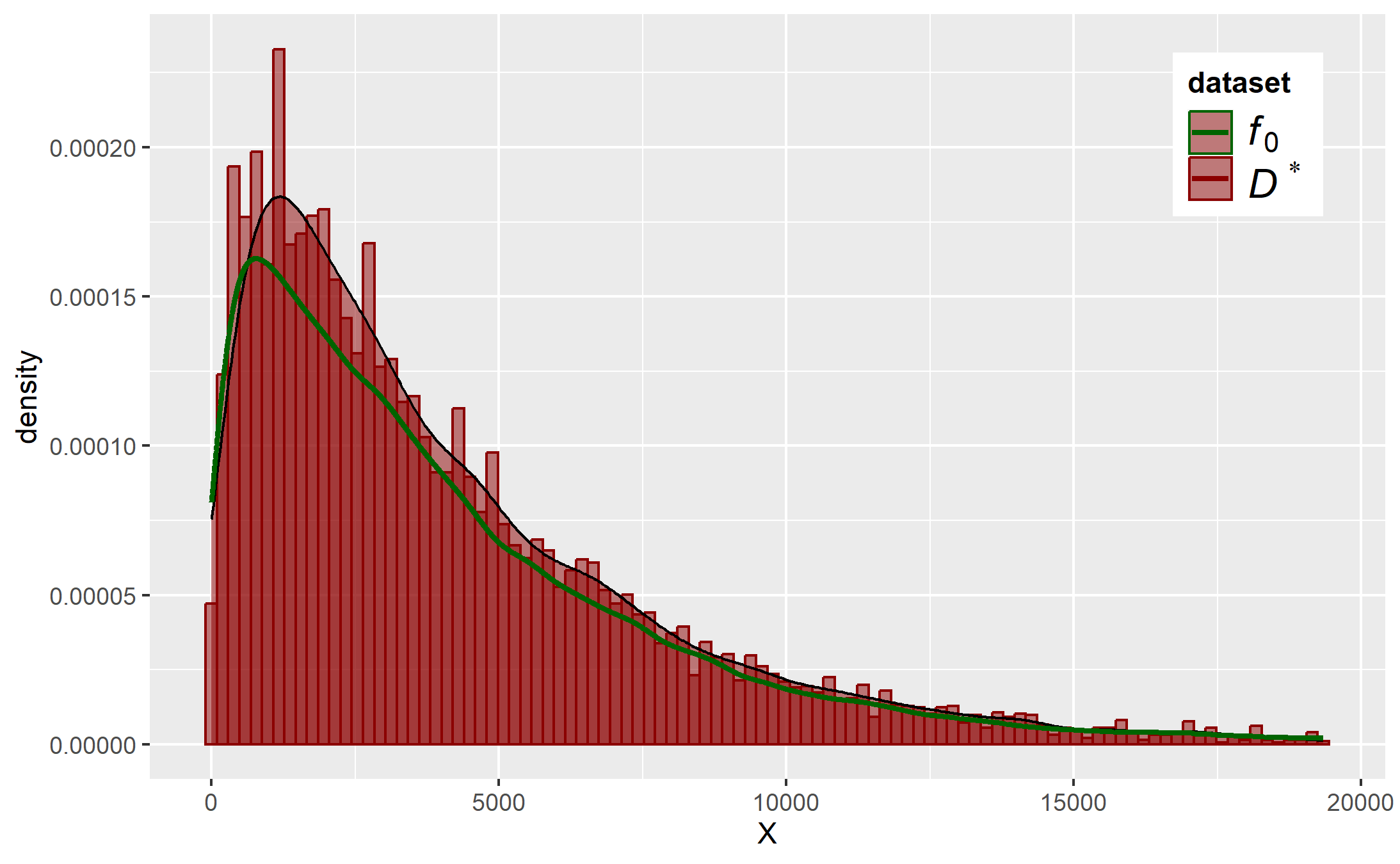}
     \caption{ROSE}
  \end{subfigure}
  \begin{subfigure}[b]{0.3\linewidth}
    \includegraphics[width=\linewidth]{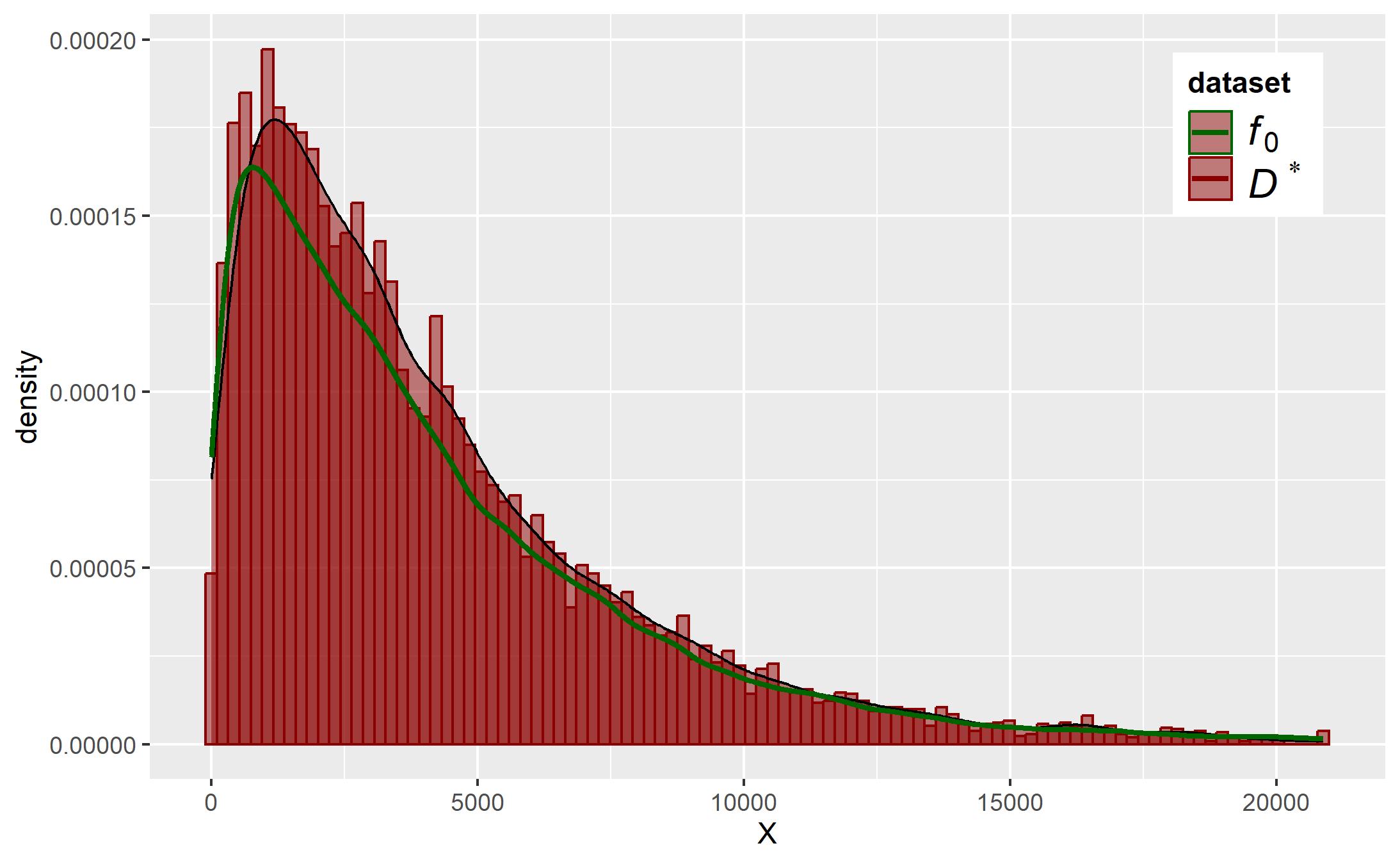}
    \caption{KDE}
  \end{subfigure}
  \begin{subfigure}[b]{0.3\linewidth}
    \includegraphics[width=\linewidth]{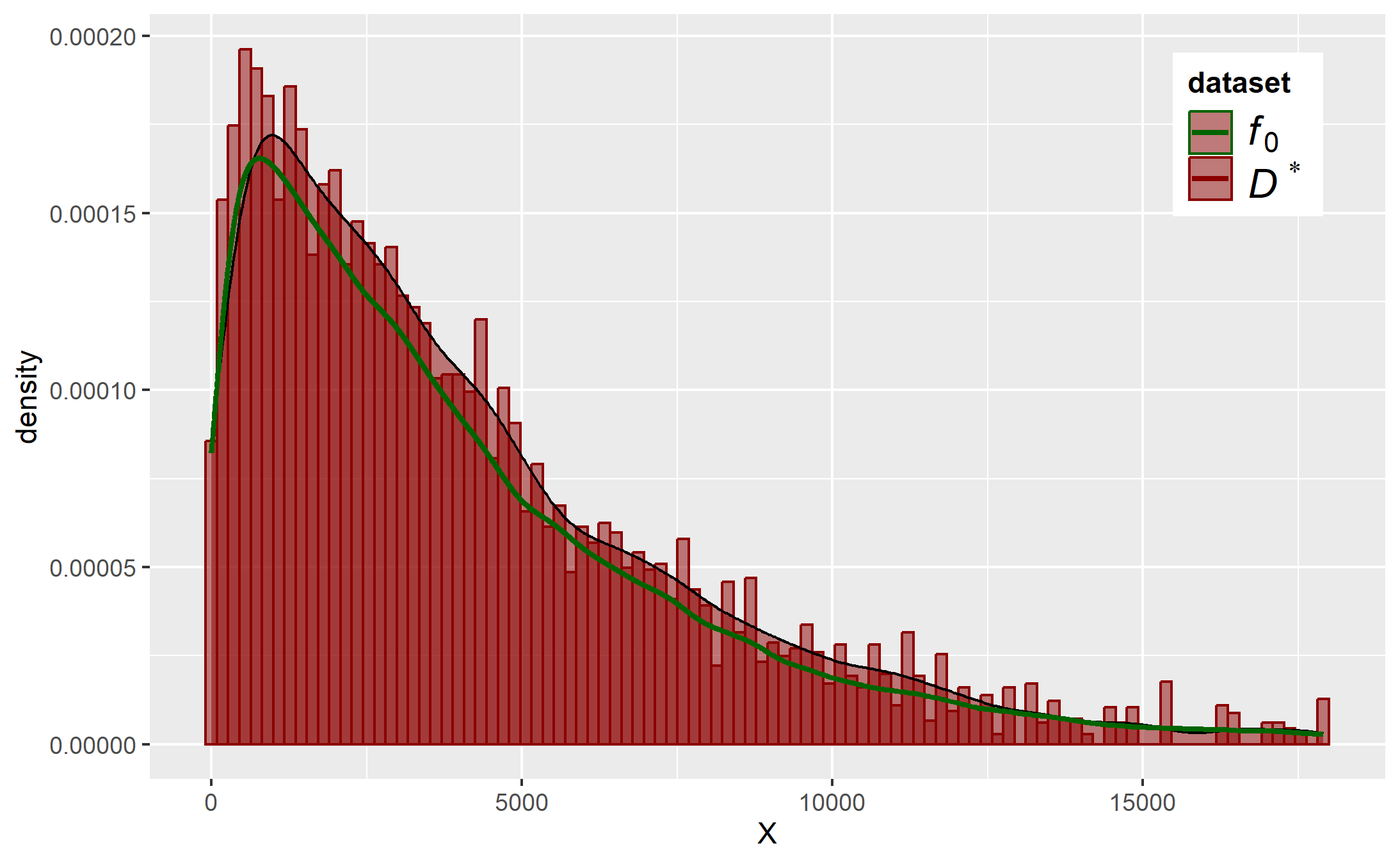}
    \caption{RF - GMM}
  \end{subfigure}
  \caption{Histogram of $X$ obtained in new samples (new) vs target (f0)}
  \label{Appli_Hist_X_ech_XXX-vs-Tgt}
\end{figure}

\textbf{Histogram of $X$ obtained in new samples vs WR}

\begin{figure}[H]
  \centering

  \begin{subfigure}[b]{0.3\linewidth}
    \includegraphics[width=\linewidth]{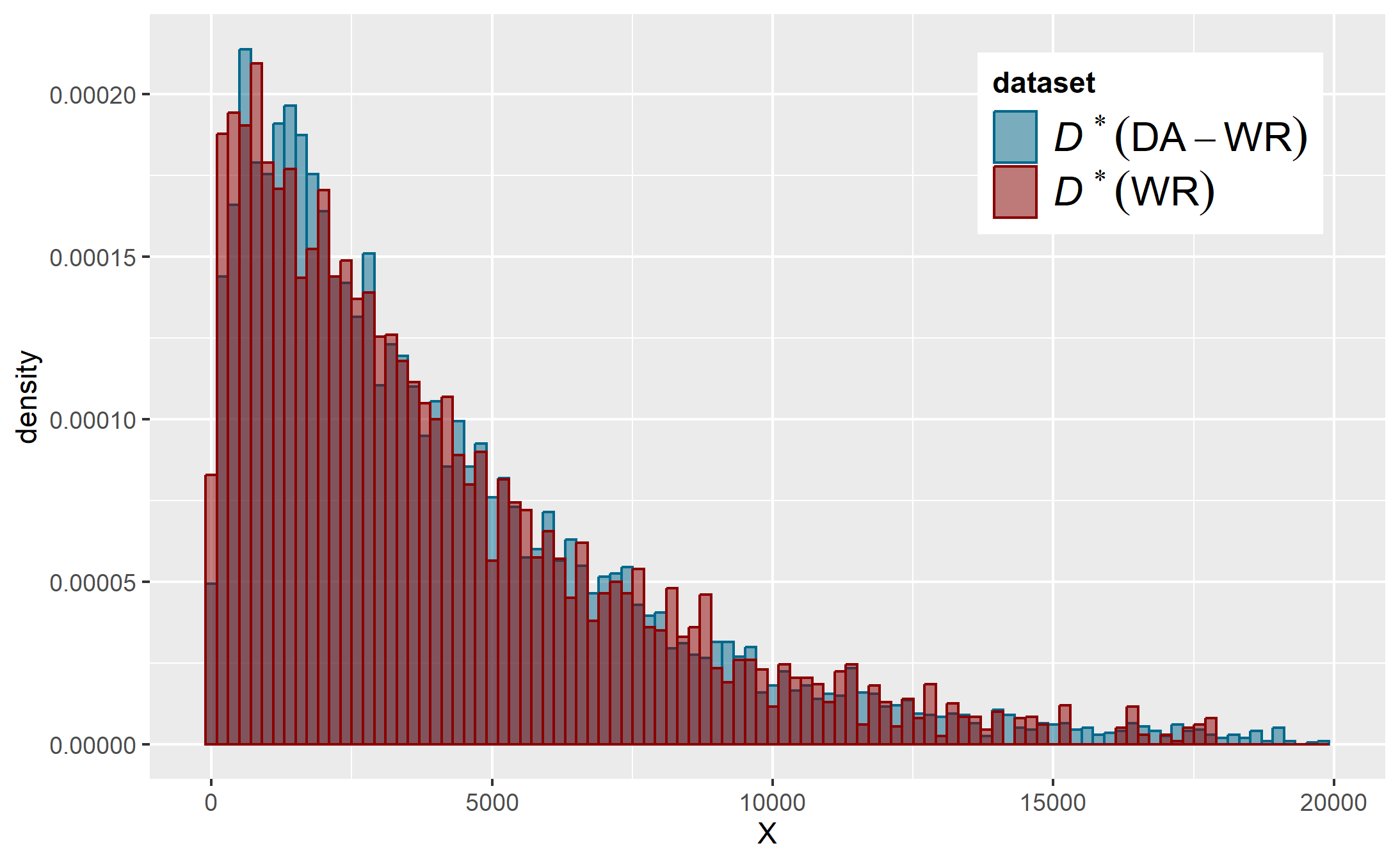}
     \caption{GN}
  \end{subfigure}
  \begin{subfigure}[b]{0.3\linewidth}
    \includegraphics[width=\linewidth]{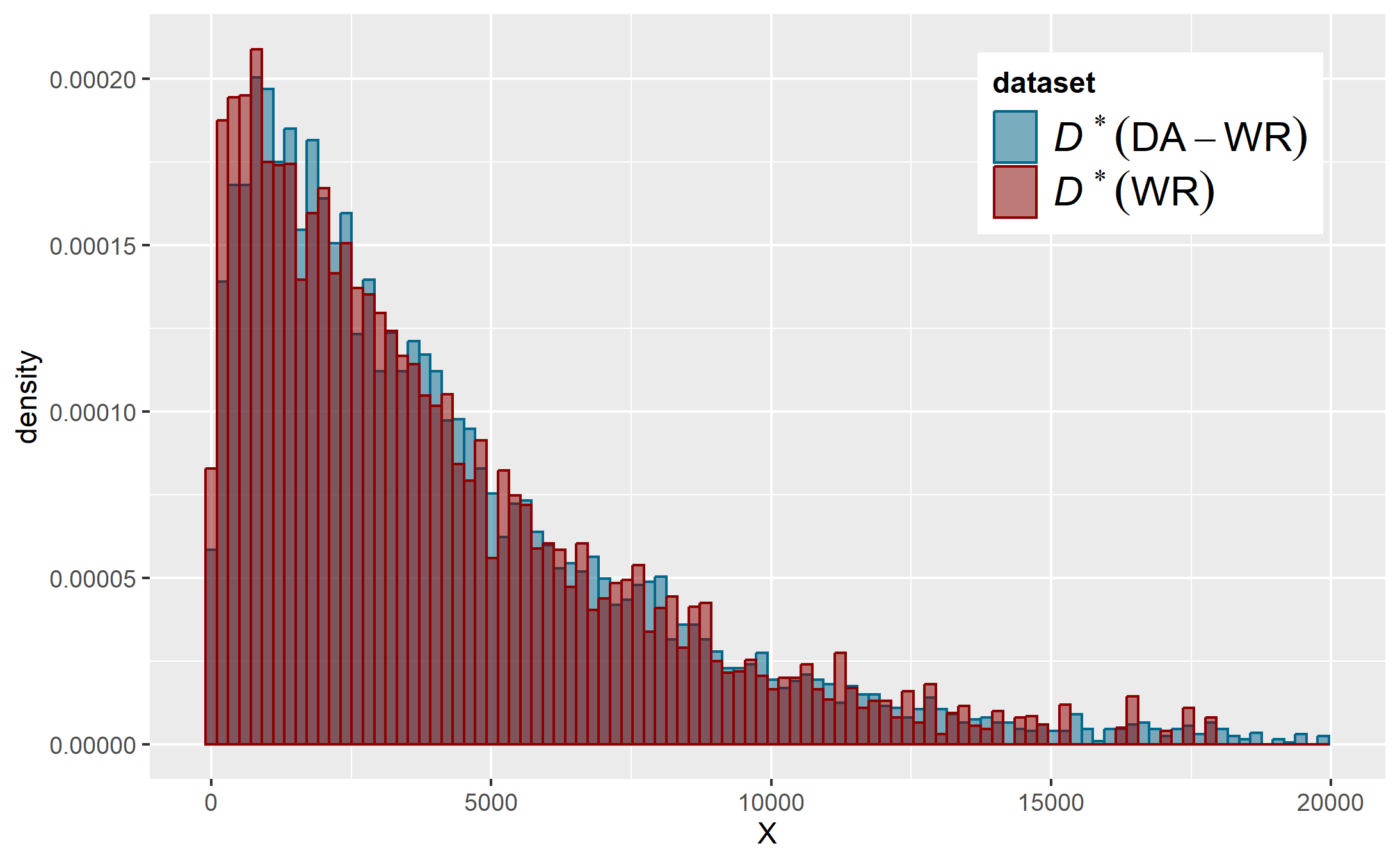}
     \caption{GN - GMM}
  \end{subfigure}
  \begin{subfigure}[b]{0.3\linewidth}
    \includegraphics[width=\linewidth]{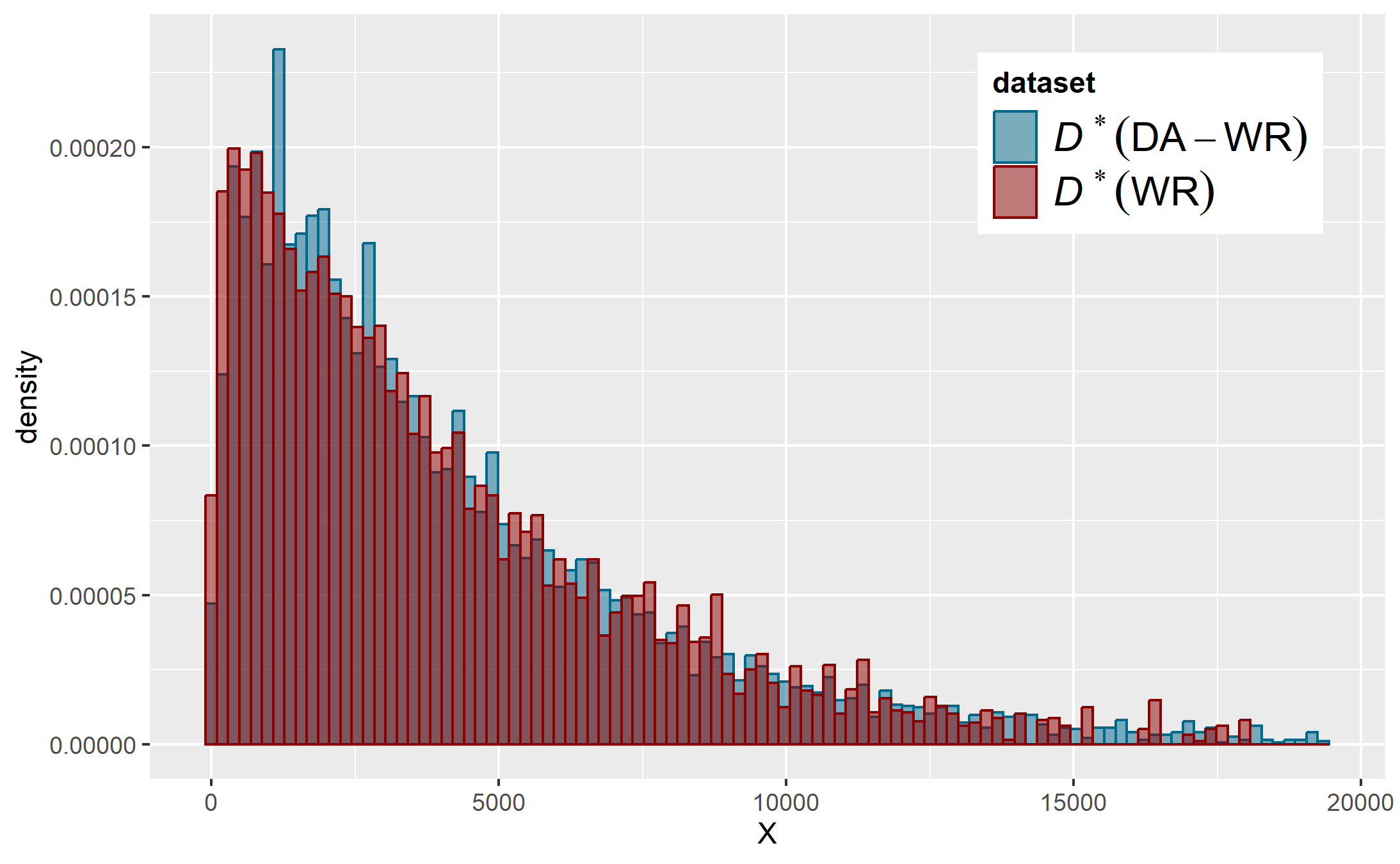}
    \caption{ROSE}
  \end{subfigure}
  \begin{subfigure}[b]{0.3\linewidth}
    \includegraphics[width=\linewidth]{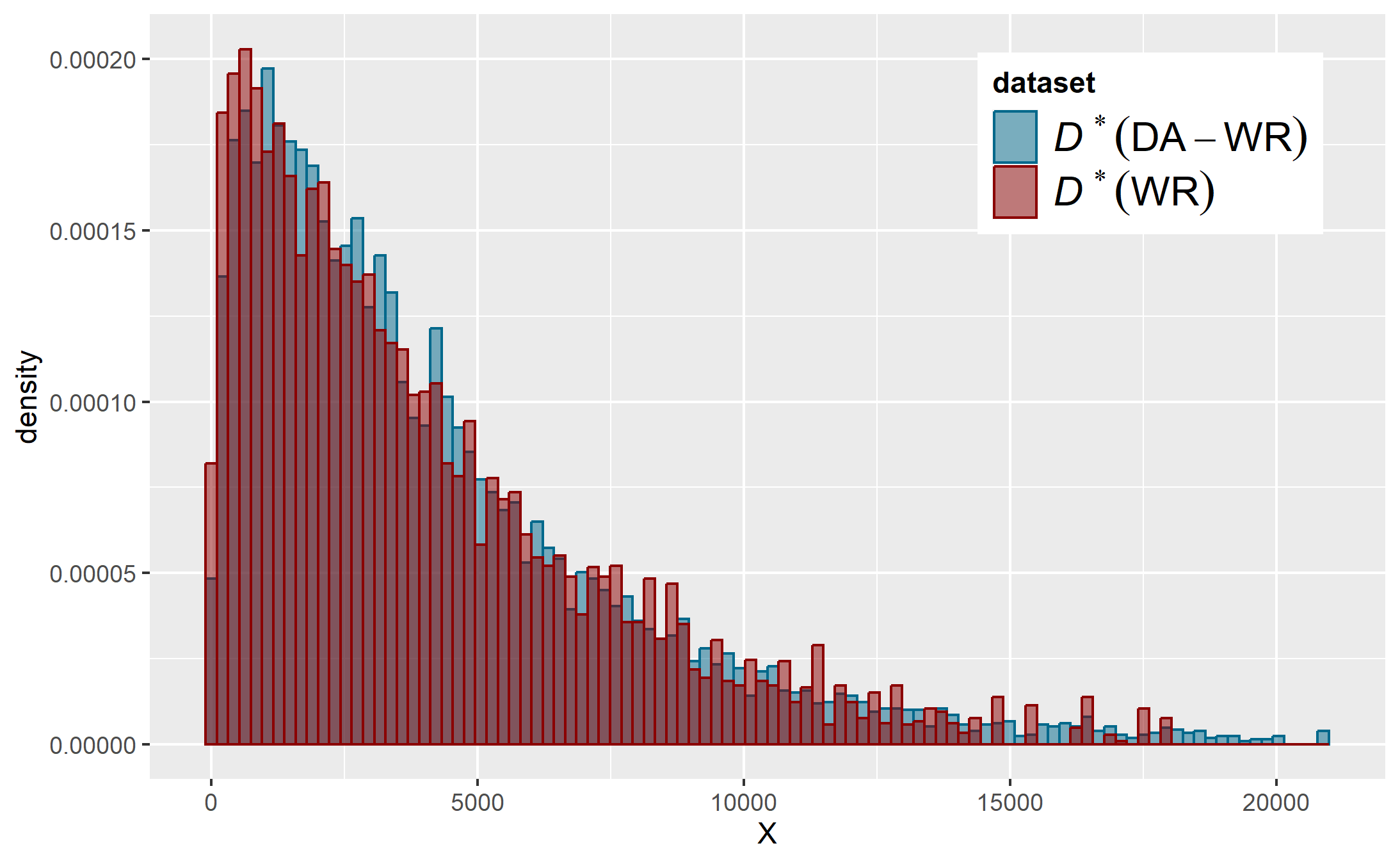}
    \caption{KDE}
  \end{subfigure}
  \begin{subfigure}[b]{0.3\linewidth}
    \includegraphics[width=\linewidth]{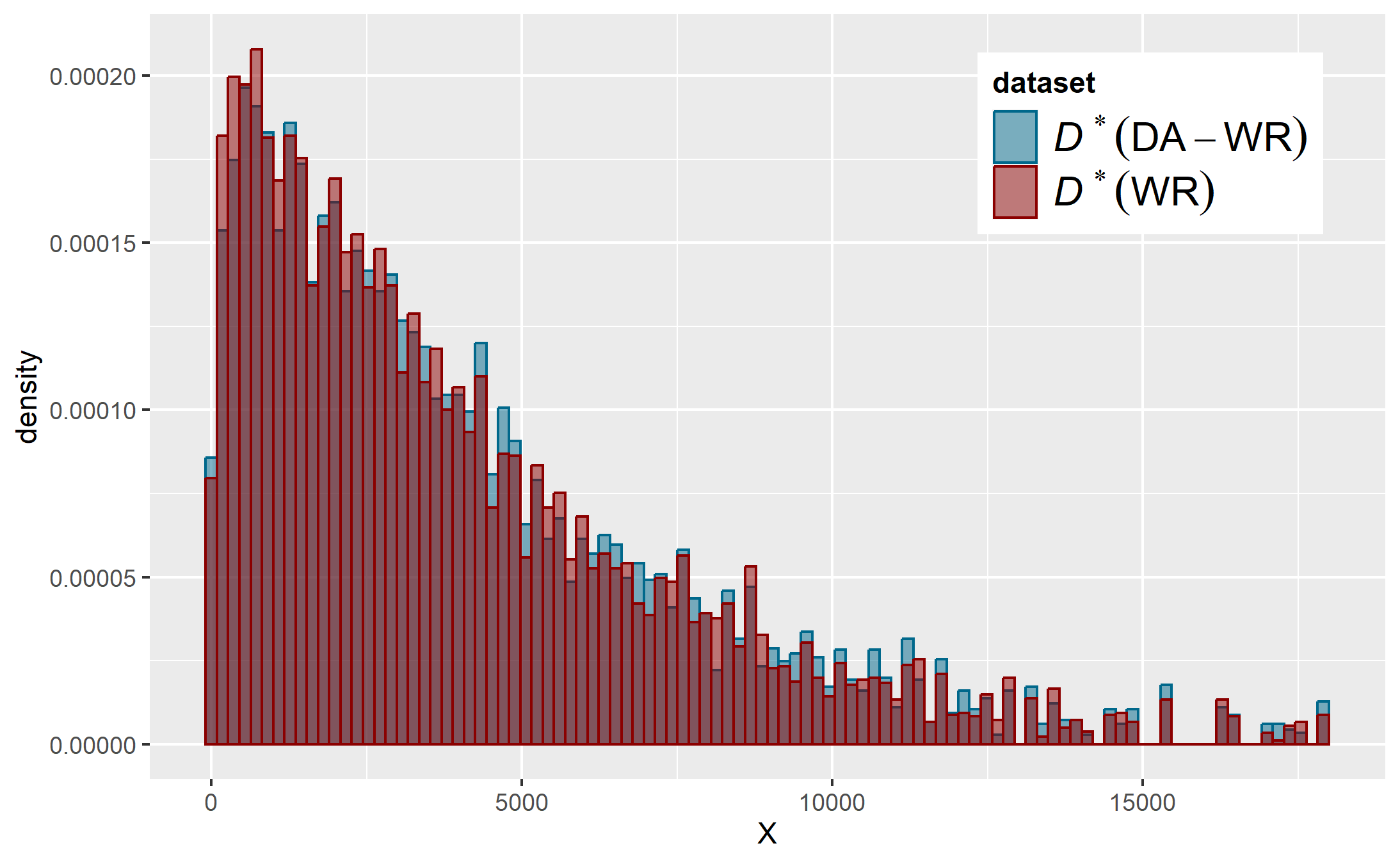}
    \caption{RF - GMM}
  \end{subfigure}
  \caption{Histogram of $X$ obtained in new samples (new) vs WR}
  \label{Appli_Hist_X_ech_XXX-vs-ech_add}
\end{figure}

\subsection{Predictions}

\textbf{Generalized Additive Model predictions}

\begin{figure}[H]
  \centering

  \begin{subfigure}[b]{0.3\linewidth}
    \includegraphics[width=\linewidth]{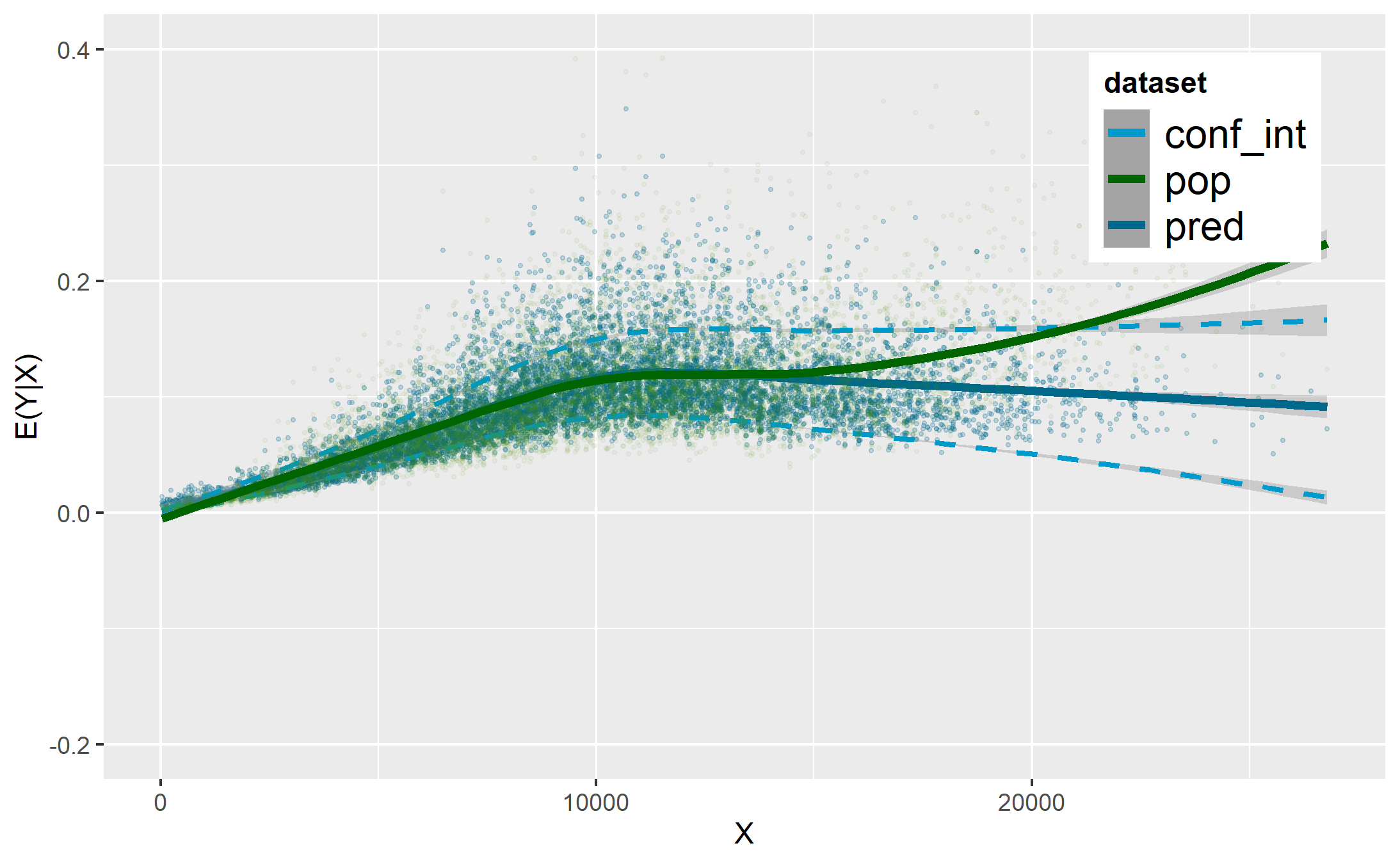}
     \caption{$\mathcal{D}^b$}
  \end{subfigure}
  \begin{subfigure}[b]{0.3\linewidth}
    \includegraphics[width=\linewidth]{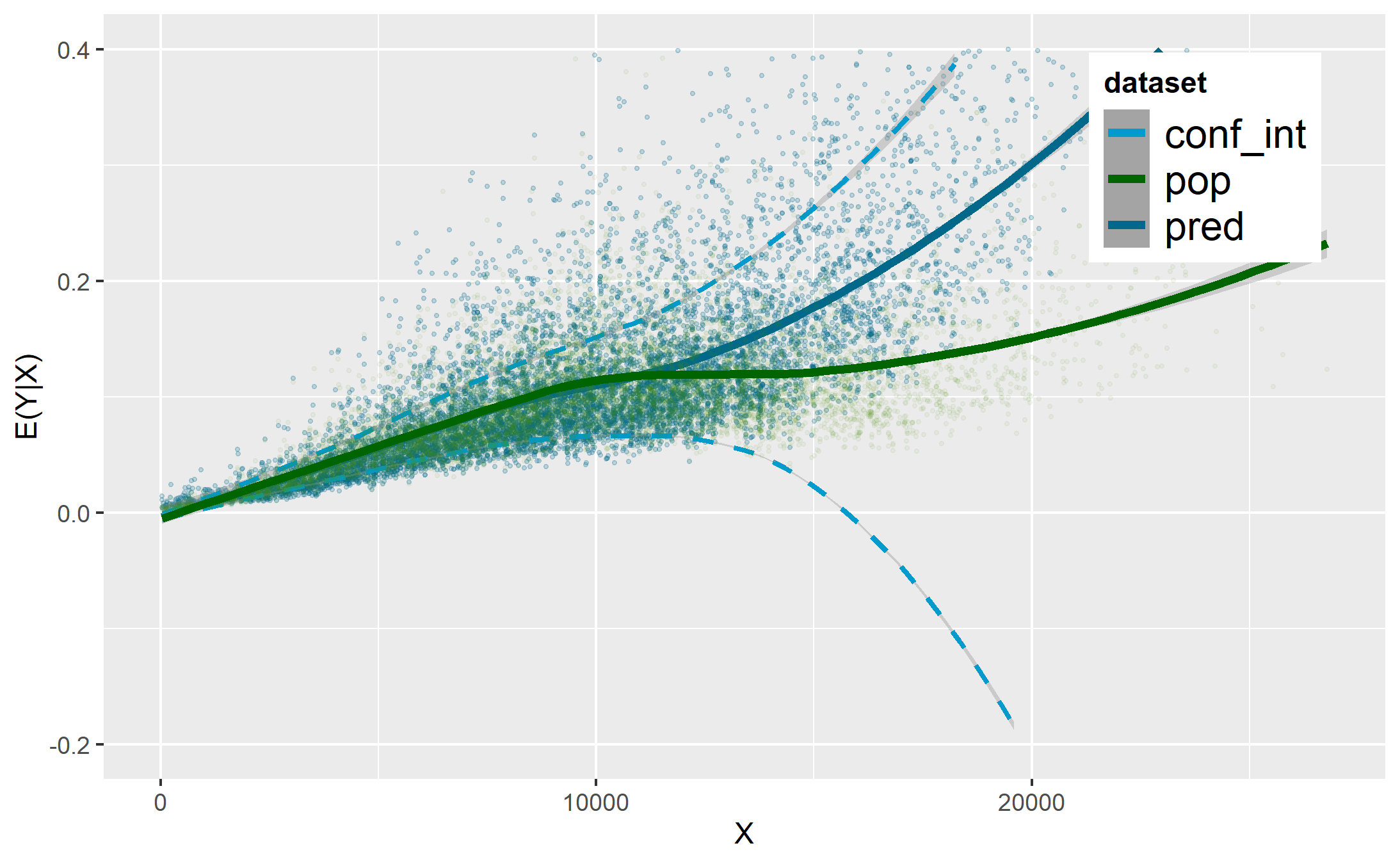}
    \caption{$\mathcal{D}^i$}
  \end{subfigure}
  \begin{subfigure}[b]{0.3\linewidth}
    \includegraphics[width=\linewidth]{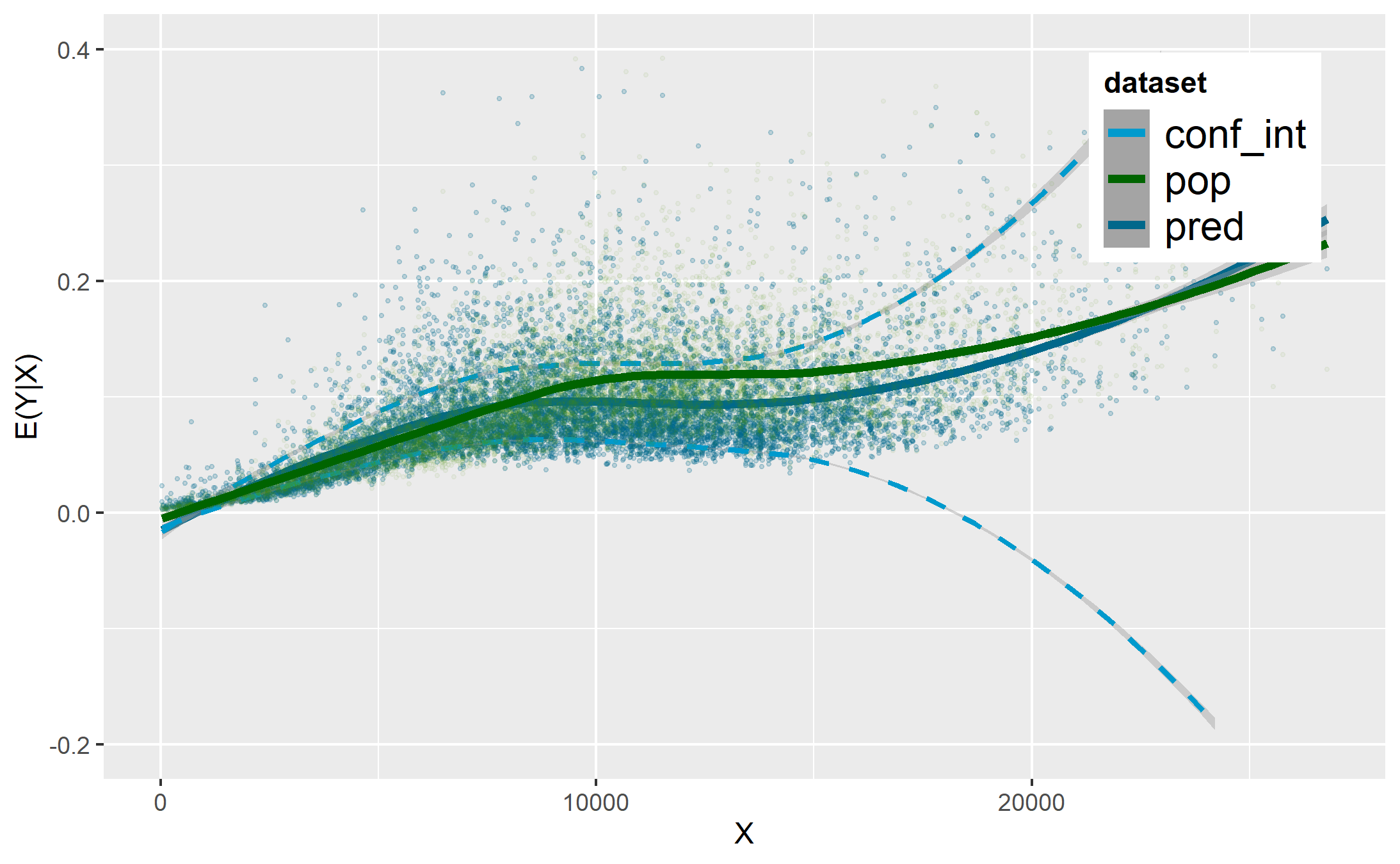}
    \caption{WR}
  \end{subfigure}
  \begin{subfigure}[b]{0.3\linewidth}
    \includegraphics[width=\linewidth]{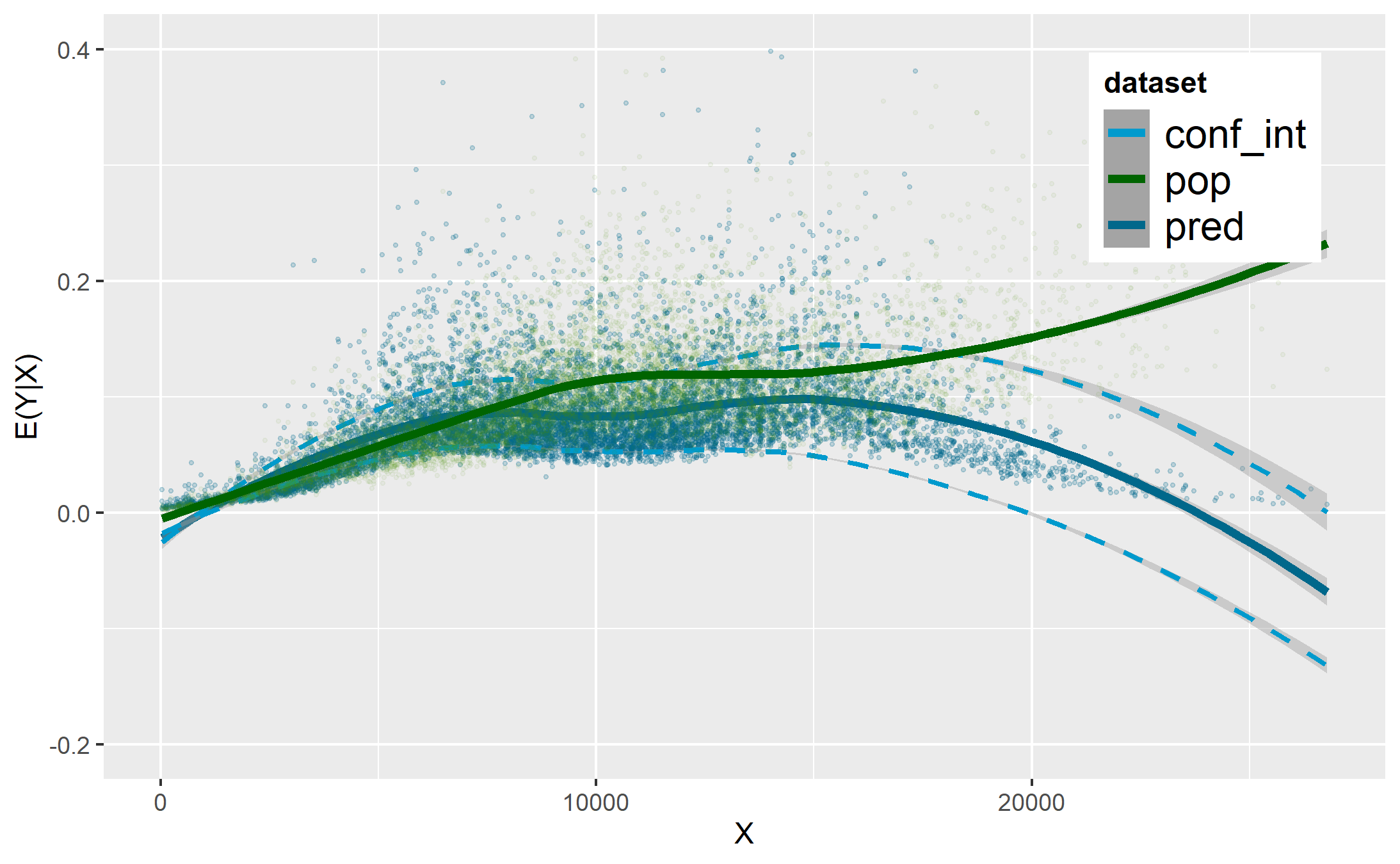}
     \caption{GN}
  \end{subfigure}
  \begin{subfigure}[b]{0.3\linewidth}
    \includegraphics[width=\linewidth]{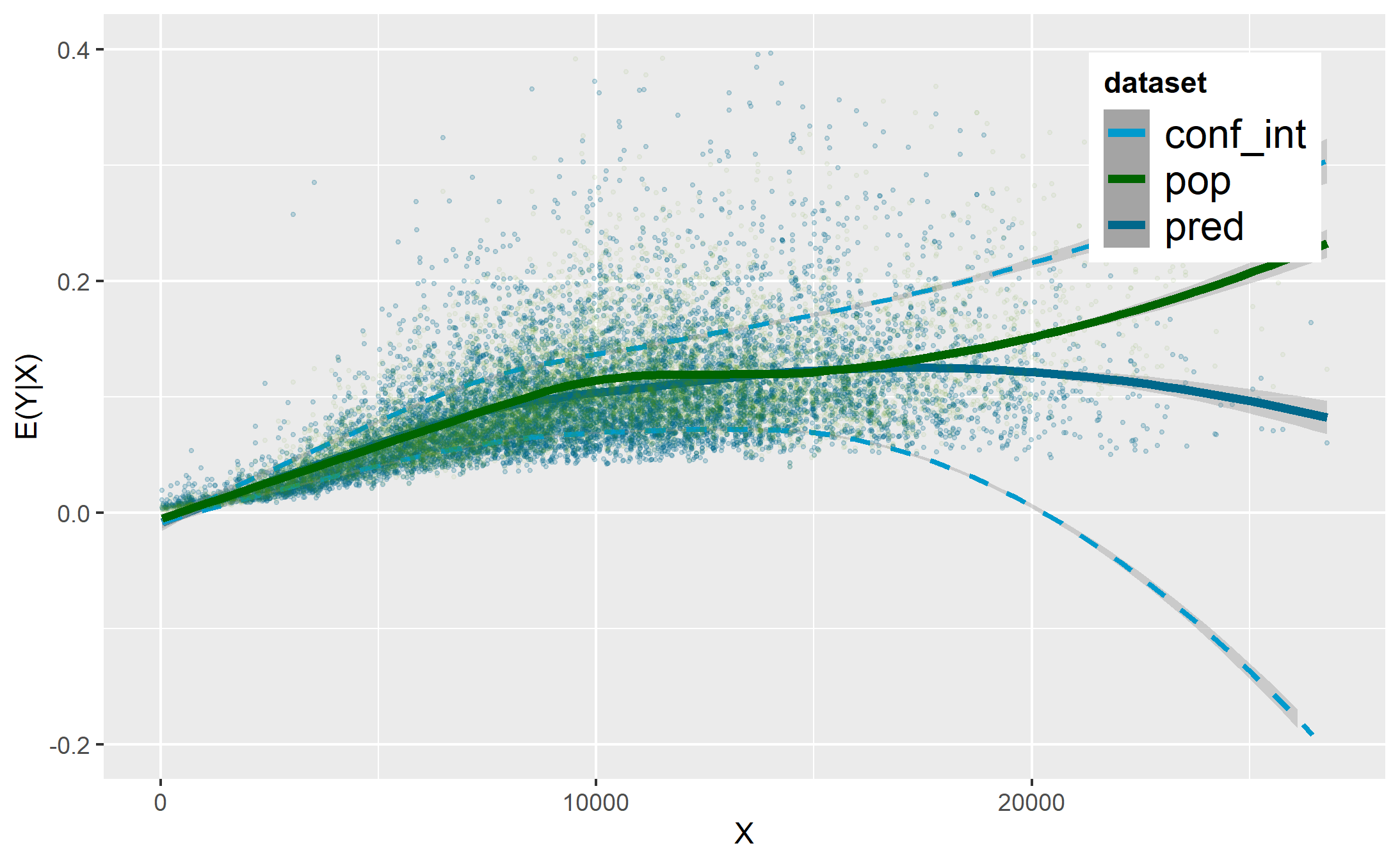}
    \caption{GN - GMM}
  \end{subfigure}
  \begin{subfigure}[b]{0.3\linewidth}
    \includegraphics[width=\linewidth]{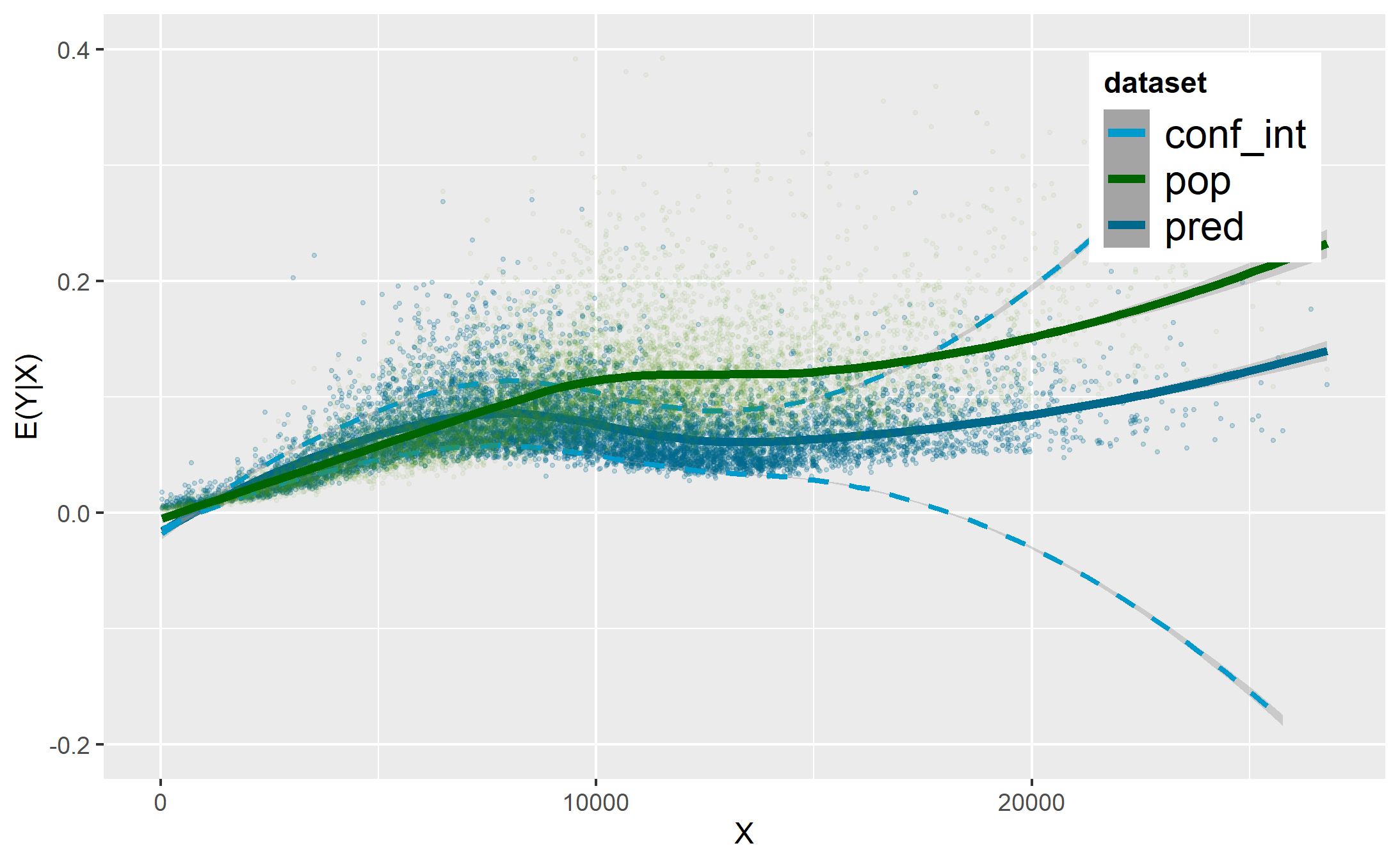}
    \caption{ROSE}
  \end{subfigure}
  \begin{subfigure}[b]{0.3\linewidth}
    \includegraphics[width=\linewidth]{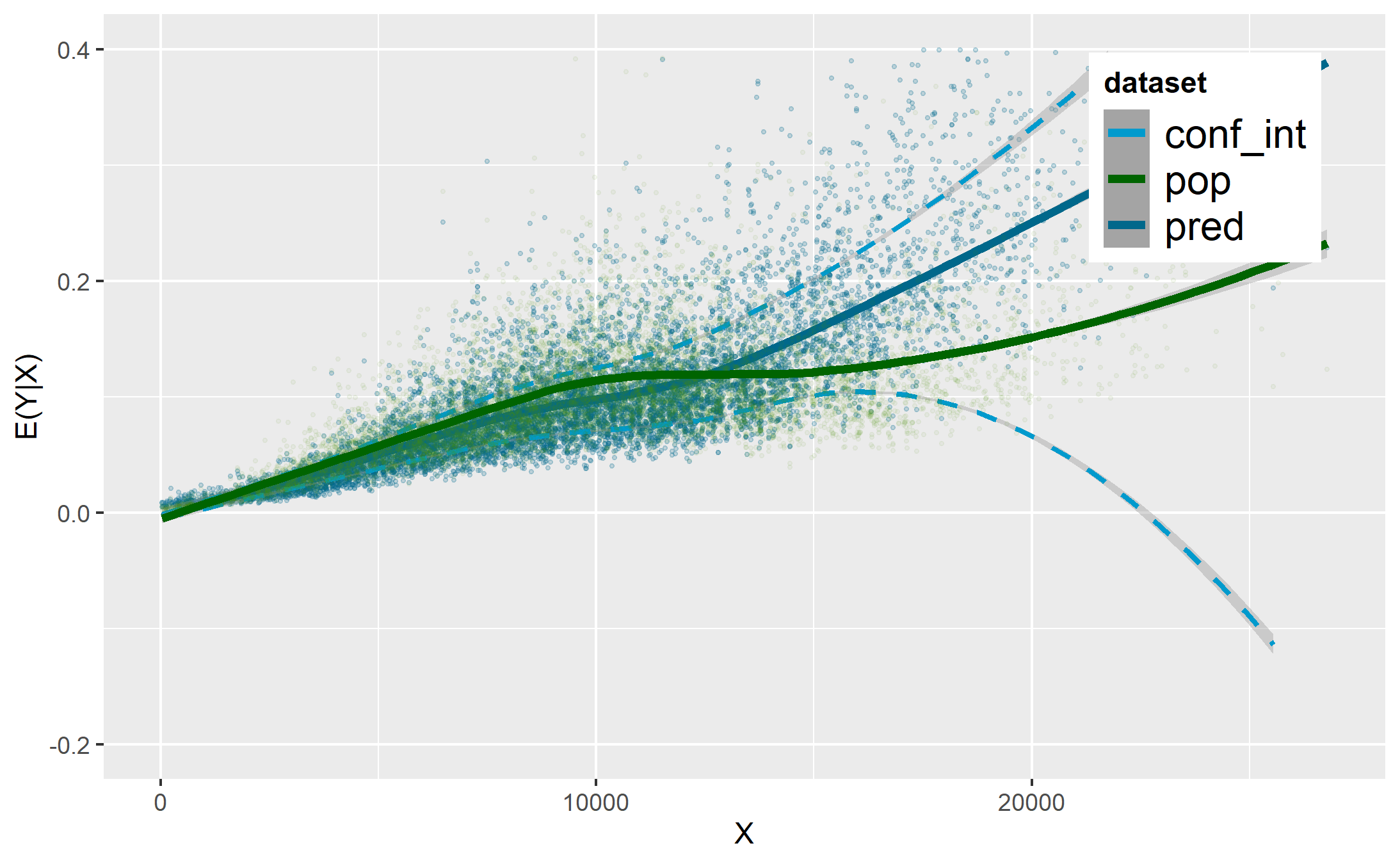}
    \caption{KDE}
  \end{subfigure}
  \begin{subfigure}[b]{0.3\linewidth}
    \includegraphics[width=\linewidth]{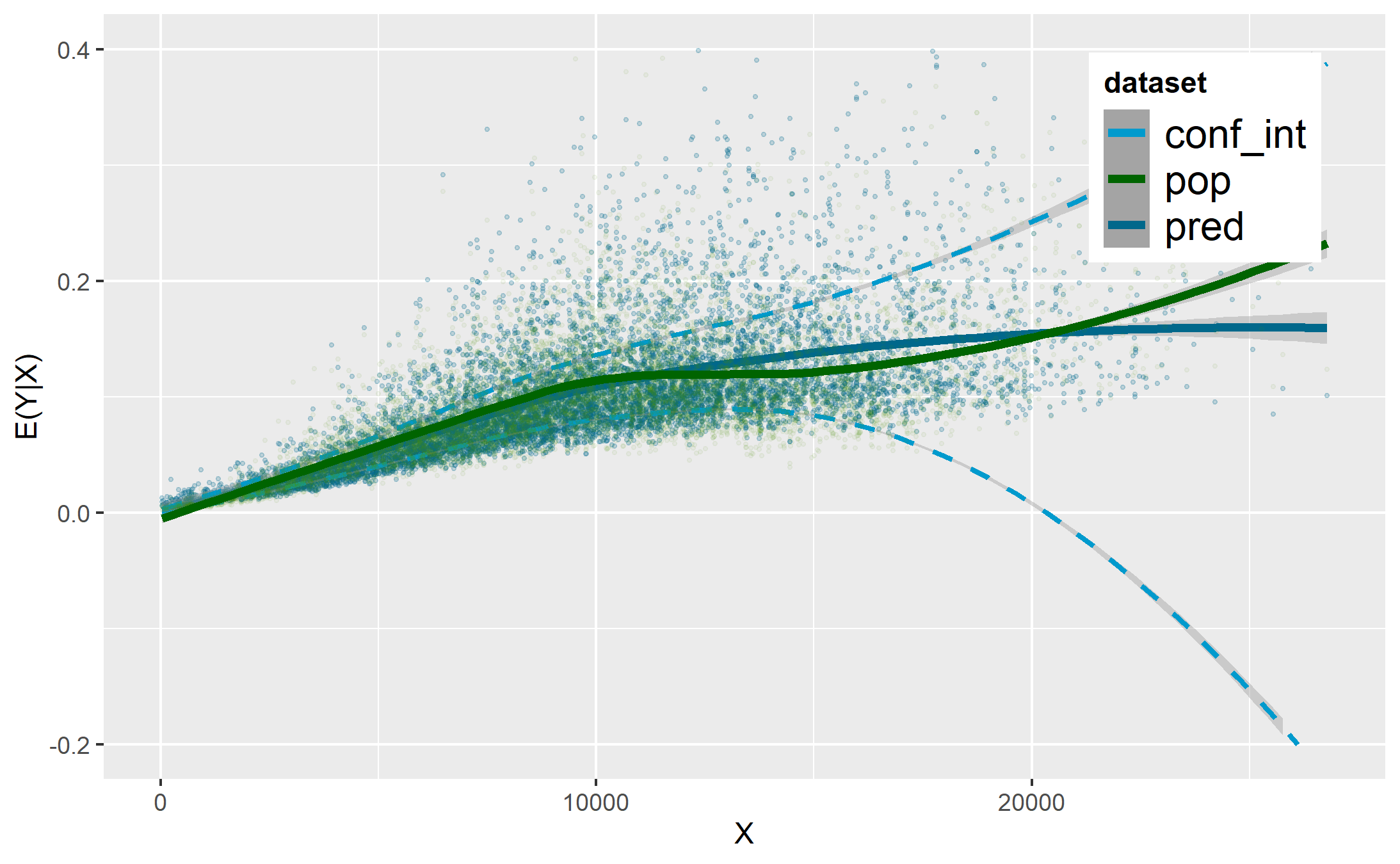}
    \caption{RF - GMM}
  \end{subfigure}  
  \caption{Smoothed predictions with GAM-ZIP}
  \label{Appli_pred_Y_Pois_ech_XXX-vs-test}
\end{figure}

\textbf{RMSE results}

\begin{table}[H]
\centering
\includegraphics[width=0.75 \textwidth]{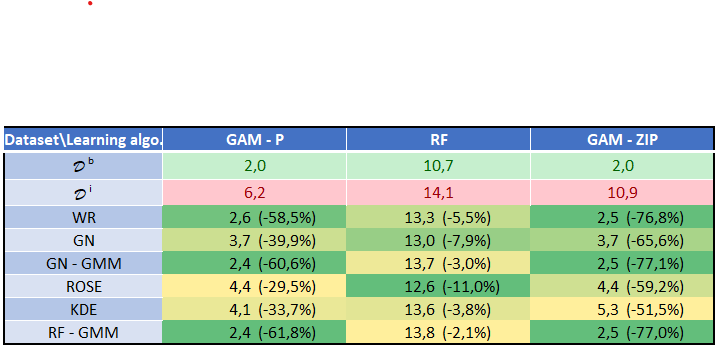}
\caption{RMSE on test dataset prediction, relative to the reference values}
\label{Appli_recap_RMSE_ref}
\end{table}

\subsection{Analysis of distribution of X}

\begin{figure}[H]
\centering
\includegraphics[width= 0.9 \textwidth]{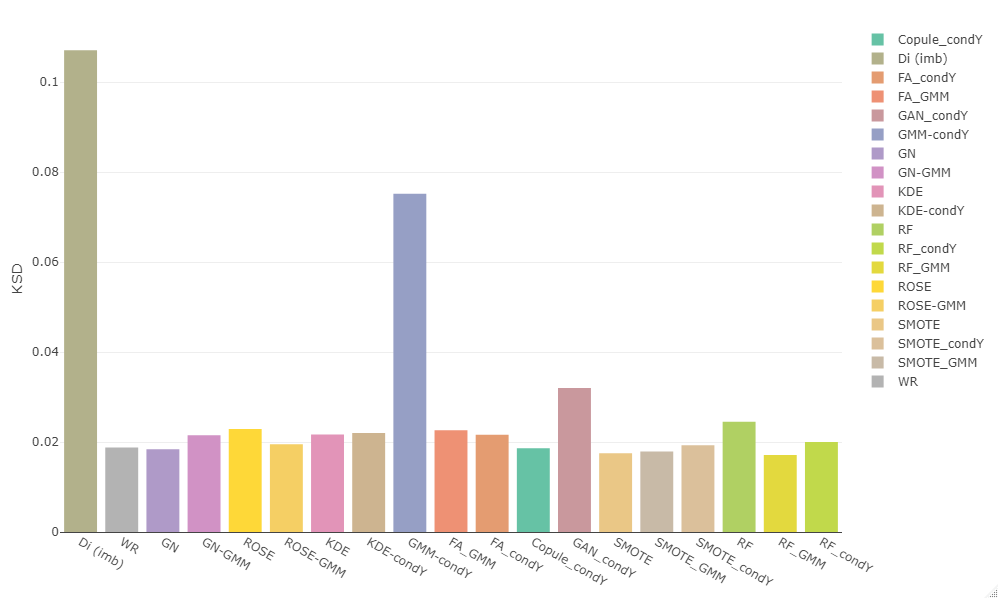}
\caption{Kolmogorv-Smirnov distance between the distribution of $X$ in the balanced sample and train samples}
\label{appli_KS_dist_X}
\end{figure}

\section{\textsc{Proof of Proposition \ref{propDAWR}}}

\begin{proof}
Writing $F^*$ and $ F^{**}$ the cdf associated to the WR and DA-WR procedures, respectively.
We have: 
\begin{align*}
   & \lim_{n \rightarrow \infty}|F^*(x)-F^{**}(x)|
    \  \\ & 
\ \ \    \leq 
    \di \sum_{i=1}^n q_i | \ind_{X_i \leq x} -\ind_{X_i'\leq x} 
     +
     \di \sum_{i=1}^n |q_i -q_i'|   \ind_{X_i' \leq x}| 
 \\ &
\ \ \  \leq 
n \max_{i=1,\cdots,n} \ind_{X_i\leq x} - 
\ind_{X_i'\leq x}|
 + 
n \max_{i=1,\cdots,n} |q_i -q_i' |, 
\end{align*}
which tends to zero and the result follows from Proposition \ref{propWR}. 
\end{proof}

\section{\textsc{Learning algorithms and evaluation metrics}}

We evaluate the impacts of the DA-WR algorithm by analysing the predictions of the test sample with three learning algorithms (only the first two for the application): one parametric and two non-parametric: 
\bit 
\item GAM: Generalized Additive Models\\
GAM represent the model class that generalizes the Generalized Linear Models approaches by extending the relationship between the covariates $X_1,\cdots,X_p$ and variable of interest $Y$ via functions such as $\mathbb{E}(Y|X) = \beta_0 + f_1(X_1) + \cdots + f_p(X_p)$. The links between $X$ and $Y$ are thus adjusted. We used the function \textit{gam} of the R-package \textit{MGCV} (the function \textit{zeroinfl} of the R-package \textit{PSCL} was also tested).

\item RF: Random Forest \\
RF is a natural generalization of Classification And Regression Trees (CART). We used the function \textit{randomForest} of the R-package \textit{randomForest}.

\item MARS: Multivariate Adaptative Regression Splines \\
MARS define a relationship between $X$ and $Y$ using hinge functions in order to capture the non-linear links and variable interactions. Indeed, hinge functions can break the range of  
X into bins. Then, for each bins, an effect is estimated (by coefficient). We can write that model as:   $g(\mathbb{E}(Y|X)) = \beta_0 + \beta_1 B_1(X) + \cdots + \beta_d B_d(X)$. $d$ is the knot number. $B_j(X)$ is a hinge function of $X$ and take the following form: $max(0, x-k)$ or $max(0,k-x)$, $k$ being a knot (constant). But it can be a product of two or more hinge functions, for example $B_j(X):= B_l(X) \times B_m(X)$ for two degrees of interaction. MARS automatically define variables and knot values of the hinge functions. We used the function \textit{earth} of the R-package \textit{earth}.

\eit

We evaluate the prediction results of the test sample through the performance indicator Root Mean Square Error (RMSE): $RMSE(Y, \widehat{Y}) := \left( \frac{\sum_{i=1}^n (y_i - \widehat{y_i})^2}{n} \right)^{1/2}$. For the illustration, $y_i$ is observed on the test sample and $\widehat{y_i}$ is the $y_i$ prediction obtained with the training sample.
For the application, $\widehat{y_i}$ is actually an estimate of $\mathbb{E}(Y|X)$ prediction. So, we preferred used the estimate of $\mathbb{E}(Y|X)$ obtained with using the remaining population as $y_i$ and the estimate of $\mathbb{E}(Y|X)$ obtained with using the training sample as $\widehat{y_i}$.

\section{\textsc{Data generation methods}} \label{generatorMethod}

Below, we briefly describe the different generators used in the illustration and the application. 
\bit 
\item  Perturbation approaches: \emph{Gaussian Noise} and \emph{Smoothed Boostrap}  \\
The idea of these methods is to simulate $N$ synthetic data by adding a noise on the initial observations. At first, an initial observation $(x_{.,i},y_i), i=1,\cdots,n$, called seed, is selected from the WR sample (or from the imbalanced sample by weighting the observations according to the WR method). Then, a synthetic data $(x_{.,m}^*,y_m^*), m=1,\cdots,N$ is generated as follows: $x_{.,m}^* = x_{.,i} + \epsilon(i)_{m}^x, y_m^* = y_i + \epsilon(i)_m^y $. The both methods are slightly different in the generation of the $\epsilon(i)_m$. The \emph{Gaussian Noise} method assumes  $\epsilon(i)_m^x \sim \mathcal{N}_p(0,\delta \times \Sigma_p)$, $\delta$ being set by the user and $\Sigma_p = diag(\widehat{\sigma}^2_1,\cdots,\widehat{\sigma}^2_p)$ a diagonal matrix $p \times p$, $\widehat{\sigma}^2_j$ being the estimated variance of $X_j$ in the initial sample. We have $\epsilon(i)_m^y \sim \mathcal{N}(0,\delta \times \sigma^2_Y)$. The \emph{Smoothed Boostrap} method suggests to define $\epsilon(i)_m$ according to a multivariate kernel density estimate $K_H(.,x_i)$ centered on the seed. The bandwidth matrix $H$ is defined according to the proposal of \cite{Bowman1999AppliedST}, $H = diag \left ( \left( \frac{4}{(p+2)n}\right)^{\frac{1}{p+4}} \widehat{\sigma_j} \right)$ or the proposal of \cite{Silverman86}, $H = \left( \left(\frac{4}{(p+2)n}\right)^{\frac{1}{p+4}} \widehat{\Sigma} \right)$, $\widehat{\Sigma}$ being the empirical covariance matrix from the initial sample.        

\item The interpolation approaches: \emph{k Nearest Neighbors} inspired by  \cite{Chawla2002Smote} \\
The purpose of this kind of methods is to create $N$ synthetic data by interpolation between two nearest neighbors. 
At first, an initial observation $(x_{.,i},y_i), i=1,\cdots,n$ is selected from the WR sample. One of the $k$ nearest neighbors of this seed is then drawn uniformly from the sample: $(x_{.,j},y_j)$. A synthetic data $(x_{.,m},y_m)$ is generated as follows: $x_{.,m} := x_{.,i} + \lambda \times (x_{.,j} - x_{.,i}), y_m := y_i + \lambda \times (y_j - y_i), \lambda \sim \mathcal{U}([0,1])$.

\item The latent structure approaches: \emph{Gaussian Mixture Models}, \emph{Factor Analysis}\\
As the kernel density estimate, the \emph{Gaussian Mixture Models} method suggests to estimate the density and it can be used as a generative data model. The model assumes that the distribution of observations can be specified by a multivariate density defined as a mixture model $G$ of: $f(z_i, \Phi) = \sum_g^G \pi_g f_g(z_i, \theta_g)$ where $z_i = (x_{.,i},y_i)$ ; $\Phi = {\pi_1, \cdots,\pi_G,\theta_1,\cdots,\theta_G}$. The mixture model parameters are such as $\pi_g > 0 , \forall g=1,\cdots,G$ and $\sum_g \pi_g = 1$ ; $\theta_g = (\mu_g,\Sigma_g)$ the parameter of the Gaussian distribution $f_g$. 
As the method \emph{GMM}, the \emph{Factor Analysis} technique allows to obtain a generative data model. This extension of the probabilistic model of principal component analysis proposes to decompose the observations $Z = (X,Y)$ such as $z_i = Wh_i + \mu + \epsilon$ where the vector $h_i$ is a latent vector supposed to be Gaussian: $h \sim \mathcal{N}(0,I)$ ; $\mu$ the paramter of the position (mean) ; $\epsilon$ the noise term distributed according to a centered Gaussian with a covariance matrix  $\Phi = diag(\phi_1,\cdots,\phi_n)$ i.e. $\epsilon \sim \mathcal{N}(0,\Phi)$ ; $W$ the " factor loading matrix" allowing to link the latent factor and the data. This model gives the following density form: $z$: $f(z) := \mathcal{N}(\mu, WW^T+\Phi)$.

\item A copula approach: \emph{Gaussian Copula Model} \cite{PatkiWV16}\\
The Gaussian copula is a distribution function defined on the unit hypercube and built from the multivariate Gaussian distribution. A copula $C$ allows to describe the joint distribution function of several random variables $F(x_1,\cdots,x_p)$ based on the dependence of the marginal distributions $F_1,\cdots,F_p$ such as $F(x_1,\cdots,x_p) = C(F_1(x_1),\cdots,F_p(x_p))$. 
This model allows also to generate some data from the copula: $C(u_1,\cdots,u_p) = \Phi_R(\Phi^{-1}(u_1),\cdots,\Phi^{-1}(u_p))$ where $\Phi^{-1}$ is the inverse distribution function of the univariate standard Gaussian distribution ; $\Phi_R$ the joint distribution function of a centered Gaussian and covariance matrix equal to the correlation matrix $R$. 

\item A \emph{Conditional Generative Adversarial Networks} approach \cite{DBLP:journals/corr/abs-1907-00503}\\
A GAN is a technique based on the competition between two networks: the generator, trying to replicate the observations as close as possible, and its adversary, the discriminator, trying to detect if the observations are real or simulated. The competition process allows them to improve their respective behaviors. The $CGAN$ allows to apply the method conditionally to a variable. 

\item A \emph{Ramdom forest} approach  \cite{Nowok2016synthpopBC} \\
The method \emph{Ramdom forest} consists to train several decision trees on various sets of observations and variables. This technique can be used to generate synthetic observations.  
\eit

\end{document}